\documentclass{article}
\usepackage[final,nonatbib]{include/neurips_2021}

\def\final{0}

\usepackage{amsmath,amsthm,amssymb, enumerate}
\usepackage[utf8]{inputenc} 
\usepackage[T1]{fontenc}    
\usepackage[breaklinks=true,bookmarks=false,colorlinks=true,linkcolor=blue]{hyperref}       
\usepackage{url}            
\usepackage{booktabs}       
\usepackage{amsfonts}       
\usepackage{amsmath,amsthm,amssymb,enumerate}
\usepackage{thm-restate}
\usepackage{amsxtra}
\usepackage{mathtools, cuted}
\usepackage{esint}
\usepackage{nicefrac}       
\usepackage{microtype}      
\usepackage{epsfig}
\usepackage{color}
\usepackage[dvipsnames]{xcolor}
\usepackage{cleveref}
\usepackage[normalem]{ulem}
\usepackage{graphicx}
\usepackage{appendix}
\usepackage{subcaption}
\usepackage{tcolorbox}
\usepackage{listings}
\usepackage{enumitem}

\usepackage{parskip}
\usepackage{ifthen}

\usepackage[export]{adjustbox}
\usepackage{dsfont}
\usepackage[ruled]{algorithm2e} 
\usepackage{algorithmic}
\usepackage{microtype}

\newcommand{\rednote}[1]{\red{#1}}

\newcommand{\diag}{\operatorname{diag}}
\newcommand{\abs}[1]{\left\lvert #1 \right\rvert}

\newcommand{\norm}[1]{\left\lVert #1 \right\rVert}

\newcommand{\rank}{\operatorname{rank}}

\renewcommand{\Pr}{\mathbb{P}}
\newcommand{\Exp}{\mathbb{E}}

\newcommand{\UNIFORM}{\operatorname{Unif}}

\newcommand{\NORMAL}{\operatorname{\mathcal{N}}}

\newcommand{\distas}{\mathbin{\sim}}

\newcommand{\bE}{\boldsymbol{E}}

\newcommand{\ind}[1]{\mathds{1}_{#1}}

\newcommand{\avg}{\operatorname{avg}}

\renewcommand{\phi}{\varphi}
\newcommand{\bzeta}{\boldsymbol{\zeta}}
\newcommand{\bzero}{\boldsymbol{0}}
\newcommand{\bone}{\boldsymbol{1}}
\newcommand{\tp}[1]{#1 ^ {\top}}

\newcommand\eu{\mathrm{e}}
\newcommand\dd{\,\mathrm{d}}

\newcommand{\bSigma}{\boldsymbol{\Sigma}}
\newcommand{\bLambda}{\boldsymbol{\Lambda}}

\newcommand{\ld}{\lambda}

\newcommand{\bLd}{\boldsymbol{\bLambda}}

\newcommand{\ip}[2]{\left\langle{#1,#2}\right\rangle}
\DeclareMathOperator*{\argmax}{argmax}

\newcommand{\softmax}{\operatorname{softmax}}
\newcommand{\perr}{p_{\mathrm{err}}}

\newcommand{\geomnist}{\textsc{GeoMNIST}}
\newcommand{\cifar}{\textsc{CIFAR}}
\newcommand{\alexnet}{\textsc{AlexNet}}

\newcommand{\mH}{\textsf{L}-model}
\newcommand{\mL}{\textsf{L}-model}
\newcommand{\mI}{\textsf{I}-model}

\newcommand{\sR}{\mathbb R}

\renewcommand{\c}{{\boldsymbol c}}
\renewcommand{\d}{{\boldsymbol d}}
\newcommand{\e}{{\boldsymbol e}}
\newcommand{\f}{{\boldsymbol f}}

\newcommand{\s}{{\boldsymbol s}}

\renewcommand{\u}{{\boldsymbol u}}
\renewcommand{\v}{{\boldsymbol v}}
\newcommand{\w}{{\boldsymbol w}}
\newcommand{\x}{{\boldsymbol x}}

\newcommand{\z}{{\boldsymbol z}}
\newcommand{\A}{{\boldsymbol A}}
\newcommand{\B}{{\boldsymbol B}}

\newcommand{\E}{{\boldsymbol E}}

\renewcommand{\H}{{\boldsymbol H}}
\newcommand{\I}{{\boldsymbol I}}

\newcommand{\M}{{\boldsymbol M}}

\renewcommand{\P}{{\boldsymbol P}}
\newcommand{\Q}{{\boldsymbol Q}}

\renewcommand{\S}{{\boldsymbol S}}

\newcommand{\W}{{\boldsymbol W}}
\newcommand{\X}{{\boldsymbol X}}
\newcommand{\Y}{{\boldsymbol Y}}

\newcommand{\bnu}{{\boldsymbol \nu}}

\DeclareFontFamily{U}{mathx}{\hyphenchar\font45}
\DeclareFontShape{U}{mathx}{m}{n}{
      <5> <6> <7> <8> <9> <10>
      <10.95> <12> <14.4> <17.28> <20.74> <24.88>
      mathx10
      }{}
\DeclareSymbolFont{mathx}{U}{mathx}{m}{n}
\DeclareFontSubstitution{U}{mathx}{m}{n}
\DeclareMathAccent{\widecheck}{0}{mathx}{"71}
\DeclareMathAccent{\wideparen}{0}{mathx}{"75}

\renewcommand{\tilde}[1]{\widetilde{#1}}

\renewcommand{\hat}[1]{\widehat{#1}}

\renewcommand{\check}[1]{\widecheck{#1}}

\newcommand{\reviewer}[3]{
  \expandafter\newcommand\csname #1\endcsname[1]{
    \ifthenelse{\equal{\final}{1}} {
      \textcolor{#3}{}
    } {
    \textcolor{#3}{[\textsf{\footnotesize \textbf{#2:} ##1]}}
    }
  }
}

\newtheorem{fact}{Fact}[section]
\newtheorem{lemma}[fact]{Lemma}
\newtheorem{theorem}[fact]{Theorem}
\newtheorem{definition}[fact]{Definition}
\newtheorem{corollary}[fact]{Corollary}
\newtheorem{proposition}[fact]{Proposition}

\newtheorem*{remark}{Remark}

\newtheorem{assumption}[fact]{Assumption}
\theoremstyle{definition}
\newtheorem{problem}[fact]{Problem}
\newenvironment{namedthm}[1]{\newline
    \textbf{Theorem #1.}\itshape
}{
\newline}
\newtheorem{example}[fact]{Example}

\crefname{equation}{equation}{equations}
\Crefname{equation}{Equation}{Equations}
\crefname{theorem}{theorem}{theorems}
\Crefname{theorem}{Theorem}{Theorems}
\crefname{assumption}{assumption}{assumptions}
\Crefname{assumption}{Assumption}{Assumptions}
\crefname{lemma}{lemma}{lemmas}
\Crefname{lemma}{Lemma}{Lemmas}
\crefname{definition}{definition}{definitions}
\Crefname{definition}{Definition}{Definitions}
\crefname{corollary}{corollary}{corollaries}
\Crefname{corollary}{Corollary}{Corollaries}
\crefname{proposition}{proposition}{propositions}
\Crefname{proposition}{Proposition}{Propositions}
\crefname{claim}{claim}{claims}
\Crefname{claim}{Claims}{Claims}
\crefname{problem}{problem}{problems}
\Crefname{problem}{Problem}{Problems}
\crefname{solution}{solution}{solutions}
\Crefname{solution}{Solution}{Solutions}
\crefname{proof}{proof}{proofs}
\Crefname{proof}{Proof}{Proofs}
\crefname{proofof}{proof}{proofs}
\Crefname{proofof}{Proof}{Proofs}
\crefname{algocf}{algorithm}{algorithms}
\Crefname{algocf}{Algorithm}{Algorithms}

\newcommand{\btcb}{\begin{tcolorbox}}
\newcommand{\etcb}{\end{tcolorbox}}
\newcommand{\bbm}{\begin{bmatrix}}
\newcommand{\ebm}{\end{bmatrix}}
\newcommand{\bassume}{\begin{assumption}}
\newcommand{\eassume}{\end{assumption}}
\newcommand{\be}{\begin{equation}}
\newcommand{\ee}{\end{equation}}
\newcommand{\ben}{\begin{equation*}}
\newcommand{\een}{\end{equation*}}
\newcommand{\bea}{\begin{aligned}}
\newcommand{\eea}{\end{aligned}}
\newcommand{\ba}{\begin{equation}\begin{aligned}}
\newcommand{\ea}{\end{aligned}\end{equation}}
\newcommand{\bd}{\begin{definition}}
\newcommand{\ed}{\end{definition}}
\newcommand{\bprop}{\begin{proposition}}
\newcommand{\eprop}{\end{proposition}}
\newcommand{\bt}{\begin{theorem}}
\newcommand{\et}{\end{theorem}}
\newcommand{\bcor}{\begin{corollary}}
\newcommand{\ecor}{\end{corollary}}
\newcommand{\beg}{\begin{example}}
\newcommand{\eeg}{\end{example}}
\newcommand{\bnt}[1]{\begin{namedthm}{#1}}
\newcommand{\ent}{\end{namedthm}}
\newcommand{\blm}{\begin{lemma}}
\newcommand{\elm}{\end{lemma}}
\newcommand{\bp}{\begin{proof}}
\newcommand{\ep}{\end{proof}}
\newcommand{\bpb}{\begin{problem}}
\newcommand{\epb}{\end{problem}}
\newcommand{\benum}{\begin{enumerate}}
\newcommand{\eenum}{\end{enumerate}}
\newcommand{\bitem}{\begin{itemize}}
\newcommand{\eitem}{\end{itemize}}
\definecolor{firebrick}{RGB}{178,34,34}

\def\brst\begin{restatable}
\newcommand{\erst}{\end{restatable}}

\renewcommand{\rednote}[1]{\textcolor{black}{#1}}
\newcommand{\toappendix}[1]{}
\graphicspath{{graphics/pdf/}}

\reviewer{weijie}{Weijie}{BrickRed}
\reviewer{hua}{Hua}{Peach}
\reviewer{jiayao}{Jiayao}{MidnightBlue}
\reviewer{TODO}{TODO}{Violet}

\hypersetup{
    colorlinks = false,
    linkbordercolor = {white},
}

\title{Imitating Deep Learning Dynamics via Locally Elastic Stochastic Differential Equations}

\author{Jiayao Zhang \quad\quad\quad Hua Wang \quad\quad\quad Weijie J.~Su \\[.5em]
University of Pennsylvania \\[.5em]
\texttt{\{zjiayao,wanghua,suw\}@wharton.upenn.edu}
}

\begin{document}
\maketitle

\begin{abstract}

Understanding the training dynamics of deep learning models is perhaps a necessary step toward demystifying the effectiveness of these models. In particular, how do data from different classes gradually become separable in their feature spaces when training neural networks using stochastic gradient descent? In this study, we model the evolution of features during deep learning training using a set of stochastic differential equations (SDEs) that each corresponds to a training sample. As a crucial ingredient in our modeling strategy, each SDE contains a drift term that reflects the impact of backpropagation at an input on the features of all samples. Our main finding uncovers a sharp phase transition phenomenon regarding the \textit{intra-class} impact: if the SDEs are \textit{locally elastic}~\cite{he2019le} in the sense that the impact is more significant on samples from the same class as the input, the features of the training data become linearly separable, meaning vanishing training loss; otherwise, the features are not separable, regardless of how long the training time is. Moreover, in the presence of local elasticity, an analysis of our SDEs shows that the emergence of a simple geometric structure called the neural collapse of the features. Taken together, our results shed light on the decisive role of local elasticity in the training dynamics of neural networks. We corroborate our theoretical analysis with experiments on a synthesized dataset of geometric shapes and CIFAR-10.

\end{abstract}


\section{Introduction} \label{sec:introduction}
   
    Deep learning models have achieved significant
    empirical success over the past decade across a wide spectrum of domains spanning computer vision, natural language processing, and reinforcement learning~\cite{lecun2015deep,silver2016mastering,young2018recent}. Despite these remarkable achievements at the empirical level, there is still much to learn about deep neural networks, as
    evidenced by the fact that almost all important advances concerning architecture design and optimization for deep learning are based on heuristics, without much input from a theoretical perspective~\cite{he2016deep, devlin2018bert,ioffe2015batch,kingma2014adam}.
    
    An important step toward opening these black-box models and unveiling their formidable details is to quantitatively understand the impact of backpropagation in deep learning training. While there has been a continued effort to demystify how simple optimization methods give rise to impressive generalization performance, for example, \cite{zhang2016understanding, keskar2016large, arora2019fine}, this is by no means an easy problem, perhaps because of the daunting nonconvex nature of neural networks. Accordingly, for near-term purposes, a more practical approach is to take a phenomenological viewpoint by relating simple empirical patterns to the effectiveness of deep learning models.

In this spirit, 
we are interested in how data from different classes gradually become separable in their feature space by repetitively calling backpropagation. From a phenomenological viewpoint, this question can be addressed by first analyzing the impact of a single update using a stochastic gradient on the performance of the neural networks. More precisely, imagine that the gradient is evaluated in an image of a cat, how does the hidden representation of another image---say, an image of another cat or an image of a plane---evolves because of the backpropagation? Recent studies answer this question by introducing a phenomenon called \emph{local elasticity}, which, roughly speaking, means that the impact is generally larger on a similar sample (an image of another cat) than on a dissimilar sample (an image of a plane) \cite{he2019le}.

Motivated by the phenomenon of local elasticity, we propose a model that captures the interaction between different training samples during deep learning training using a set of stochastic differential equations (SDEs) that reflect local elasticity in neural networks. Characterizing the \emph{intra-class} and \emph{inter-class} effects is an essential component of our modeling strategy, each of which contains a drift term that imitates the impact of backpropagation on specific training data 
of all samples.

\begin{figure*}[t]
    \centering
    \begin{subfigure}[t]{0.25\textwidth}
	    \centering
	    \includegraphics[width=\linewidth]{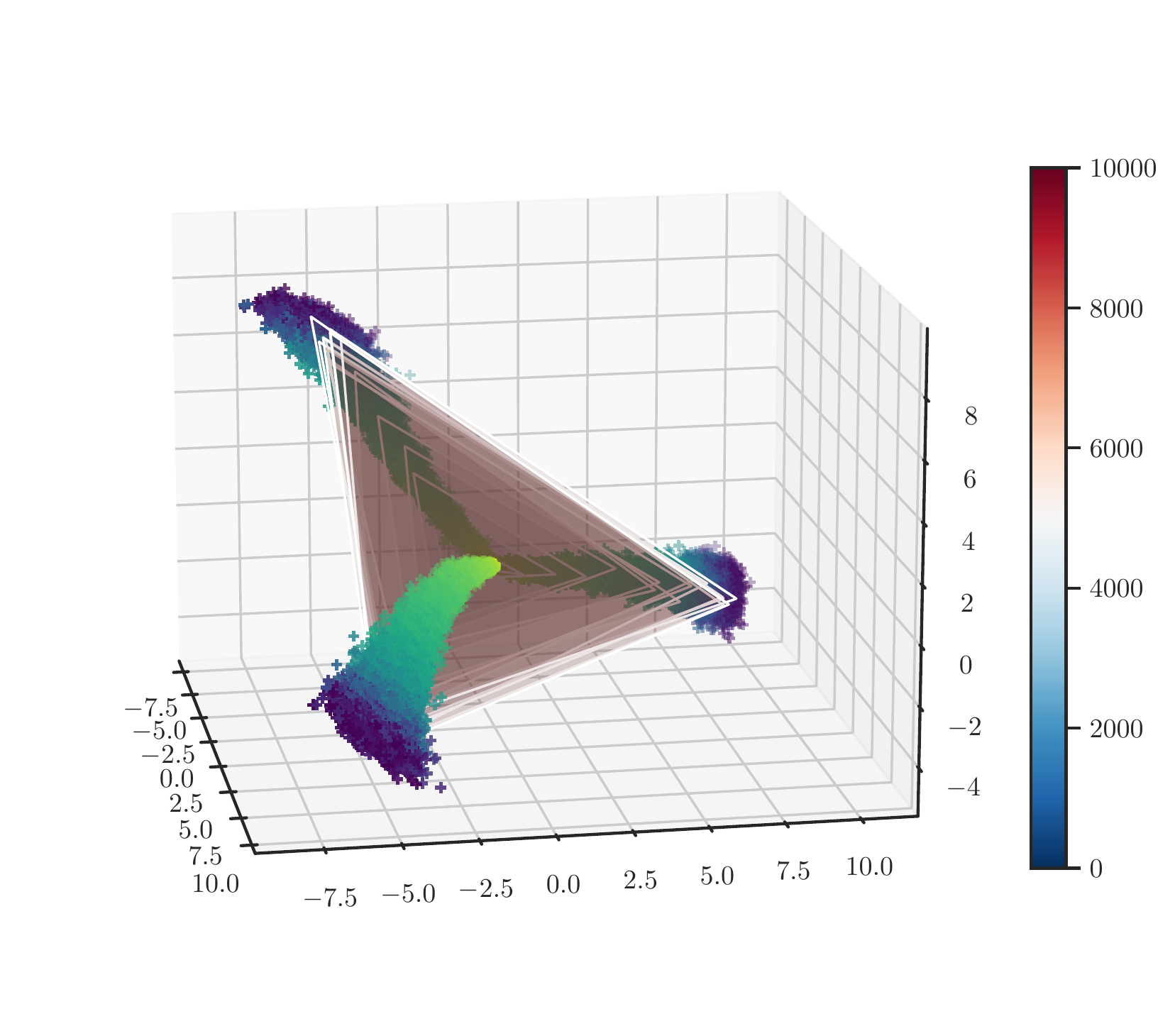}
	    \caption{\geomnist{} in $\sR^3$.} 
	\end{subfigure}%
	\begin{subfigure}[t]{0.25\textwidth}
	    \centering
	    \includegraphics[width=\linewidth]{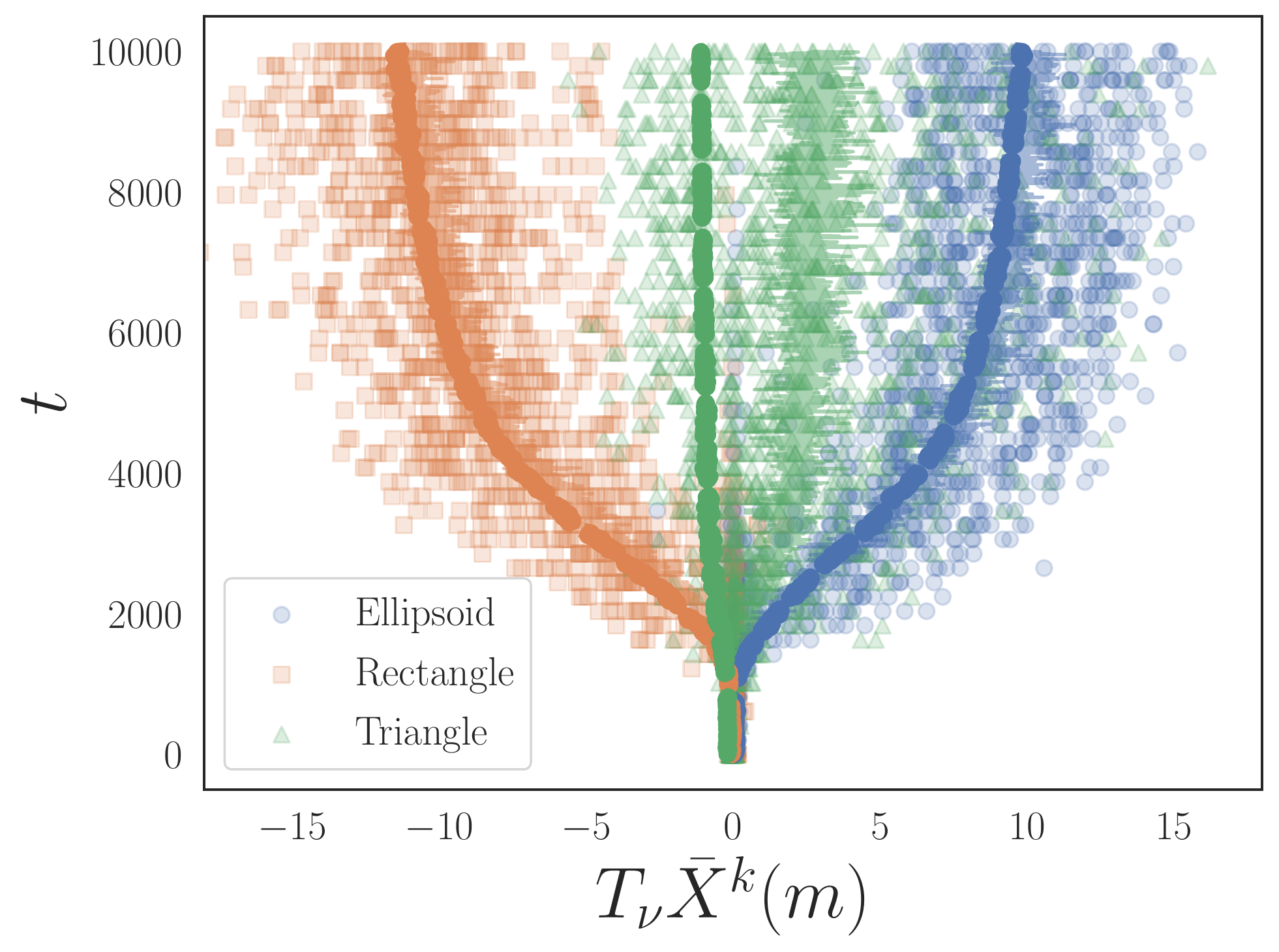}
	    \caption{\geomnist{} in $\sR$.} 
	\end{subfigure}%
	\begin{subfigure}[t]{0.25\textwidth}
	    \centering
	    \includegraphics[width=\linewidth]{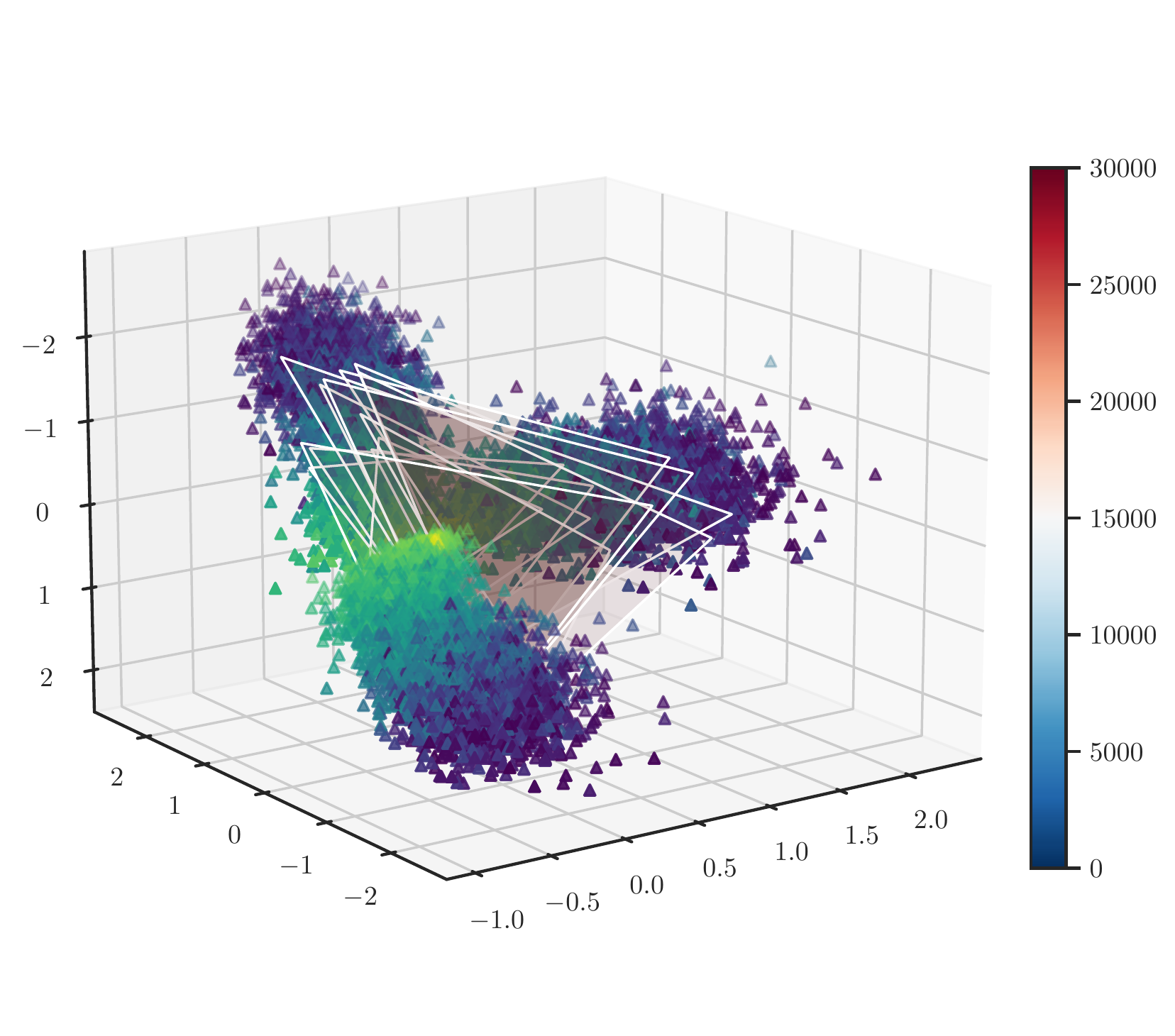}
	    \caption{\cifar{} in $\sR^3$.} 
	\end{subfigure}%
	\begin{subfigure}[t]{0.25\textwidth}
	    \centering
	    \includegraphics[width=\linewidth]{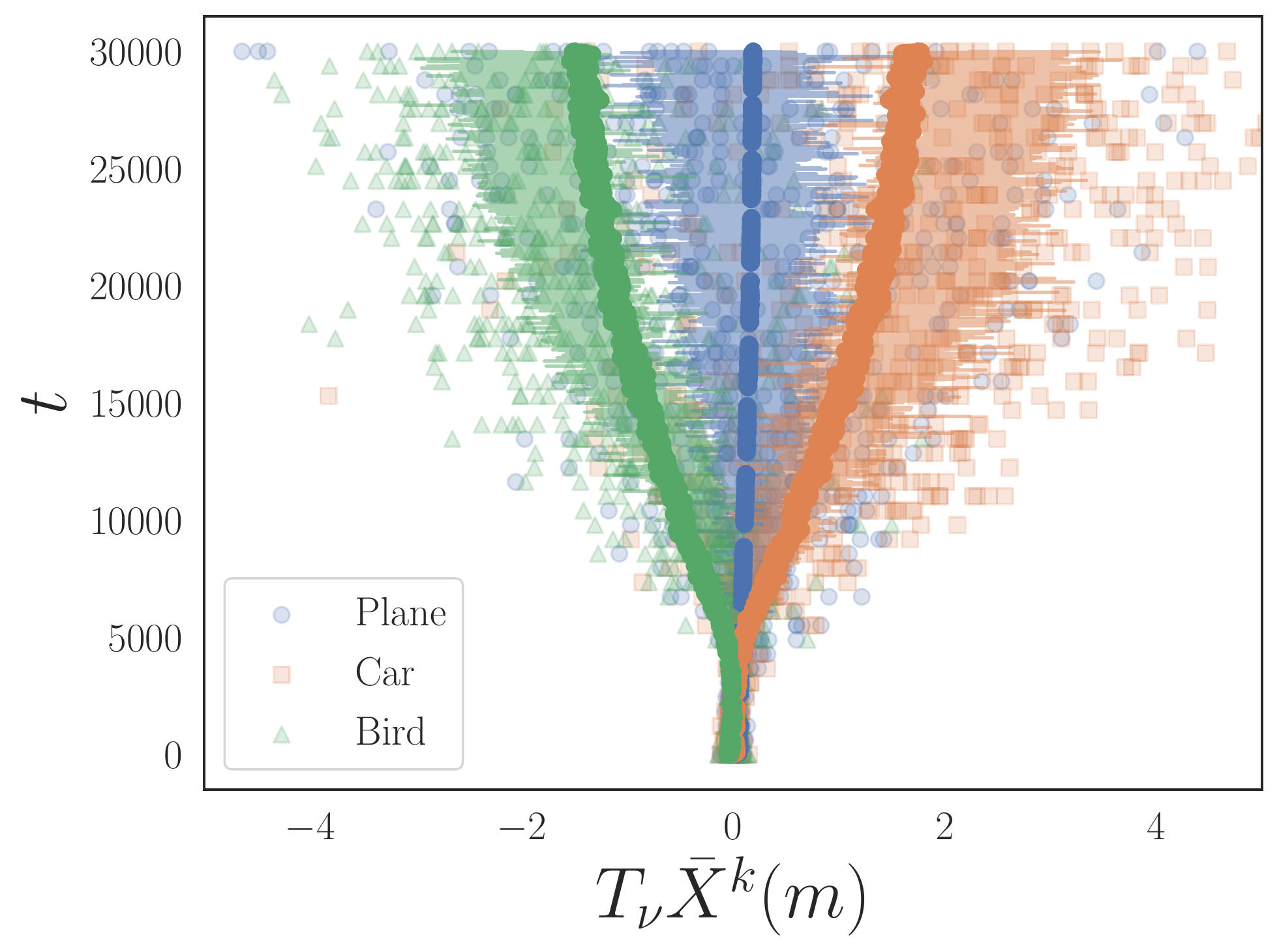}
	    \caption{\cifar{} in $\sR$.} 
	\end{subfigure}%
        \caption{\textbf{Separation of features (logits).}
        \geomnist{} dataset ((a)---(b))) and \cifar{} dataset
        ((c)---(d)) trained using $K=3$ classes
        on a variant of the \alexnet{} model.
        Separation in $\sR$ is done by projecting to $\bnu\in\sR^3$,
        which is set to be the difference between a pair of 
        class means $\bar{\X}^k(T)-\bar{\X}^l(T)$ for a large
        $T$, where $k,l$ are chosen heuristically. 
        The dashed lines in (b) and (d) are simulated paths from \Cref{eq:leode:sol:H}
        using estimated $\hat{\alpha}(t)$ and $\hat{\beta}(t)$.
        More details are given in \Cref{sec:experiments} and \Cref{sec:app_sim:results}.
        }
        \label{fig:teaser:sep}
        \vspace{-0.1cm}
\end{figure*}

   Our main finding uncovers a sharp phase transition phenomenon regarding the {intra-class} and {inter-class} impact; if the SDEs are \textit{locally elastic} in the sense that the impact is more significant on samples from the same class as the input (the intra-class effect is strictly greater than the inter-class effect), the features of the training data are guaranteed to be linearly separable, meaning vanishing training loss; otherwise, the features are not separable, no matter how long the training time is. This result provides convincing theoretical evidence for the presence of \textit{local elasticity} in deep learning \cite{he2019le}. Our model is also quite accurate in simulating the feature dynamics of deep learning. As shown in \Cref{fig:teaser:sep}, the dynamics of the predicted logits are quite close to the real dynamics of deep learning on both synthetic and real datasets, indicating a well-suited model for theoretical and practical purposes. Moreover, in the presence of local elasticity, our SDEs also predict the emergence of a simple geometric structure called neural collapse of features \cite{papyan2020collapse}. 
   
   Taken together, our results shed light on the decisive role of local elasticity in the training dynamics of neural networks. We corroborate our theoretical analysis with experiments on a synthetic dataset of geometric shapes, as well as on CIFAR-10. The experimental evidence consistently supported our model,
   which provides new insights into the dynamics of deep learning training.

\subsection{Related Work}\label{sec:related}


\paragraph{Dynamics in Deep Neural Nets.}
    Many properties of linear deep neural nets are relatively well understood,
    such as the loss landscapes \cite{kawaguchi2016deep},
    trajectory-based convergence \cite{arora2018convergence, du2019width}, 
    and
    implicit acceleration \cite{arora2018optimization}. Exact
    solutions of the training dynamics can be obtained in certain
    initialization schemes  \cite{saxe2013exact, saxe2019mathematical, lampinen2018analytic}.
    In the presence of non-linearity, various assumptions are generally made.
    \cite{saad1996dynamics, goldt2019dynamics} studied the dynamics
    of shallow neural nets with non-linearity and the
    neural tangent kernel
    (NTK) literature
    \cite{jacot2018neural,arora2019fine,du2019graph}
    linearizes the network function of an infinitely wide neural net
    at initialization, which is similar to that of the deep Gaussian process
    literature \cite{damianou2013deep,hazan2015steps,lee2017deep}
    (a treatise comparing and contrasting them can be found in
    \cite{yang2019scaling}).
    Although as approximations, NTKs are generally used 
    when studying optimization trajectories of neural nets
    such as \cite{li2018learning}, which
    also appears implicitly in many works when studying the optimization trajectories of neural net training
    \cite{soudry2018implicit,allen2019convergence,du2019gradient,ji2019polylogarithmic,chen2020generalized}.

\paragraph{SGD as SDEs in Neural Nets.}
    The study of dynamics or trajectories of weights in deep neural nets
    via SDEs  relies on the more precise
    characterization of stochasticity
    \cite{li2017stochastic,mandt2017stochastic}. 
    Built on top of this formalism, \cite{smith2017don} studied
    the trade-off between batch size and learning rate, \cite{jastrzkebski2017three},
    analyzed factors influencing the quality of local minima,
    \cite{chaudhari2018stochastic} studied the behavior of the SGD
    near local minima, and
    \cite{shi2020learning} studied the effect of learning rates.
    Although SGD-SDE approximation requires an infinitesimal learning rate,
    \cite{li2021validity} verified that the SDE approximation can be
    meaningful in practical settings and obtained
    necessary conditions for the validation of such approximation.
    
\paragraph{Local Elasticity and Phenomenological Models.}
    Local elasticity is proposed in \cite{he2019le} as a 
    phenomenological approach to reasoning the behaviors of neural networks. 
    This phenomenon has inspired several works on generalization bounds \cite{deng2020toward} and an improvement on the NTK \cite{chen2020labelaware}.

\section{Binary Separation via LE-SDE}\label{sec:binary_and_notation}
\subsection{Setup, Notations and Assumptions}\label{sec:notation}
Throughout the paper, we work with the following setup and assumptions. 
For ease of reading, vectors and matrices are written in boldface
and we denote by $[n]$ the set $\{1,\ldots,n\}$.
When there is no ambiguity, we will write both $X(t)$ and $X_t$ for a
continuous-time (possibly stochastic) process.

\textbf{Classification Problem.} Consider a $K$-class classification problem with $K\ge 2$, with each class having $n$ training examples. We denote by $\z^k_i\in \sR^d$ the $i$-th sample of the $k$-th class, and $y^k_i\in [K]$ its label, where  $i\in[n], k\in [K]$. A neural net is a function $f:\sR^d\to \sR^K$ that maps the samples to logits ( pre-activation of the softmax).

\textbf{Feature Vectors.} We denote by $\X^k_i(m)\in \sR^p$ a $p$-dimensional feature of the $i$-th sample in the $k$-th class learned by the neural net at iteration $m$. For example, 
it can be the logits or the output of the second-to-last layer. 
Assume that the initial values $\X^{k}_i(0)$ are i.i.d.~samples from some distribution for each $i\in[n], k\in [K]$. 
We use $i, j\in[n]$ as indices for an individual sample, $k, l\in[K]$ for classes, and capital letters $J$ and $L$ to indicate random samples from $\UNIFORM([n])$ and $\UNIFORM([K])$, respectively.

\textbf{Training Dynamics.} We model the training dynamics in neural nets under SGD with an emphasis on \textit{local elasticity}. At the $m$-th iteration, the $J_m$-th sample is sampled from the $L_m$-th class, where $J_{m}\sim\UNIFORM([n])$ and $L_m\sim \UNIFORM([K])$. 
Training on $\z_{I_m}^{J_m}$ affects the features of another data sample $\z_i^k$ 
in the form of
    \ba\label{eq:LE-SDE_raw_def}
    \X^k_i(m)-\X^k_i(m-1) = h\cdot E_{k, L_m}(m) \X^{L_m}_{J_m}(m-1) + \sqrt{h} \bzeta_{i}^k(m-1)
    \ea
    where $i\in[n], k \in [K]$, $h$ is the step size, and $\bzeta_{i}^k(m)$ is the noise term that is modeled as Gaussian noise. The scalar $E_{k, L_m}$ measures the strength of 
local elasticity that $\z_{J_m}^{L_m}$ exerts on $\z_i^k$ at iteration $m$.
We assume $\X^{k}_i(0)$, $\bzeta_{i}^k(m)$ are jointly independent.

\paragraph{Local Elasticity.} 
Clearly, by writing $E_{k, l}(m)$, we assume that this effect depends only on the class $l$, $k$, and time $m$.  We write the effect matrix as $\bE(m) = \left(E_{k, l}(m)\right)_{k,l=1}^K$. 
For ease of exposition, we assume $\E$ only
consists of two values $\alpha(m)$ and $\beta(m)$,
with $\alpha(m)$ representing the \emph{intra-class}
effect and $\beta(m)$ the \emph{inter-class} effect. 
To this end, we assume the \emph{effective training assumption}, that is,
as training progresses, the features become more discriminative: features from the same class are more similar, whereas those from different classes are more distinct, as measured by some similarity measure in the feature space.
We also assume that the LE effect is ``\textit{proportional}'' to the feature $\X^{L_m}_{J_m}(m)$ itself. We generalize this point in \Cref{sec:K-dim} by introducing a transformation matrix $\H$ on the features. 

\subsection{Binary LE-SDE}\label{sec:binary}
Our construction of \cref{eq:LE-SDE_raw_def}
emphasizes the effect of intra- and inter-class effects on the dynamics of \textit{features}, and thus differs from
the usual weight dynamics
that is common in the literature. 
Before deriving the general form of our locally elastic SDE (LE-SDE),
we shall familiarize the reader with our model by
demonstrating this in the case of binary classification ($K = 2$) with a one-dimensional
features ($p=1$) --- the output of the model to be fed into the softmax function,
also called the \emph{logit}. 

Let the intra-class effect be
$E_{11} = E_{22} = \alpha$, and the
inter-class effect is $E_{12} = E_{21} = \beta$, both of which are time-independent.
Expanding \cref{eq:LE-SDE_raw_def}, for $1 \leq i, j \leq n$ and $m\ge 0$,
when we train the model on the $J_m$-th training example from the $L_m$-th class,
we have 
\begin{align*}
    \begin{cases}
   X^{1}_i(m)& =~~~X^{1}_i(m-1)+h\cdot \alpha           
        X^{L_m}_{J_{m}}(m-1)+\sqrt{h}\cdot \zeta^{L_m}_{i}(m-1), \\
    X^2_{j}(m)&=~~~X^2_{j}(m-1)+h\cdot \beta 
        X^{L_m}_{J_{m}}(m-1)+\sqrt{h}\cdot \zeta^{L_m}_j({m-1}).
    \end{cases}
\end{align*}
In the limit of $h\to 0$, we can show that $X^k_{i}(m)$ approximates some
continuous-time stochastic processes $X^k_{i}(t)$ 
(under the identification of $t=mh$) governed by the
set of stochastic differential equations as follows:
\begin{align}\label{eq:simple_binary_SDE}
    \dd X^k_{i}(t)=
        \left(\frac{\alpha}{2} \bar{X}^k(t)+\frac{\beta}{2} \bar{X}^{3-k} (t)\right) \dd t+\sigma \dd W^k_{i}(t), \quad t \ge  0,k\in[K], i\in[n]
\end{align}
where $\bar{X}^k(t)\coloneqq  \left(X^{k}_1(t)+\cdots+X^k_{n}(t)\right) / n$, and $W^{k}_i$ are independent standard Wiener  processes. The
detailed derivation is given in \Cref{sec:app_binary_continuous}.

Now averaging over $i$ for each $k$ in \cref{eq:simple_binary_SDE},
we obtain the following set of 
two ordinary differential equations (ODEs)
governing the per-class means that $\bar{X}^k(t)$ for $k = 1, 2$:
\begin{align*}
    \dd \bar{X}^k(t)=\left(\frac{\alpha}{2} \bar{X}^k(t)+\frac{\beta}{2} \bar{X}^{3-k}(t)\right) \dd t+\sigma \dd \frac{W^{k}_i(t)+\cdots+W^{k}_i(t)}{n}.
\end{align*}
Taking the limit of $n \rightarrow \infty$, we observe that $\sigma \mathrm{d} \frac{W^{1}(t)+\cdots+W^{n}(t)}{n} \Rightarrow 0$. 
Thus, the above display converges weakly to the following ODE:
\ba
    \frac{\dd \bar{X}^k(t)}{\mathrm{d} t}=\frac{\alpha}{2} \bar{X}^k(t)+\frac{\beta}{2} \bar{X}^{3-k}(t).
\ea
With the initial conditions $\mathbb{E} X^{k}_i(0)=c_{k}$, the
solution to the above ODE is
\begin{align*}
    \bar{X}^1(t)=\frac{c_{1}-c_{2}}{2} \mathrm{e}^{\frac{\alpha-\beta}{2} t}+\frac{c_{1}+c_{2}}{2} \mathrm{e}^{\frac{\alpha+\beta}{2} t},\quad 
    \bar{X}^2(t)=-\frac{c_{1}-c_{2}}{2} \mathrm{e}^{\frac{\alpha-\beta}{2} t}+\frac{c_{1}+c_{2}}{2} \mathrm{e}^{\frac{\alpha+\beta}{2} t}.
\end{align*}
In the finite-sample setting, 
we may replace $\bar{X}^1, \bar{X}^2$ in the SDE \eqref{eq:simple_binary_SDE} by 
their deterministic solutions and obtain
\begin{align*}
\begin{cases}
    X^1_{i}(t)&= ~~\frac{c_{1}-c_{2}}{2} \eu^{\frac{\alpha-\beta}{2} t}+\frac{c_{1}+c_{2}}{2} \eu^{\frac{\alpha+\beta}{2} t}
    -c_{1}+X^1_{i}(0)+\sigma W^1_{i}(t),\\
    X^2_{j}(t)&=~~-\frac{c_{1}-c_{2}}{2} \mathrm{e}^{\frac{\alpha-\beta}{2} t}+\frac{c_{1}+c_{2}}{2} \eu^{\frac{\alpha+\beta}{2} t}-c_{2}+X^2_{j}(0)+\sigma W^2_{j}(t).
\end{cases}
\end{align*}
We are now ready to
derive the condition under which
these $2 n$ feature vectors
become \textit{asymptotically separable}, that is, 
$\min _{i} X^1_{i}(t)>\max _{j} X^{2}_j(t)$ or $\max_{i} X^1_{i}(t)<\min_{j} X^2_{j}(t)$ as $t\to\infty$.
\begin{restatable}[Separation in Binary Classification]{theorem}{thmbinaryseparation}
\label{thm:binary_separation}
Given the feature vectors $X^1_i(t)$, $X^2_j(t)$ for $i, j\in[n]$, as $t\to\infty$ and large $n$,
\begin{enumerate}
    \item if $\alpha >\beta$, they are asymptotically separable with probability tending to one,
    \item if $\alpha \le \beta$, they are asymptotically separable with probability tending to zero.
\end{enumerate}
\end{restatable}
This result indicates a sharp phase transition 
when $\alpha$ is \emph{just} above $\beta$, 
that is, in the regime of \textit{local elasticity}.
As long as the intra-class effect is slightly greater than the inter-class effect,  separation is guaranteed.
This simple model already captures local elasticity and reveals
the important role it plays in the \textit{perfect separation} of training samples. 
We can generalize this model to more realistic settings: 
when there are multiple classes, when features are high-dimensional, and when the LE matrix $\E$ is time-dependent. 
In the next section, we discuss each of these three
generalizations in more depth.
\section{General LE-SDE Model} \label{sec:K-dim} \label{sec:lesde:general}

Now, we consider the general case where $K\ge 2$ and the feature vectors are $p$-dimensional with $p\ge K$. Inquisitive readers may have already noticed that \Cref{thm:binary_separation} only asserts the \emph{emergence} of the separation of features, while being inconclusive to their relative orders at separation,
    that is, \emph{which class converges to where}?
    This drawback is intrinsic to the toy model as
    neither intra-class nor inter-class effect identifies different classes. In this section, we introduce the general LE-SDE model
    that alleviates this difficulty with the help of an extra block
    matrix $\H$ with the $(i,j)$-th block $\H_{i,j}$ models
    how features in the $j$-th class affect those in the $i$-th class,
    which also partially defines how classes are separated in higher
    dimensions. 
    In the local elasticity formalism, $\H_{i,j}$ can be viewed
    as inducing a metric on the feature space  under which
    local elasticity manifests.

    As hinted before, in the case of multiple-class features in higher
    dimensions, we want to guarantee
    a stronger separation: to know which class converges to where,
    thus incorporating supervision from label information.
    For example, when the features are logits (outputs of the neural nets)
    and the model is trained under the softmax cross-entropy loss,
    previous work suggests they separate according to specific
    geometric structures \cite{papyan2020collapse}.
    To this end, 
    we need to adjust the raw feature vectors $\X^{k}_{i}$ with a
    proper transformation that incorporates the label information into 
    the dynamics. 
    This motivates the following modification of the dynamics \eqref{eq:LE-SDE_raw_def} by adding an extra transformation $\H_{k,L_m}\in \sR^{p\times p}$ to the features. For $k\in [K]$, $i\in[n]$, and at iteration $m$, we have the following: 
    \ba \label{eq:lesde:single}
        \X^k_i(m) = \X^k_i(m-1) + h\cdot E_{k,L_m}(m) \H_{k,L_m}(m) \X^{L_m}_{J_m}(m-1)
        +\sqrt{h} \bzeta^k_i(m-1).
    \ea
    The $\H_{k, L_m}(m)$ term models the LE effect as \textit{proportional} to a linear ``transformation'' of the features. The dynamics in \Cref{eq:LE-SDE_raw_def} are special cases when
    $\H_{k, l}(m)\equiv \I_p$ for all $k,l\in[K]$. 
    By specifying a proper $\H$, 
    we can overcome the limitation in our toy example of not knowing
    which class converges to where.
    We specify interesting choices of $\H$ in \Cref{sec:twomodels}.
        
    A further step of abstraction is to 
    write $\tilde{\X}^k(m)$ instead of $\X_i^k(m)$,
    to indicate one generic sample from the distribution
    $\mathcal{D}^{k}(m)$ 
    of all the features of class $k$ at iteration $m$.
    As in \Cref{sec:binary}, we can derive the continuous dynamics of \cref{eq:lesde:single} in the limit of $h\to 0$ in the same way as \cref{eq:simple_binary_SDE}. 
    Similar to writing $\tilde{\X}=(\tilde{\X}^k)_{k=1}^K \in \sR^{Kp}$ for the concatenation of
    per-class features, $\bar{\X}=(\bar{\X}^k)_{k=1}^K \in \sR^{Kp}$  is the concatenation of per-class mean features. Our model \eqref{eq:lesde:single} approximates the following SDE with identification $t = mh$ as $h\to 0$. We term this model LE-SDE:
    \ba \label{eq:LE-SDE_general_continuous}
        \dd\tilde{\X}_t =  \M_t \bar{\X}_t \dd t
        + \bSigma^{\frac{1}{2}}_t \dd\W_t, 
    \ea
    where $\W_t$ is the standard Wiener process in $\sR^{Kp}$, $\bSigma_t$ is the covariance matrix,
    and $\M_t\in \sR^{Kp\times Kp}$ is a $K\times K$ block matrix, with each block of size $p\times p$. The $(k, l)$th block 
    of $\M_t$ is ${E_{k, l}(t)} \H_{k, l}(t)/K$ when $l\ne k$, and ${E_{l, l}}(t) \H_{l, l}(t)/K$ when $l = k$.
    The rationale for dividing $K$ is that we assume that the data are balanced; therefore, each of the $K$ possible classes has an equal chance of being sampled,
    as proved in \Cref{eq:simple_binary_SDE}, where $K = 2$. In \Cref{sec:app_ext},
    we discuss how we can
    generalize this to model SGD with mini-batches, imbalanced data,
    and label corruptions.

    Taking expectation with respect to the randomness arising
    from sampling $\tilde{\X}_t$ from its distribution,
    the per-class mean $\bar{\X}_t$ satisfies the following system, which we term the LE-ODE:
    \ba \label{eq:leode}
        \bar{\X}'_t = \M_t \bar{\X}_t.
    \ea
    Under the assumptions in \Cref{sec:notation}, we define $
    \gamma(t) = \min\left\{ \alpha(t) - \beta(t), \alpha(t)
        +(K-1)\beta(t)\right\},$
    \ba
        A(t) = \int_0^t \alpha(\tau) \dd \tau, \quad
        B(t)  = \int_0^t \beta(\tau) \dd \tau,\quad
        \Gamma(t) =\min\left\{ A(t) - B(t), A(t)
        +(K-1)B(t)\right\}.
    \ea

\subsection{The Separation Theorem}
\label{sec:separation_thm}
    Similar to the discussions in \Cref{thm:binary_separation}, 
    the LE-SDE allows us to derive the separability result
    for a general $K$ and $p\ge K$.
    We say the feature vectors
    $\left\{(\X^k_i)_{i\in[n]}\right\}_{k\in[K]}$ are \textit{separable}
    if for any two classes $k\ne l$, there exists a hyperplane in $\sR^p$ that linearly separates the features of the two classes. 
    To characterize the separation as in \Cref{thm:binary_separation}, 
    we need conditions on $\alpha(t), \beta(t)$ as therein.
    Intuitively, when $\gamma(t) \le 0$, 
    the classes cannot be separated, even in a pairwise manner. Therefore, we focus on a more interesting case when $\gamma(t) > 0$.
    We now state the following characterization theorem of separability for general LE-SDE dynamics:
    \begin{restatable}[Separation of LE-SDE]{theorem}{thmseparationmean}
    \label{thm:separation:mean}
        Under our working assumptions in \Cref{sec:notation}, and in the case of local elasticity (i.e., $\gamma(t)>0),$
        assume $\H = \left(\H_{ij}\right)_{ij}$ is positive semi-definite (PSD)
        with positive diagonal entries. As $t\to\infty$, we have\footnote{Here, $\gamma(t) = \omega\left(1/t\right)$ stands for $\gamma(t) \gg 1/t$ as $t\to\infty$. For example,
        $1 / t^{0.5}= \omega\left(1/t\right)$ and
        $(t\ln t)^{-1} = o\left(1/t\right)$ as $t\to\infty$.}:
        \begin{enumerate}[leftmargin=0.8cm]
            \item if $\gamma(t) = \omega\left(1/t\right)$,
        the features are separable with probability tending to $1$;
       \item if $\gamma(t)= o\left(1/t\right)$, and the number of per-class-feature $n$ tending to $\infty$ at an arbitrarily slow rate, the features are asymptotically pairwise separable with probability $0$.

        \end{enumerate} 
    \end{restatable}
    
    This theorem sheds light on the crucial impact of the local elasticity effect 
    for separation in a general case. The proof to \Cref{thm:separation:mean} as well as discussions on the empirically best ways of choosing the universal direction $\bnu$ (i.e., a direction that does not depend on the class index) are detailed in \Cref{sec:app:pf}. 
    
    \subsection{Two Specific Models} \label{sec:twomodels}
    We next discuss two specific choices of the $\H$ matrix that
    allows us to analyze $\tilde{\X}$ precisely. 
    \subsubsection{Isotropic Feature Learning Model}
        \label{sec:model:I}
        As a straightforward extension to \Cref{sec:binary}, 
        we can simply choose $\H_{lk} = \I_p$ to be the identity matrix. 
        This choice of $\H$ is PSD, and thus, we can apply \Cref{thm:separation:mean} to obtain the conditions for asymptotic separation. 
        In this case, the solution $\bar{\X}(t)$ to the LE-ODE
        can be computed analytically as given in the following
        proposition.
        \begin{restatable}[\mI]{proposition}{thmimodel}
        \label{thm:imodel}
        Let $\H_{k,l} = \I_p$, then the solution to the LE-ODE \eqref{eq:leode} is given by
        \ba \label{eq:leode:sol:id}
            \bar{\X}(t) =  \c \eu^{\frac{1}{K}A(t) - \frac{1}{K}B(t)} +  
            \left(\bone_K\otimes \c_0\right) \eu^{\frac{1}{K}A(t) + \frac{K-1}{K}B(t)},
        \ea
        where $\c=(\c_k)_{k=1}^K\in\sR^{Kp}$ and $\c_0\in\sR^p$ are constants with $\sum_{k=1}^K \c_k = \bzero\in\sR^p$ and $ \bar{\X}(0) = \c + \c_0$.
        \end{restatable}
        The derivation of \Cref{eq:leode:sol:id} is deferred to
        \Cref{sec:app:pf:imodel}. 
        From \cref{eq:leode:sol:id}, we can easily reconstruct \Cref{thm:separation:mean} in this special case.
        The difference between a feature vector from class $k$ 
        and that from class $l$ at time $t$ is given by $(\c_k - \c_l)\eu^{\frac{1}{K}A(t) - \frac{1}{K}B(t)} + \bSigma^{1/2}(\W^k_t - \W^l_t)$, provided that the first deterministic term dominates the random second term, thus ensuring separation,
        which are precisely 
        the conditions specified in \Cref{thm:separation:mean}.
        We term this model as the \emph{isotropic feature model},
        or \mI{} for short; as the $\H$ matrix has identity matrices
        as its blocks and consequently
        the dynamics do not
        prescribe any preferred directions for each class.
        
    \subsubsection{Logits-as-Features Model}
        \label{sec:model:H}
        An important type of features in neural nets is the
        \emph{logits}, the outputs of the neural net before
        the softmax layer.
        A logit vector (or logits) 
        is $K$-dimensional, and in this model we identify
        $\tilde{\X}^k(t)$ as the logits at time $t$ of a generic
        sample from the $k$-th class.
        In a well-trained neural net,
        the logits of a learned data instance from the $k$-th class 
        should have its $k$-th logit being the largest, and heuristically, the other coordinates should be approximately
        equal and negative. 
        As we shall detail in \Cref{sec:app:dynamics}, the exact
        dynamics of neural net training pushes the logits $\tilde{\X}^k$
        by its \emph{margin},
        $\d_k\coloneqq \e_k - \softmax (\tilde{\X}^k)$, 
        which roughly aligns with the direction of $\e_k - \bone_p/K$.
        This suggests us how to choose the metric under which
        local elasticity acts: we can choose $\H_{l, k}$ such that it always aligns $\tilde{\X}^k$ in the direction of $\d_k$,
        that is,
        \ba \label{eq:H:ce}
            \H_{ij} = \bar{\H}^j \coloneqq 
            \frac{\d_j\tp{\d}_j}{ \norm{\d_j}_2^2} \in \sR^{p \times p},
            \quad 
            \d_j \coloneqq \e_j - \frac{1}{K} \bone_p \in \sR^p, \quad j\in[K].
        \ea
        Roughly speaking, the map $\x \mapsto \bar{\H}^j\x$ projects $\x$ 
        in the direction of
        $\d_j$ and ideally aligns $\x$ with $\d_j$ after iterative applications; hence, $\bar{\H}^j$ can be viewed as an approximation of the nonlinear transformation
        in the exact dynamics in the sense that the direction
        of their stationary point coincides. Furthermore,
        $\bar{\H}^j$ thus defined has operator norm $1$; thus, it does not
        affect the magnitudes, but only directions.
        Note that $\H$ does not satisfy the condition
        in \Cref{thm:separation:mean} as it is not symmetric; 
        yet the separation theorem can be easily extended in light
        of the following proposition.
        \begin{restatable}[\mH]{proposition}{thmhmodel}
        \label{prop:leode:logit} \label{thm:hmodel}
            Let $\H$ be the same as in \cref{eq:H:ce}, then the solution to the LE-ODE \eqref{eq:leode} is given by
            \ba \label{eq:leode:sol:H}
            \bar{\X}(t) =  \c_0
                + C_1 \d \eu^{\frac{1}{K}A(t) - \frac{1}{K}B(t)}+
            \left(\sum_{l=1}^{K-1} C_{2l} \f_l\right)
                \eu^{\frac{1}{K}A(t) + \frac{1}{K(K-1)}B(t)},
            \ea
            where
            $\f_l$'s are fixed vectors in $\sR^{K^2}$,
            $\c_0\in\sR^{K^2}$ is a constant vector
            with $K(K-1)$ degrees of freedom,
            and $C_1, C_{2l}\in\sR$, for $l \in [K-1]$ are constants.
        \end{restatable}
        The specific form of $\f_l$ is not the focus here; \cref{eq:leode:sol:H} allows us to prove the statement of \Cref{thm:separation:mean} under this choice of $\H$. 
        The proof of \Cref{thm:hmodel} is 
        deferred to \Cref{sec:app:pf:hmodel},
        where we also provide the analytical solution of $\f_l$'s when $K= 3$.
        
        We term this model as the logits-as-features model, or \mH{} for short, because it is an elaborate model specifically for logits.
        In \Cref{sec:experiments}, we provide concrete demonstrations of our abstract feature vector $\tilde{X}$ as logits under \mH{}. 
        We numerically simulated our LE-ODE and compared its predicted dynamics with real deep learning training dynamics. 
        The experimental results provide strong empirical support for the validity of \mH{}.

\subsection{Connection with Neural Collapse}
    Neural collapse is a recent phenomenological finding on the geometry 
    of the logits learned by deep neural nets at convergence  with the cross-entropy loss \cite{papyan2020collapse} (see an explanation of neural collapse in \cite{fang2021layer}).
    Simply speaking, taking our \mH{} as an example, with balanced training samples, this model asserts that the logit
    vectors from different classes at convergence
    form an equiangular tight frame (ETF). ETFs are the best
    configuration to spread $K$ unit vectors in an ambient space of $p$ dimensions.
    Formally, we say a set of vectors $\{\s_i\}_{i=1}^K$ form an ETF
    in $\sR^p$ if they are the columns of a matrix
    \ba
        \S =\sqrt{\frac{K}{K-1}} \Q \left(\I_K - \frac{1}{K}\bone_K\tp{\bone}_K \right),
    \ea where $\Q\in\sR^{p\times K}$, and  $\tp{\Q}\Q = \I_K.$
    As a direct corollary of \Cref{prop:leode:logit},
    when $\H$ is set according to \cref{eq:H:ce}, we find that our \mH{} also predicts the existence of neural collapse from the local elasticity point of view.
    \begin{restatable}[Neural Collapse of the LE-ODE]{proposition}{thmcollapse}
    \label{thm:collapse}
        Under \mH{} and the same setup as in Theorem~\ref{thm:separation:mean}, if $\gamma(t)> 0$ and
        there exists some $T>0$
        such that $B(t) < 0$ for $t\ge T$, then $\left\{\bar{\X}^k(t)/\|\bar{\X}^k(t)\|\right\}_{k=1}^K$ forms an ETF as $t\to\infty.$
    \end{restatable}


\section{Experiments}\label{sec:experiments}

We perform various experiments to test our theory, where we choose \textit{logits} as our protagonist\footnote{Code for reproducing
our experiments is publicly available at \href{https://github.com/zjiayao/le_sde}{\texttt{github.com:zjiayao/le\_sde.git}}.}.

\subsection{Setup}
    
\begin{figure*}[t]
        \centering
        \includegraphics[width=0.8\linewidth]{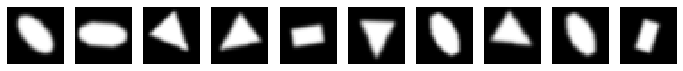}
        \caption{{Samples from \textsc{GeoMNIST} dataset}.} \label{fig:geomnist:smp}
        \vspace{-0.6cm}
    \end{figure*}

\paragraph{Datasets and Models.}
    We perform experiments on a synthesized
    dataset called \geomnist{} containing
    $K=3$ types of geometric shapes 
    (\textsc{Rectangle},
    \textsc{Ellipsoid}, and \textsc{Triangle})
    and on CIFAR-10 (\cite{CIFAR}, denoted by \cifar) with
    $K\in[2,3]$ classes. 
    A few samples from \geomnist{} are shown in \Cref{fig:geomnist:smp}.
    We vary
    the number of training samples per class and
    label pollution ratio $\perr$ and use
    variants of the \alexnet{} (\cite{krizhevsky2012imagenet}) model.
    More details can be found in
    the Appendix.
\toappendix{
    We generate a dataset consisting of $K=3$ simple
    geometric shapes (\textsc{Rectangle},
    \textsc{Ellipsoid}, and \textsc{Triangle})
    that are rotated to various
    angles and applied Gaussian blurring,
    which we conveniently name Geometric-MNIST or \geomnist{} for short.
    A few samples from \geomnist{} are shown in \Cref{fig:geomnist:smp}.
    We use a varying number of training samples per class $n_{\mathrm{tr}}\in\{80, 480, 600, 4800\}$ with the validation sample per class being $n_{\mathrm{val}}=\{20, 120, 400,1200\}$. We also pollute each label class by randomly choosing $\perr\cdot n_{\mathrm{tr}}$ samples to flip the label to another class.
    (uniform across all other classes). In this setup, we fix
    $n_{\mathrm{tr}}=n_{\mathrm{val}}=500$ and set
    $\perr\in\{0.1,0.2,\ldots, 0.8\}$.
    In addition to \geomnist{},
    we use \cifar-10 (\cite{CIFAR}, denoted
    by \cifar) for a more realistic scenario with $5000$ training
    samples and $1000$ validation samples per class.
    We vary the total number of classes $K\in [2,3]$.
    Variants of the \alexnet{} model (\cite{krizhevsky2012imagenet})
    are used, which consists
    of two convolutional layers and three fully-connected layers
    activated by the \texttt{ReLU} function.
}
    
    \paragraph{Training Configurations.}
    All models are trained for $T=10^5$ iterations (for \geomnist)
    or $T=3\times 10^5$ iterations (for \cifar)
    with a learning rate of $0.005$ and
    a batch size of $1$ under the softmax cross-entropy loss.
    Models on \geomnist{} converged with training and validation losses
    to zero, and those on \cifar{} to validation accuracies greater than $90\%$.

    \paragraph{Estimation Procedures.}
        Each experiment is repeated for $n_{\mathrm{trial}} = 100$
        independent runs to estimate $\bar{\X}(t)$.
        We use both the isotropic feature learning model
        (\Cref{sec:model:I}) and the 
        logits-as-features model (\Cref{sec:model:H}),
        denoted by \mH{} and \mI{} respectively,
        to estimate $\alpha(t)$ and $\beta(t)$.
        The \mH{} is used only when $K=3$.
        To estimate $\alpha(t)$ and $\beta(t)$, we first
        estimate $A(t)$ and $B(t)$ by 
        \ba \label{eq:estimation:AB}
        (\text{\mI}) &\quad
        \begin{cases}
        \hat{A}(t) &= \avg \avg_k{
            \log \abs{\frac{\check{\X}(\bar{\X}^k - \check{\X})^{K-1}}{\c_0 \c_k^{K-1}}}
        }, \\
        \hat{B}(t) &=
        -\avg \avg_k{
            \log \abs{\frac{\c_0}{\c_k} \frac{\bar{\X}^k - \check{\X}}{\check{\X}}}
        }, 
        \end{cases}
        \quad \check{\X}_t \coloneqq \avg_l \bar{\X}^l_t, \\
        (\text{\mH}) &\quad
        \begin{cases}
        \hat{A}(t) &= A'(t)+2B'(t), \\
        \hat{B}(t) &= 2(B'(t)-A'(t)), \\
        \end{cases} \quad
        \begin{cases}
            A'(t) &\coloneqq \log \abs{\left\langle\tp{\bar{\X}}\v_1 -1\right\rangle}, \\
            B'(t) &\coloneqq \log \abs{\left\langle
            \tp{\bar{\X}}\left(\v_2 - \frac{4}{3}\v_1\right)
            \right\rangle},
        \end{cases}
        \ea
    where vector division is interpreted entry-wise.
    We write $\avg_l$ for averaging over the class index,
    $\avg$ for averaging over the coordinates, and define $\langle \X\rangle(t) \coloneqq \X(t)/\X(0)$. We explain how and why to choose the vectors $\v_1$ and $\v_2$ in \Cref{sec:app_sim:estimation}. \rednote{The main idea is to view
    the eigenvectors of the $Kp$-by-$Kp$ drift matrix as a concatenation of $K$
    vectors of dimension $p$ and construct their linear combinations
    such that one or more independent components in the solution vanishes.}
    With $A(t)$ and $B(t)$ estimated, we use the Savitzky - Golay filter
    to obtain $\hat{\alpha}(t)$ and $\hat{\beta}(t)$ through numerical
    differentiation. For \geomnist{} and \cifar{} datasets,
    we choose window sizes of this filter as
    $191$ and $551$, respectively, in \Cref{fig:exp:ABab},
    and $21$ and $21$, respectively, in \Cref{fig:exp:sde}.
    
    We assess the tail of $\hat{\alpha}(t)$ and $\hat{\beta}(t)$
    by a \emph{tail index} defined as 
    $r_{\alpha} \coloneqq  \sup_s \left\{s:\lim_{t\to\infty} \alpha(t) \cdot t^s < \infty \right\}$
    and $r_{\beta}$ is defined similarly.
    We  estimate $\hat{r}_{\alpha}$ by fixing an interval $[T_1,T_2]$
    with $T_1<T_2$ sufficiently large such that we may ignore terms
    with smaller order and have
    $\hat{r}_{\alpha}
        = 1-\avg_{T_1 \le t \le T_2} \frac{\log \alpha(t)}{\log (1+t)}$,
    and similarly for $\hat{r}_{\beta}$. We use the estimates
    from the last $1000$
    iterations for averaging in our experiments.

\subsection{Results}
    \begin{figure*}
    \centering
    \begin{subfigure}[t]{0.25\textwidth}
	    \centering
	    \includegraphics[width=\linewidth]{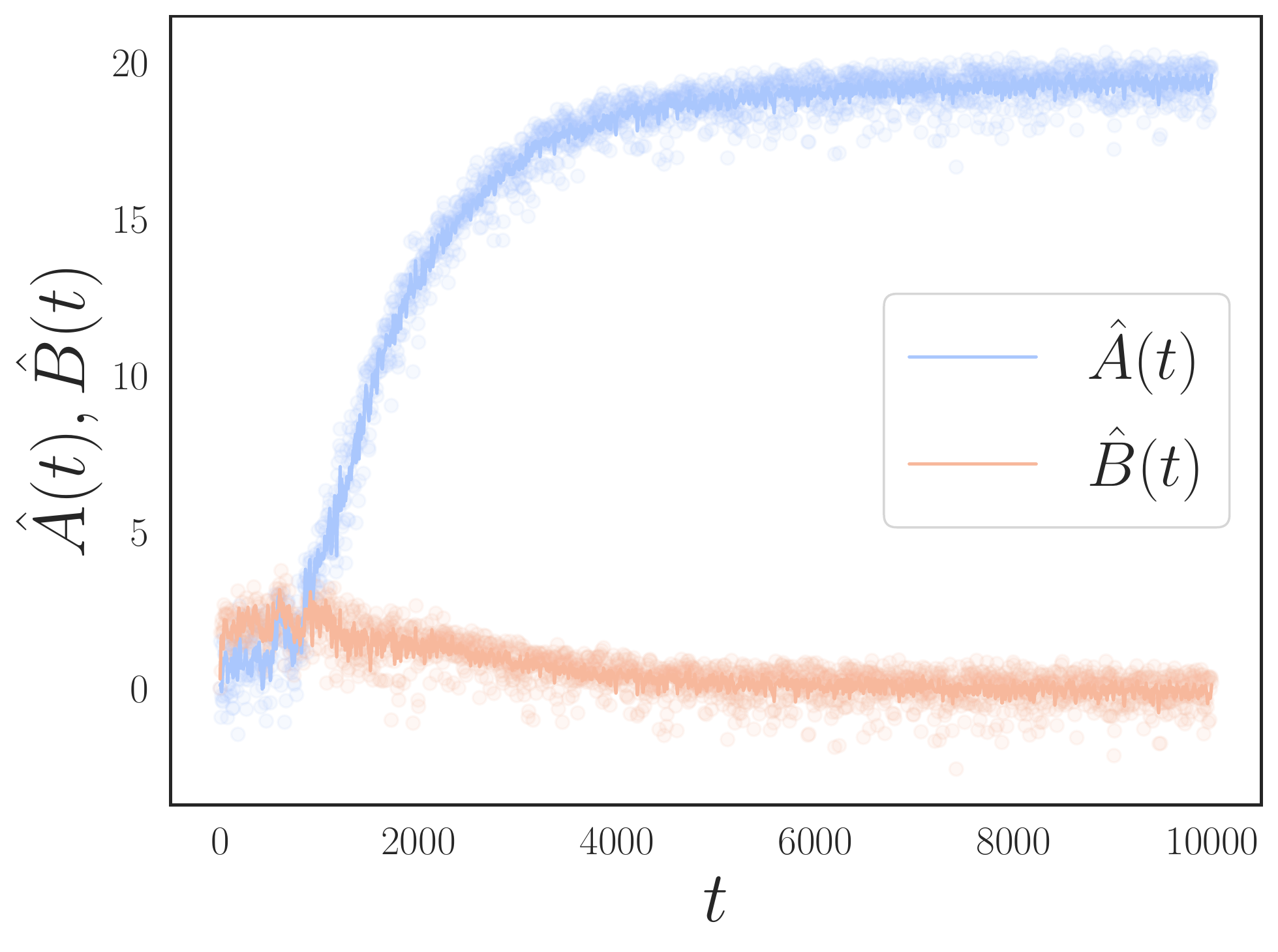}
	    \caption{\geomnist{} (\mI{}). } 
	\end{subfigure}%
	   \begin{subfigure}[t]{0.25\textwidth}
	    \centering
	    \includegraphics[width=\linewidth]{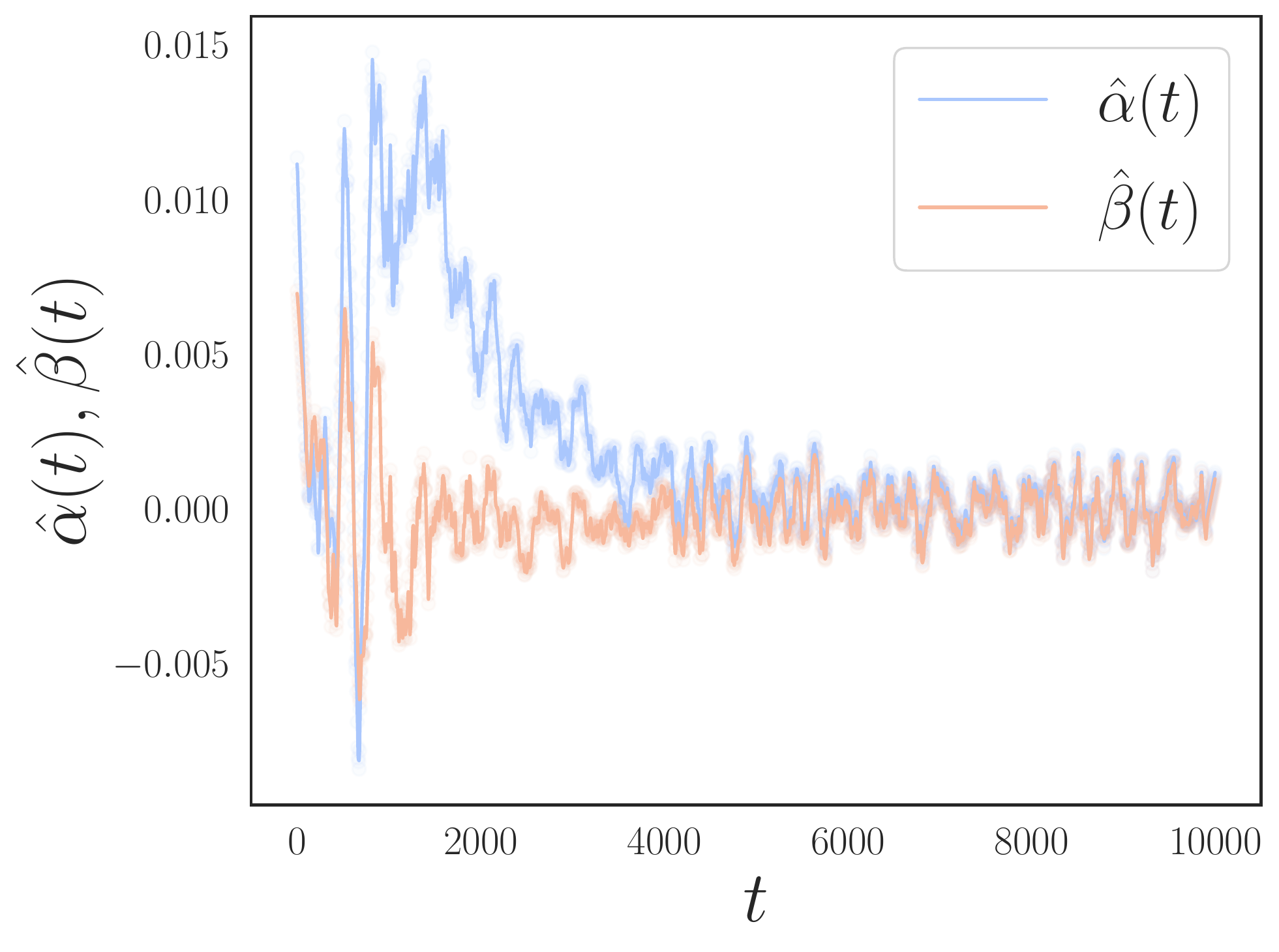}
	    \caption{\geomnist{} (\mI{}).} 
	\end{subfigure}%
    \begin{subfigure}[t]{0.25\textwidth}
	    \centering
	    \includegraphics[width=\linewidth]{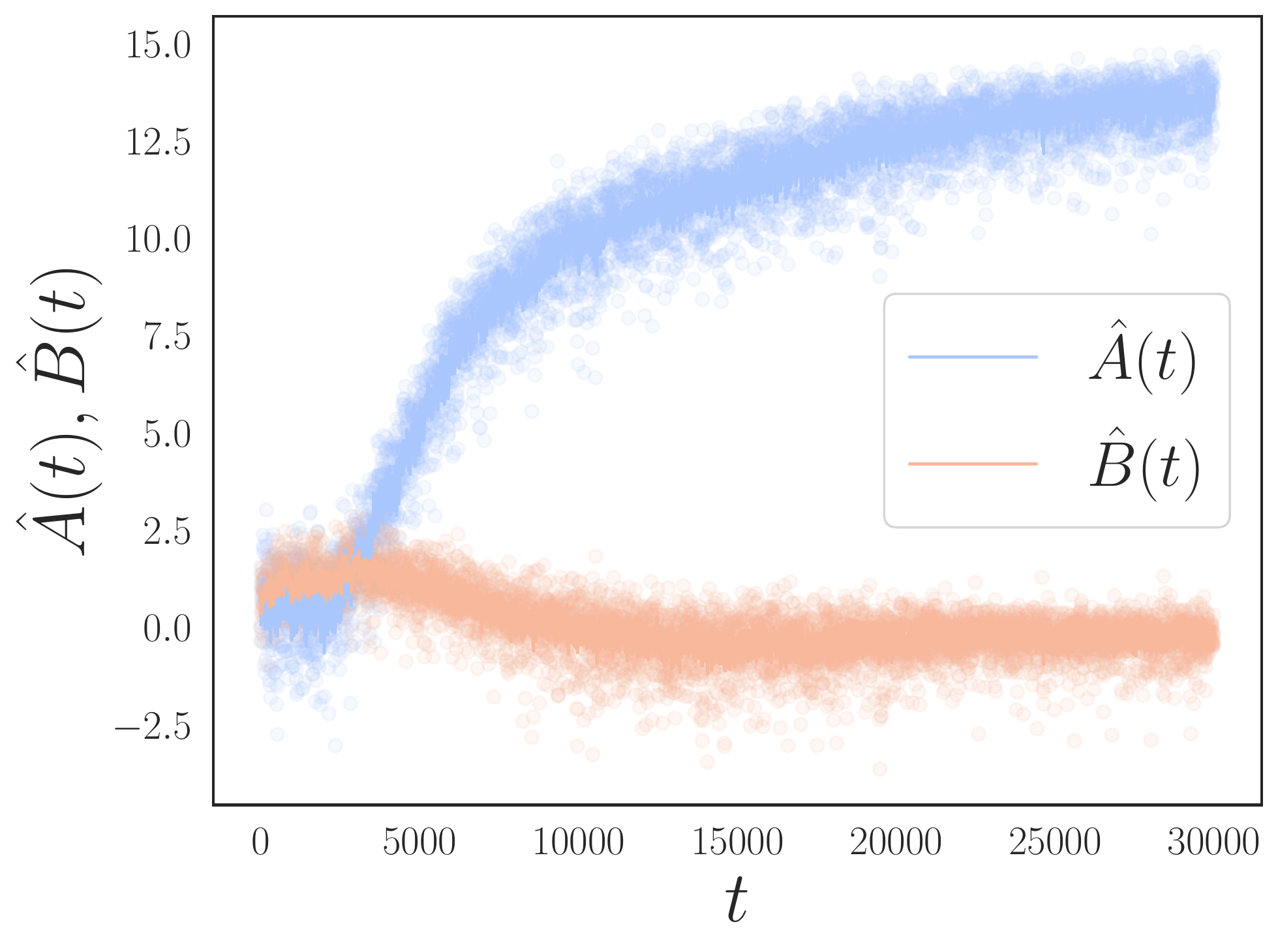}
	    \caption{\cifar{} (\mI).} 
	\end{subfigure}%
    \begin{subfigure}[t]{0.25\textwidth}
	    \centering
	    \includegraphics[width=\linewidth]{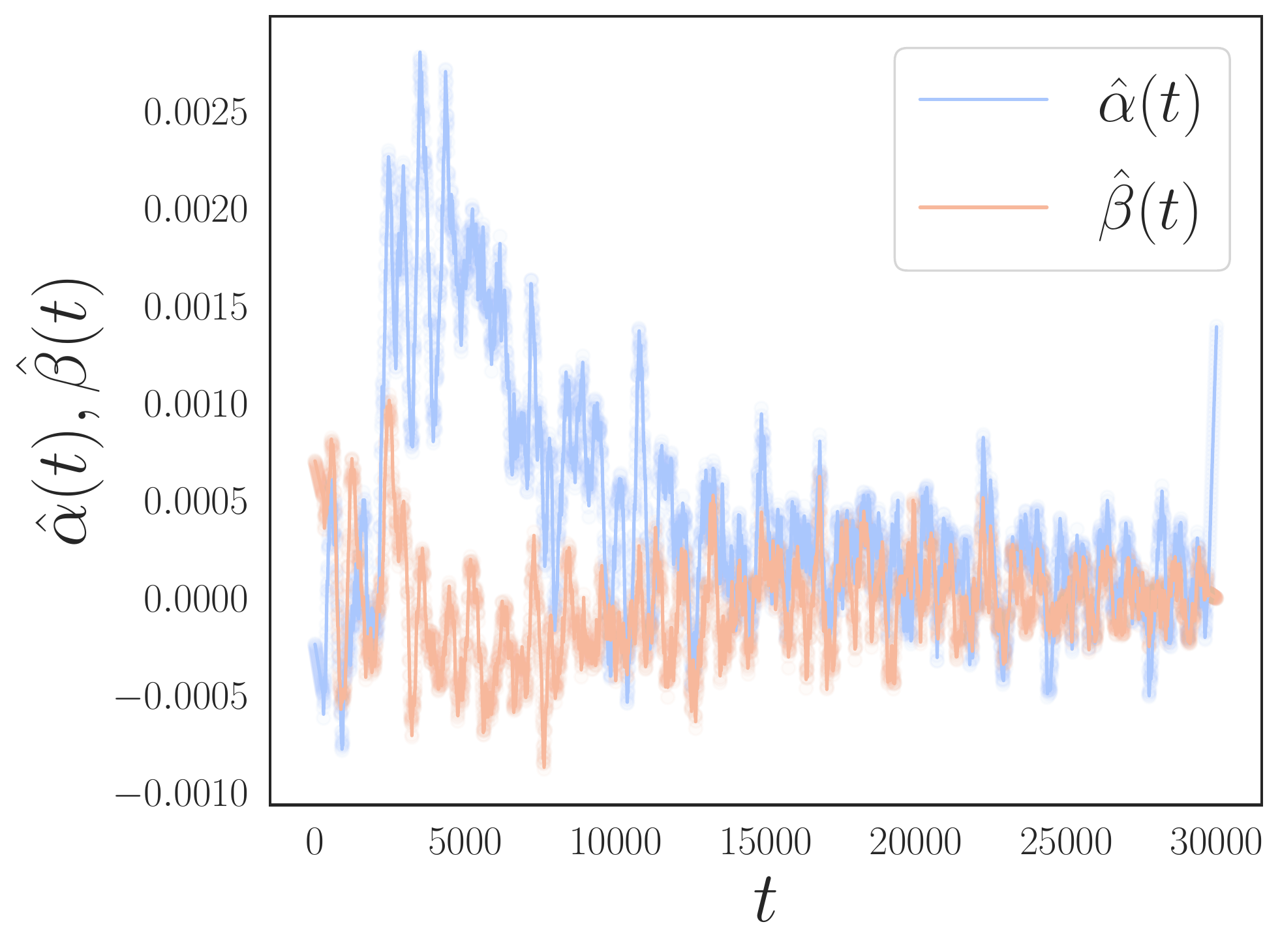}
	    \caption{\cifar{} (\mI).} 
	\end{subfigure}\\
    \begin{subfigure}[t]{0.25\textwidth}
	    \centering
	    \includegraphics[width=\linewidth]{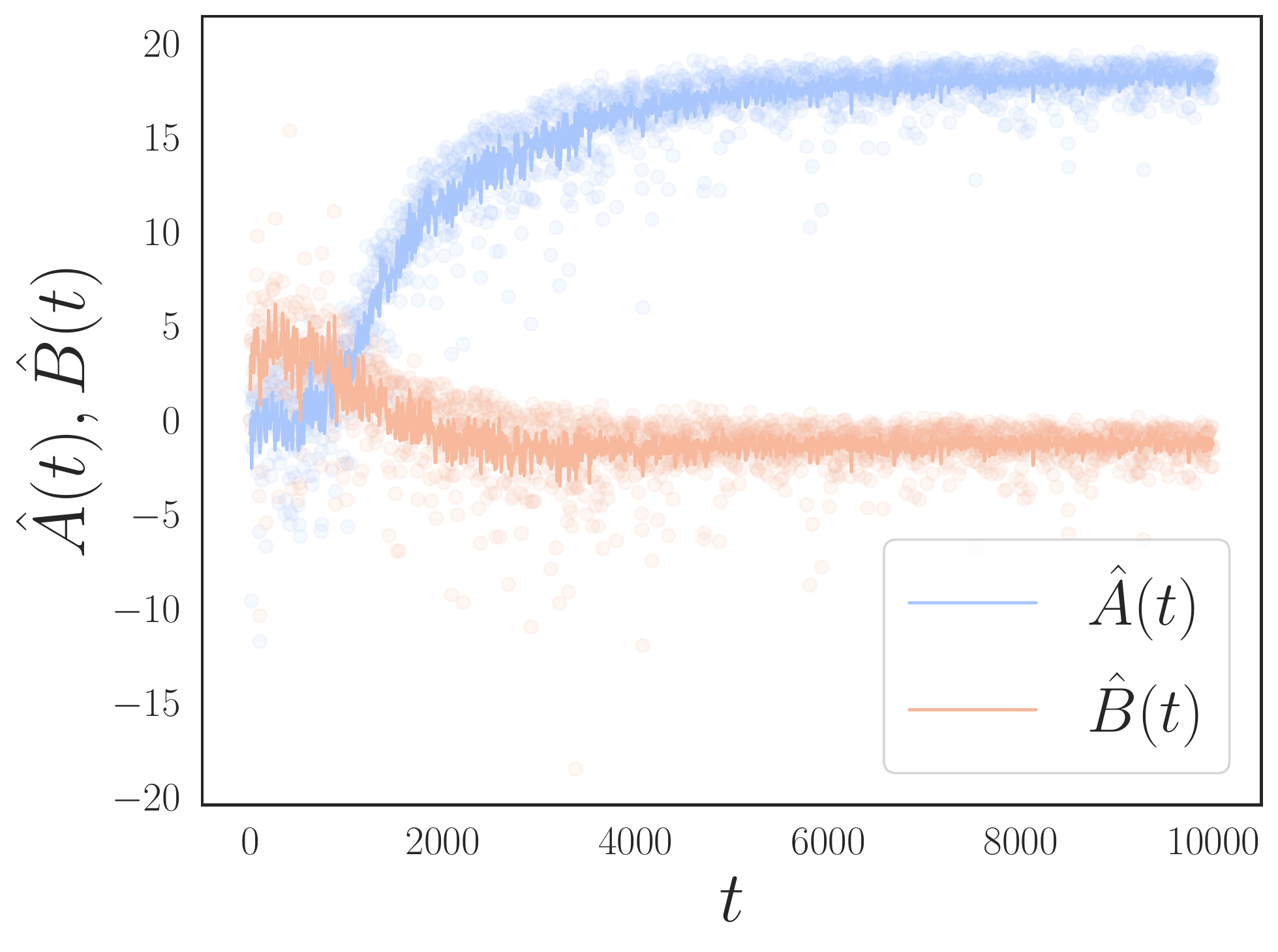}
	    \caption{\geomnist{} (\mH{}). } 
	\end{subfigure}%
    \begin{subfigure}[t]{0.25\textwidth}
	    \centering
	    \includegraphics[width=\linewidth]{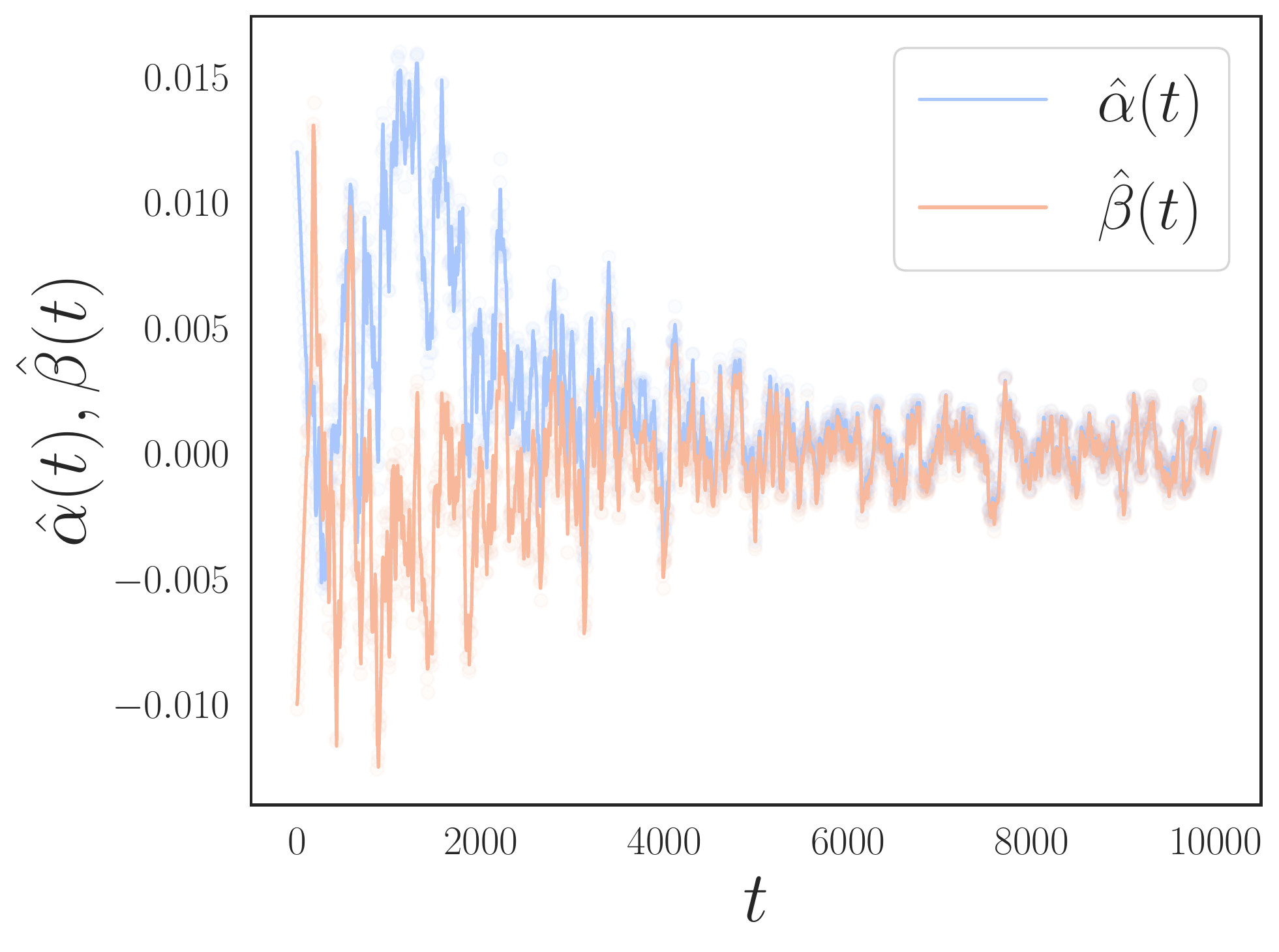}
	    \caption{\geomnist{} (\mH{}). } 
	\end{subfigure}%
    \begin{subfigure}[t]{0.25\textwidth}
	    \centering
	    \includegraphics[width=\linewidth]{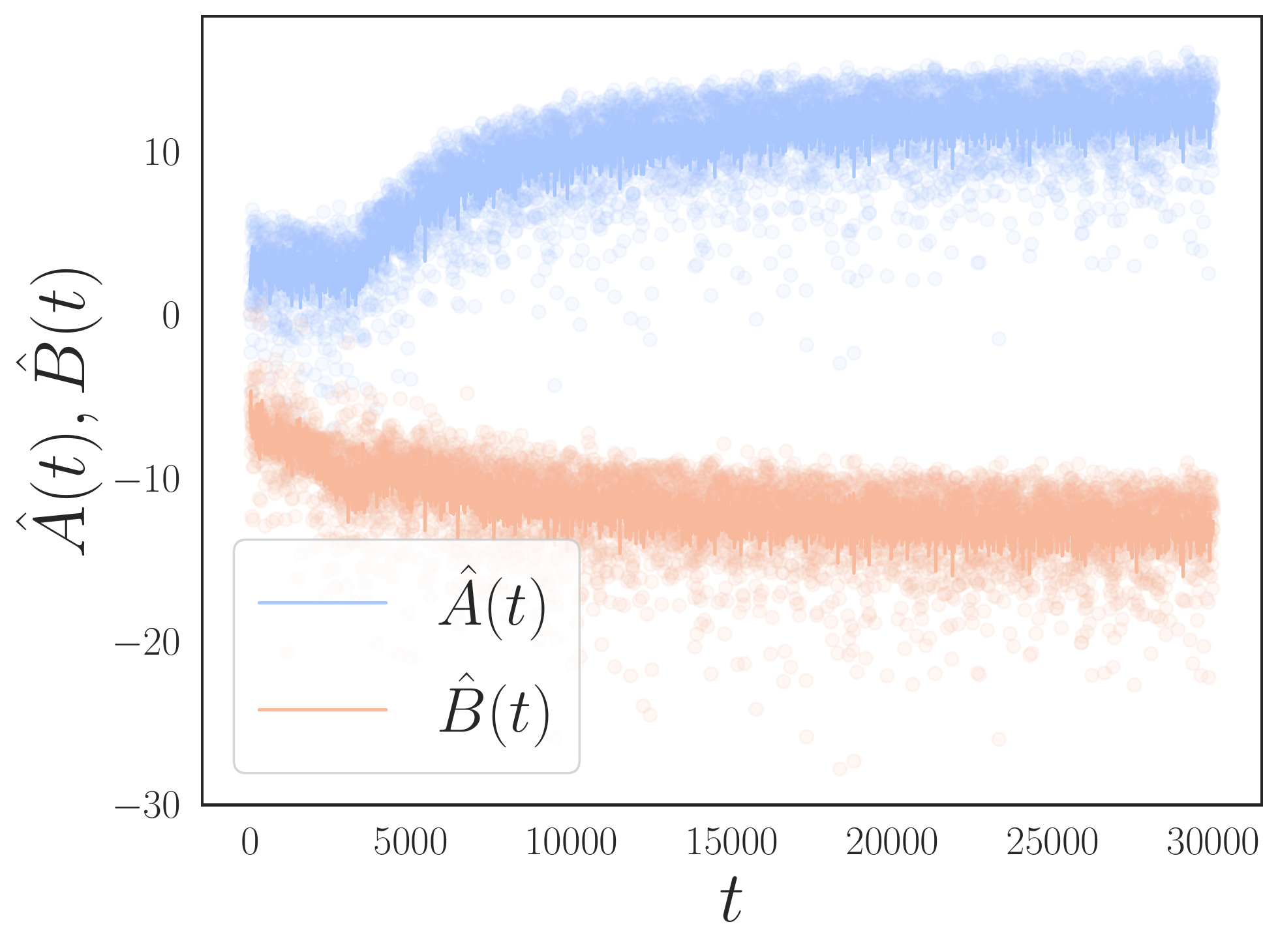}
	    \caption{\cifar{} (\mH). } 
	\end{subfigure}%
	   \begin{subfigure}[t]{0.25\textwidth}
	    \centering
	    \includegraphics[width=\linewidth]{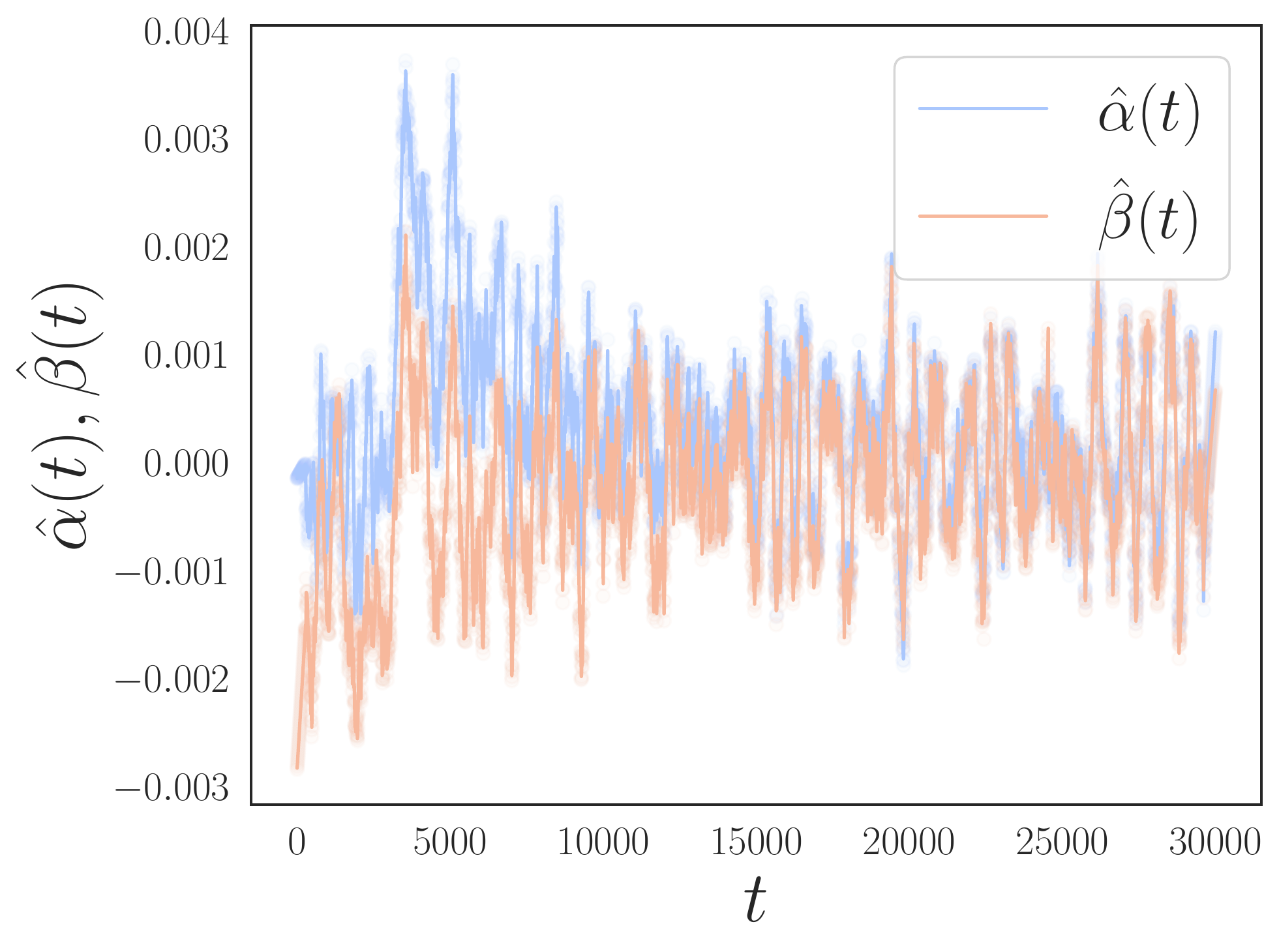}
	    \caption{\cifar{} (\mH). } 
	\end{subfigure}%
        \caption{
            \textbf{Estimated $\hat{A}(t)$, $\hat{B}(t)$, $\alpha(t)$,
            and $\beta(t)$.} The first row was estimated using
            \mI{} and the second \mH; the first two columns are on
            \geomnist{} and the last two on \cifar{}. The first and
            third rows show $\hat{A}(t)$ and $\hat{B}(t)$ and the 
            other two rows $\hat{\alpha}(t)$ and $\hat{\beta}(t).$}
        \label{fig:exp:ABab}
        \vspace{-0.6cm}
\end{figure*}

\paragraph{Local Elasticity in Neural Net Training.}
    Local elasticity manifests from our model as the 
    heaviness of the tail
    of $\gamma(t) = \alpha(t) - \beta(t)$, and in \Cref{fig:exp:ABab},
    we plot
    the estimations $\hat{A}(t)$, $\hat{B}(t)$, $\hat{\alpha}(t)$,
    and $\hat{\beta}(t)$ using both \mI{} and \mH{}.
    We note that
    (i) The estimations from the two models are visually similar, especially
    in the late stage of training when $t$ is large;
    (ii) The major difference lies in the initial
    stage, where the estimates from \mH{} behave slightly     wilder. This is not surprising because of the effect
    of the unknown constant offset $\c_0$ in the \mH{};
    (iii) Both $\hat{\alpha}(t)$ and $\hat{\beta}(t)$ behave similarly
    on both datasets.
    \toappendix{
    There are several common 
    aspects in which $\hat{\alpha}(t)$ and $\hat{\beta}(t)$ 
    behave similarly on both datasets: (i) $\alpha(t)$ rises
    as training progresses and peaks at the time when the training loss
    drops the fastest; (ii) both $\alpha(t)$ and $\beta(t)$ exhibited
    convergent behavior when $t$ is large, and $\alpha(t)$
    dominates $\beta(t)$ eventually.
}
    
\begin{figure*}[t]
    \centering
    \begin{subfigure}[t]{0.33\textwidth}
	    \centering
	    \includegraphics[width=\linewidth]{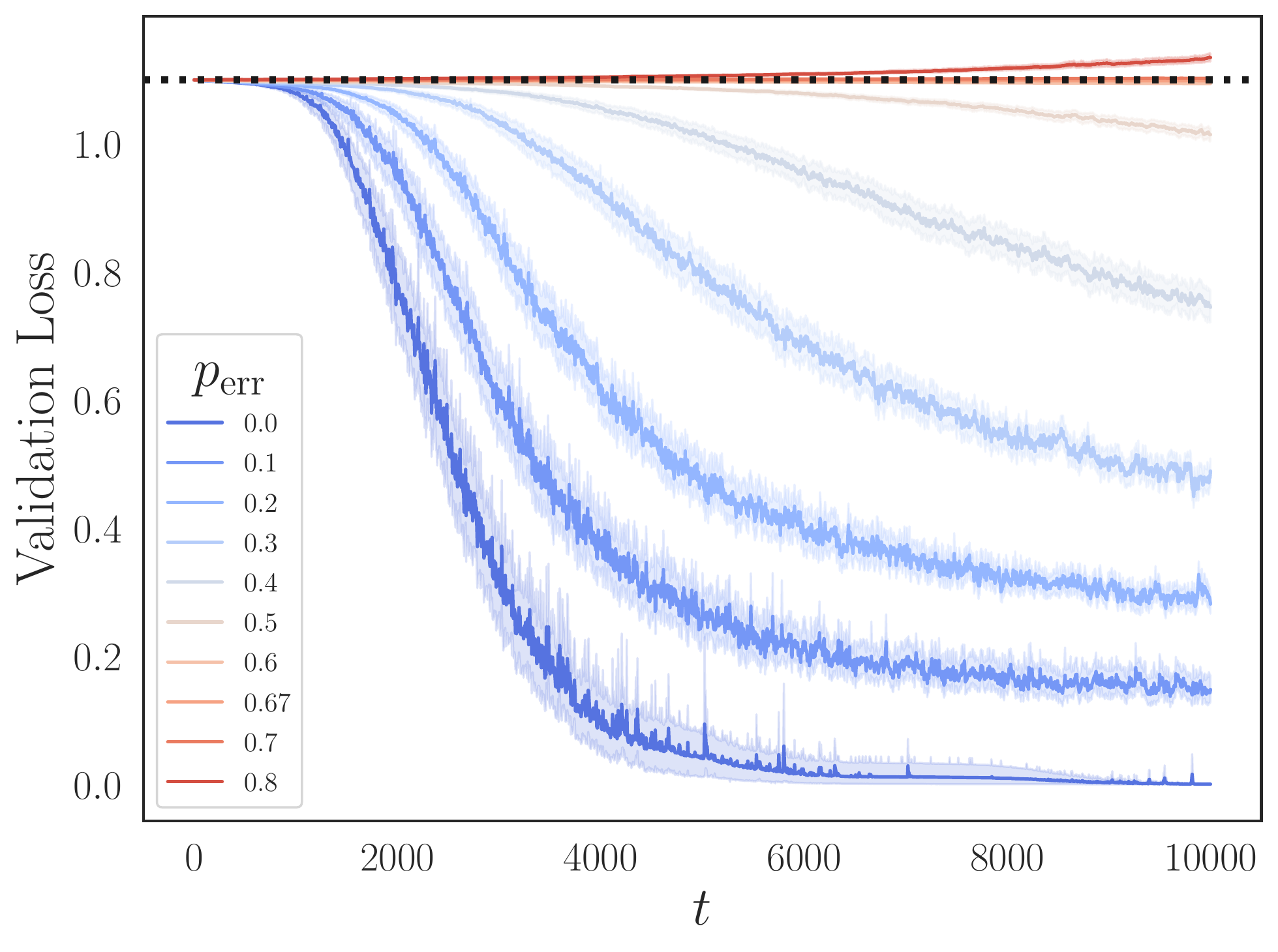}
	    \caption{Validation loss versus $\perr$.} \label{fig:exp:phase:valloss}
	\end{subfigure}~
    \begin{subfigure}[t]{0.33\textwidth}
	    \centering
	    \includegraphics[width=\linewidth]{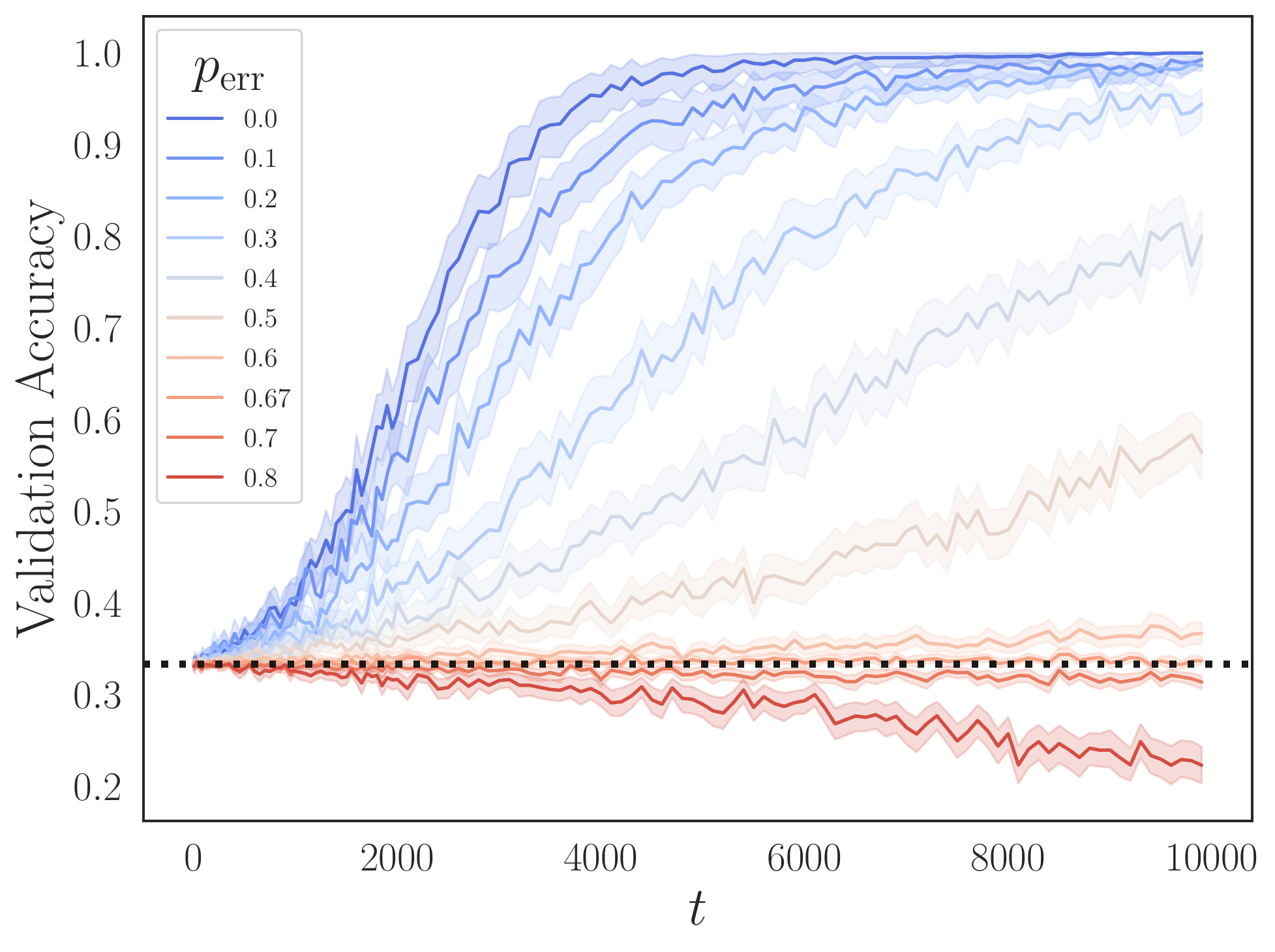}
	    \caption{Validation accuracy versus $\perr$.} \label{fig:exp:phase:valacc}
	\end{subfigure}~
	\begin{subfigure}[t]{0.33\textwidth}
	    \centering
	    \includegraphics[width=\linewidth]{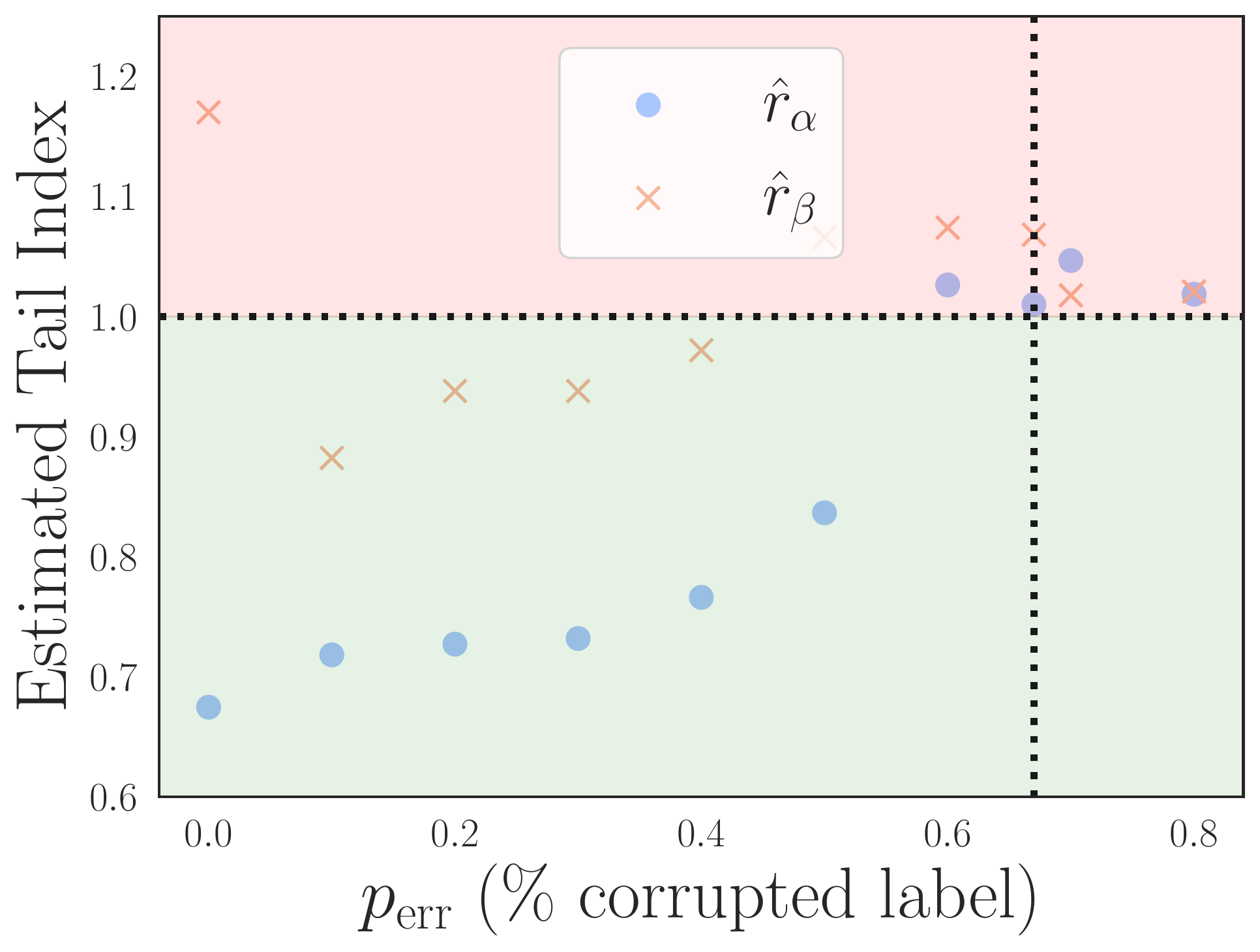}
	    \caption{Tail index versus $\perr$.} \label{fig:exp:phase:tail}
	\end{subfigure}%
        \caption{\textbf{Phase transition
        of separability.} \textbf{(a)---(b)} Validation loss and accuracy suggest
        separation fails for $\perr \ge \perr^*= 2/3$. 
        The dashed line in (a) carries the value at initialization
        and overlaps with the case where $\perr=0.6$;
        the dashed line in (b) is $\perr^*=2/3$, when
        labels are assigned completely at random.
        \textbf{(c)} Tail indices of $\alpha(t)$ and $\beta(t)$.
        Note that
        $\gamma(t) = \alpha(t)-\beta(t)$ crosses the horizontal line $r=1$,
        entering the non-separable regime (shaded in red) from the 
        separable regime (shaded in green), 
        around the same $\perr$ that cross the dashed lines
        in (a) and (b), as predicted by \Cref{thm:separation:mean}.}
        \label{fig:exp:phase}
        \vspace{-0.3cm}
\end{figure*}

\paragraph{Phase Transition of Separability.}
    \Cref{thm:separation:mean} states that separation of
    features under the LE-SDE takes place when
    $\gamma(t) = \alpha(t) - \beta(t) =\omega(1/t)$,
    or roughly speaking, when $r_{\gamma} = \min\{r_{\alpha}, r_{\beta}\} <1$.
    Although we cannot directly control $\alpha(t)$ and $\beta(t)$,
    we can bias them by tuning the label corruption
    ratio $\perr$. When $\perr \approx \perr^*\coloneqq 2/3$,
    we are in effect assigning labels completely at random and thus
    we expect a phase transition of separability should happen around $\perr^*$.
    This is indeed the story depicted in \Cref{fig:exp:phase}: 
    \Cref{fig:exp:phase:valloss,fig:exp:phase:valacc} show that
    the validation loss and accuracy for $p \ge \perr$ are not increasing
    over time and in \Cref{fig:exp:phase:tail}
    we observe the minimum tail index of $\alpha(t)$ and $\beta(t)$
    crosses $1$ from below around $\perr = 0.6$, entering
    the non-separable regime (shaded in red) from the separable regime
    (shaded in green), given in \Cref{thm:separation:mean}.
    
\begin{figure*}
    \centering
    \begin{subfigure}[t]{0.25\textwidth}
	    \centering
	    \includegraphics[width=\linewidth]{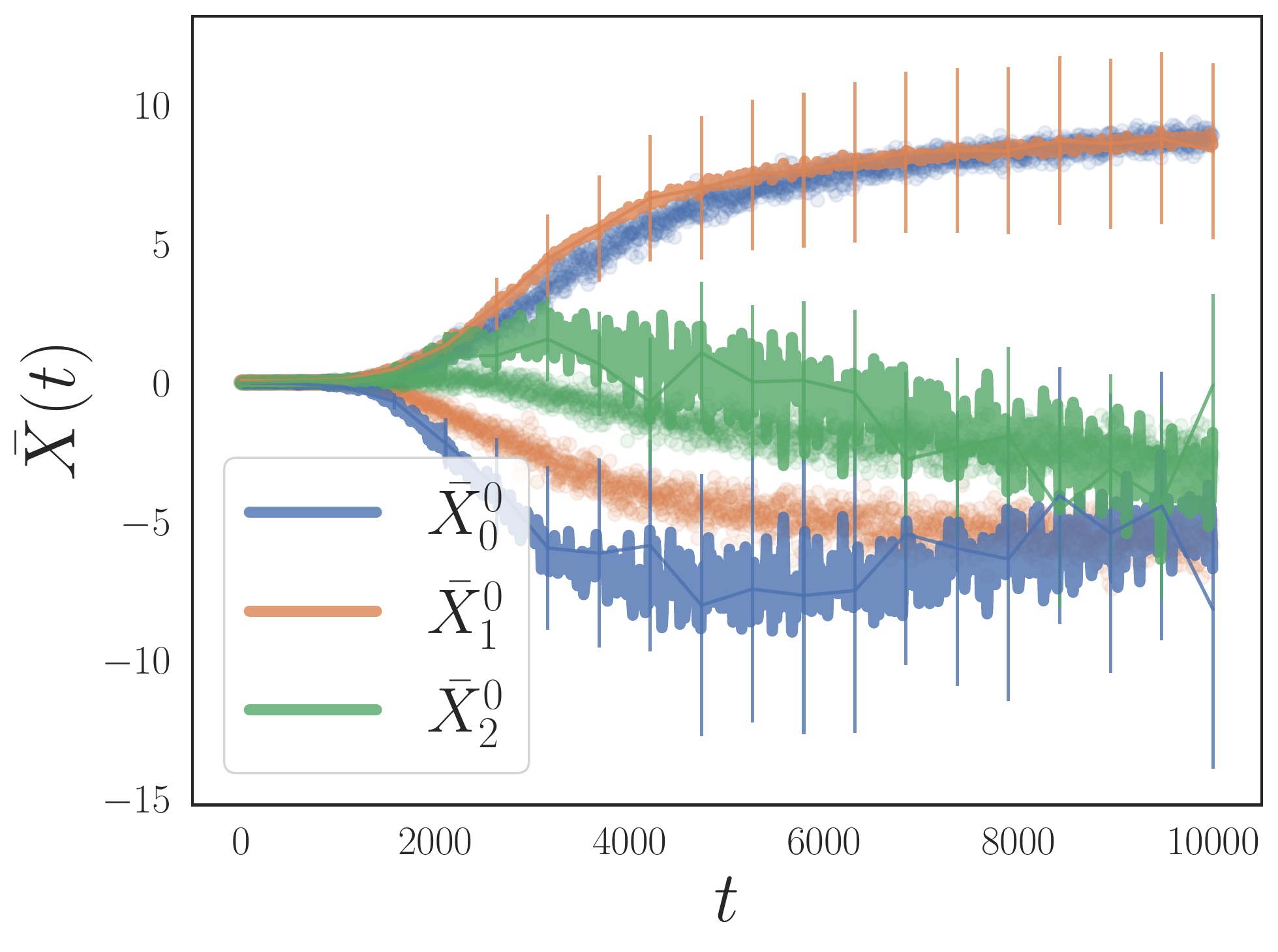}
	    \caption{\geomnist{} (\mI{}). } \label{fig:exp:sde:geomI}
	\end{subfigure}%
	   \begin{subfigure}[t]{0.25\textwidth}
	    \centering
	    \includegraphics[width=\linewidth]{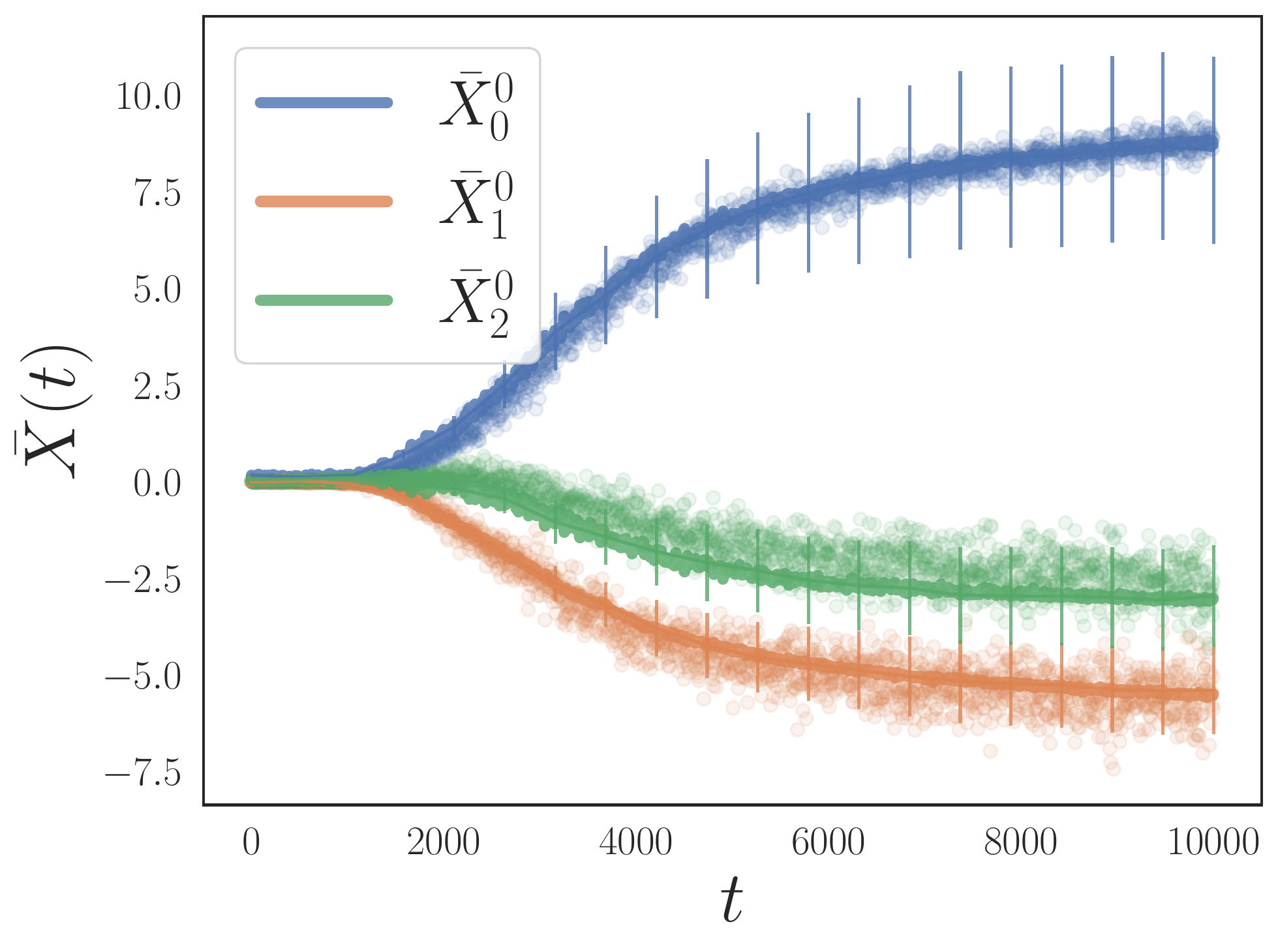}
	    \caption{\geomnist{} (\mH{}).} 
	\end{subfigure}%
    \begin{subfigure}[t]{0.25\textwidth}
	    \centering
	    \includegraphics[width=\linewidth]{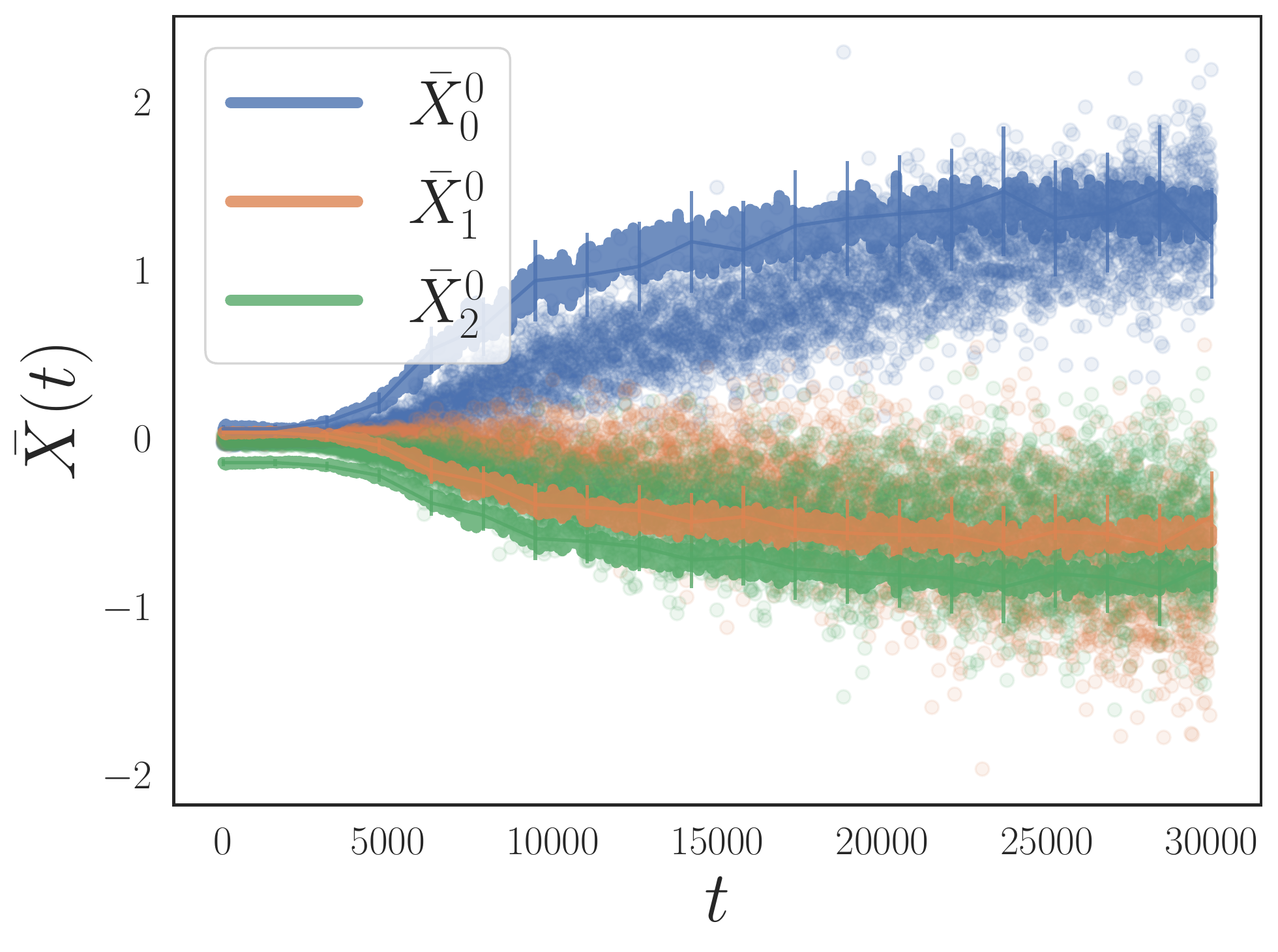}
	    \caption{\cifar{} (\mI).}
	\end{subfigure}%
    \begin{subfigure}[t]{0.25\textwidth}
	    \centering
	    \includegraphics[width=\linewidth]{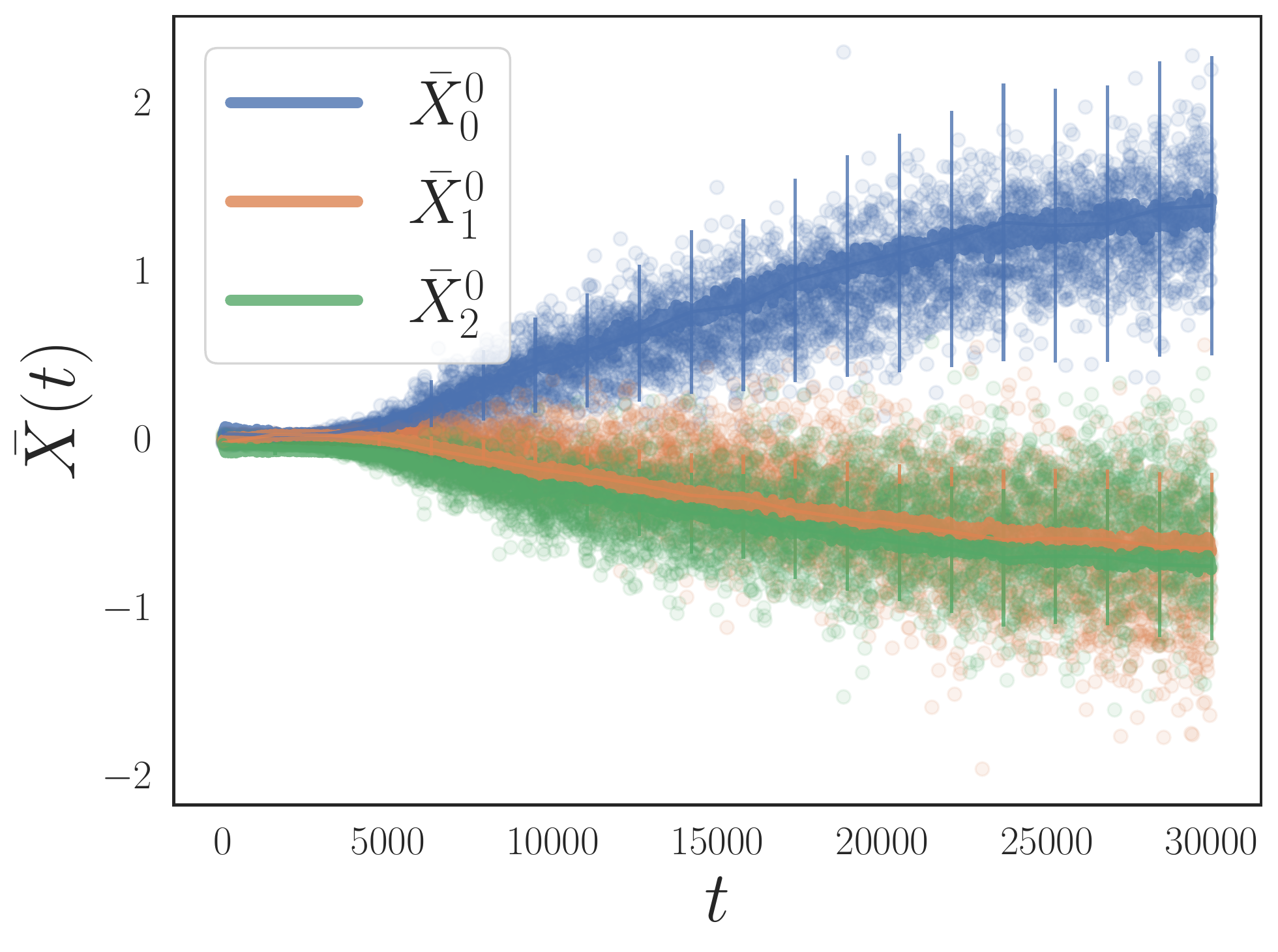}
	    \caption{\cifar{} (\mH).} 
	\end{subfigure}
        \caption{
            \textbf{Simulated LE-ODE solutions versus genuine
            dynamics.} We use $\hat{\alpha}(t)$ and $\hat{\beta}(t)$
            estimated from \mI{} ((a) and (c)) or \mH{}, 
            ((b) and (d)) and numerically simulate
            the solution under the \mH{}. The results were
            overlaid with
            true dynamics from neural nets.
	    We note \mL{} in general imitated true dynamics
	    reasonably well.
            }
        \label{fig:exp:sde}
        \vspace{-0.6cm}
\end{figure*}

\paragraph{Simulating DNN Dynamics via LE-ODE.}
    Having estimated $\alpha(t)$ and $\beta(t)$, it is natural to ask,
    to what capacity can our LE-ODE models recover the real dynamics
    of deep neural nets? We use the forward Euler method to simulate
    the \mH{} using $\hat{\alpha}(t)$ and $\hat{\beta}(t)$ estimated
    from either \mI{} or \mH{}, 
\toappendix{
    Initial value for the $k$th class
    is set to be $\bzeta^K\distas \NORMAL_K(\bzero, \sigma_k\I_{K^2})$
    for all $k\in [K]$ with $\sigma_k = \norm{\bar{\X}^k(0)}_2/\sqrt{K}$
    with $\bar{\X}^k(0)$ sampled from simulations in a DNN.
}
    We choose $K=3$ and
    show in \Cref{fig:exp:sde} the simulated solution
    (solid line) with error bars depicting
    one standard deviation over $500$ independent runs,
    overlaying on the real dynamics
    from DNNs in the background (shaded transparent markers). 
    As the moving average may reduce
    the magnitudes of $\alpha(t)$ and $\beta(t)$, we rescale
    the simulated paths such that its first coordinate 
    is approximately equal to the ground truth at convergence.
    Note that estimations from \mH{} can faithfully
    recover the genuine dynamics from neural nets, whereas
    those from \mI{} fail, notably in \Cref{fig:exp:sde:geomI},
    where the simulated paths preserve the relative magnitude but
    fail to identify the correct order of three logits.
\toappendix{
    This is due to the nature of \mI{}, which does not distinguish
    logits from different classes, 
    and thus \mI{}-based estimations would arguably
    perform poorly when there is supervision.
    We provide more experimental data in \Cref{sec:app_sim:results},
    including simulations under the \mI{} and the results
    from all other classes.
}

\section{Discussion and Future Works} \label{sec:discussions}
In this study, we introduce LE-SDE/ODE models that draw inspiration
from the local elasticity phenomenon. Conditions for
sharp phase transition of separability of features are derived.
We also show that once the elasticity strengths $\alpha(t)$ and
$\beta(t)$ are well estimated, our model can faithfully simulate
the dynamics of neural nets. 
\toappendix{
    This link between a tractable
mathematical model (LE-SDE/ODE) and an expressive yet complicated
black box would potentially yield greater benefits toward
demystifying the magic of deep models. }
We outline a few
interesting problems for future research
while leaving the details in the Appendix.
\textbf{(i) General LE Matrix.} A similar result as in \Cref{thm:separation:mean} may be expected for symmetric but no necessarily semi-definite LE
matrices $\E(t)$.
\textbf{(ii) Mini-batch Training, Imbalanced Datasets, and Label Corruptions.} Generalizing
the drift matrix to $\M_t = \left(\E_t \otimes \P\right)\circ \H / K$ for a $K$-by-$K$ doubly
stochastic matrix $\P$ can be used to model various
sampling effects.
\textbf{(iii) Beyond \mL{} for Imitating Genuine Dynamics of DNNs.} 
    Although the \mL{} is shown to be able
    to mimic the real dynamics reasonably well,
    we postulate that a more precise model
    might have its $(i,j)$-th block encode the
    other directions other than $\d_j$.

\section*{Acknowledgements}



This work was supported in part by NSF through CCF-1934876, an Alfred Sloan Research Fellowship, the Wharton Dean's Research Fund, and ONR Contract N00014-19-1-2620. We would like to thank Dan Roth and the Cognitive Computation Group
at the University of Pennsylvania for stimulating discussions and for providing computational resources.


{\footnotesize
\nocite{cohen2021gradient}
\bibliography{references}
\bibliographystyle{abbrv}
}



\clearpage

\setcounter{section}{1}\setcounter{equation}{0}
\numberwithin{equation}{section}
\counterwithin{figure}{section}

\appendix

\section{Derivation of Continuous Dynamics of Binary Case}
\label{sec:app_binary_continuous}
We will now derive continuous dynamics \eqref{eq:simple_binary_SDE}
in the main paper.
Let $\ind{m} = 1$ if class $1$ is selected at iteration $m$
and $\ind{m} = 0$ otherwise. 
Chaining the dynamic \eqref{eq:simple_binary_SDE} $r$ times, we have
\begin{align*}
    &X^1_{i}(m-1+r)-X^1_{i}(m-1) \\
=&\sum_{q=1}^{r}\left(\ind{m+q-1} \alpha h X^1_{I_{m+q-1}}(m+q-2)+\left(1-\ind{m+q-1}\right) \beta h X^2_{I_{m+q-1}}(m+q-2)+\zeta_{m+q-2}^{i}\right).
\end{align*}
When $m \gg r$, we have approximately
\begin{align*}
&\sum_{q=1}^{r}\left(\ind{m+q-1} \alpha h X^1_{I_{m+q-1}}(m+q-2)+\left(1-\ind{m+q-1}\right) \beta h X^2_{I_{m+q-1}}(m+q-2)\right) \\
\approx &\sum_{q=1}^{r}\left(\ind{m+q-1} \alpha h X^1_{I_{m+q-1}}(m-1)+\left(1-\ind{m+q-1}\right) \beta h X^2_{I_{m+q-1}}(m-1)\right) \\
=&\sum_{q=1}^{r} \ind{m+q-1} \alpha h X^1_{I_{m+q-1}}(m-1)+\sum_{q=1}^{r}\left(1-\ind{m+q-1}\right) \beta h X^2_{I_{m+q-1}}(m-1) \\
\approx &~r \cdot \frac{1}{2} \alpha h \frac{\sum_{i=1}^{n} X^1_{i}(m-1)}{n}+r \cdot \frac{1}{2} \beta h \frac{\sum_{j=1}^{n} X^2_{j}(m-1)}{n} \\
\approx& ~\frac{\alpha hr}{2 } \bar{X}(m-1)+\frac{\beta hr}{2 } \bar{Y}(m-1).
\end{align*}
Next, observe that
\begin{align*}
\sum_{q=1}^{r} \zeta_{m+q-2}^{i} \sim \mathcal{N}\left(0, \sigma^{2} r h\right),
\end{align*}
hence taken together, the calculations above give
\begin{align*}
X^1_{i}(m-1+r)-X^1_{i}(m-1) \approx \frac{\alpha h r}{2 } \bar{X}(m-1)+\frac{\beta hr}{2 } \bar{Y}(m-1)+\mathcal{N}\left(0, \sigma^{2} r h\right).
\end{align*}
Writing $\Delta t=r h$ and $t=(m-1) h$, we have
\begin{align*}
X^1_{i}(t+\Delta t)-X^1_{i}(t) \approx \frac{\alpha}{2} \bar{X}(t) \Delta t+\frac{\beta}{2} \bar{Y}(t) \Delta t+\sigma \mathcal{N}(0, \Delta t),
\end{align*}
which is the discretization of
\begin{align*}
\mathrm{d} X^1_{i}(t)=\left(\frac{\alpha}{2} \bar{X}(t)+\frac{\beta}{2} \bar{Y}(t)\right) \mathrm{d} t+\sigma \mathrm{d} W^{i}(t).
\end{align*}
Likewise, we can obtain the dynamics of $X^2_{j}$ similarly.
We will next prove the separation theorem in binary classification, \Cref{thm:binary_separation}.

\thmbinaryseparation*

\begin{proof}[Proof of \Cref{thm:binary_separation}]
Note that whenever $\alpha \leq \beta$, we have $\frac{c_{1}-c_{2}}{2} 
\eu^{\frac{\alpha-\beta}{2} t} \rightarrow 0$ as $t \rightarrow \infty$,
thus $X^1_{i}$ and $X^2_{j}$ are interspersed and separation happens with probability tending to zero. This also aligns with our intuition that the intra-class 
effect should be stronger than its inter-class counterpart.

On the other hand, when $\alpha > \beta$, ignoring a null set we may
assume $c_1 > c_2$ without loss of generality.
\rednote{ To see this,
    note that by definition $c_k = \Exp_{\text{data}}[ X^k(0) | \theta(0) =\theta_0 ]$ for $k\in\{1,2\}$ where $\theta(0)$ is all parameters of the neural net at initialization and $\theta_0$ is a particular realization given the initialization scheme. Here the expectation is taken with respect to the data
    distribution, and
    when we ignore a null set of neural net with respect to the probability measure
    induced by the initialization scheme, $c_1 \ne c_2$ holds.
    In other words,
    this statement can be interpreted as ``the expected feature at initialization from the first class is different from that from the second class for a neural
    net, except possibly on a null set in the space of neural nets with respect to the probability measure induced by the parameter initialization scheme.''
}
It suffices to show that
\begin{align*}
\min_{1 \leq i \leq n} \frac{c_{1}-c_{2}}{2} \eu^{\frac{\alpha-\beta}{2} t}+\frac{c_{1}+c_{2}}{2} & \eu^{\frac{\alpha+\beta}{2} t}
    -c_{x}+X^1_{i}(0)+\sigma W^1_{i}(t) \\
>& 
    \max_{1 \leq j \leq n}-\frac{c_{1}-c_{2}}{2} \eu^{\frac{\alpha-\beta}{2} t}+\frac{c_{1}+c_{2}}{2} \eu^{\frac{\alpha+\beta}{2} t}-c_{2}+X^2_{j}(0)+\sigma W^2_{j}(t),
\end{align*}
which is equivalent to
\begin{align*}
\left(c_{1}-c_{2}\right) \eu^{\frac{\alpha-\beta}{2} t}-c_{1}+\min _{1 \leq i \leq n} X^1_{i}(0)+\sigma W^1_{i}(t)>-c_{2}+\max_{1 \leq j \leq n} X^2_{j}(0)+\sigma W^2_{j}(t).
\end{align*}
But the above display happens with probability tending to one
provided $\alpha > \beta$, thus completing the proof.
\end{proof}

\section{Further Details on Drift Modeling}
\label{sec:app:dynamics}
\subsection{Dynamics of the Logits-as-Features Model}
    This section provides more details on why 
    the construction of $\H$ in \Cref{sec:model:H} is probably a
    good choice
    for modeling the dynamics of logits in deep neural nets.
    Given $K \ge 2$ classes with $n$ training examples per class,
    the feature vectors are the logits $\X^k(m)\in \sR^K$
    for all $k\in[K]$, where $m$ is the iteration number.
    When the neural net is trained
    under the softmax cross-entropy loss $L$,
    at the $m$-th iteration, 
    if the $J_m$-th sample from the $L_m$-th class  is sampled,
    the dynamics of the logits $\X^k_i$
    should be governed by 
    \ba\label{eq:exact_dynamic}
       \X^k_i(m)-\X^k_i(m-1) \approx & h \left[\frac{\partial \X^k_i(m-1) }{\partial \w} \frac{\partial \X^{L_m}_{J_m}}{\partial \w}^\top 
       \left(\e_{L_m} - \softmax (\X^{L_m}_{J_m})\right)
        \right].
    \ea
    The derivation of \cref{eq:exact_dynamic}
    is a straightforward computation from the Taylor approximation
    \ba
    \X^k_i(m)-\X^k_i(m-1) \approx \nabla_\w \X^k_i(m-1) \Delta\w(m-1),
    \ea
    where we observe that 
    \ba
    \Delta\w(m-1) = -h\frac{\partial L(\w(m-1))}{\partial \w} =  h\frac{\partial \X^{L_m}_{J_m}}{\partial \w}^\top  \left(\e_{L_m} - \softmax (\X^{L_m}_{J_m})\right).
    \ea
    
    However, as \cref{eq:exact_dynamic} is highly non-linear,
    $J_m$ and $L_m$  are random,
    and the Gram matrix $\frac{\partial \X^k_i(m-1) }{\partial \w} \frac{\partial \X^{L_m}_{J_m}}{\partial \w}^\top $ is also
    time-dependent, direct analyses and simulation
    of the exact dynamics are difficult. Note that
    this Gram matrix is also the key element in the NTK literature \cite{jacot2018neural, du2018gradient, arora2018convergence},
    which is treated as roughly fixed during the lazy training process.
    
    Recall that we consider the per-class mean of logits, $\bar{\X}^k=\Exp \tilde{\X}^k$, where $\tilde{\X}^k$ is a random sample,
    for each class $k$.
    In the limit of $h\to 0$ with the identification of time $t = mh$, 
    the above modeling allows us to re-write \eqref{eq:exact_dynamic} 
    in the per-class mean $\bar{\X}^k$ and logits from a generic sample
    $\tilde{\X}^k$ in this continuous limit as
    \be  \label{eq:exact:sde}
        {\footnotesize
        \begin{aligned}
        \dd\tilde{\X}^k_t 
        &\approx \Exp_{L\sim \UNIFORM([K])}\left[\Exp_{\tilde{\X}\sim \mathcal{D}_t^{L}} \left[\frac{\partial \X^k_i(m-1) }{\partial \w} \frac{\partial \tilde{\X}}{\partial \w}^\top  \left(\e_L - \softmax (\tilde{\X})\right)\right]
        \right] \dd t  +\bSigma^{\frac{1}{2}}_t \dd \W_t,\\
        &\approx \frac{1}{K} \sum_{L} \left(\left[\Exp_{\tilde{\X}^\prime\sim \mathcal{D}_t^{k}, \tilde{\X}\sim \mathcal{D}_t^{L}} \frac{\partial \tilde{\X}^\prime }{\partial \w} \frac{\partial \tilde{\X}}{\partial \w}^\top \right] \left(\e_L - \softmax (\bar{\X}^L_t)\right) \right)\dd t+ \bSigma^{\frac{1}{2}}_t \dd W_t, \\
        &= \frac{1}{K} \sum_{L} \left(\Theta_{k, L} \left(\e_L - \softmax (\bar{\X}^L_t)\right) \right)\dd t+ \bSigma^{\frac{1}{2}}_t \dd \W_t.
        \end{aligned}
        }
    \ee
    The presence of expectation over non-linearity posed considerate
    difficulties of using \cref{eq:exact:sde} 
    to analyze neural nets; however, it motivates our LE-SDE model as
    a linear approximation to it while explicitly encodes the local elasticity into the dynamics.
    More precisely, given a sample instance $\z^k$ from the $k$-th class,
    in a well-trained neural network, 
    the probability vector associated with $\z^k$,
    i.e., the output from the softmax applied to its logits,
    should have its $k$-th entry as the largest and the other entries as
    roughly equal. Hence $\e_k - \softmax(\bar{X}^k_t)$
    should be roughly in the direction of $\e_k - \frac{1}{K}\bone$.
    To approximate this limit by the limit from a linear map, 
    we may use our choice of
    $\H_{kl} = \bar{\H}^l \coloneqq \frac{\d_l\d_l^\top}{\|\d_l\|^2}$ for all $k,l\in[K]$, such
    that in the limit of $t\to\infty$, we expect
    \ba \label{app:eq:Lmodel_approx}
        \e_k - \softmax\left(\bar{\X}^k_t\right) \approx C 
        \bar{\H}^k \bar{\X}^k_t,
    \ea
    for any $k\in[K]$ and some positive constant $C$.
    This is the intuition behind our \mH. To summarize,
    in our model, the local elasticity matrix $E$ 
    describes the effect $\Theta_{k, L}$, and $\H_{k, l}$ 
    transforms the mean logit $\bar{\X}^L_t$ 
    to the direction governed by the supervision.
    
    \subsection{Discussions on Linearization} \label{sec:app:dynamics:linear}
   We provide more details of the rationale behind our choice of $M(t) X(t)$, a seemingly linear term, as the surrogate for the non-linear drift in the dynamics given
   in equation~\ref{eq:exact:sde}.  Our following argument can be extended to any post-activation features. 
   Writing equation~\ref{eq:exact:sde} in terms of $X(t) \in \mathbb{R}^{Kp}$ (recall in this case we have $p=K$), the concatenation of $K$ per-class feature vectors $X^k(t) \in \mathbb{R}^p$ for $k \in [K]$, and denoting by $\sigma:\mathbb{R}^K \to \mathbb{R}^K$ the softmax function for simplicity, we can express the drift term as
    \ba
    F(\tilde{\X}(t), t) \coloneqq \Theta(t) \left( \left[ e_k - \sigma(\tilde{\X}^k(t)) \right]_{k=1}^K \right),
    \ea
    where we wrote $[\cdot]$ for vector concatenation and $\Theta(t) \in \mathbb{R}^{(Kp)\times (Kp)}$ for the Gram matrix. A commonly used linearization scheme in SDE for non-linear drifts by the filtering community (cf. Chapter 9.1 of \cite{solin2019asde}) 
    is to linearize $F$ for each $t$ at the mean $\phi(t) =(\phi_k(t))_{k=1}^K \equiv \bar{X}(t) :=\mathbb{E}_B X(t)$ where the expectation is taken with respect to
    the diffusion. Concretely, we have
    \ba
        F(\tilde{\X}(t), t) \approx \tilde{F}(\tilde{\X}(t), t) := F(\phi(t), t) + \nabla_X F(\phi(t), t)\left( \tilde{\X}(t) - \phi (t)\right),
    \ea
    where $\nabla_X F$ denotes the Jacobian of $F$ with respect to the spacial variable $\tilde{\X}$. For notation completeness, we introduce
    \ba
        p=(p_k)_{k=1}^K \in \mathbb{R}^{Kp}, \quad p_k := \sigma (\tilde{\X}^k(t)) \in \mathbb{R}^p, 
        \bar{p} = (\bar{p}_k) _{k=1}^K \in \mathbb{R}^{Kp}, \quad k \in [K],
    \ea
    and similarly
    \ba
         \bar{p}_k := \sigma (\bar{\X}^k(t))  \in \mathbb{R}^p,
        \quad k \in [K].
    \ea
    We write the per-class Jacobians as
    \ba
        J_{kk} = J_k := \operatorname{diag}(\bar{p}_k)-\bar{p}_k\bar{p}_k^T.
    \ea

    Clearly, the Jacobian $\nabla F(\phi, t) = J(t)$
    can be written as a block-diagonal matrix $J(t) = (J_{kk})_{k=1}^K$ consisting of per-class Jacobians. Now continuing linearization, we can write
    \ba
        \tilde{F}(\tilde{\X}(t), t) &= \Theta(t) \left( [e_k - \bar{p}_k]_k + J(t)(\tilde{\X}(t) - \phi(t))\right) \\
        &= \Theta(t) \left( J(t) X(t) + \left[ e_k - \bar{p}_k + J_k \phi_k(t) \right]_k\right).
    \ea
    Define $\Psi: \mathbb{R}^{Kp} \to \mathbb{R}^{Kp} : z \mapsto [e_k - \sigma(z_k)]_k$ and write $\Psi_k: \mathbb{R}^{p} \to \mathbb{R}^p$ to be the $k$-th component of $\Psi$, using Taylor's theorem to expand $\Psi(z)$ around $\phi(t)$ for each $t$, we have
    \ba
        \Psi = \Psi(\phi) + J(t)\phi - J(\phi) z + o\left(\norm{z-\phi}\right),
    \ea
    or
    \ba
        \Psi(\phi) + J(t) \phi = \Psi(z) + J(\phi) z + o\left(\norm{z-\phi}\right).
    \ea
    This implies that
    \ba \label{app:eq:linear_residue}
        \tilde{F} = \Theta(t) J(t) \tilde{\X}(t) + \Theta(t) R(t), \quad R(t;z) := \Psi(z) + J(t) z + o\left(\norm{z-\phi(t)}\right),
    \ea
    where $R(t;z)$ is the residue that depends on the choice of $z$ around which $\Psi(\phi)$ is expanded. Note that the first term is a time-varying linear term in $X(t)$. As long as the residue term is negligible, we get exactly the time-varying linear map $M(t)$ in the LE-SDE model. 

    By choosing different $z$'s in equation~\ref{app:eq:linear_residue}, we can focus on different stages of the real dynamics using the LE-SDE. Two particular choices are of great interest so as to make the residue term vanishing:
    \bitem
        \item \textbf{Around initialization.} Let $z=u := c \cdot [\boldsymbol{1}_K/K] _{k=1}^K$  be a scaling of vectors of ones where $c$ is some fixed constant. Then each of the $K$ components of $\sigma(u)$ assigns approximately the same probability ($1/K$) for every label. Furthermore, $u \in \operatorname{Ker} J(t)$ for all $t$ hence the residue $R(t; u) = \Psi(u) + o\left(\norm{z-\phi(t)}\right)$ is a constant vector (which is colinear with $\d=(\d_j)_{j=1}^K$ defined by $\d_j=\e_j-\bone_K/K$ when
        we discussed the \mL). Note that this approximation works best when the model has not learned much about the data (in the ``first stage'' as we call it) since $\norm{u-\phi(t)}$ is small in this regime.

        \item \textbf{Around convergence.} Given that the model converges, $\phi_{\infty} \coloneqq \phi(\infty)$ is finite. Let $z=\phi_{\infty}$, under the effective training assumption, $\norm{\Psi(\phi_{\infty})} \approx 0$ by construction. Hence the residue $R(t; \phi_ {\infty}) =  J(t)\phi_ {\infty} + o\left(\norm{\phi(t)-\phi_{\infty})}\right)$. Here the $o(\cdot)$ term converges to $0$ as training progresses, leaving us a term that is asymptotically equivalent to $v = (v_k) _ {k=1}^K \coloneqq J(\phi _ {\infty}) \phi _ {\infty} \in \mathbb{R}^{K^2}$, where $v_k = [(z_{k,i}- \sum_{j=1}^K p _ {k,j} z_{k, j}) p _ i] _ {i=1}^K \in \mathbb{R}^K$. Again, under the effective training assumption $z_k$ has its $k$-th entry $z_{k,k}$ the largest, and $p_k$ has its $k$-th entry close to $1$ while the others to zero. Thus $v_{k,i} \approx 0_K$. We see that in this regime, the approximation $\Theta(t) J(t) X(t)$ is only off by a residue $o\left(\norm{\phi(t) - \phi_{\infty}}\right)$ that eventually vanishes.
\eitem
    As discussed above, we choose to linearize the drift at convergence instead of around initialization in the \mL{} given by equation~\ref{app:eq:Lmodel_approx} such that the residue vanishes under the effective training assumption.


\section{Miscellaneous Proofs}\label{sec:app_proofs}
\subsection{Various Definitions of Separability} \label{sec:app:def:sep}

   We first give formal definitions of separability which generalize \Cref{thm:binary_separation} in \Cref{sec:binary} to our $K$-class, $p$-dimensional feature setting. In the beginning of \Cref{sec:separation_thm}, we state the definition of separability in natural language, and we formalize the definition therein as follows. It is the most natural definition in terms of linear separation by a hyperplane.
   \bd[Pairwise Separation]
   \label{def:ps1}
   We say the feature vectors $\left\{(\X^k_i)_{i\in[n]}\right\}_{k\in[K]}$ is pairwisely separable at time $t$ if for each pair of $1\le k< l \le K$, there exists a direction $\bnu_{k, l}$ such that
    \begin{align}\label{eq:sep_nu_condition}
            \min_{i} \ip{\bnu_{k, l}}{\X^k_{i}(t)}>\max _{j} \ip{\bnu_{k, l}}{\X^{l}_j(t)}.
    \end{align}
    \ed
    In the \Cref{thm:separation:mean}, we claim that when $\gamma(t) = \omega(1/t)$, the pairwise separation (\Cref{def:ps1}) happens with probability tending to $1$ as $t\to\infty.$ This is a notion of asymptotic separability stated in \Cref{thm:separation:mean}, yet it is a weaker notion of separation in the following sense: the hyperplane may depend on the classes $l, k$, and the hyperplane may depend on time $t$. We state the following asymptotic  separable definitions of increasingly stronger guarantees, and will remark on how to obtain them in our proof of \Cref{thm:separation:mean}.
    
    The direct application of \Cref{def:ps1} gives us the following (weakest) definition.
    \bd[Asymptotic Pairwise Separation]\label{def:pairwise_sep}
        We say the feature vectors $\left\{(\X^k_i)_{i\in[n]}\right\}_{k\in[K]}$ are asymptotically pairwisely separable if for each pair of $1\le k< l \le K$, there exist directions $\bnu_{k, l}(t)$ such that 
        \begin{align}\label{eq:sep_nu_condition_prob}
            \Pr\left(\min_{i} \ip{\bnu_{k, l}(t)}{\X^k_{i}(t)}>\max _{j} \ip{\bnu_{k, l}(t)}{\X^{l}_j(t)}\right)\to 1.
        \end{align}
    \ed 
    Requiring all the classes to be separable with the same hyperplane give us the following universal separation.
    \bd[Asymptotic Universal Separation]\label{def:universal_sep}
    We say the feature vectors $\left\{(\X^k_i)_{i\in[n]}\right\}_{k\in[K]}$ are asymptotically universally separable if it is asymptotically pairwisely separable, and there exists $\nu$, such that either $\bnu = \bnu_{k, l}$ or $\bnu = - \bnu_{k, l}$, for all $1\le k< l \le K$ in \eqref{eq:sep_nu_condition_prob}.
    \ed
    Note that we allow the universal direction differs in sign for different pair of classes. This is because we do not require separating the $k$ class in any specific order.
    
    We note that the above definitions are in the sense of 
    separation ``in probability'',
    which only asserts separation at an arbitrarily fixed large time $t$. Specifically, it does not guarantee the existence of a fixed direction that can always separate a pair of classes for all sufficiently large $t$. We now state the almost sure definition of asymptotic pairwise separation, which guarantees the same fixed direction separates a pair of classes for all large enough $t$. And \Cref{thm:separation:mean} also holds for such definition. 
 \bd[Uniform Asymptotic Pairwise Separation]\label{def:uniform_sep}
        We say the feature vectors $\left\{(\X^k_i)_{i\in[n]}\right\}_{k\in[K]}$ are uniformly asymptotically pairwisely separable if for each pair of $1\le k< l \le K$, there exists a direction
        $\bnu_{k, l}$ such that 
        \begin{align}\label{eq:C4_sep_cond}
            \liminf_{T\to\infty}\min_{k, l}\Pr\left(\min_{i} \ip{\bnu_{k, l}}{\X^k_{i}(t)}>\max _{j} \ip{\bnu_{k, l}}{\X^{l}_j(t)}, \text{ for all }t>T\right) = 1.
        \end{align} 
    \ed 
    We define uniform asymptotic universal separation the same way as its non-uniform counterpart.
    \bd[Uniform Asymptotic Universal Separation]\label{def:uniform_sep2}
        We say the feature vectors $\left\{(\X^k_i)_{i\in[n]}\right\}_{k\in[K]}$ are uniformly asymptotically universally separable if it is uniformly asymptotically pairwisely separable, and there exists $\nu$, such that either $\bnu = \bnu_{k, l}$ or $\bnu = - \bnu_{k, l}$, for all $1\le k< l \le K$ in \eqref{eq:C4_sep_cond}.
    \ed 
    The asymptotic inseparability definitions can be similarly stated. 

\subsection{Proofs in \Cref{sec:lesde:general}} \label{sec:app:pf}
    
    Before proving the main result, \Cref{thm:separation:mean},
    it is convenient to first prove \Cref{thm:imodel,thm:hmodel}.

    \subsubsection{Proof of \Cref{thm:imodel}} \label{sec:app:pf:imodel}
        \thmimodel*


    \bp[Proof of \Cref{thm:imodel}.]
        When $\alpha(t) = \alpha$ and $\beta(t) = \beta$ are
        constants, $\E$ has two distinct
        eigenvalues: $\ld_0 = (\alpha-\beta)/K$ with multiplicity
        $K-1$ and $\ld_1 = (\alpha+(K-1)\beta)/K$ with multiplicity one,
        and the eigendecomposition $E = \Q\bLd \Q^{-1}$ is given
        by
        \ba
            \bLd = \diag(\ld_1, \ld_0, \ldots, \ld_0) \in \sR^{K\times K}, \quad
            \Q = \bbm
                1  & 1 & 1 & \cdots & 1 \\
                1  & -1& 0 & \cdots & 0 \\
                1  & 0 & -1& \cdots & 0 \\
                \vdots & & & & \vdots \\
                1 &  0 & 0 & \cdots & -1 \\
            \ebm.
        \ea
        Hence the eigenpairs of $\E \otimes \I_p$ are
        $(\e_1-\e_k)\otimes \e_j$ for $1 <k \le K$ and $1 \le j \le p$
        with eigenvalue $\ld_0$
        and $(\bone_K\otimes \e_j)$ for $1\le j \le p$ with eigenvalue
        $\ld_1$.
        Reparameterizing the solution such that the initial
        values are $\bar{\X}^k(0) = \c_0 + \c_k$ for all $k\in[K]$,
        we have
        \ba
            \bar{\X}_t = 
                \c \eu^{\ld_0 t} +  
                \left(\bone_K\otimes \c_0\right) \eu^{\ld_1 t} ,
        \ea
        for $\c_0\in\sR^p$ and $\c=(\c_k)_{k=1}^K$
        with $\sum_{k=1}^K\c_k = \bzero$.
        Based on this solution, it is not difficult to show that the general
        solution with varying $\alpha(t)$ and $\beta(t)$ is given by
        \ba
            \bar{\X}_t =  \c \eu^{\frac{1}{K}A(t) - \frac{1}{K}B(t)} +  
            \left(\bone_K\otimes \c_0\right) \eu^{\frac{1}{K}A(t) + \frac{K-1}{K}B(t)},
        \ea
        where $\sum_{k=1}^K \c_k = \bzero$.
        \ep

    \subsubsection{Proof of \Cref{thm:hmodel}} \label{sec:app:pf:hmodel}
        \thmhmodel*
        \bp[Proof of \Cref{thm:hmodel}.]
            Recall that the $(k,l)$-th block of the $\H$ matrix is
            $\bar{\H}^l\coloneqq \d_l\tp{\d}_l/\tp{\d}_l\d_l$.
            When $\alpha(t) = \alpha$ and $\beta(t) = \beta$ are constants, we immediately find the matrix $(\E\otimes \I_K)\circ \H$ has an
            eigenvalue of $\alpha-\beta$ with multiplicity one
            whose  eigenvector is $\d$, and an eigenvalue of $0$ with multiplicity $K(K-1)$ -- 
            since the null spaces of $\H_{kl}$ all have dimension $K-1$.
            We also find it has another eigenvalue $\alpha + \beta/(K-1)$ with multiplicity
            $K-1$. Therefore, 
            we have the general solution \eqref{eq:leode:sol:H} by noting there is an additional $1/K$ factor
            in the definition of $\M_t$.
            
            Specifically, when $K=3$, we can explicitly write out the eigenvectors of $(\E\otimes \I_K)\circ \H$ as follows.
            There is one eigenvector $\d$ corresponding to the eigenvalue
            $\alpha-\beta$.
            The eigenvectors corresponding to the eigenvalue
            $0$ take the form of
            \ba
                \tp{\bbm w_1 + w_2, 2w_1, 2w_2, 2w_3 -w_4, w_3, w_4, -w_5-2,w_5, w_6 \ebm},
            \ea
            where $\{w_i\}_{i=1}^6$ are free parameters; the eigenvectors corresponding
            to the eigenvalue $(\alpha+\beta/(K+1))/K$ take the form of
            \ba{\footnotesize
            \tp{\bbm -\xi_1 w_1+\xi_2, \xi_2 w_1 - \xi_5, \xi_3 w_1 - \xi_4,
                          -\xi_2 w_1+\xi_4, \xi_1 w_1 - \xi_3, -\xi_3 w_1 + \xi_6,
                          1-w_1, w_1, w_2\ebm},
            }\ea
            where $w_1$ and $w_2$ are free parameters and
            \begin{equation}
            \begin{gathered}
                \xi_1 = \frac{2\alpha+\beta}{3\beta}, \quad
                \xi_2 = \frac{\alpha+2\beta}{3\beta}, \quad
                \xi_3 = \frac{\alpha-\beta}{3\beta},  \\
                \xi_4 = \frac{\alpha^2+\alpha\beta+7\beta^2}{6\alpha\beta+3\beta^2}, \quad 
                \xi_5 = \frac{\alpha^2+4\alpha\beta-5\beta^2}{6\alpha\beta+3\beta^2},\quad
                \xi_6 = \frac{\alpha^2-2\alpha\beta-8\beta^2}{6\alpha\beta+3\beta^2}.
            \end{gathered}
            \end{equation}
        \ep

    \subsubsection{Proof of \Cref{thm:separation:mean}} \label{sec:app:pf:separation}
            We are now ready to prove \Cref{thm:separation:mean}.
        We first recall the theorem statement.
        \thmseparationmean*
     Recall that in the main text, we only consider the most natural (pairwise) separation \Cref{def:ps1}, which corresponds to \Cref{def:pairwise_sep}. After the prove of this theorem, We will also remark how to extend this result to other stronger definitions, especially \Cref{def:uniform_sep2}.
        \bp[Proof of \Cref{thm:separation:mean}]
        Let $\H \in \sR^{Kp \times Kp}$ be a
        symmetric positive semi-definite
        (SPD) matrix with positive diagonal entries $(d_i)_{i=1}^{Kp}$,
        recall that we denote by $\H_{ij}\in\sR^{p\times p}$ the $(i,j)$-th
        block of $\H$ for $i,j \in [K]$. We write
        $\M_t =\frac{1}{K}\left(\E_t\otimes \I_K\right)\circ \H$, where
        $(\A\circ \B)_{ij} = A_{ij}B_{ij}$ is the Hadamard product
        between two matrices of the same size. 
        
        First, assume $\alpha(t) = \alpha$ and $\beta(t) = \beta$ are constants, then
        from \Cref{thm:imodel}, we know $\E\otimes \I_p$ has
        an eigenvalue $\ld_0 = (\alpha-\beta)/K$ with multiplicity
        $p(K-1)$ and $\ld_1 = (\alpha+(K-1)\beta)/K$ with multiplicity
        $p$. Writing $\ld_{\min}$ and $\ld_{\max}$ as
        the minimum and the maximum of $\{\ld_0, \ld_1\}$
        respectively,
        by Schur's theorem (Theorem 9.J.2, \cite{marshall11}),
        we have
        \ba \label{eq:pf:sep:eig}
           \ld_{\min}\min_{1\le i \le Kp} d_i \le \ld_i\left(\M_t\right) \le
           \ld_{\max}\max_{1 \le i \le Kp} d_i,
        \ea
        where we write $\mu_i\coloneqq \ld_i\left(\M_t\right)$
        as the $i$-th largest eigenvalue of $\M_t$, and $\u_i\in\sR^{Kp}$
        the corresponding eigenvector and recall that in this case
        $\M_t = \frac{1}{K}\left(\E \otimes \I_p\right)\circ \H$ does not depend
        on time.
        We will denote by $\u_i^k \in \sR^{p}$
        the $k$-th block of $\u_i$ for $k\in[K]$, i.e.,
        $(\u_i^k)_j = (\u_i)_{k\times K + j}$.
        
        Note the eigendecomposition of $\H$ is $\sum_{i=1}^{Kp} \mu_i \u_i\tp{\u}_i$, and thus the solution to $\bar{\X}'_t = \M_t \bar{\X}$
        is
        \ba
            \bar{\X}_t = \bar{\X}_0 + \sum_{i=1}^{Kp} c_i \u_i \eu^{\mu_i t}, \quad
            \bar{\X}_0 = \sum_{i=1}^{Kp} c_i \u_i.
        \ea
        Substituting back this solution to the LE-SDE,
        we have
        \ba \label{eq:pf:sep:sde}
            \tilde{\X}^k(t) 
            &= \tilde{\X}^k(0) + \M_t\bar{\X}(t) 
            - \Exp [\tilde{\X}^k(0)]
            + \bSigma^{\frac{1}{2}}_k(t)
                \W^k(t) \\
            &=\tilde{\X}^k(0) + \sum_{i=1}^{Kp}c_i\mu_i \u_i^k \eu^{\mu_i t} 
                 - \sum_{i=1}^{Kp}c_i \u_i^k
                + \bSigma^{\frac{1}{2}}_k\W^k(t),
        \ea
        where $\bSigma^{\frac{1}{2}}_k(t)$ is the covariance
        for this class.
        By definition, to prove separation, it suffices
        to identify a direction $\bnu$ such that
        \ba
            \ip{\tilde{\X}^k(t)-\tilde{\X}^l(t)}{\bnu} > 0
        \ea
        with probability\footnote{Here the randomness comes from the dynamics as well as the random sample $\tilde{\X}^k$.} tending to $1$ as $t\to\infty$
        for any two classes $k\ne l$. Substituting
        \cref{eq:pf:sep:sde}, we have equivalently
        \ba \label{eq:app:pf:ineq}
            \ip{\sum_{i=1}^{Kp}c_i\left(\u_i^k-\u_i^l\right)
                \left(\mu_i\eu^{\mu_i t} -1\right) }{\bnu}>
            \ip{\tilde{\X}^l(0) - \tilde{\X}^k(0)}{\bnu}
                +\ip{\bSigma^{\frac{1}{2}}_l \W^l(t)
                -\bSigma^{\frac{1}{2}}_k \W^k(t) }{\bnu}.
        \ea
        By the Gaussian tail bound,
        the right-hand side of the above display 
        is $O_{\mathbb{P}}(C_0 + C_1 \sigma_{\max} \sqrt{t}) =O_{\mathbb{P}}(\sqrt{t})$ where $C_0$ and $C_1$ are
        constants that do not depend on class labels and we assume
        $\sigma_{\max} = \max_{k\in [K]} \sup_t \norm{\bSigma_k(t)}_2$
        is finite.
        Due to randomization in the initialization, with probability zero
        $\bar{\X}^k(0) = \bar{\X}^l(0)$ and thus
        \ba \label{eq:app:pf:sep:init}
            \sum_{i=1}^{Kp} c_i(\u_i^k - \u_i^l) \ne \bzero,
        \ea
        with probability one. Now \cref{eq:pf:sep:eig}
        implies that $0 < \ld_{\min} \min_i d_i \le \mu_i$,
        thus when $t$ is sufficiently large, there must exist at least
        one index $j$ with
        \ba
            c_j (\u_j^k - \u_j^l) (\mu_j\eu^{\mu_j t} -1 ) \ne 0
        \ea
        as otherwise \cref{eq:app:pf:sep:init} is contradicted.
        Thus separation takes place provided that
        \ba
            \exp\left\{\mu_j t \right\} > C \sqrt{t}
        \ea
        holds for some constant $C$ that only depends on $C_0$, $C_1$, and
        $\sigma_{\max}$. The case where $\alpha(t)$ and
        $\beta(t)$ are time-varying (and so is $\mu_i(t)$) is similar
        with $\mu_j t$ being replaced by $\int_0^t \mu_j(s)\dd s$, i.e.,
        \ba
            \exp\left\{\int_0^t \mu_j(s)\dd s \right\} > C \sqrt{t}.
        \ea
        But \cref{eq:pf:sep:eig} implies that the order of $\mu_j(t)$
        is the same as $\gamma(t)$ for all $t$, hence the above
        condition is equivalent to
        \ba \label{eq:pf:sep:cond}
            \exp\left\{\Gamma(t) \right\} > C' \sqrt{t},
        \ea
        for some constant $C^\prime$ that only depends on 
        $C_0$, $C_1$, $\sigma_{\max}$, and the specific choice
        of $\H$. Note that to prove \cref{eq:pf:sep:cond},
        it suffices to require $\gamma(t)$ has a tail that is at least $1/t$, or
        \ba
            \gamma(t) = \omega\left(\frac{1}{t}\right)
        \ea
        as $t\to\infty$. Since the order of $\mu_1(t) = \max_i \mu_i(t)$
        is also the same as $\gamma(t)$ for all $t$ in light of
        \Cref{eq:pf:sep:eig}, whenever
        \ba
            \gamma(t) = o\left(\frac{1}{t}\right)
        \ea
        as $t\to\infty$, the probability of separation tends to zero as long as $n\to \infty$ at an arbitrary rate\footnote{Note that when $n$ is finite, even if two classes of exactly the same mean (i.e. completely intervened) have a non-zero probability of separation: consider $2n$ i.i.d.~standard Wiener processes, let $n$ of them be of class $1$, and the rest class $2$. Then at any time, the probability the two classes are separated is when the largest $n$ belongs to one of the classes, which occurs with probability $\frac{2}{\binom{2n}{n}}>0$.
        }.
        Note that when $\gamma(t) = \Theta\left(\frac{1}{t}\right)$,
        the separability depends non-trivially on the constant factors,
        and the rate of $n$ and $t$ tends to infinity. 
        
        Thus we have shown that the order of $\gamma(t)$ characterizes
        a sharp phase transition in terms of separability.
        Finally, in the above we proved for each pair of classes $k$ and
        $l$, a choice $\bnu=\bnu(k,l)$ exists that ensures separation. We remark that it is possible to
        remove class-dependence on $\bnu$ in our case, and therefore achieve a stronger sense of separability: consider the 
        $(K(K-1)/2)$-by-$K$ matrix $\Psi$ 
        with $\u_{i(k,l)}^k - \u_{i(k,l)}^l$ as its rows for all $k < l$
        where $i(k,l)\in[K]$ is the index of the dominant
        eigenvalue for the separation between these two classes as
        discussed above.
        The existence of a class-independent $\bnu$ is equivalent to 
        $\Psi \bnu \ne \bzero$. This is equivalent to that the nullity of
        $\Psi$ is less than $K$,
        which is obvious since $\rank \Psi \ge 1$ almost surely by construction.
    \ep
    \paragraph{Remark.}
        In the proof of \Cref{thm:separation:mean}, we use a simple
        consequence of the Gaussian tail bound to derive the critical order
        for separation of $\gamma(t)$, which is $1/t$. This is correct
        when we deal with asymptotic pairwise separation (\Cref{def:pairwise_sep}) or  asymptotic universal separation (\Cref{def:universal_sep}), essentially
        a law of large numbers result. To obtain 
        the same guarantee for uniform pairwise separation (\Cref{def:uniform_sep}) or uniform universal separation (\Cref{def:uniform_sep2}), we need to take extra care for the order  of $\gamma(t)$. The key observation here is that if we want to guarantee a separation direction that is independent of time $t$, it is essentially an almost sure statement. To  bound the influence of the Wiener process in \eqref{eq:app:pf:ineq}, we now need to apply the law of the iterated logarithms. Recall that given 
        a standard Wiener process $W_t$,
        \ba
            \limsup_{t\to\infty} \frac{W_t}{\sqrt{2t\log\log t} } = 1,
            \quad
            \liminf_{t\to\infty} \frac{W_t}{\sqrt{2t\log\log t} } = -1,
        \ea
        hence the conditions in \Cref{thm:separation:mean} can be generalized as follows:
        \bitem
            \item if $\gamma(t) = \omega\left( \frac{1}{t\log\log t}\right)$, the features are separable in the sense of uniform asymptotic
            universal separation;
            \item if $\gamma(t) = o\left( \frac{1}{t\log\log t}\right)$, the features are \emph{not} separable in the sense of uniform asymptotic universal separation.
        \eitem 
        Note that the difference between being not uniform asymptotic universally
        separable and uniform asymptotic universally inseparable,
        the latter of which happens when
        $\gamma(t) = o\left(1/t\right)$.
        In practice, the non-uniform version of the theorem is
        powerful enough, since we do not ask to separate in a pre-specified direction.

    \subsubsection{Proof of \Cref{thm:collapse}.} \label{sec:app:pf:collapse}
        \thmcollapse*
        
        \bp[Proof of \Cref{thm:collapse}.]
            By \Cref{prop:leode:logit}, when $\gamma(t)>0$ and $B(t)<0$ eventually, the dominating component of $\bar{X}^k_t$ is $C_1 \d_k \eu^{\frac{1}{K}A(t) - \frac{1}{K}B(t)}$. As $t\to\infty$, the unit vector in the direction of $\bar{X}^k$ tends to $\d_k$, and therefore those unit vectors form
            an ETF, since $\{d_j\}_{j=1}^K$ form an ETF.
        \ep
        
\subsection{Obtaining the Hyperplane}
    In this subsection we briefly discuss several methods for obtaining
    the class-independent $\sR^1$ projection $\bnu$ asserted by \Cref{thm:separation:mean}.
    As suggested by the proof, it is not hard to see that a random Gaussian vector 
    $\bnu$ satisfies \Cref{thm:separation:mean} (with probability one)
    and it can be used to construct the map $T_{\bnu}$
    that asymptotically separate all classes almost surely.
    However, the rate of separation will depend on the choice of $\bnu$.
    In practice, we can find a good $\bnu$ by solving the following optimization problem
    \begin{align*}
        \max_{\norm{\bnu}_2 = 1}\min_{k\ne l\in [K]}\abs{\ip{ \c_k-\c_l }{\bnu}}.
    \end{align*}
    Its solution $\bnu^*$ is plausible since informally, it is the direction that can separate all the $K$ classes in the ``shortest time'' with high probability.
    To see the intuition behind this claim, 
    consider the worst case of the right hand side of
    \cref{eq:app:pf:ineq}, which is $O_\mathbb{P}(\sqrt{t})$ and 
    the least $t$ achieving this for all $k\ne l\in [K]$ requires $\bnu$ maximizes $\min_{k\ne l\in [K]}|\ip{ \c_k-\c_l }{\bnu}|$.

    \begin{remark} In practice, estimating $\c_0 + \c_l$ from several
    independent trials incur high variance, despite using a larger number
    of trials. Although the SDE is not time-reversible, since we are mainly
    interested in the asymptotic behaviour, we can use an interval $[T_1, T_2]$
    with $T_2 > T_1 \gg 0$ when the model is almost convergent to estimate
    $\bnu$ as follows:
    \ba \label{eq:app:nuopt}
        \max_{\norm{\bnu}_2 \le 1} \min_{k < l}
        \sum_{t=T_1}^{T_2} |\ip{\bar{X}^k(t) - \bar{X}^l(t)}{\bnu}|. 
    \ea
    Note that this problem is convex and easy to solve. 
    In practice, we observe directly setting
    \ba
        \bnu = \frac{\bar{\X}^k(T) - \bar{\X}^l(T)}
            {\norm{\bar{\X}^k(T) - \bar{\X}^l(T)}}
    \ea
    for a pair $k\ne l$ and a large $T$ can obtain relatively decent separation in $\sR^1$
    compared with \cref{eq:app:nuopt}, which is the case when we construct
    \Cref{fig:teaser:sep}.
    \end{remark}


\section{More Details on Experiments}\label{sec:app_sim}
    We first recall the setup of our experiments.
    We generate a dataset consisting of $K=3$ simple
    geometric shapes (\textsc{Rectangle},
    \textsc{Ellipsoid}, and \textsc{Triangle})
    that are rotated to various
    angles and applied Gaussian blurring,
    which we conveniently name Geometric-MNIST or \geomnist{} for short.
    A few samples from \geomnist{} are shown in \Cref{fig:geomnist:smp}.
    We use a varying number of training samples per class $n_{\mathrm{tr}}\in\{80, 480, 600, 4800\}$ with the validation sample per class being $n_{\mathrm{val}}=\{20, 120, 400,1200\}$. We also pollute each label class by randomly choosing $\perr\cdot n_{\mathrm{tr}}$ samples to flip the label to another class.
    (uniform across all other classes). In this setup, we fix
    $n_{\mathrm{tr}}=n_{\mathrm{val}}=500$ and set
    $\perr\in\{0.1,0.2,\ldots, 0.8\}$.
    In addition to \geomnist{},
    we use \cifar-10 (\cite{CIFAR}, denoted
    by \cifar) for a more realistic scenario with $5000$ training
    samples and $1000$ validation samples per class.
    We vary the total number of classes $K\in [2,3]$.
    Variants of the \alexnet{} model (\cite{krizhevsky2012imagenet})
    are used, which consists
    of two convolutional layers and three fully-connected layers
    activated by the \texttt{ReLU} function.

\subsection{Estimation Procedures} \label{sec:app_sim:estimation}
    We will discuss here how we estimate several quantities
    in \eqref{sec:experiments}, including local elasticity strengths
    $\alpha(t)$ and $\beta(t)$ in the \mI{} and the \mH{}, and tail
    indices $r_{\alpha}$ and $r_{\beta}$.

    \subsubsection{Estimation of Integrated Local Elasticity Strengths $\hat{A}(t)$ and
    $\hat{B}(t)$}
        Recall that in \Cref{eq:estimation:AB} we give the following
        formulae for estimating $A(t)$ and $B(t)$:
        \ba
        (\text{\mI}) &\quad
        \begin{cases}
        \hat{A}(t) &= \avg \avg_k{
            \log \abs{\frac{\check{\X}(\bar{\X}^k - \check{\X})^{K-1}}{\c_0 \c_k^{K-1}}}
        }, \\
        \hat{B}(t) &=
        -\avg \avg_k{
            \log \abs{\frac{\c_0}{\c_k} \frac{\bar{\X}^k - \check{\X}}{\check{\X}}}
        }, 
        \end{cases}
        \quad \check{\X}_t \coloneqq \avg_l \bar{\X}^l_t, \\
        (\text{\mH}) &\quad
        \begin{cases}
        \hat{A}(t) &= A'(t)+2B'(t), \\
        \hat{B}(t) &= 2(B'(t)-A'(t)), \\
        \end{cases} \quad
        \begin{cases}
            A'(t) &\coloneqq \log \abs{\left\langle\tp{\bar{\X}}\v_1 -1\right\rangle}, \\
            B'(t) &\coloneqq \log \abs{\left\langle
            \tp{\bar{\X}}\left(\v_2 - \frac{4}{3}\v_1\right)
            \right\rangle},
        \end{cases}
        \ea
        where
         \ba
            \v_1 = \frac{1}{4}\tp{\bbm 1 , -1 , -1 , -1 , 1 , -1, -2, -2, 0\ebm},\quad
            \v_2 = \frac{1}{3}\tp{\bbm 2 ,-1 ,-1 , -1 , 2 , -1 , 0 , 0 , 0\ebm}.
        \ea
        We will now explain how it is done.
        \paragraph{Estimation in \mI.}
            Recall from \Cref{thm:imodel}, the per-class means
            $\bar{\X}_t$
            solve the LE-ODE \eqref{eq:leode:sol:id} under the \mI{}
            as
            \ba
            \bar{\X}(t) =  \c \eu^{\frac{1}{K}A(t) - \frac{1}{K}B(t)} +  
            \left(\bone_K\otimes \c_0\right) \eu^{\frac{1}{K}A(t) + \frac{K-1}{K}B(t)},
            \ea
            where $\c=(\c_k)_{k=1}^K\in\sR^{Kp}$ and $\c_0\in\sR^p$ are constants with $\sum_{k=1}^K \c_k = \bzero$ and $\bar{\X}(0) = \c + \c_0$. The specific structure of this solution
            implies that
            \ba
                \check{\X}_t \coloneqq \avg_l \bar{\X}^l_t = \frac{1}{K}\sum_{l=1}^K \bar{\X}^l_t = \c_0 \eu^{\frac{1}{K}A(t) + \frac{K-1}{K}B(t)} \in \sR^K,
            \ea
            and thus for all $k\in[K]$,
            \ba
                \bar{\X}^k_t - \check{\X}_t
                = \c_k  \eu^{\frac{1}{K}A(t) - \frac{1}{K}B(t)}.
            \ea
            It follows that
            \ba
                \abs{\frac{\c_0}{\c_k} \frac{\bar{\X}^k - \check{\X}}{\check{\X}}}
                = \abs{\bone_K \eu^{B(t)}}, \quad
                \abs{\frac{\c_0}{\c_k} \frac{\bar{\X}^k - \check{\X}}{\check{\X}}}
                = \abs{\bone_K \eu^{A(t)}},
            \ea
            for all $k\in[K]$. Taking logarithm and averaging over $K$ classes and $K$
            coordinate, we have the estimation equation of
            \mI{} in \cref{eq:estimation:AB}.
            
        \paragraph{Estimation in \mH.}
            Recall from \Cref{thm:hmodel}, the per-class means
            $\bar{\X}_t$
            solve the LE-ODE \eqref{eq:leode:sol:H} 
            under \mH{} as
            \ba
            \bar{\X}(t) =  \c_0
                + C_1 \d \eu^{\frac{1}{K}A(t) - \frac{1}{K}B(t)}+
            \left(\sum_{l=1}^{K-1} C_{2l} \f_l\right)
                \eu^{\frac{1}{K}A(t) + \frac{1}{K(K-1)}B(t)},
            \ea
            where $\c_0$ is a constant vector with $K(K-1)$ free parameters;
            $\f_l$'s are eigenvectors corresponding to the eigenvalue
            $(A(t) + B(t)/(K-1))/K$.
            Although exact solutions can be obtained for general $K$, they
            are overly complicated thus we will restrict our attention to the
            case where $K=3$.
            Define
            \ba
            \v_1 = \frac{1}{4}\tp{\bbm 1 , -1 , -1 , -1 , 1 , -1, -2, -2, 0\ebm},\quad
            \v_2 = \frac{1}{3}\tp{\bbm 2 ,-1 ,-1 , -1 , 2 , -1 , 0 , 0 , 0\ebm},
            \ea
            from the proof for \Cref{thm:hmodel} (\Cref{sec:app:pf:hmodel}),
            we immediately have
            \ba
                \tp{\v}_1 \f_j = 0, \quad j=1,2, \quad
                \tp{\v}_2 \c_0 = 0, \quad
                \tp{\bone}_K \d = 0,
            \ea
            and
            \ba
                \tp{\v}_1 \c_0 = 1-\frac{w_1+w_2+w_3}{4},  \quad
                \tp{\v}_1 \d = 1, \quad
                \tp{\v}_2 \f_j = 1, \quad
                \tp{\v}_2 \d = \frac{4}{3}.
            \ea
            Recall that we define $\left\langle\X(t)\right\rangle\coloneqq \X(t) / \X(0)$,
            with vector-division interpreted as elementwise division, writing
            $C_3=w_1 + w_2 + w_3$ with $w_i$'s being defined
            in \Cref{sec:app:pf:hmodel}, we have
            \ba
                \begin{cases}
                \tp{\bar{\X}_t}\v_1
                &= C_1 \eu^{\frac{1}{K}A(t) - \frac{1}{K}B(t)} + 1 - \frac{C_3}{4}, \\
                \tp{\bar{\X}_t}\v_2
                &= \frac{3C_1}{4}  \eu^{\frac{1}{K}A(t) - \frac{1}{K}B(t)} 
                    +(C_{21} + C_{22})\eu^{\frac{1}{K}A(t) + \frac{1}{K(K-1)}B(t)},
                \end{cases}
            \ea
            and thus
            \ba
                \begin{cases}
                \left\langle\tp{\bar{\X}_t}\v_1-1 \right\rangle
                &= \frac{C_1\eu^{\frac{1}{K}A(t) - \frac{1}{K}B(t)} - \frac{C_3}{4}}{C_1 - \frac{C_3}{4}} \\
                \left\langle\tp{\bar{\X}_t}\left(\v_2-\frac{4}{3}\v_1\right)\right\rangle
                &= \frac{(C_{21}+C_{22})\eu^{\frac{1}{K}A(t) + \frac{1}{K(K-1)}B(t)}
                    + \frac{3C_3}{16}}{C_{21} + C_{22} + \frac{3C_3}{16}}.
                \end{cases}
            \ea
            We define
            \ba
                \begin{cases}
                A'(t) &= \log\abs{\left\langle\tp{\bar{\X}_t}\v_1-1 \right\rangle},\\
                B'(t) &= \log\abs{\left\langle\tp{\bar{\X}_t}\left(\v_2-\frac{4}{3}\v_1\right)\right\rangle},
                \end{cases}
            \ea
            when $t$ is sufficiently large such that
            \ba
                \eu^{\frac{1}{K}A(t) - \frac{1}{K}B(t)} \gg \frac{C_3}{4C_1}, \quad
                \eu^{\frac{1}{K}A(t) + \frac{1}{K(K-1)}B(t)}
                    \gg \frac{3C_3}{16(C_{21}+C_{22})},
            \ea
            we have approximately
            \ba
                A'(t) \approx \frac{1}{K}A(t) - \frac{1}{K}B(t), \quad
                B'(t) \approx \frac{1}{K}A(t) + \frac{1}{K(K-1)}B(t),
            \ea
            where $K=3$. Hence $A(t)$ and $B(t)$ can be recovered by
            \ba \label{eq:app:hatAB}
                \hat{A}(t) = A'(t) + 2B'(t), \quad
                \hat{B}(t) = 2(B'(t) - A'(t)).
            \ea
            Although \cref{eq:app:hatAB} is only an approximation that 
            is precise only when $t$ is large, we will nonetheless use \cref{eq:app:hatAB}
            for estimation in the \mL{} for all $t$.

    \subsubsection{Estimation of $\alpha(t)$ and $\beta(t)$}
\begin{figure*}
        \centering
        \begin{subfigure}[t]{0.25\textwidth}
    	    \centering
    	    \includegraphics[width=\linewidth]{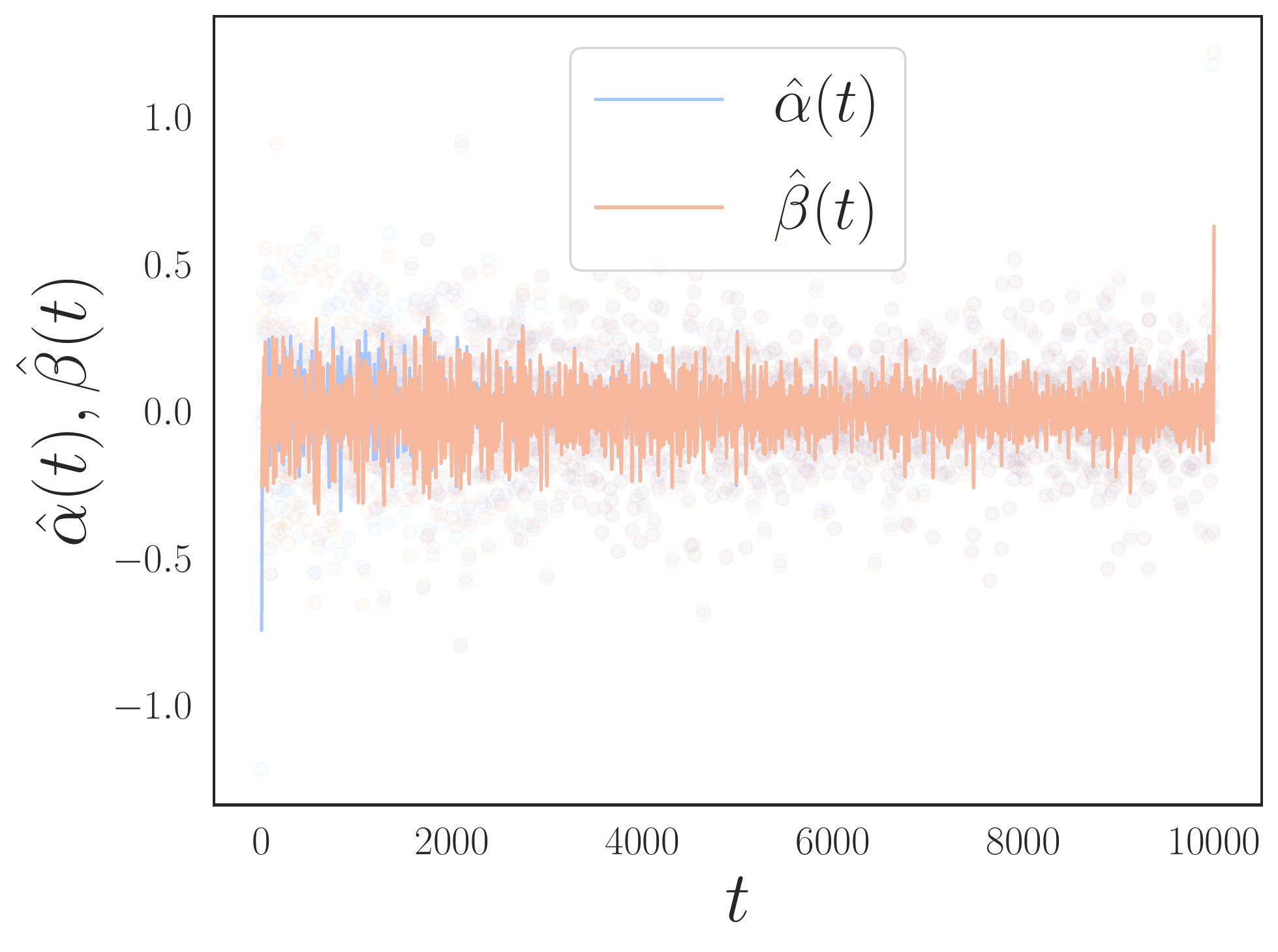}
    	    \caption{$\omega=11$.} \label{fig:app:salgov:w11}
    	\end{subfigure}%
    	\begin{subfigure}[t]{0.25\textwidth}
    	    \centering
    	    \includegraphics[width=\linewidth]{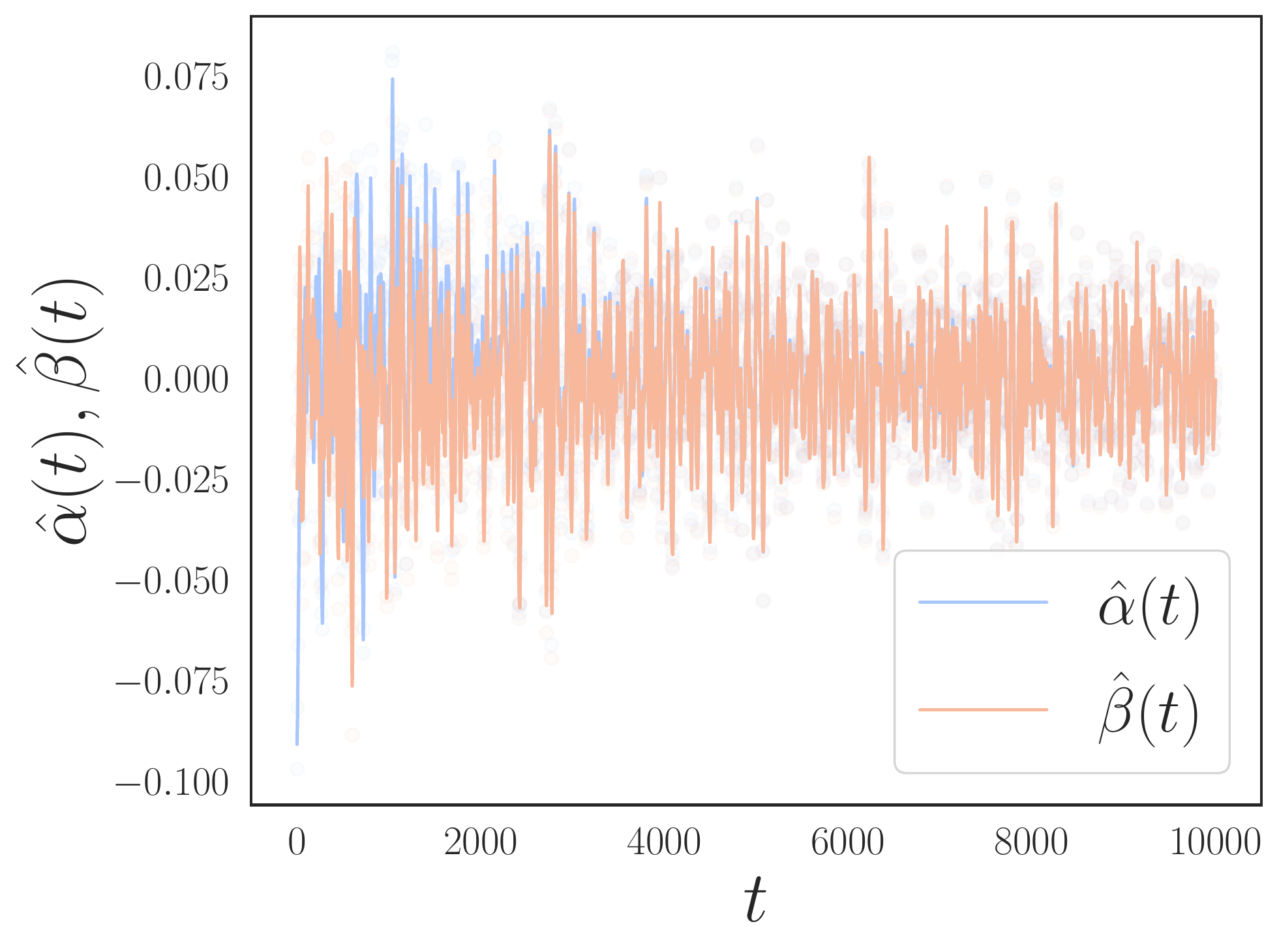}
    	    \caption{$\omega=51$.} \label{fig:app:salgov:w31}
    	\end{subfigure}%
    	\begin{subfigure}[t]{0.25\textwidth}
    	    \centering
    	    \includegraphics[width=\linewidth]{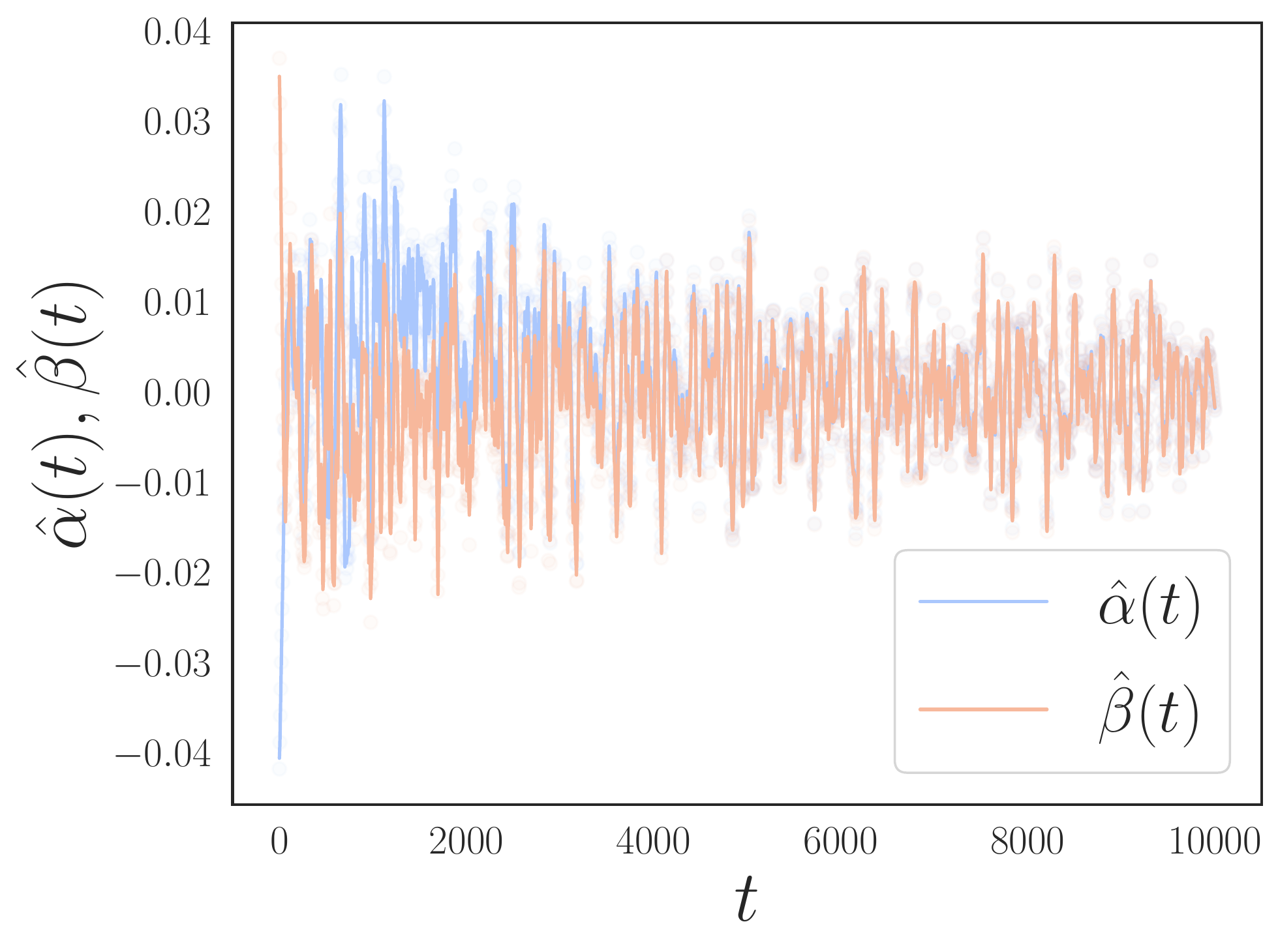}
    	    \caption{$\omega=101$.} \label{fig:app:salgov:w101}
    	\end{subfigure}%
    	\begin{subfigure}[t]{0.25\textwidth}
    	    \centering
    	    \includegraphics[width=\linewidth]{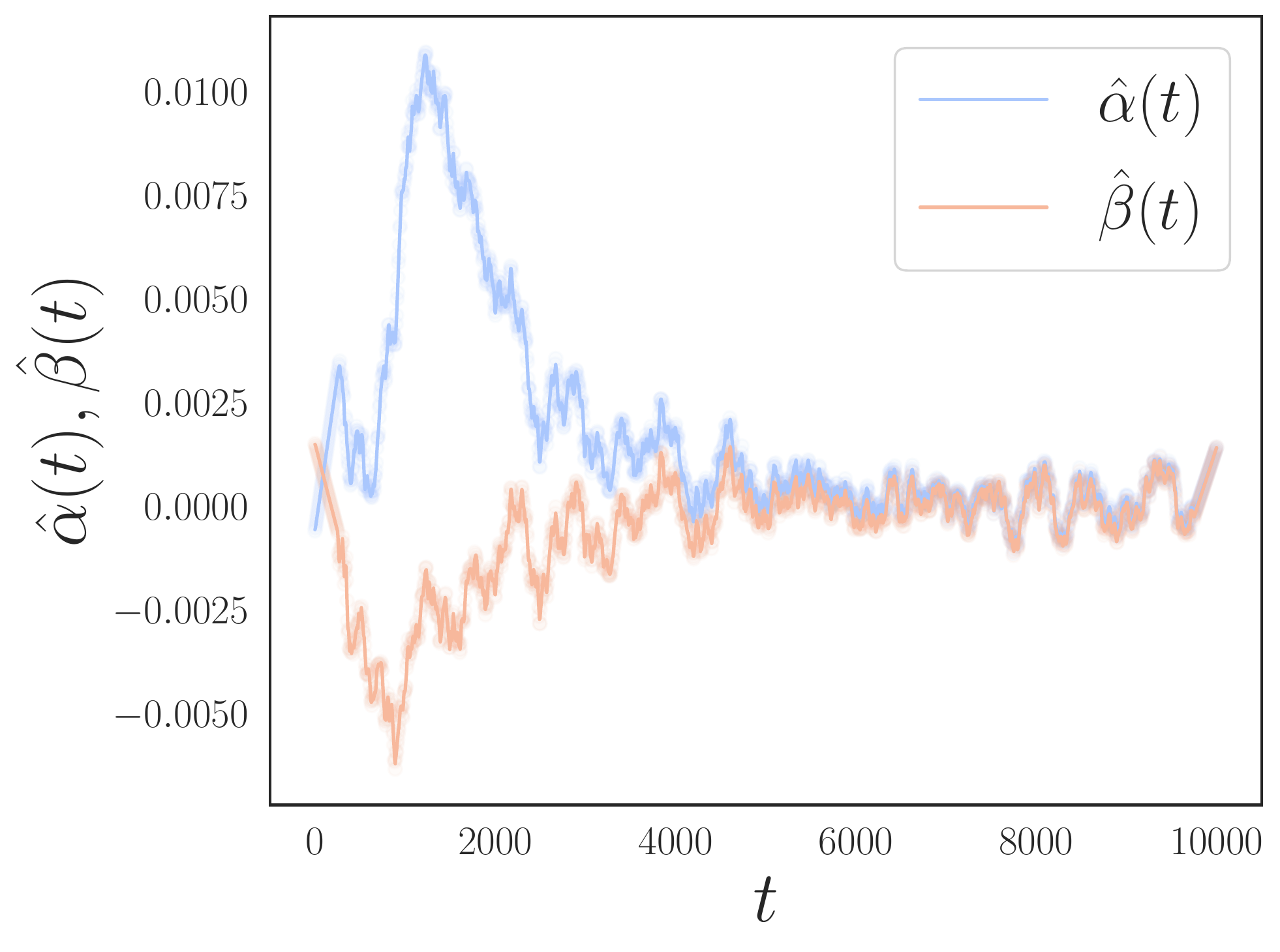}
    	    \caption{$\omega=501$.} \label{fig:app:salgov:w501}
    	\end{subfigure}%
            \caption{\textbf{Effect of the window size
            $\omega$ in the Savitzky–Golay
            filter.} These estimations are performed on the \geomnist{}
            dataset with $\perr=0$ and $n_{\mathrm{tr}}=3000$.}
            \label{fig:app:savgol}
    \end{figure*}

        Once we have estimates for $A(t)$ and $B(t)$, namely
        $\hat{A}(t)$ and $\hat{B}(t)$, we may numerically differentiate
        these estimates to obtain $\hat{\alpha}(t)$ and $\hat{\beta}(t)$.
        Although the composition of finite difference quotient and 
        moving average
        yields visibly well results,
        we shall use the well-established Savitzky–Golay filter
        for this purpose. There is a window size parameter $\omega$
        in this filter which roughly corresponds to the window
        size in moving averages: a smaller $\omega$ preserves
        more fluctuations in the original data and a larger $\omega$
        smooths the data more. As a rule of thumb, we test on a set
        of different values for $\omega$ and choose one that is both informing
        and not losing too much detail in our presentations.
        Specifically, in the main paper,
        we use $\omega=191$ for experiments on \geomnist{} and
        $\omega=551$ for those on \cifar{}. In the case of simulating LE-ODE
        solutions, we chose $\omega=21$ to preserve  finer details.
        We show in \Cref{fig:app:savgol} estimated $\hat{\alpha}(t)$
        and $\hat{\beta}(t)$  trained on \geomnist{} with $\perr=0$
        and $n_{\mathrm{tr}} = 3000$ samples per class under various
        window sizes $\{11,51,101,301\}$ for the Savitzky–Golay filter.
        Note that the general trend is not discovered until
        the window size $\omega$ is reasonably large.
        
    \subsubsection{Estimation of Tail Index}
        We are interested in the tail behavior of $\alpha(t)$ and $\beta(t)$.    Taking $\alpha(t)$ as an example, suppose
        \ba
            \alpha(t) \sim \frac{\alpha_0}{(1+t)^r},
        \ea
        for some $r$, which is the tail index of $\alpha(t)$, we have
        \ba
            A(t) =\begin{cases} 
            \frac{\alpha_0(1+t)^{1-r}}{1-r}, & 0 < r < 1, \\
            \alpha_0\log(1+t), & r = 1.
        \end{cases}
        \ea
        Hence with $t$ being sufficiently large, we can estimate $r$
        using
        \ba \label{eq:app:tailest}
            \frac{\log A(t)}{\log(1+r)}
            =(1-r) + \frac{\log \alpha_0}{\log(1+t)} - \frac{\log(1-r)}{\log(1+t)}
            \approx 1-r,
        \ea
        and $\hat{r}_{\alpha} = 1- \avg_{t\ge T_0} \log A(t)/ \log(1+t)$
        for some sufficiently large $T_0$. We estimate $\hat{r}_{\beta}$ similarly.
        Although this estimator suffers from large bias when the true model
        has a constant offset, i.e., when $\alpha(t) \sim \alpha_1 + \alpha_0/(1+t)^r$, we choose it over other estimators based on $\alpha(t)\sim \alpha_1 +\alpha_0/(1+t)^r$ as it is simpler and it is
        directly based on the integrated local elasticity strength $A(t)$ without
        the need to perform numerical differentiation beforehand.

        \begin{figure*}
        \centering
        \begin{subfigure}[t]{0.45\textwidth}
    	    \centering
    	    \includegraphics[width=\linewidth]{figs/geom_tailidx_I}
    	    \caption{\mI{}.} \label{fig:app:perr:I}
    	\end{subfigure}\hfill
    	\begin{subfigure}[t]{0.45\textwidth}
    	    \centering
    	    \includegraphics[width=\linewidth]{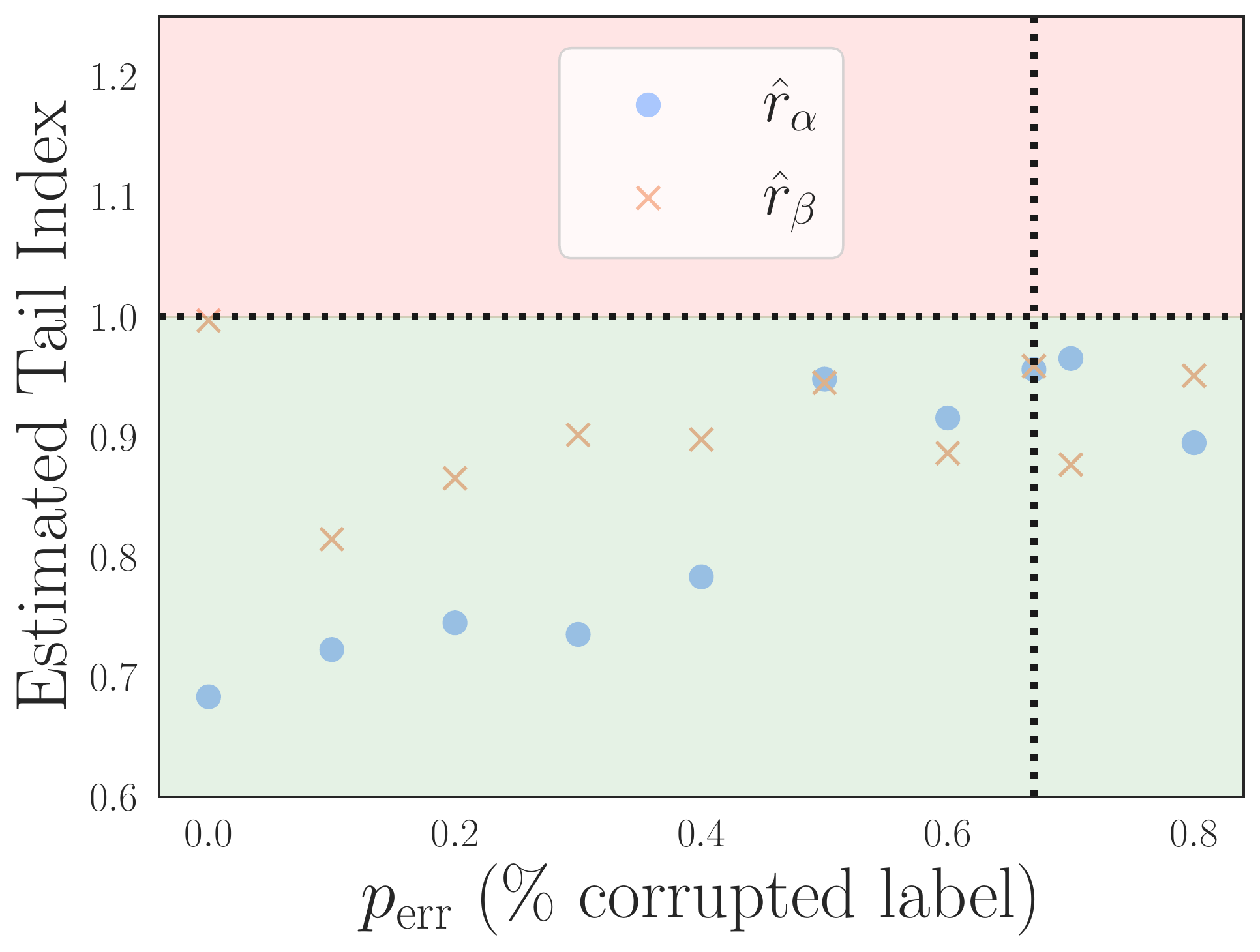}
    	    \caption{\mH{}.} \label{fig:app:perr:L}
    	\end{subfigure}%
            \caption{\textbf{Estimated tail indices versus label corruption ratio $\perr$.} The tail indices are estimated using \cref{eq:app:tailest}
            with $\hat{\alpha}(t)$ and $\hat{\beta}(t)$
            estimated via (a) \mI{}, and (b) \mH{}.
            Although the case for the \mH{} does not exhibit
            a clear phase transition, we note around $\perr\approx 2/3$,
            the tail index of $\hat{\beta}(t)$ begins to dominate
            that of $\hat{\alpha}(t)$.}
            \label{fig:app:perr}
    \end{figure*}

\subsection{More Results from Experiments} \label{sec:app_sim:results}
\paragraph{Effects of Training Sample Size $n$.}

\begin{figure*}
    \centering
    \begin{subfigure}[t]{0.25\textwidth}
	    \centering
	    \includegraphics[width=\linewidth]{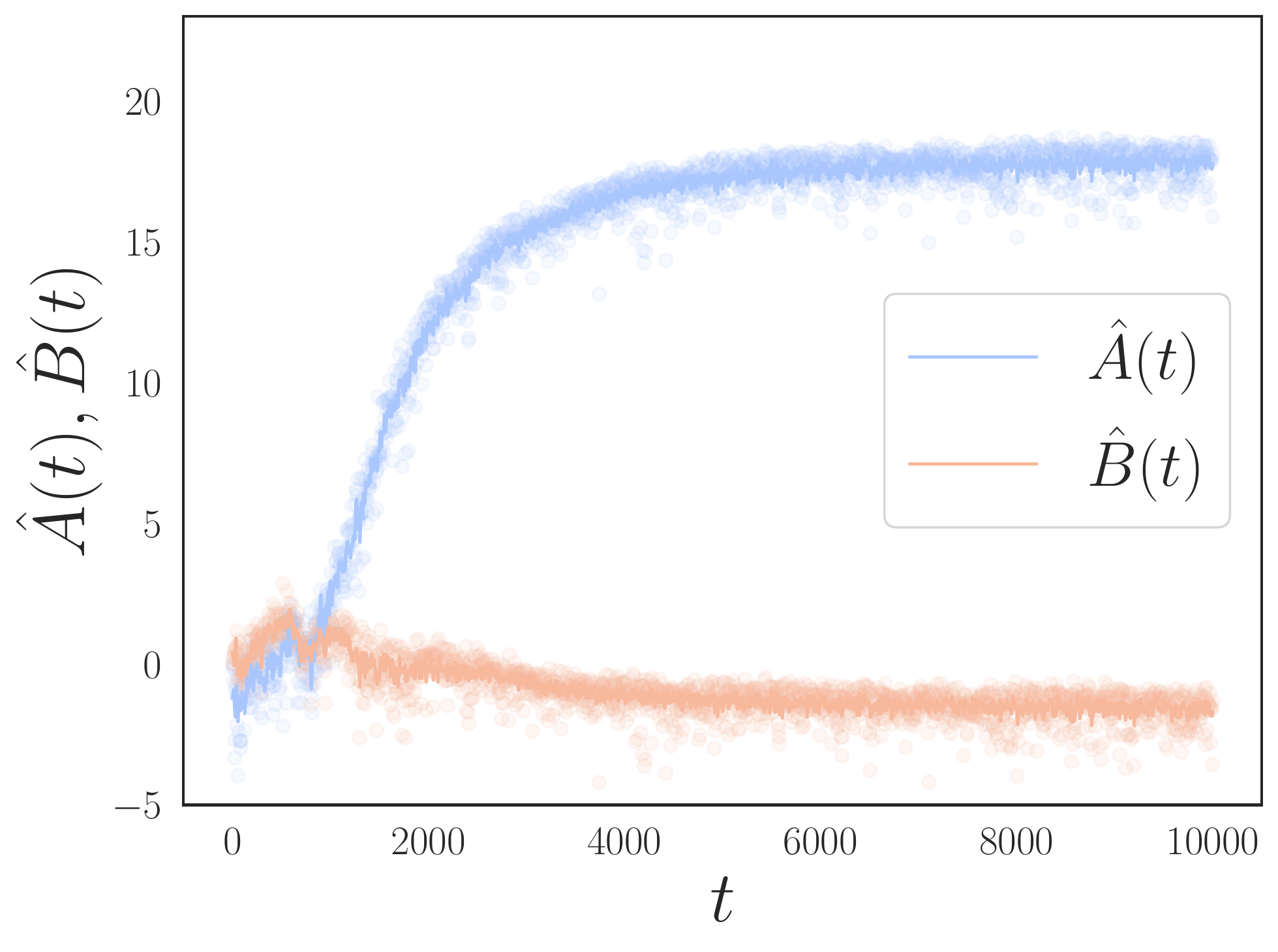}
	    \caption{$n=80$ (\mI).} \label{fig:fig:supp:geomN:AB:I:100}
	\end{subfigure}%
	\begin{subfigure}[t]{0.25\textwidth}
	    \centering
	    \includegraphics[width=\linewidth]{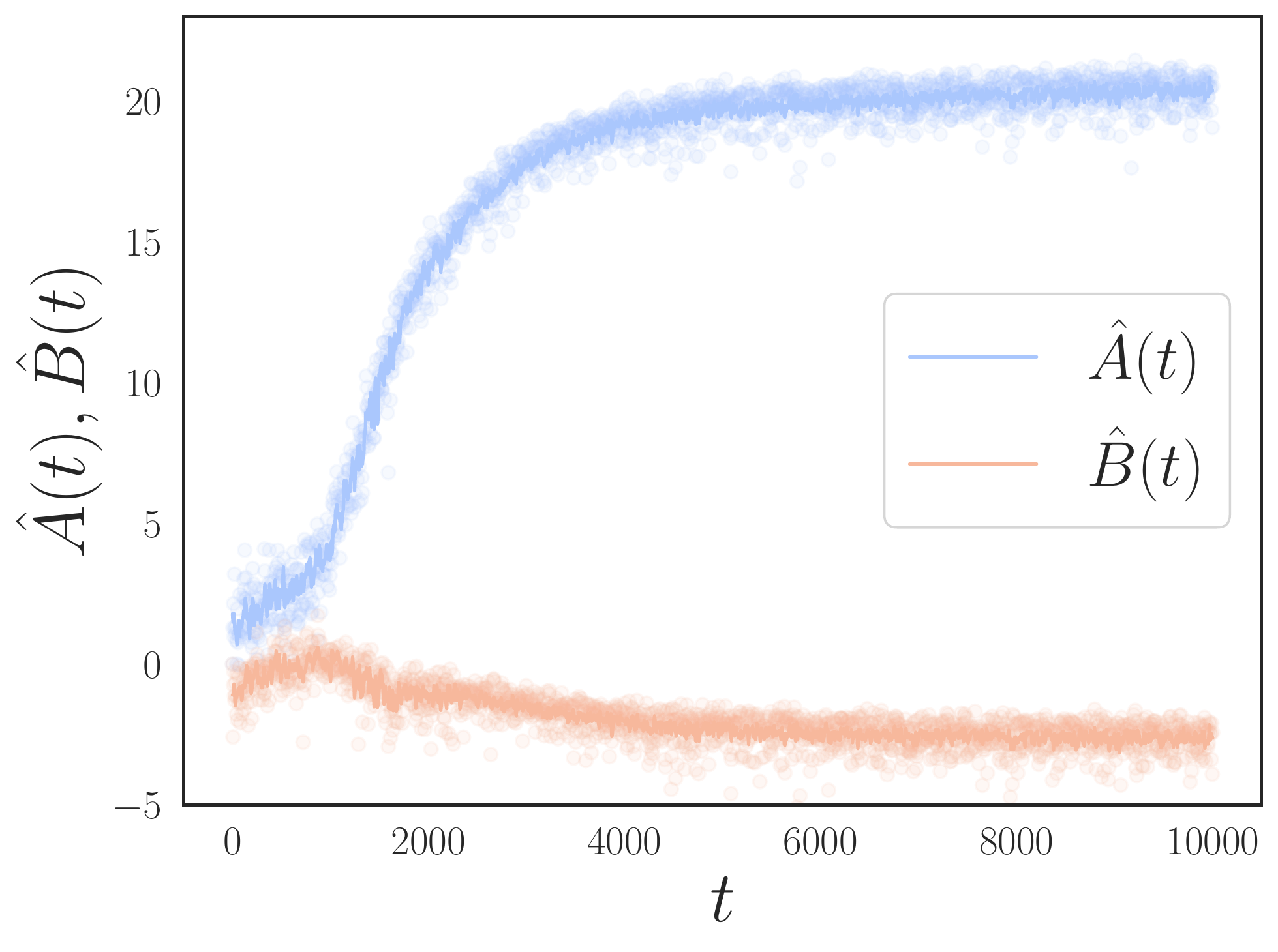}
	    \caption{$n=480$ (\mI).} \label{fig:fig:supp:geomN:AB:I:600}
	\end{subfigure}%
	\begin{subfigure}[t]{0.25\textwidth}
	    \centering
	    \includegraphics[width=\linewidth]{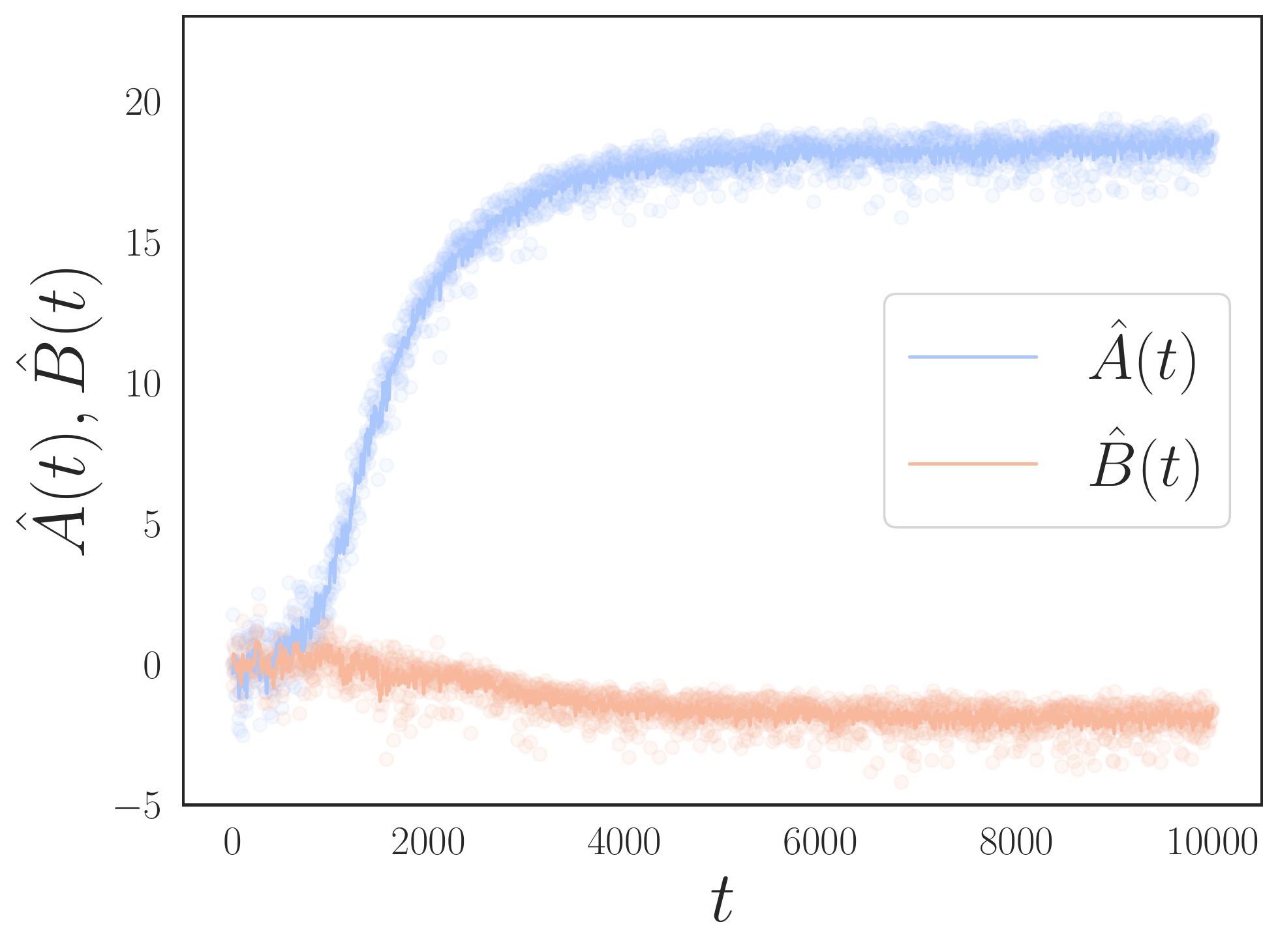}
	    \caption{$n=600$ (\mI).} \label{fig:fig:supp:geomN:AB:I:1000}
	\end{subfigure}%
	\begin{subfigure}[t]{0.25\textwidth}
	    \centering
	    \includegraphics[width=\linewidth]{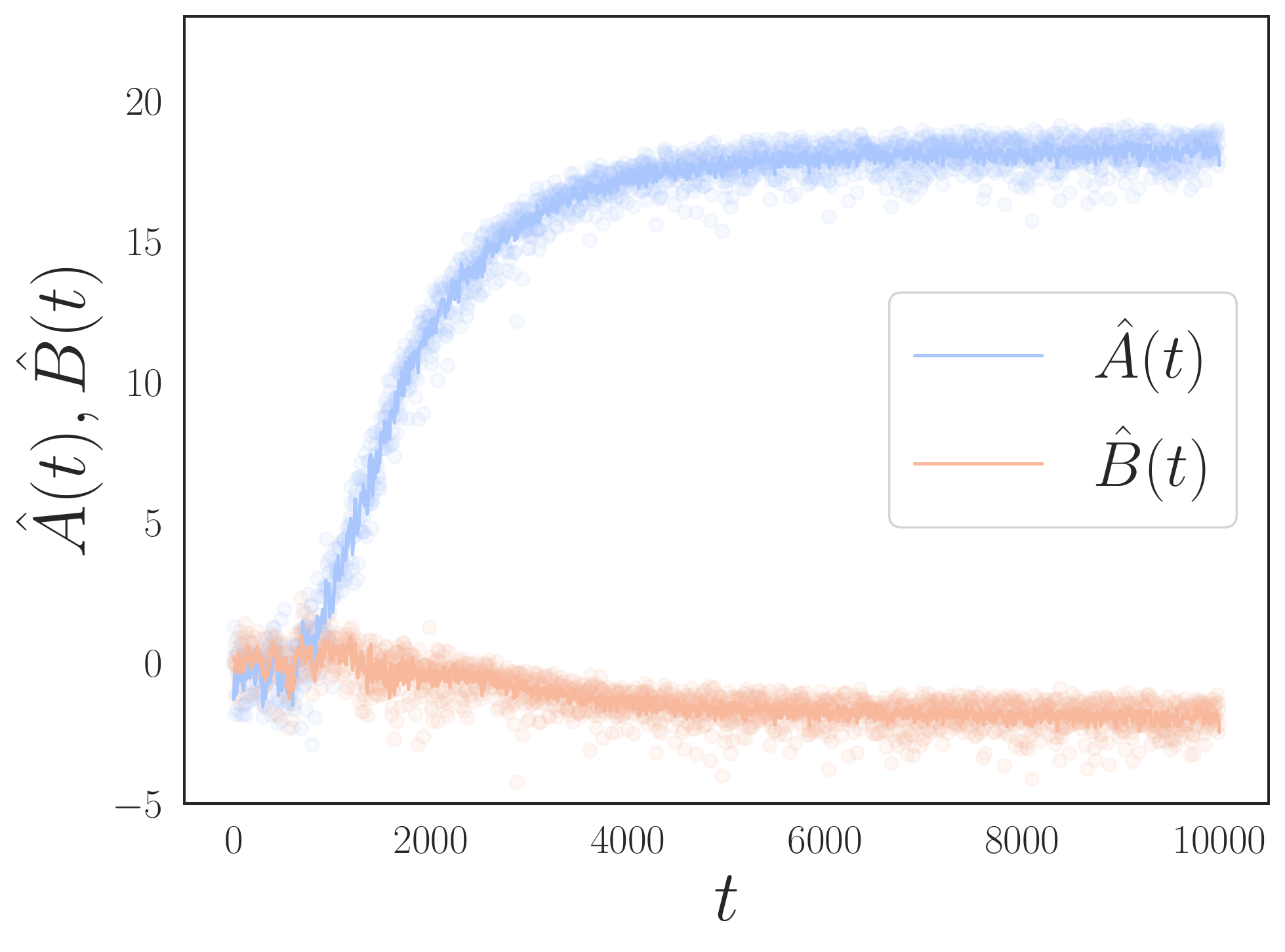}
	    \caption{$n=4800$ (\mI).} \label{fig:fig:supp:geomN:AB:I:6000}
	\end{subfigure}\\
	\begin{subfigure}[t]{0.25\textwidth}
	    \centering
	    \includegraphics[width=\linewidth]{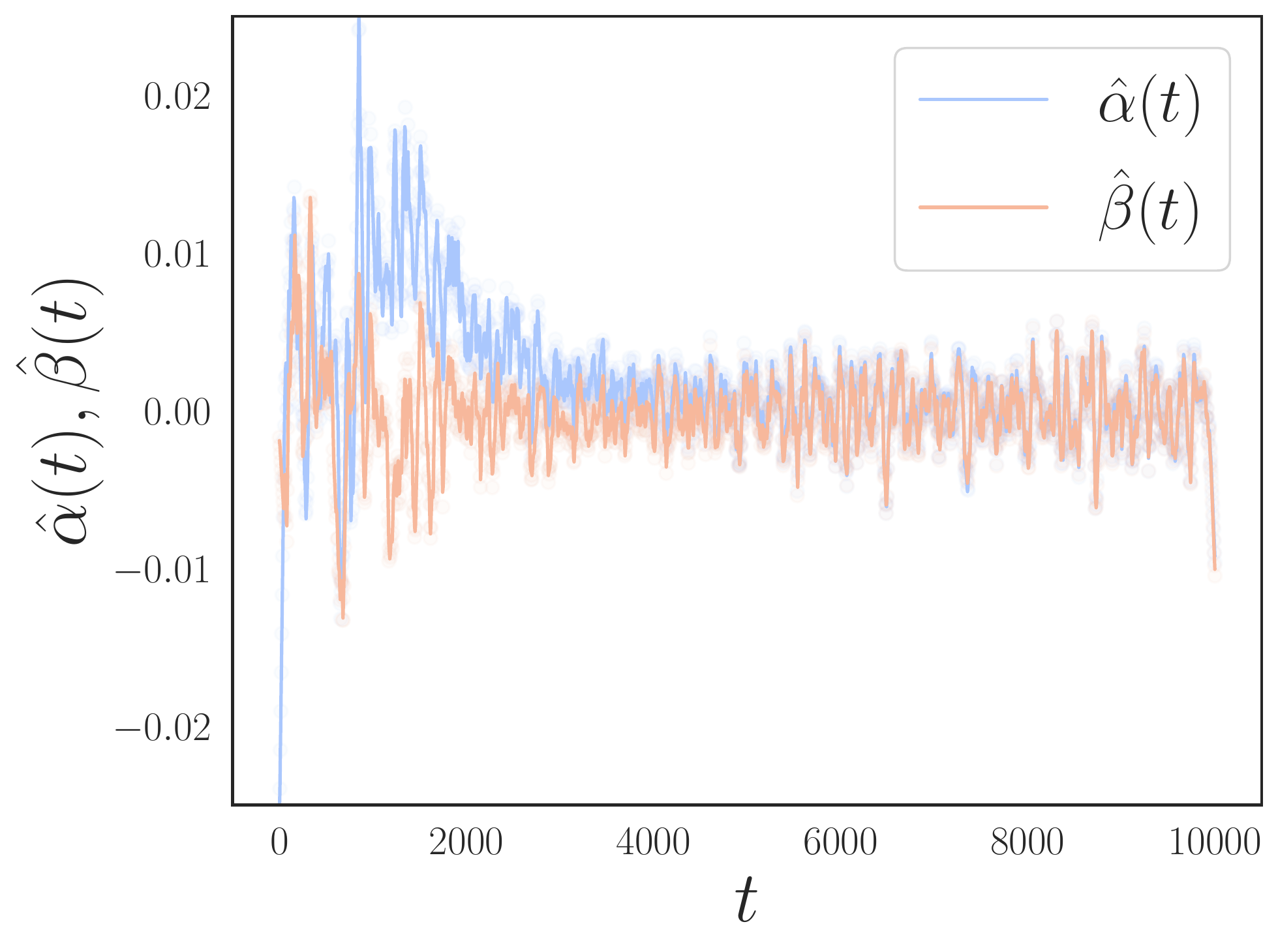}
	    \caption{$n=80$ (\mI).} \label{fig:fig:supp:geomN:dAB:I:100}
	\end{subfigure}%
	\begin{subfigure}[t]{0.25\textwidth}
	    \centering
	    \includegraphics[width=\linewidth]{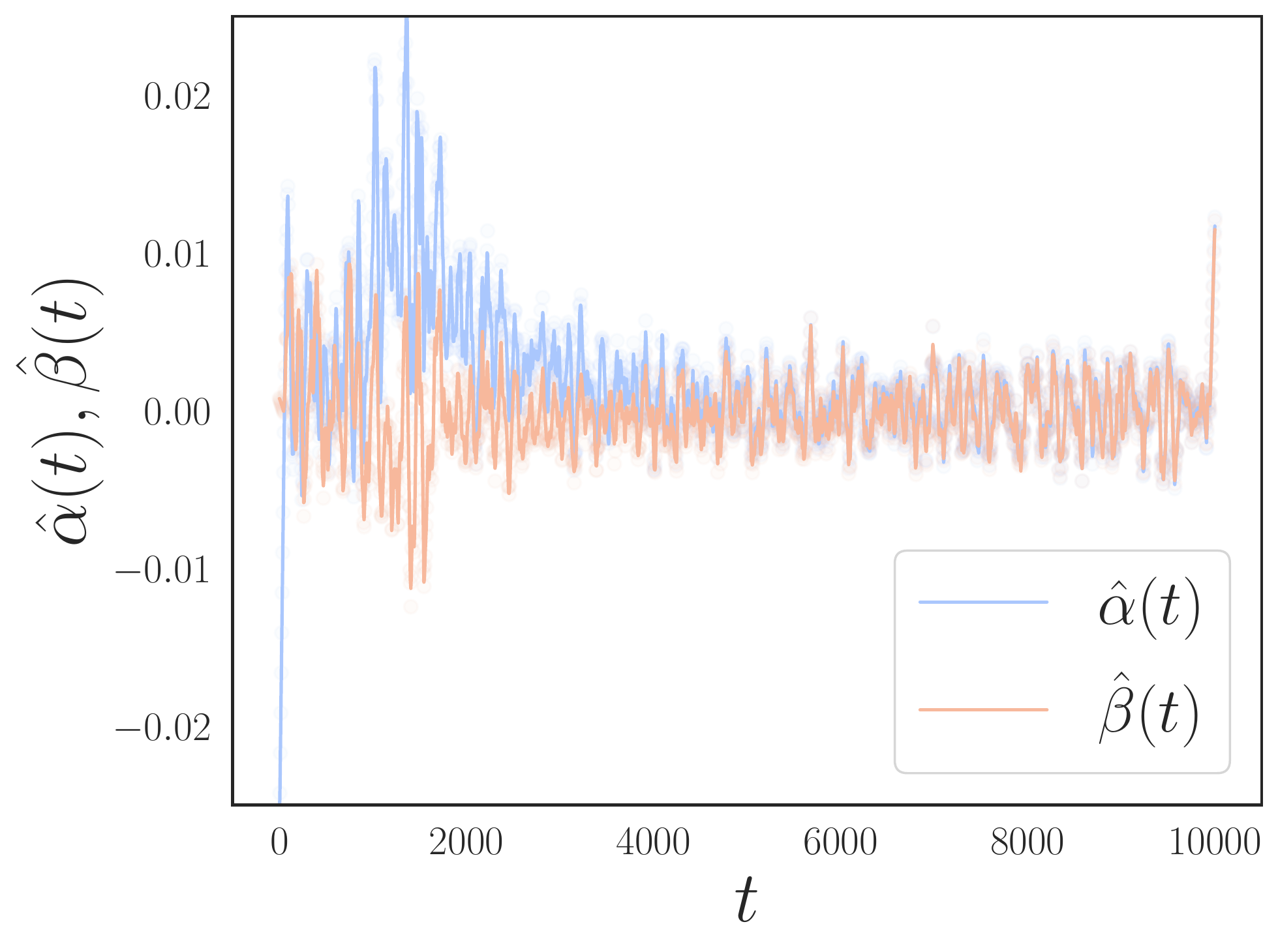}
	    \caption{$n=480$ (\mI).} \label{fig:fig:supp:geomN:dAB:I:600}
	\end{subfigure}%
	\begin{subfigure}[t]{0.25\textwidth}
	    \centering
	    \includegraphics[width=\linewidth]{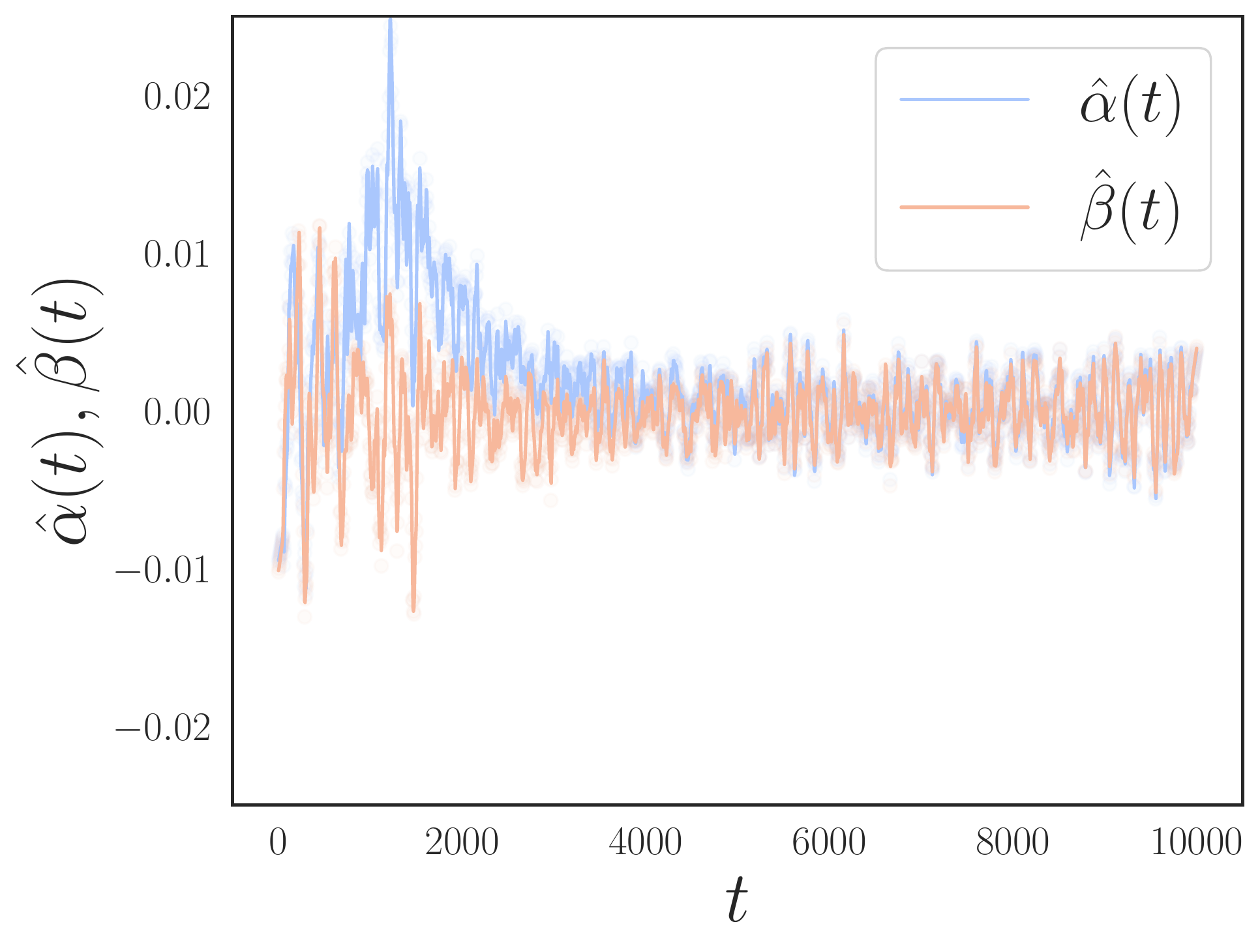}
	    \caption{$n=600$ (\mI).} \label{fig:fig:supp:geomN:dAB:I:1000}
	\end{subfigure}%
	\begin{subfigure}[t]{0.25\textwidth}
	    \centering
	    \includegraphics[width=\linewidth]{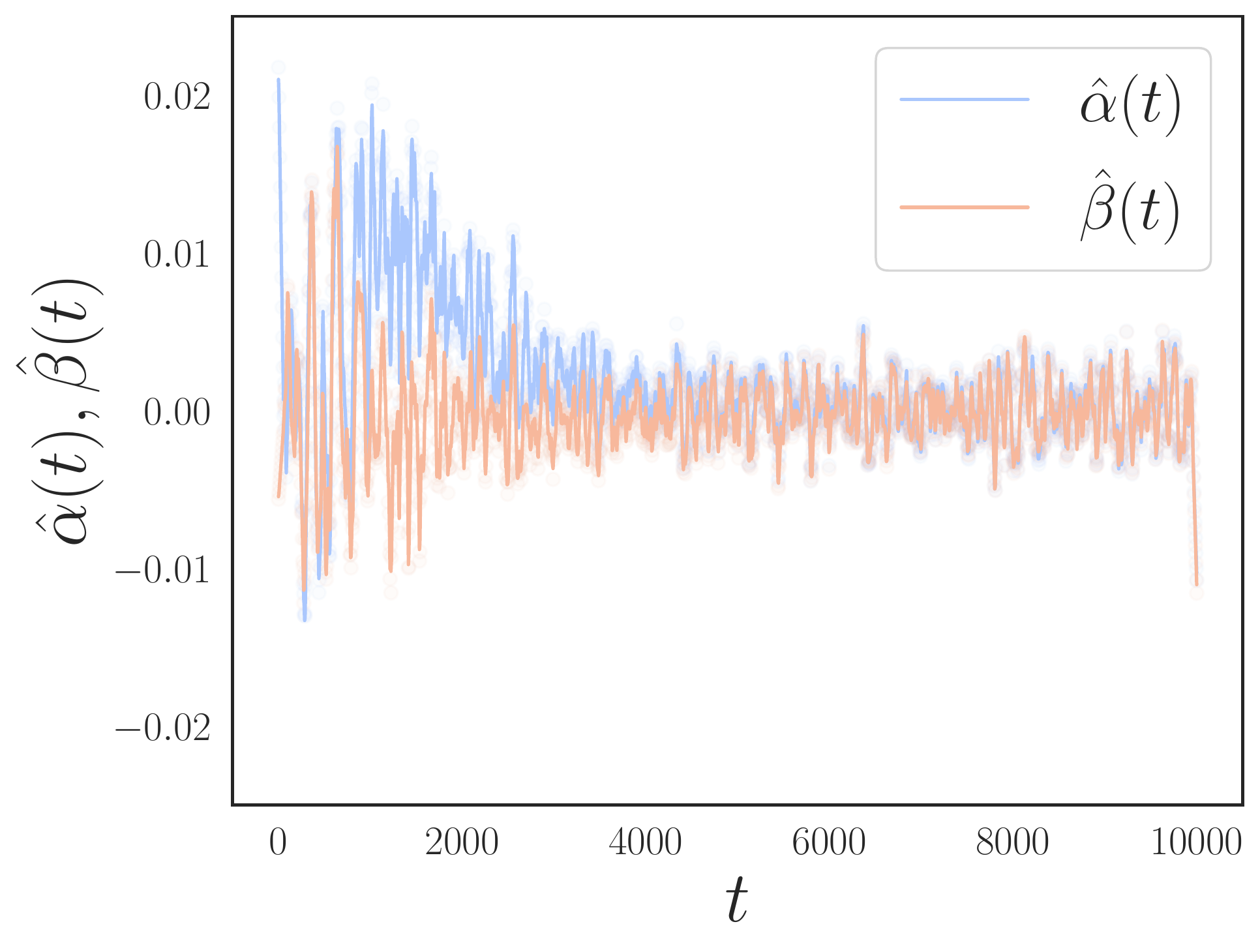}
	    \caption{$n=2400$ (\mI).} \label{fig:fig:supp:geomN:dAB:I:6000}
	\end{subfigure}\\
	\begin{subfigure}[t]{0.25\textwidth}
	    \centering
	    \includegraphics[width=\linewidth]{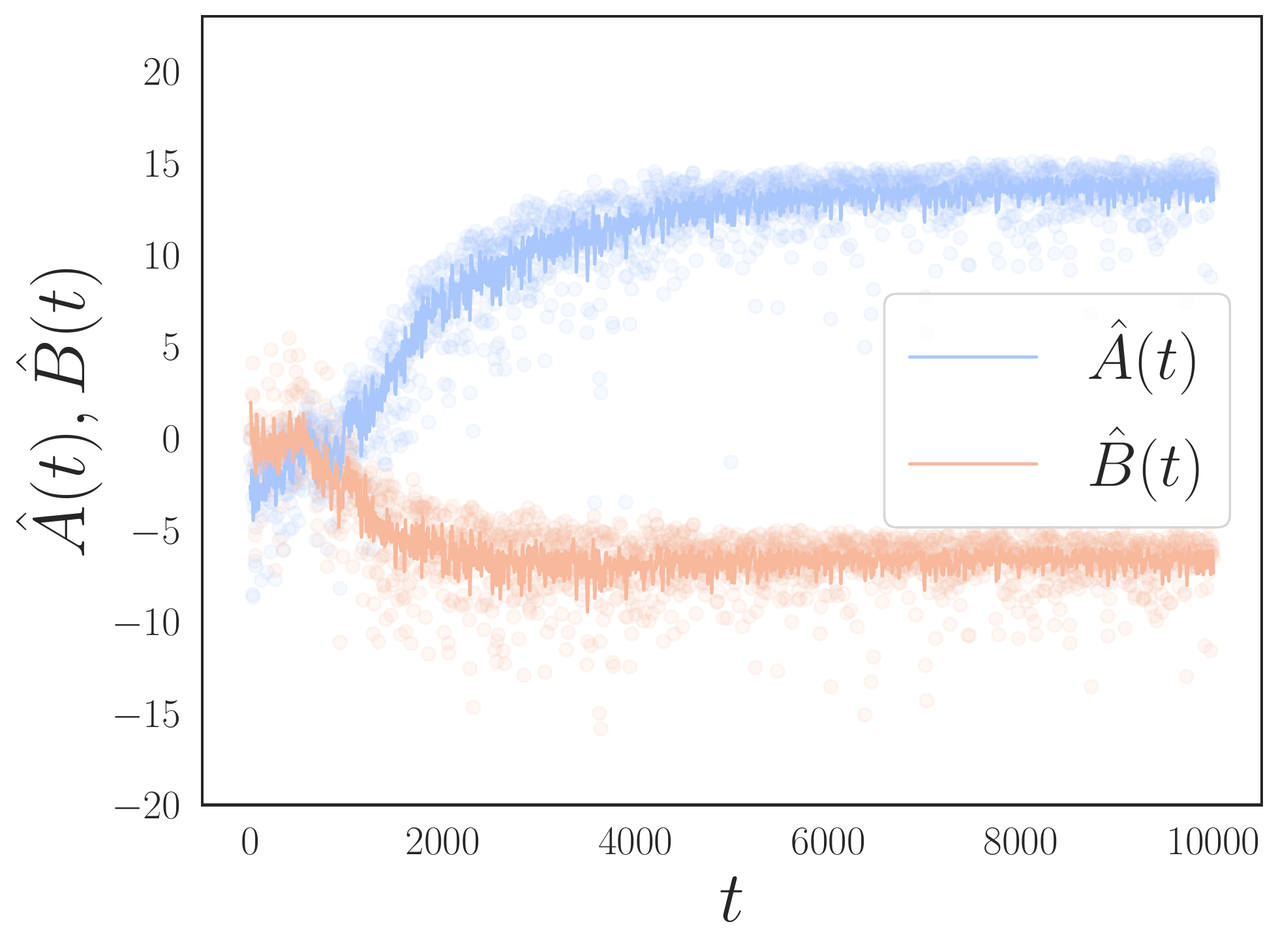}
	    \caption{$n=80$ (\mL).} \label{fig:fig:supp:geomN:AB:L:100}
	\end{subfigure}%
	\begin{subfigure}[t]{0.25\textwidth}
	    \centering
	    \includegraphics[width=\linewidth]{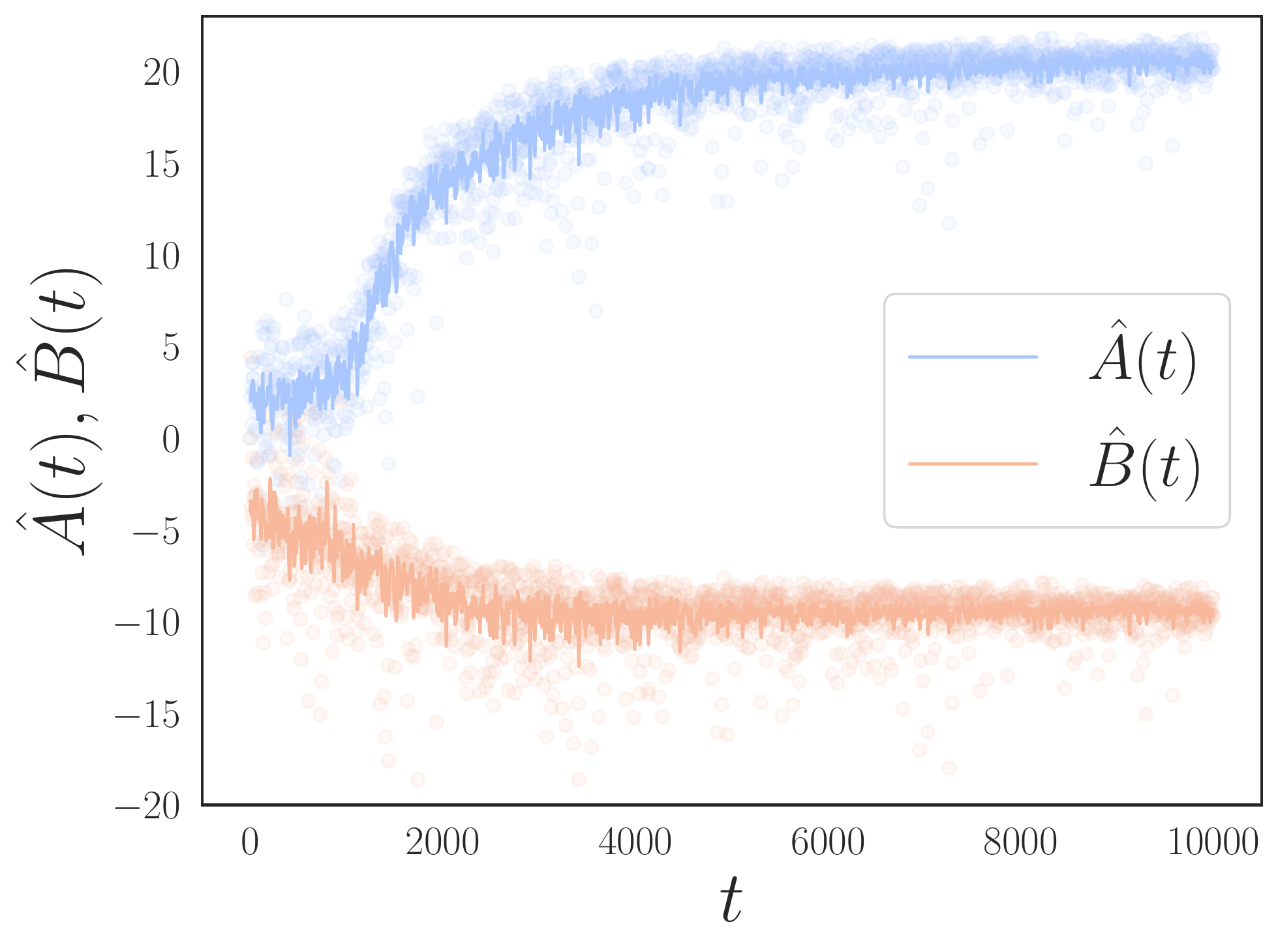}
	    \caption{$n=480$ (\mL).} \label{fig:fig:supp:geomN:AB:L:600}
	\end{subfigure}%
	\begin{subfigure}[t]{0.25\textwidth}
	    \centering
	    \includegraphics[width=\linewidth]{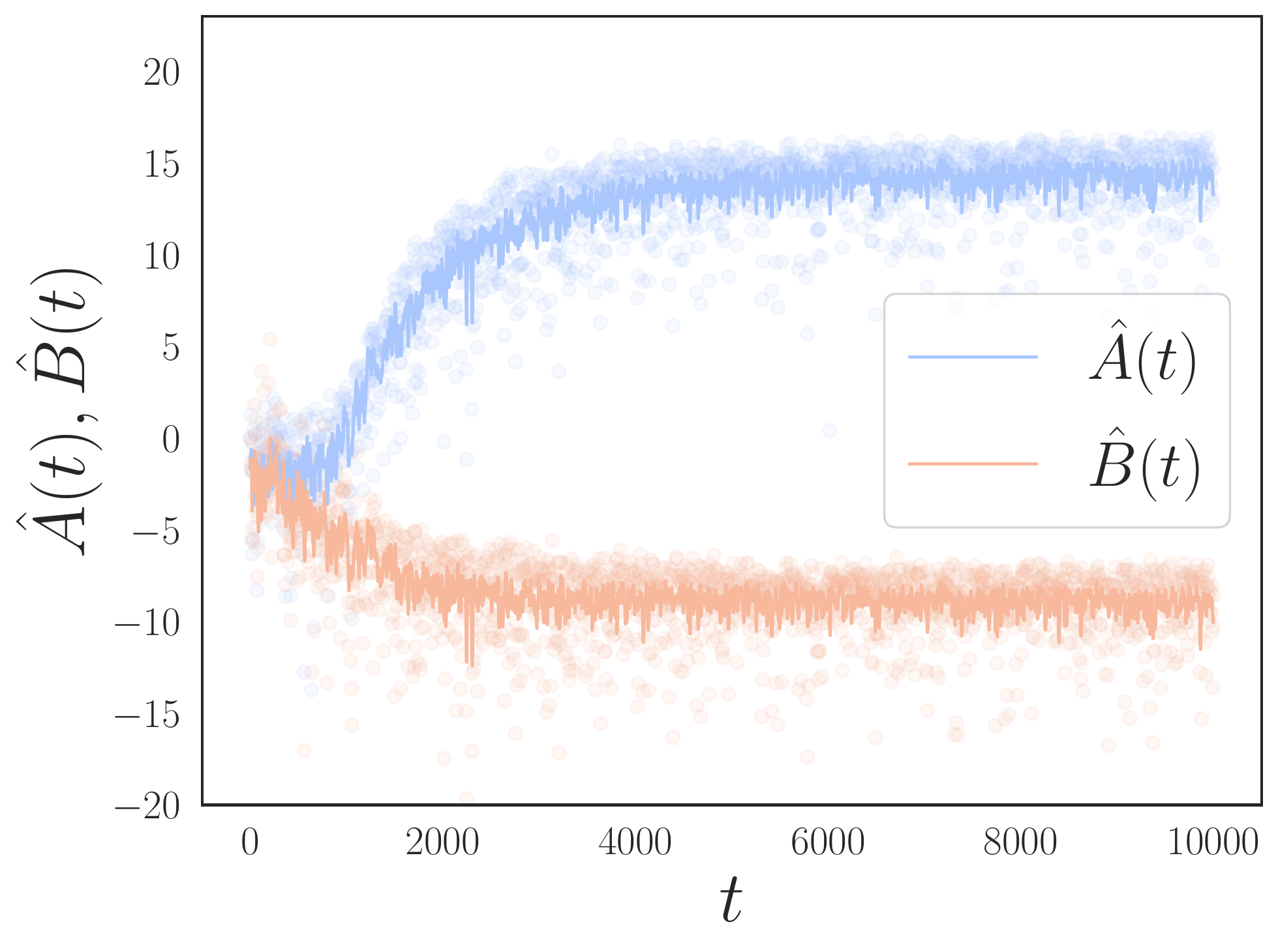}
	    \caption{$n=600$ (\mL).} \label{fig:fig:supp:geomN:AB:L:1000}
	\end{subfigure}%
	\begin{subfigure}[t]{0.25\textwidth}
	    \centering
	    \includegraphics[width=\linewidth]{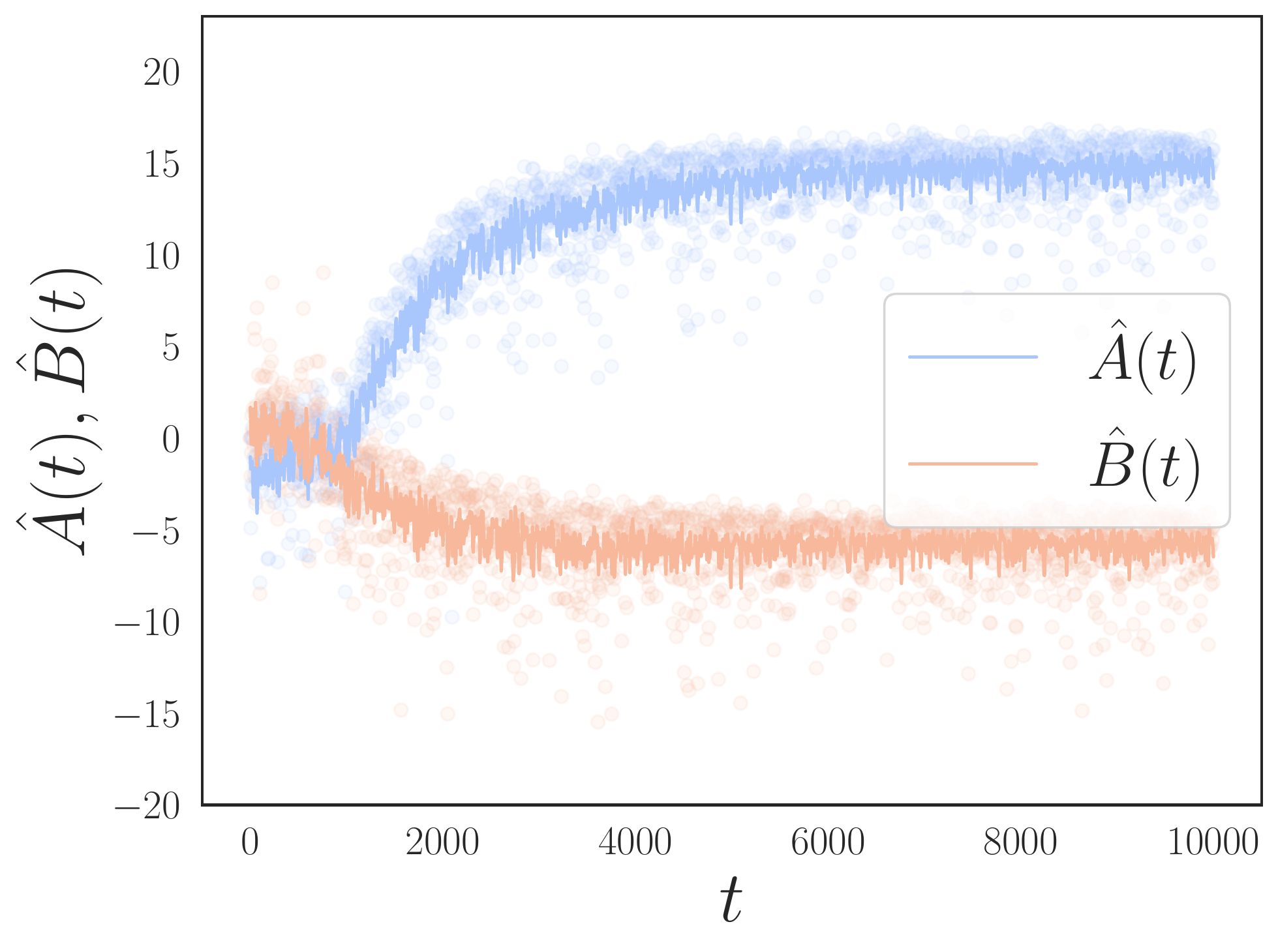}
	    \caption{$n=4800$ (\mL).} \label{fig:fig:supp:geomN:AB:L:6000}
	\end{subfigure}\\
	\begin{subfigure}[t]{0.25\textwidth}
	    \centering
	    \includegraphics[width=\linewidth]{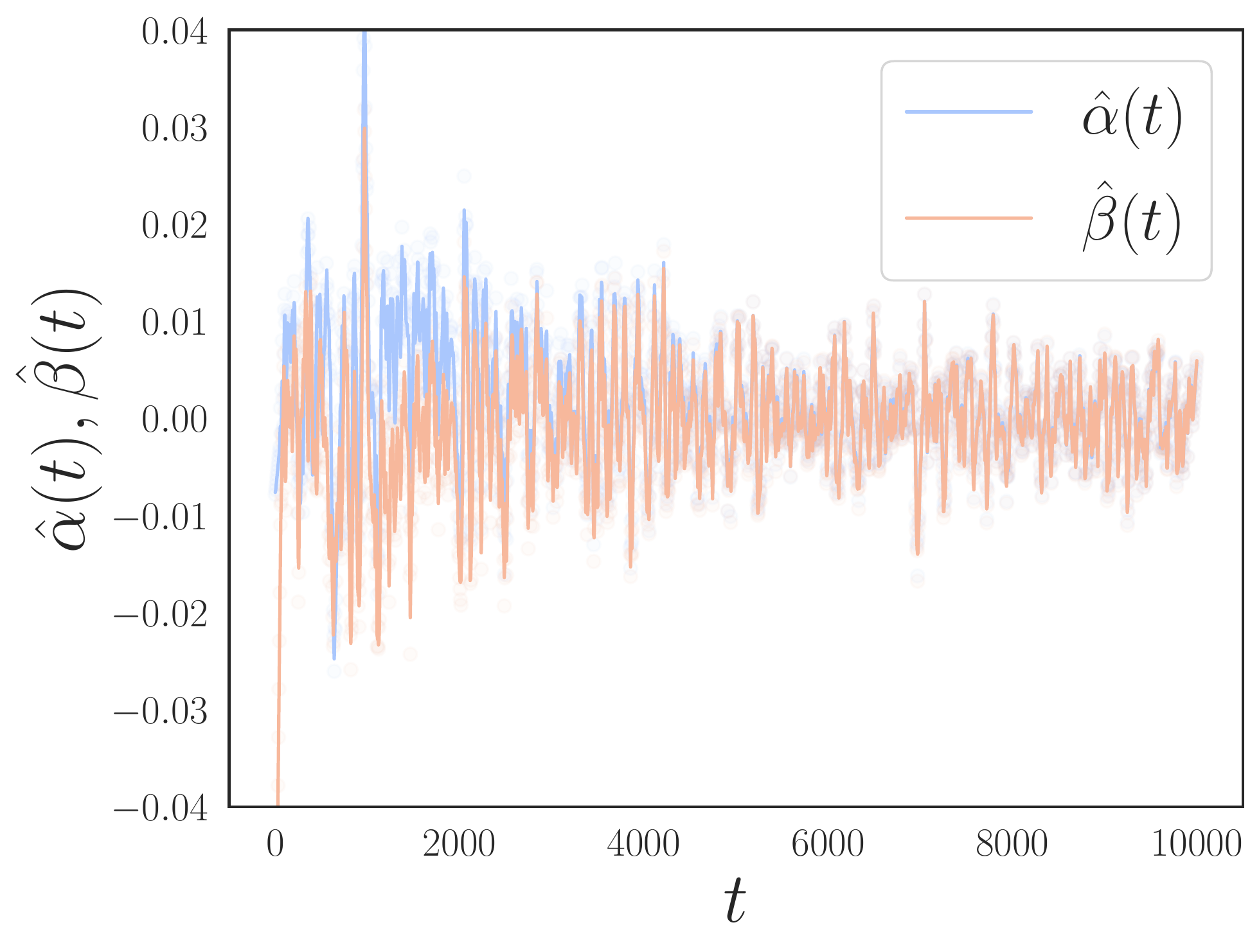}
	    \caption{$n=80$ (\mL).} \label{fig:fig:supp:geomN:dAB:L:100}
	\end{subfigure}%
	\begin{subfigure}[t]{0.25\textwidth}
	    \centering
	    \includegraphics[width=\linewidth]{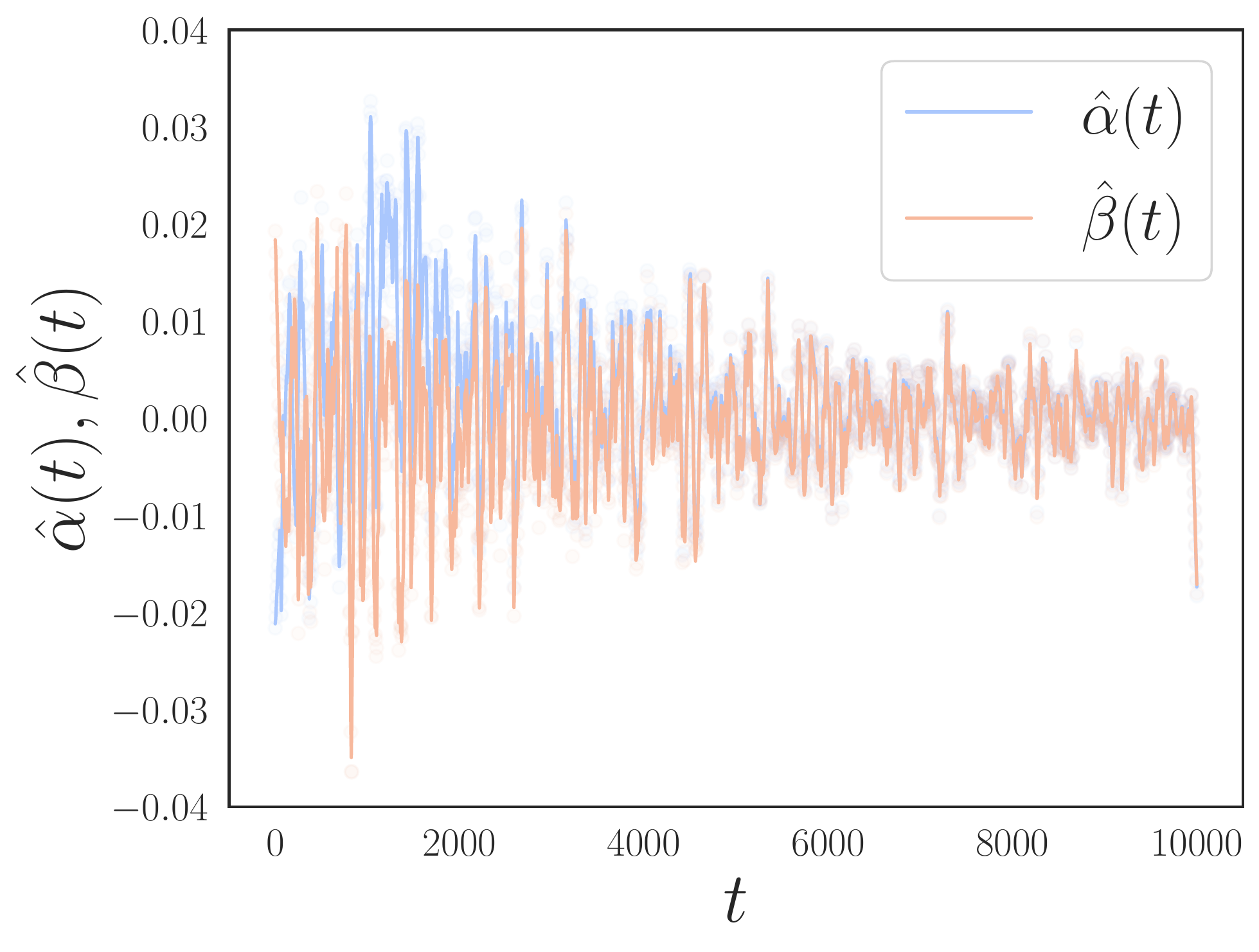}
	    \caption{$n=480$ (\mL).} \label{fig:fig:supp:geomN:dAB:L:600}
	\end{subfigure}%
	\begin{subfigure}[t]{0.25\textwidth}
	    \centering
	    \includegraphics[width=\linewidth]{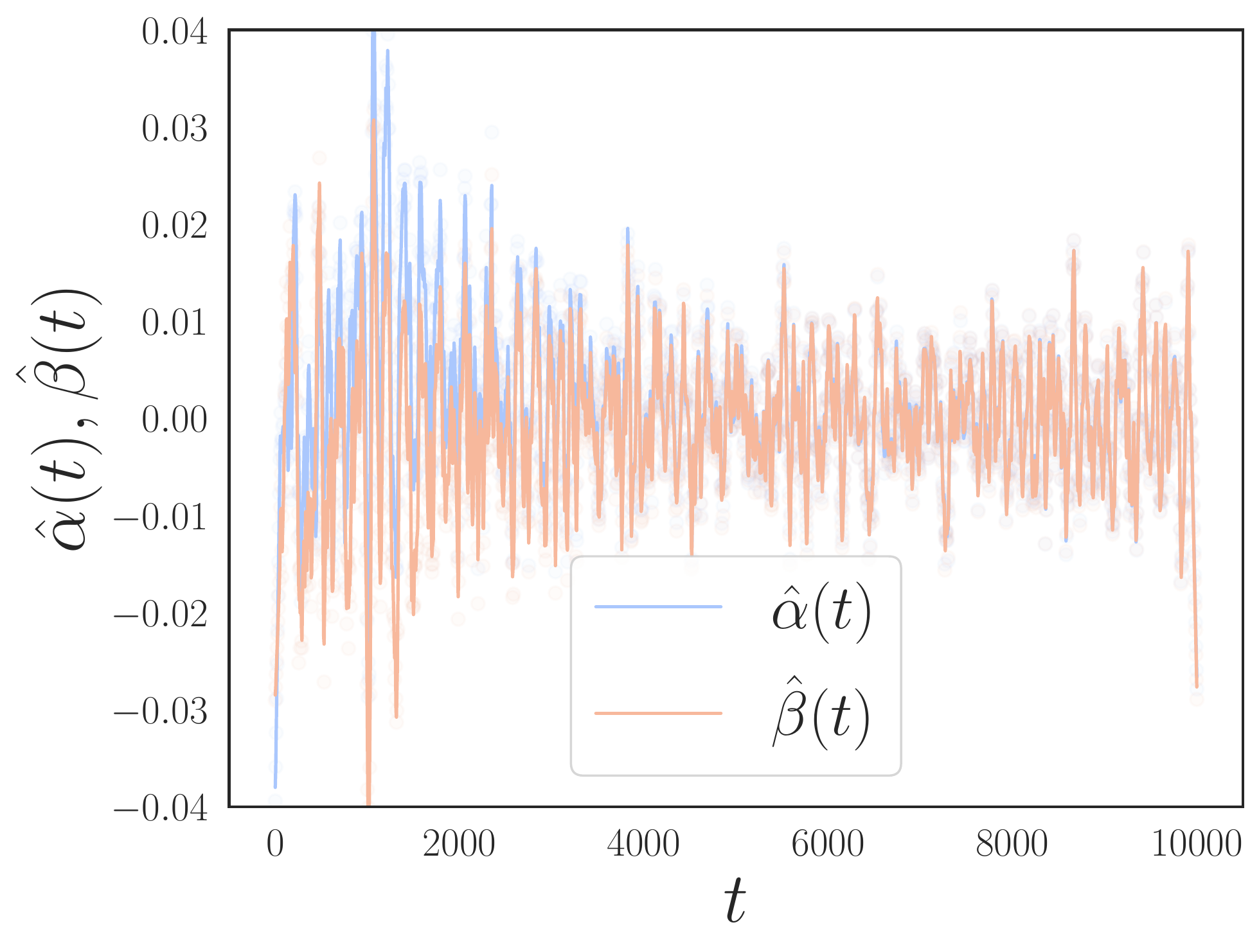}
	    \caption{$n=600$ (\mL).} \label{fig:fig:supp:geomN:dAB:L:1000}
	\end{subfigure}%
	\begin{subfigure}[t]{0.25\textwidth}
	    \centering
	    \includegraphics[width=\linewidth]{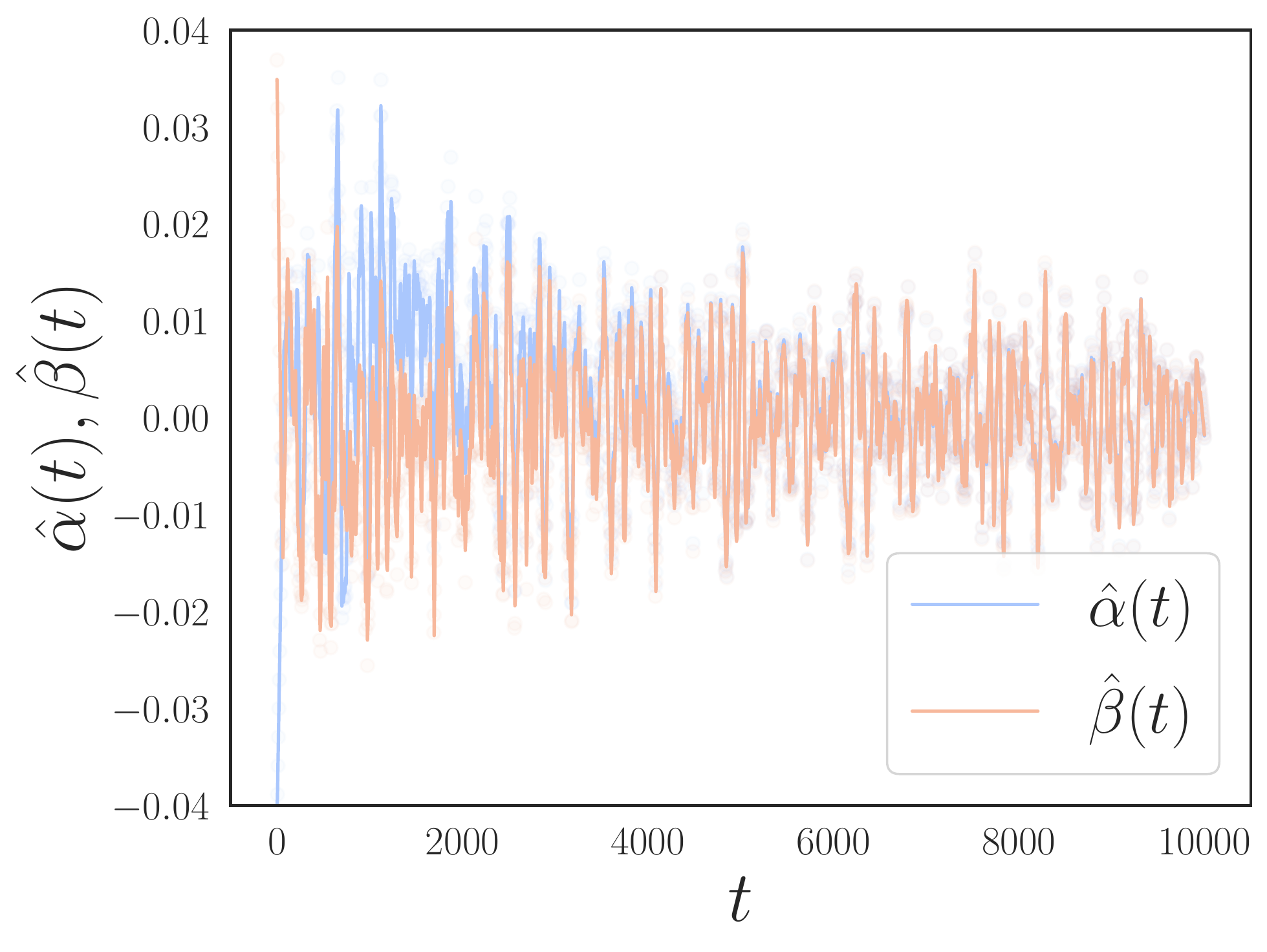}
	    \caption{$n=4800$ (\mL).} \label{fig:fig:supp:geomN:dAB:L:6000}
	\end{subfigure}\\
        \caption{\textbf{Effect of per-class sample size $n$.}
            Estimated $\hat{A}(t)$, $\hat{B}(t)$ (the first and the third rows)
            and $\hat{\alpha}(t)$, $\hat{\beta}(t)$ (the second and the fourth rows), under \mI{} (the first two rows) and \mL{} (the last two rows). Note that $n$ appears to be insignificant
            in determining the shapes and magnitudes of the curves.}
        \label{fig:supp:geomN}
\end{figure*}

    We show in \Cref{fig:supp:geomN} estimated $\hat{A}(t)$, $\hat{B}(t)$,
    $\hat{\alpha}(t)$ and $\hat{\beta}(t)$ versus training time $t$ on both
    \geomnist{} and \cifar{} under both \mI{} and \mL{} with various
    per-class sample size $n=n_{\mathrm{tr}} \in\{80,480,600,4800\}$.
    We choose the window size $\omega=151$ when applying the Savitzky–Golay filter.
    Note that the estimates under both \mI{} and \mL{} do not vary significantly under different choices of $N$ and share similar trends: (i) $\beta(t)$ is dominated
    by $\alpha(t)$; (ii) $\alpha(t)$ has an initial increasing stage and
    a second stage converging to the vicinity around zero.
    This is expected since our theory is independent of
    $n$ once $n$ is reasonably large.

\paragraph{Effects of Number of Classes $K$.}

\begin{figure*}
    \centering
    \begin{subfigure}[t]{0.25\textwidth}
	    \centering
	    \includegraphics[width=\linewidth]{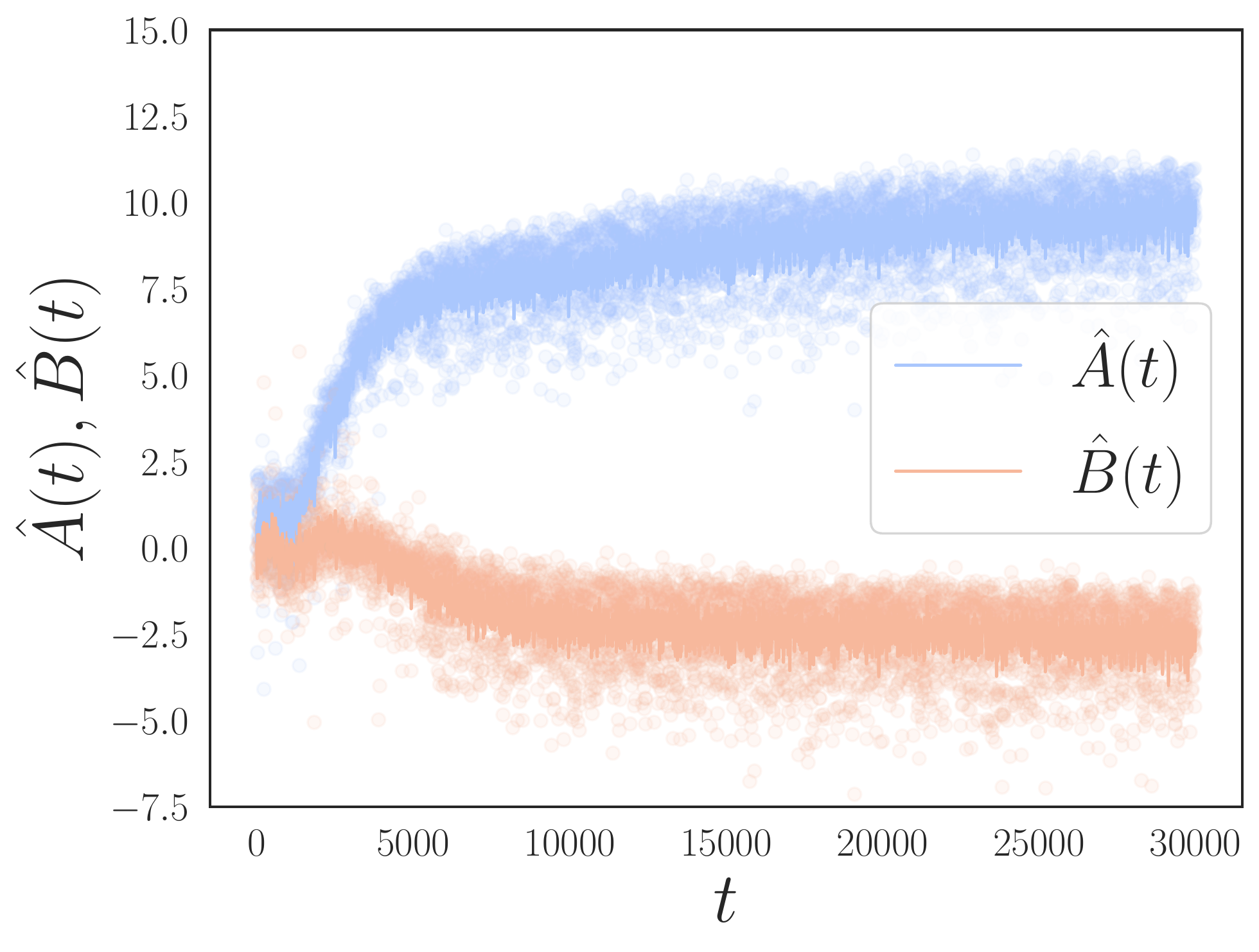}
	    \caption{$K=2$.} \label{fig:supp:cifar10:K2:AB:I}
	\end{subfigure}%
	\begin{subfigure}[t]{0.25\textwidth}
	    \centering
	    \includegraphics[width=\linewidth]{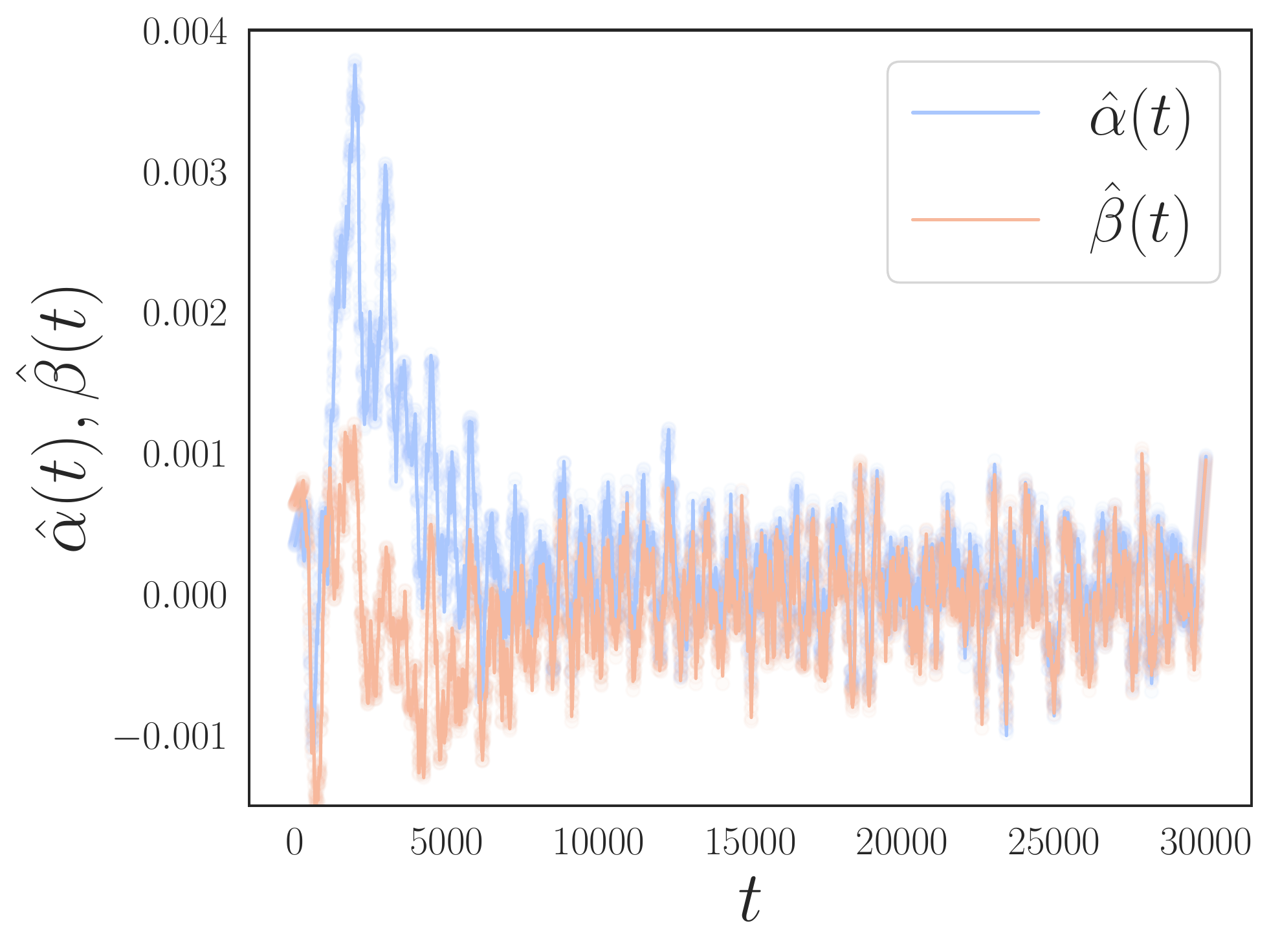}
		\caption{$K=2$.} \label{fig:supp:cifar10:K2:dAB:I}
	\end{subfigure}%
	\begin{subfigure}[t]{0.25\textwidth}
	    \centering
	    \includegraphics[width=\linewidth]{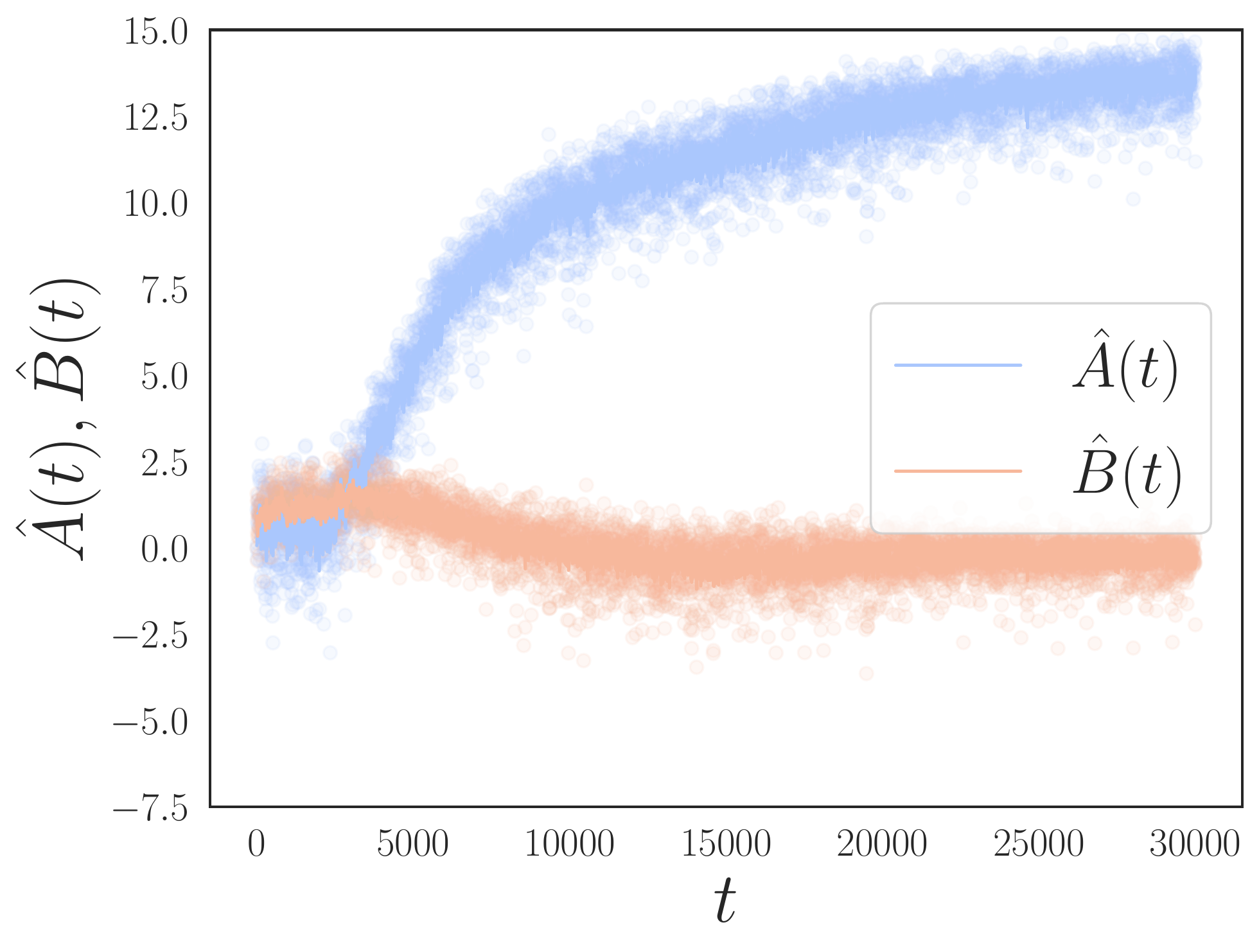}
	    \caption{$K=3$.} \label{fig:supp:cifar10:K3:AB:I}
	\end{subfigure}%
	\begin{subfigure}[t]{0.25\textwidth}
	    \centering
	    \includegraphics[width=\linewidth]{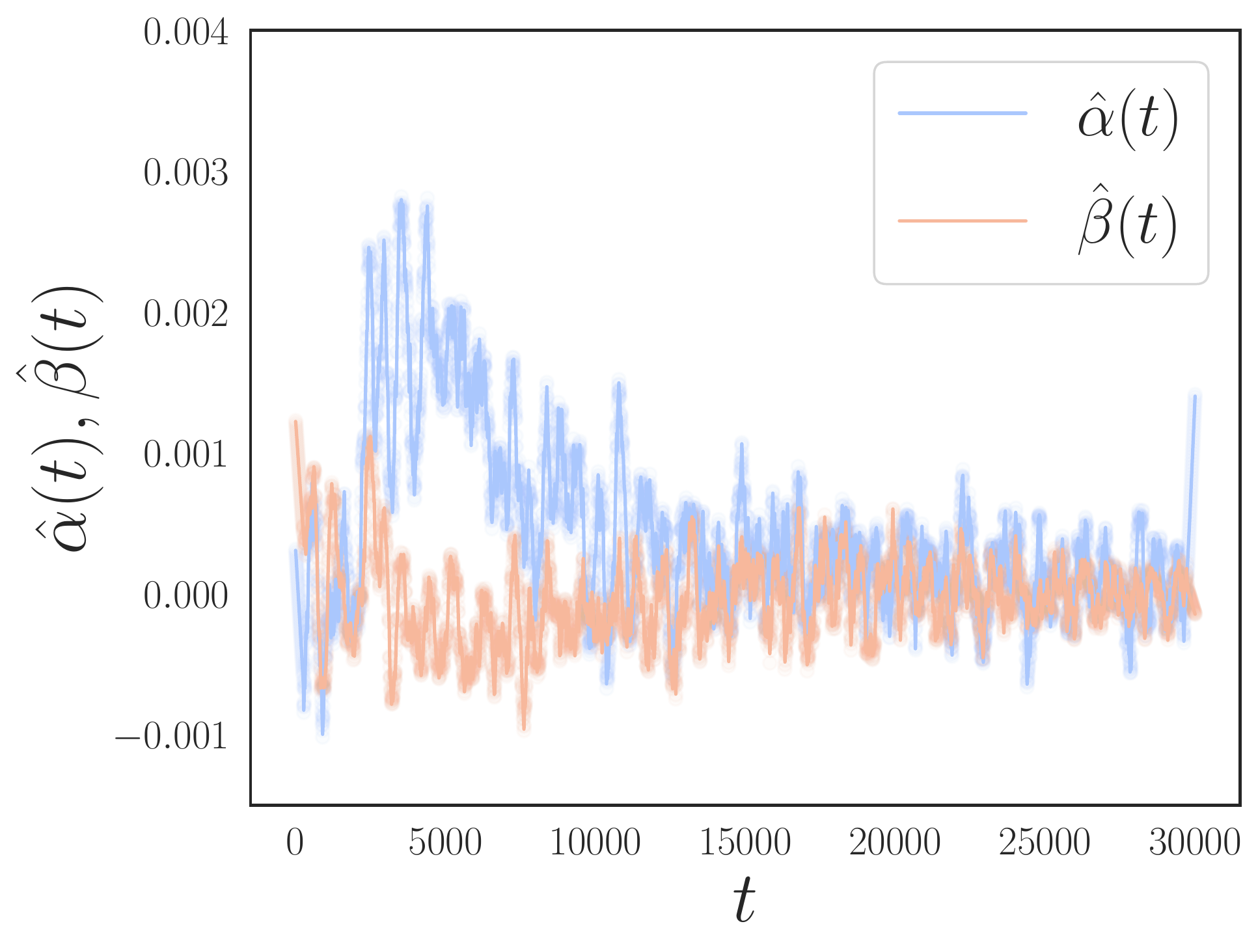}
	    \caption{$K=3$.} \label{fig:supp:cifar10:K3:dAB:I}
	\end{subfigure}\\
        \caption{\textbf{Effect of number of classes $K$.}
            Estimated $\hat{A}(t)$, $\hat{B}(t)$ ((a) and (c))
            and $\hat{\alpha}(t)$, $\hat{\beta}(t)$ ((b) and (d))
            on \cifar{} with $K=2$ ((a)-(b)) and $K=3$ ((c)-(d)),
            all under \mI{}. Note that the general trends
            with different $K$'s are similar.
        }
        \label{fig:supp:cifar10K}
\end{figure*}

    In \Cref{fig:supp:cifar10K} we show the estimated $\hat{A}(t)$, $\hat{B}(t)$,
    $\hat{\alpha}(t)$ and $\hat{\beta}(t)$ versus training time $t$ on 
    \cifar{} under the \mI{} with number of classes $K\in\{2,3\}$. We note
    that in both cases, generally speaking, the shapes of $A(t)$ and $B(t)$
    are similar:
    $A(t)$ increases while $B(t)$ decreases, suggesting this
    behavior is more general and is likely independent of the number of classes. On the other hand, note that the number of classes
    affects the scale of $A(t)$ and $B(t)$.

\paragraph{Effects of Label Corruption Ratio $\perr$.}

\begin{figure*}
    \centering
    \begin{subfigure}[t]{0.25\textwidth}
	    \centering
	    \includegraphics[width=\linewidth]{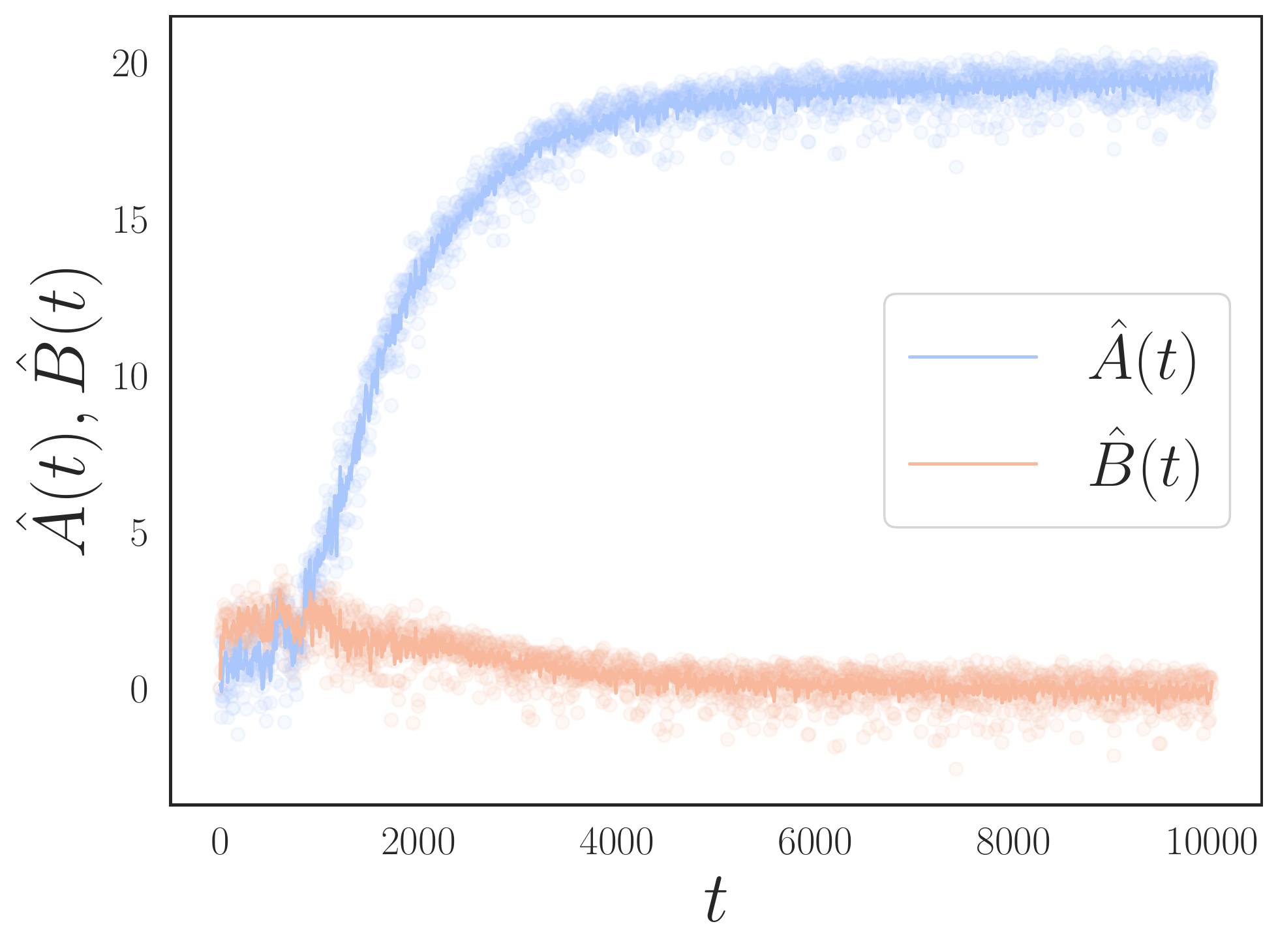}
	    \caption{$\perr=0$ (\mI).} \label{fig:supp:geomperr:AB:I:0}
	\end{subfigure}%
	\begin{subfigure}[t]{0.25\textwidth}
	    \centering
	    \includegraphics[width=\linewidth]{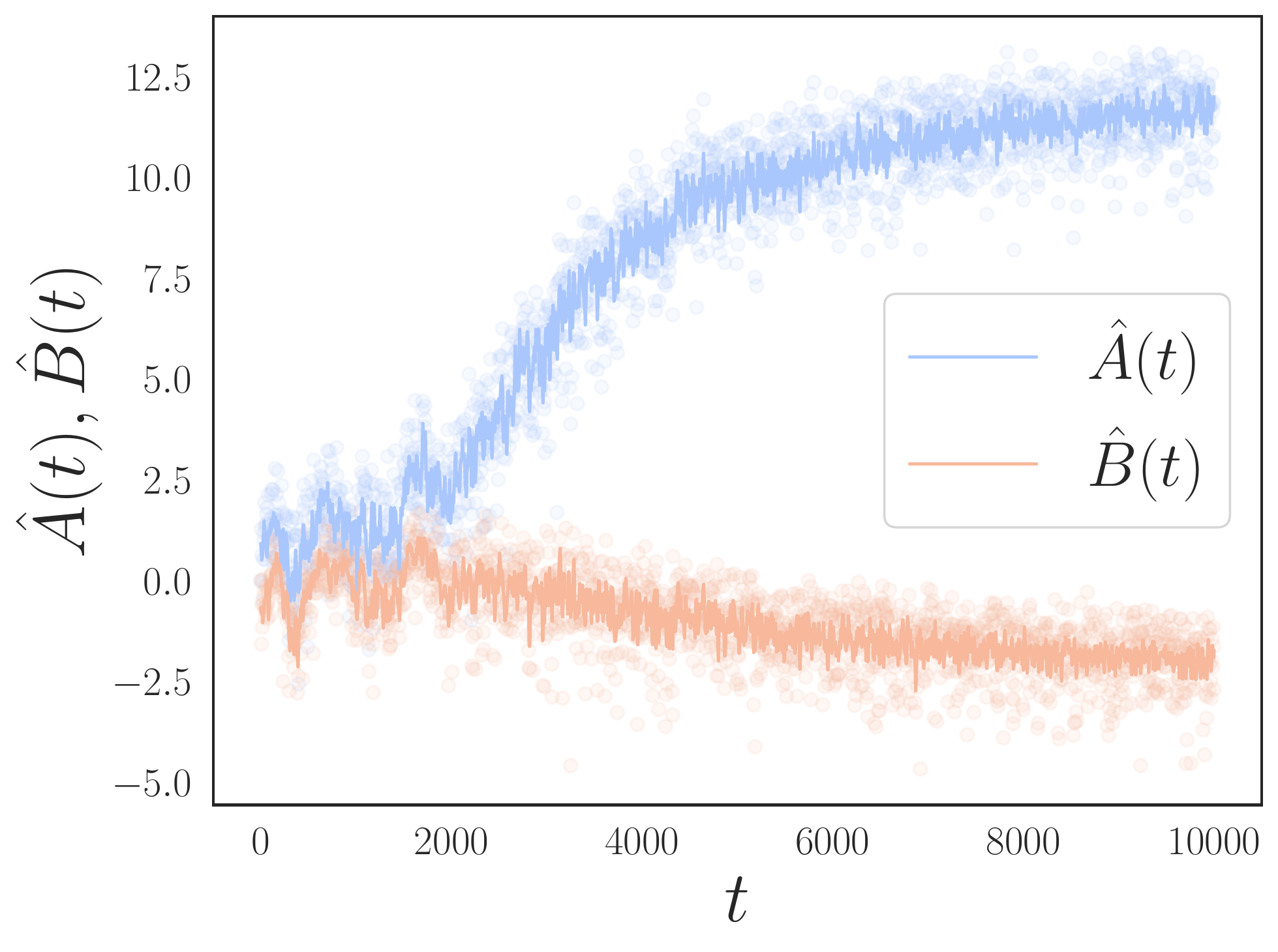}
	    \caption{$\perr=0.3$ (\mI).} \label{fig:supp:geomperr:AB:I:3}
	\end{subfigure}%
	\begin{subfigure}[t]{0.25\textwidth}
	    \centering
	    \includegraphics[width=\linewidth]{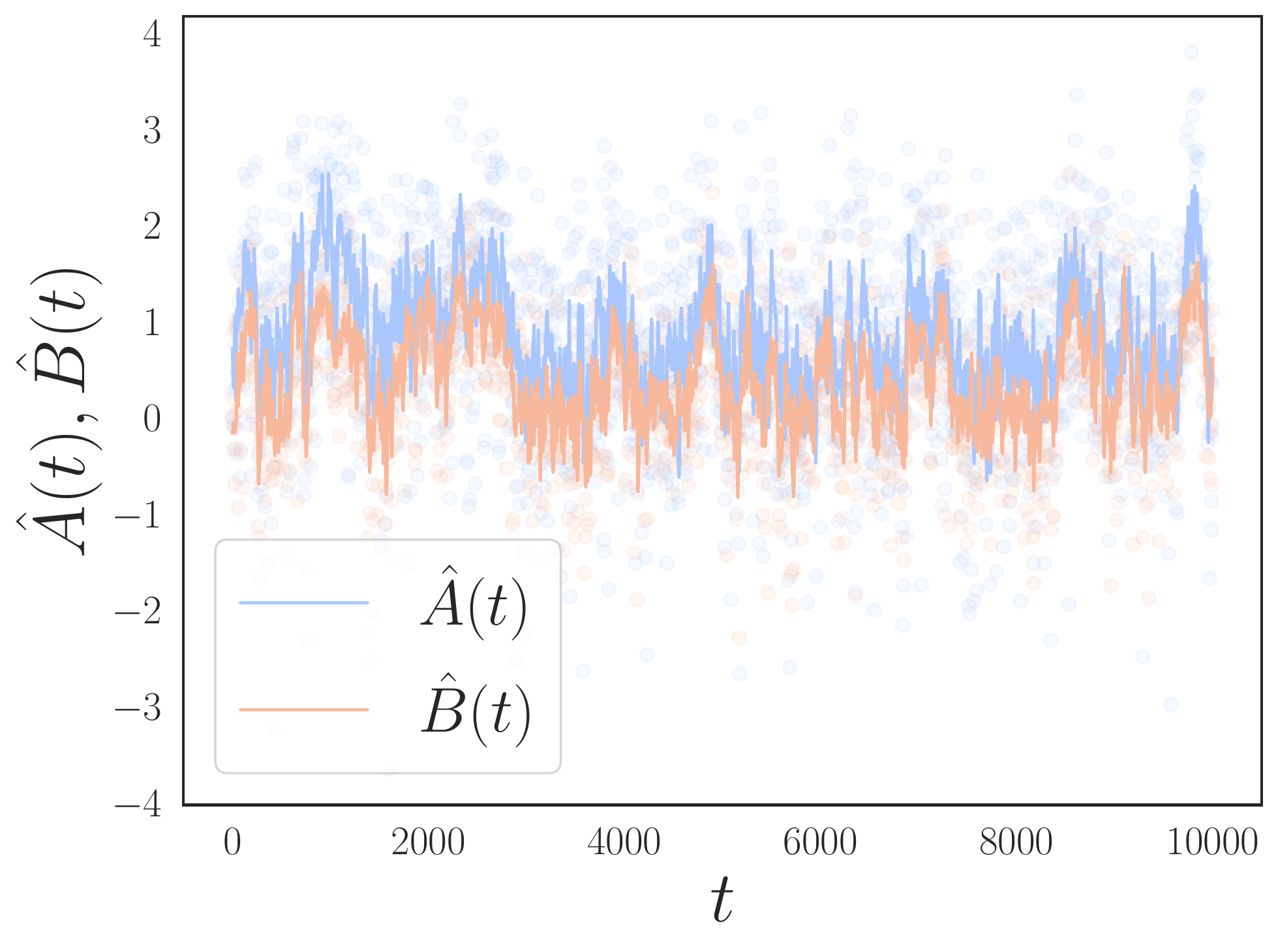}
	    \caption{$\perr=\frac{2}{3}$ (\mI).} \label{fig:supp:geomperr:AB:I:6}
	\end{subfigure}%
	\begin{subfigure}[t]{0.25\textwidth}
	    \centering
	    \includegraphics[width=\linewidth]{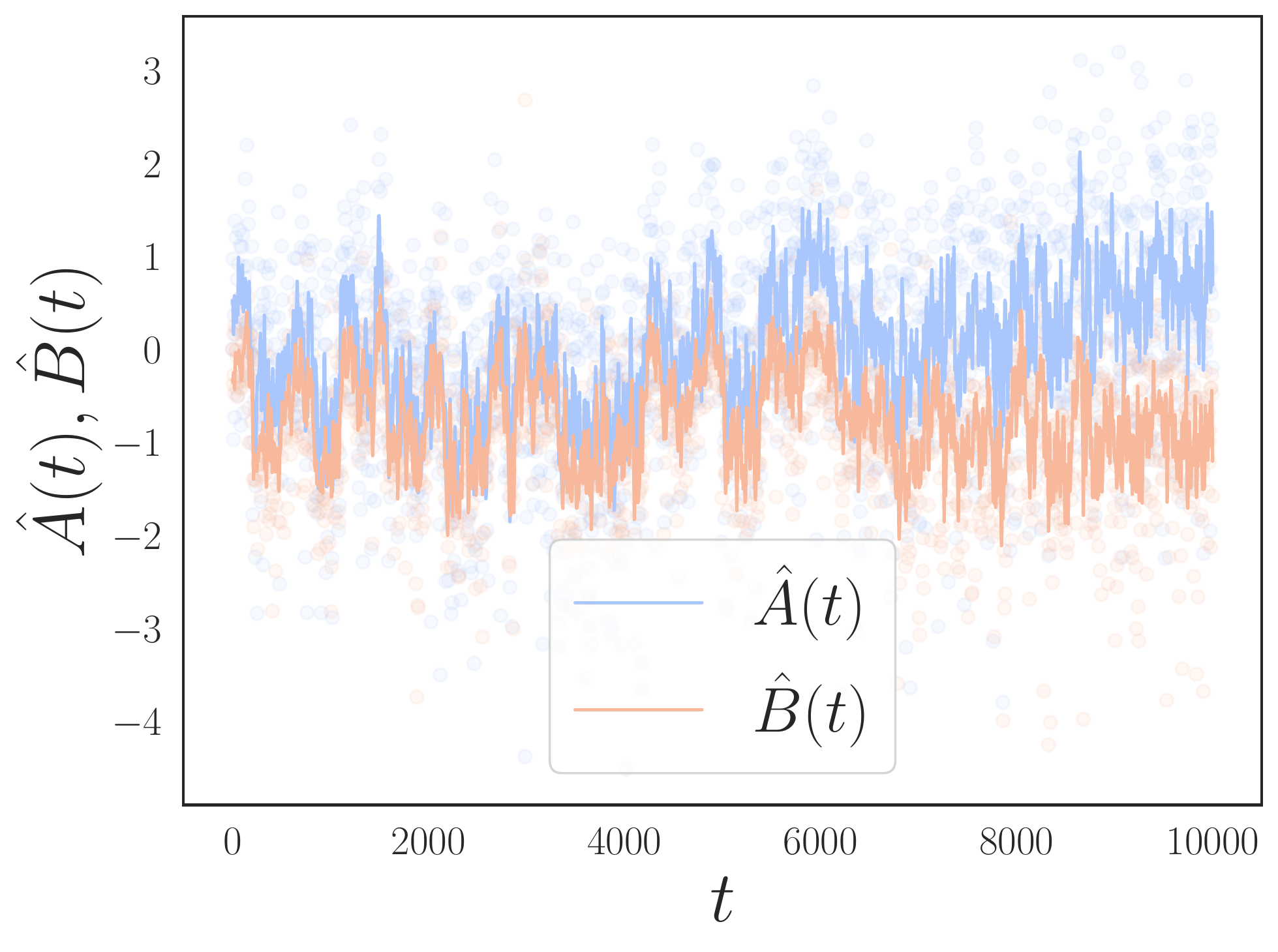}
	    \caption{$\perr=0.8$ (\mI).} \label{fig:supp:geomperr:AB:I:8}
	\end{subfigure}\\
	\begin{subfigure}[t]{0.25\textwidth}
	    \centering
	    \includegraphics[width=\linewidth]{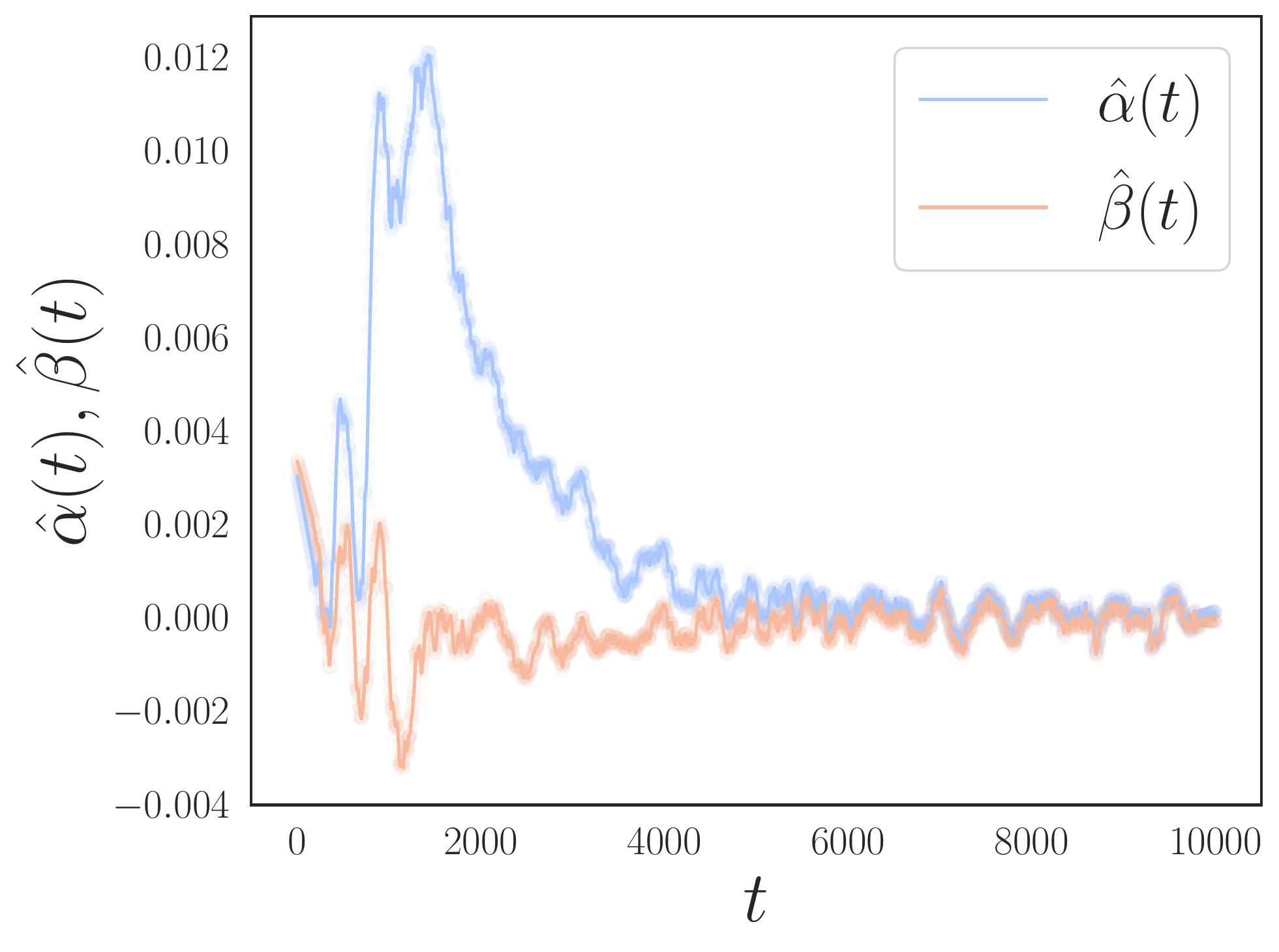}
	    \caption{$\perr=0$ (\mI).} \label{fig:supp:geomperr:dAB:I:0}
	\end{subfigure}%
	\begin{subfigure}[t]{0.25\textwidth}
	    \centering
	    \includegraphics[width=\linewidth]{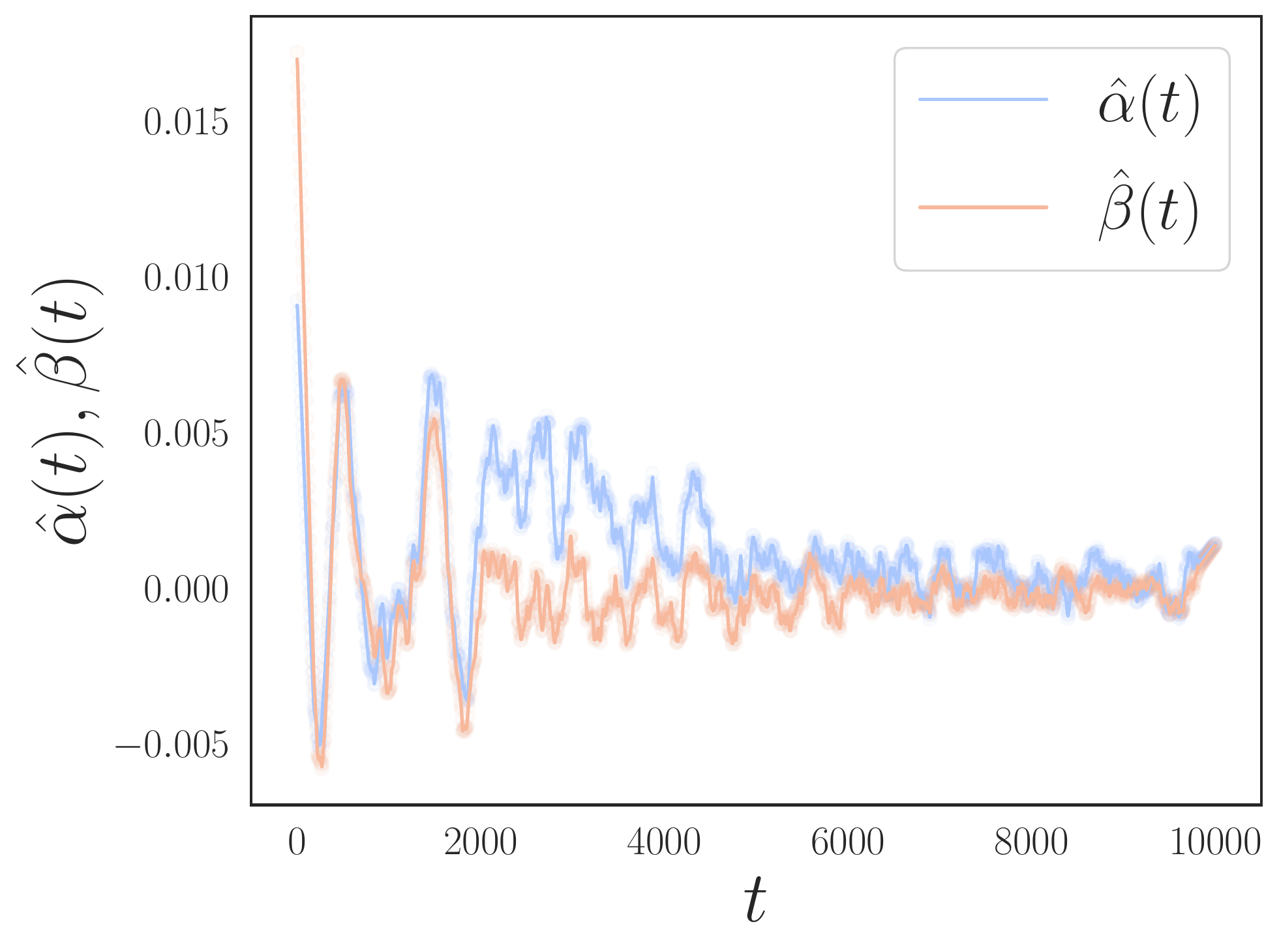}
	    \caption{$\perr=0.3$ (\mI).} \label{fig:supp:geomperr:dAB:I:3}
	\end{subfigure}%
	\begin{subfigure}[t]{0.25\textwidth}
	    \centering
	    \includegraphics[width=\linewidth]{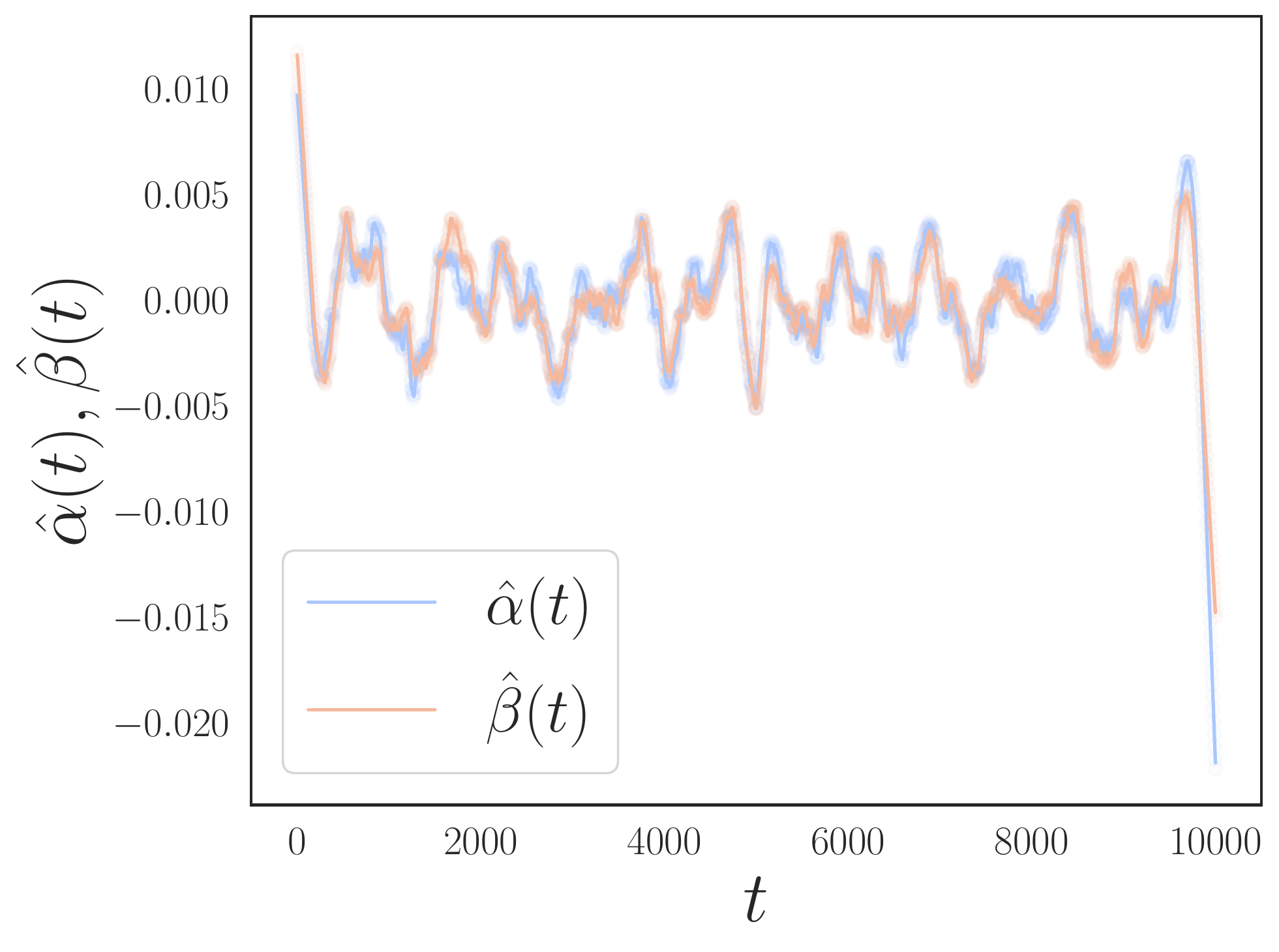}
	    \caption{$\perr=\frac{2}{3}$ (\mI).} \label{fig:supp:geomperr:dAB:I:6}
	\end{subfigure}%
	\begin{subfigure}[t]{0.25\textwidth}
	    \centering
	    \includegraphics[width=\linewidth]{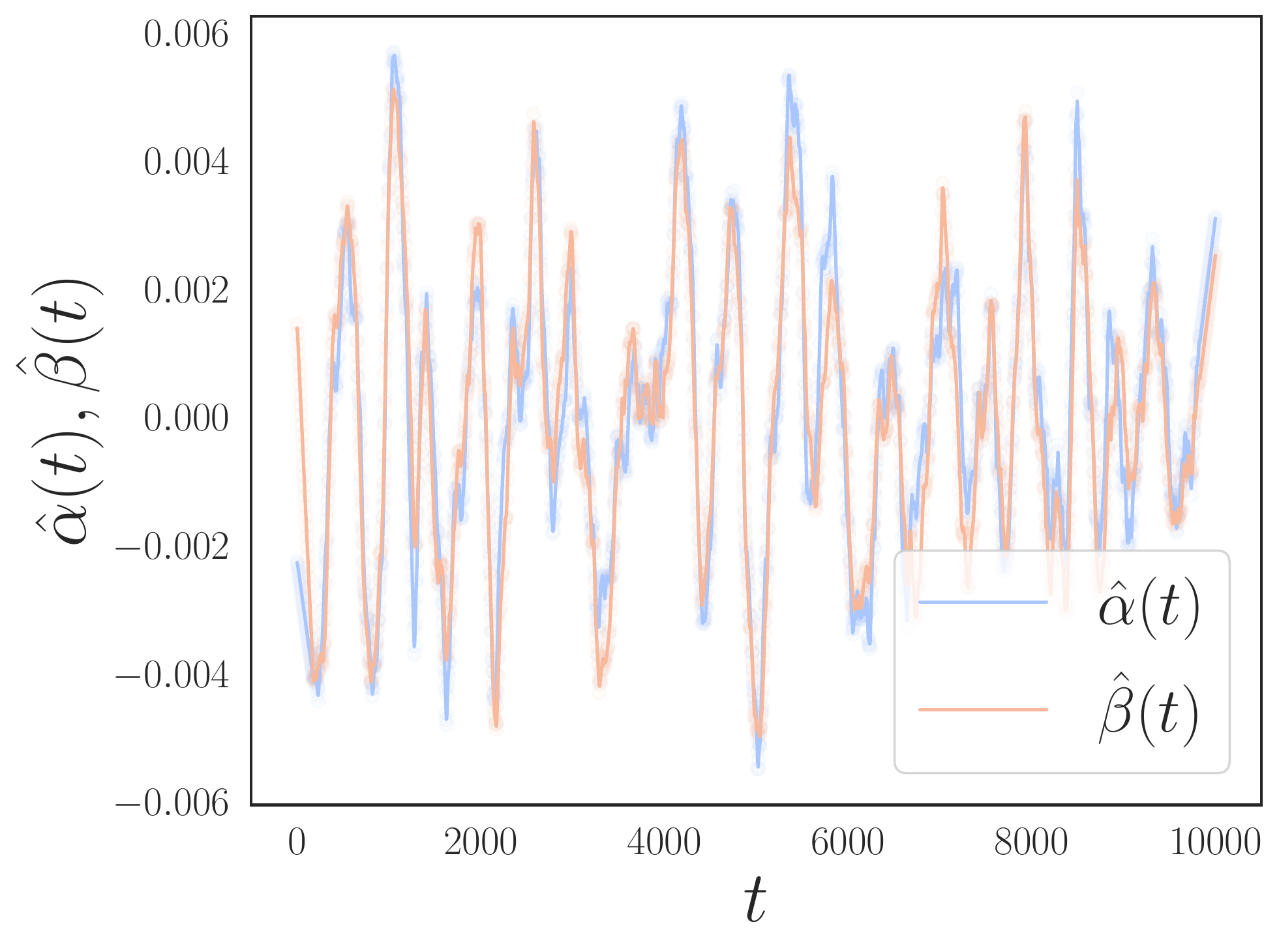}
	    \caption{$\perr=0.8$ (\mI).} \label{fig:supp:geomperr:dAB:I:8}
	\end{subfigure}\\
	\begin{subfigure}[t]{0.25\textwidth}
	    \centering
	    \includegraphics[width=\linewidth]{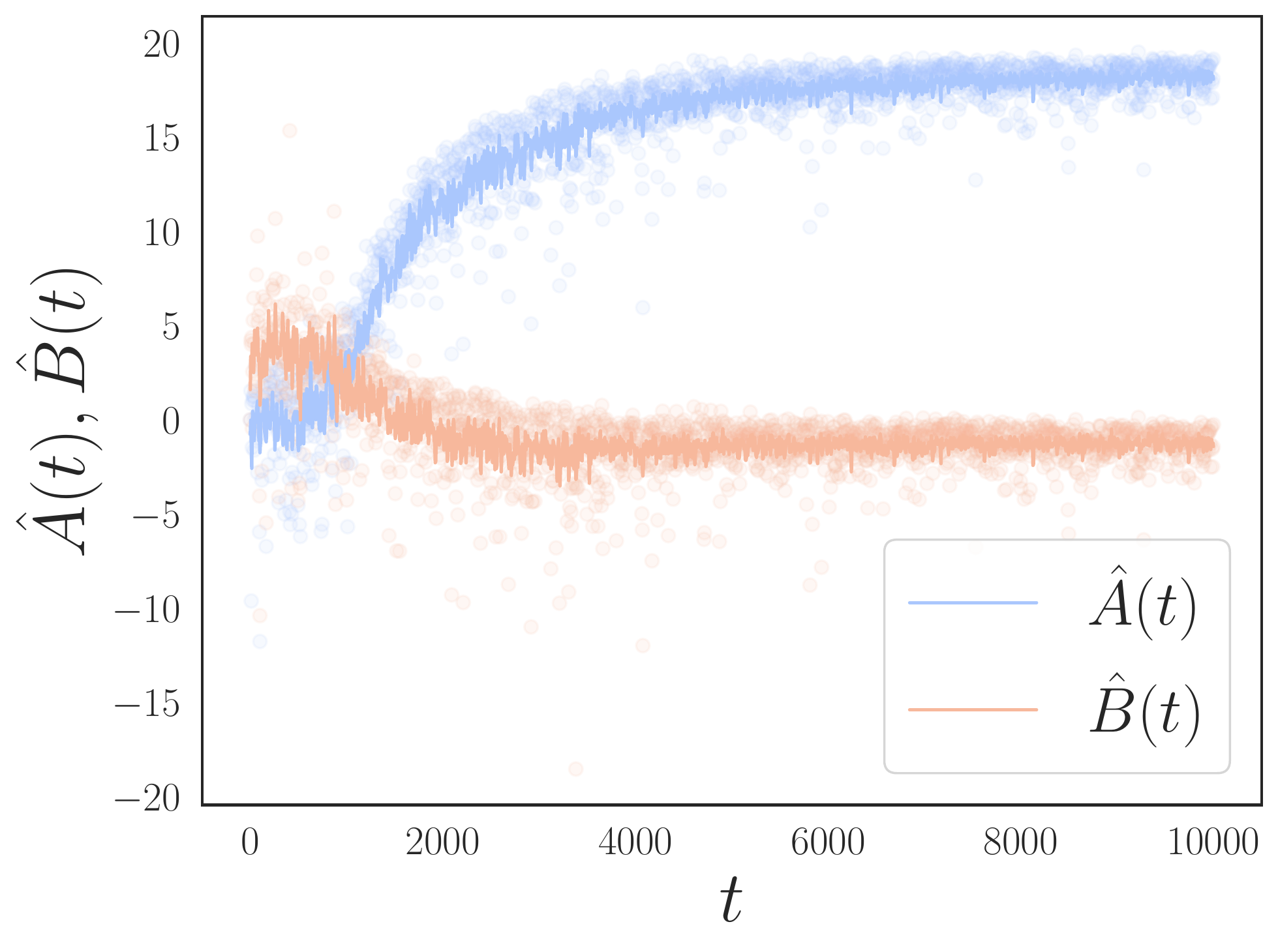}
	    \caption{$\perr=0$ (\mL).} \label{fig:supp:geomperr:AB:L:0}
	\end{subfigure}%
	\begin{subfigure}[t]{0.25\textwidth}
	    \centering
	    \includegraphics[width=\linewidth]{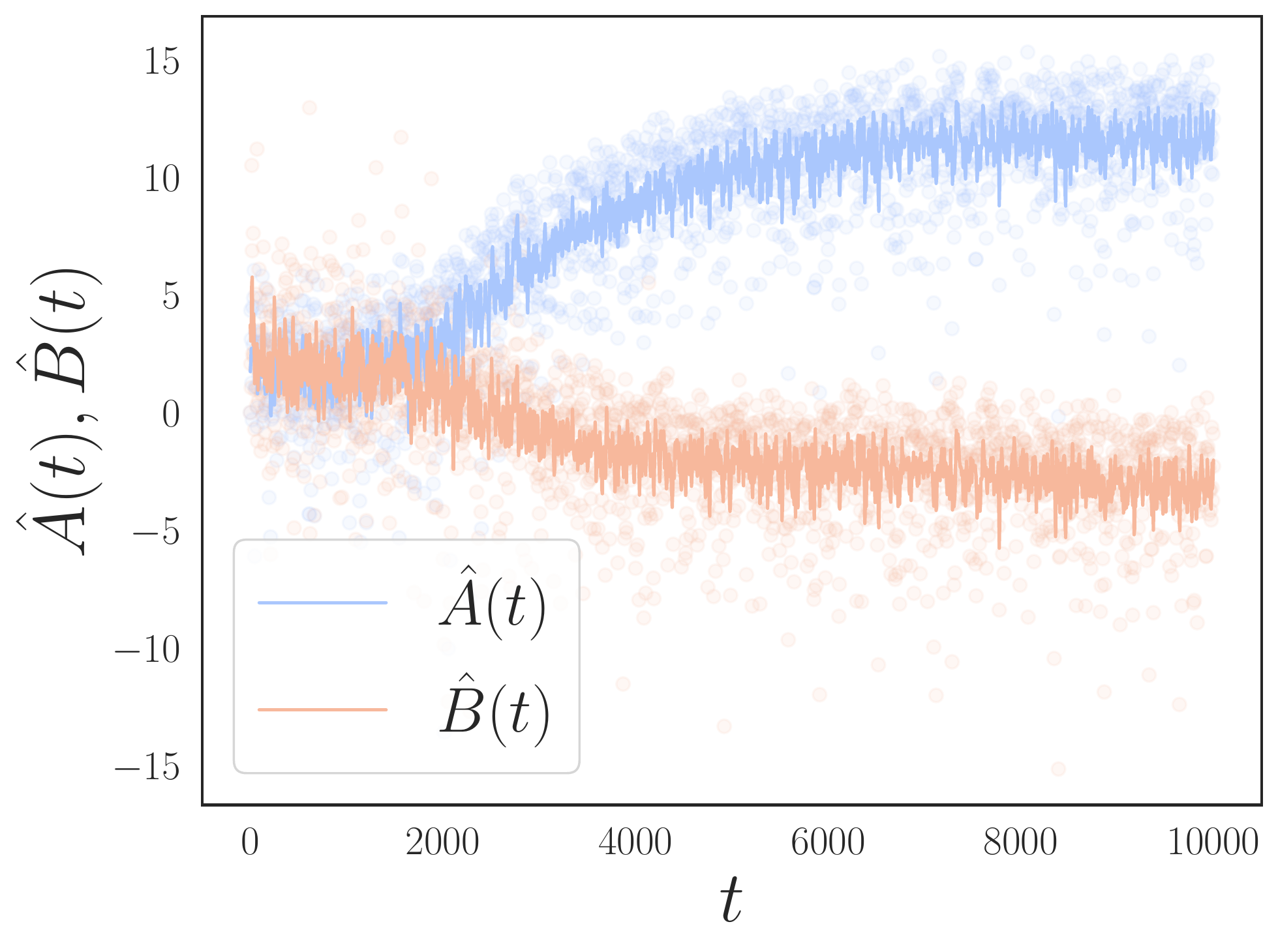}
	    \caption{$\perr=0.3$ (\mL).} \label{fig:supp:geomperr:AB:L:3}
	\end{subfigure}%
	\begin{subfigure}[t]{0.25\textwidth}
	    \centering
	    \includegraphics[width=\linewidth]{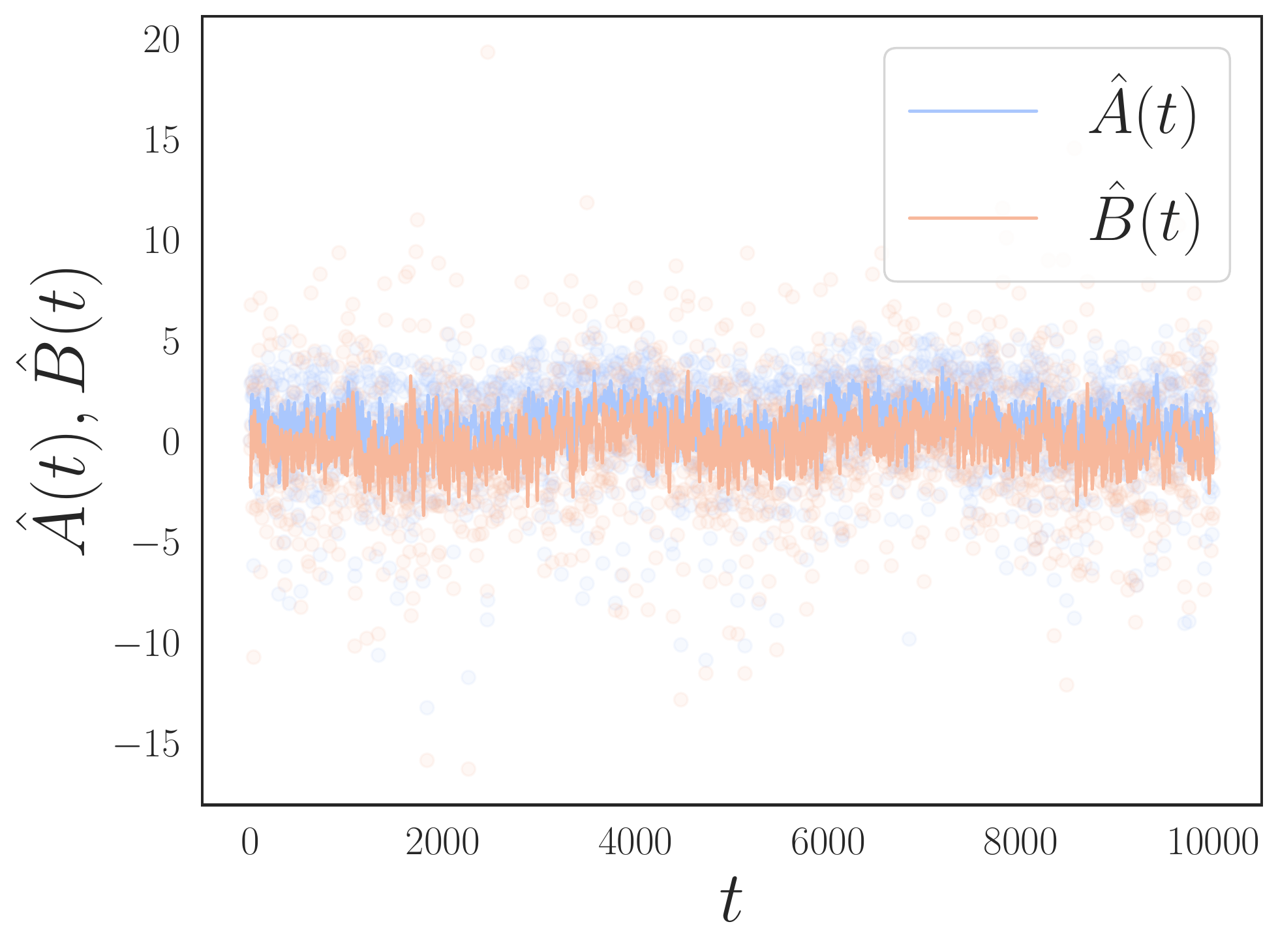}
	    \caption{$\perr=\frac{2}{3}$ (\mL).} \label{fig:supp:geomperr:AB:L:6}
	\end{subfigure}%
	\begin{subfigure}[t]{0.25\textwidth}
	    \centering
	    \includegraphics[width=\linewidth]{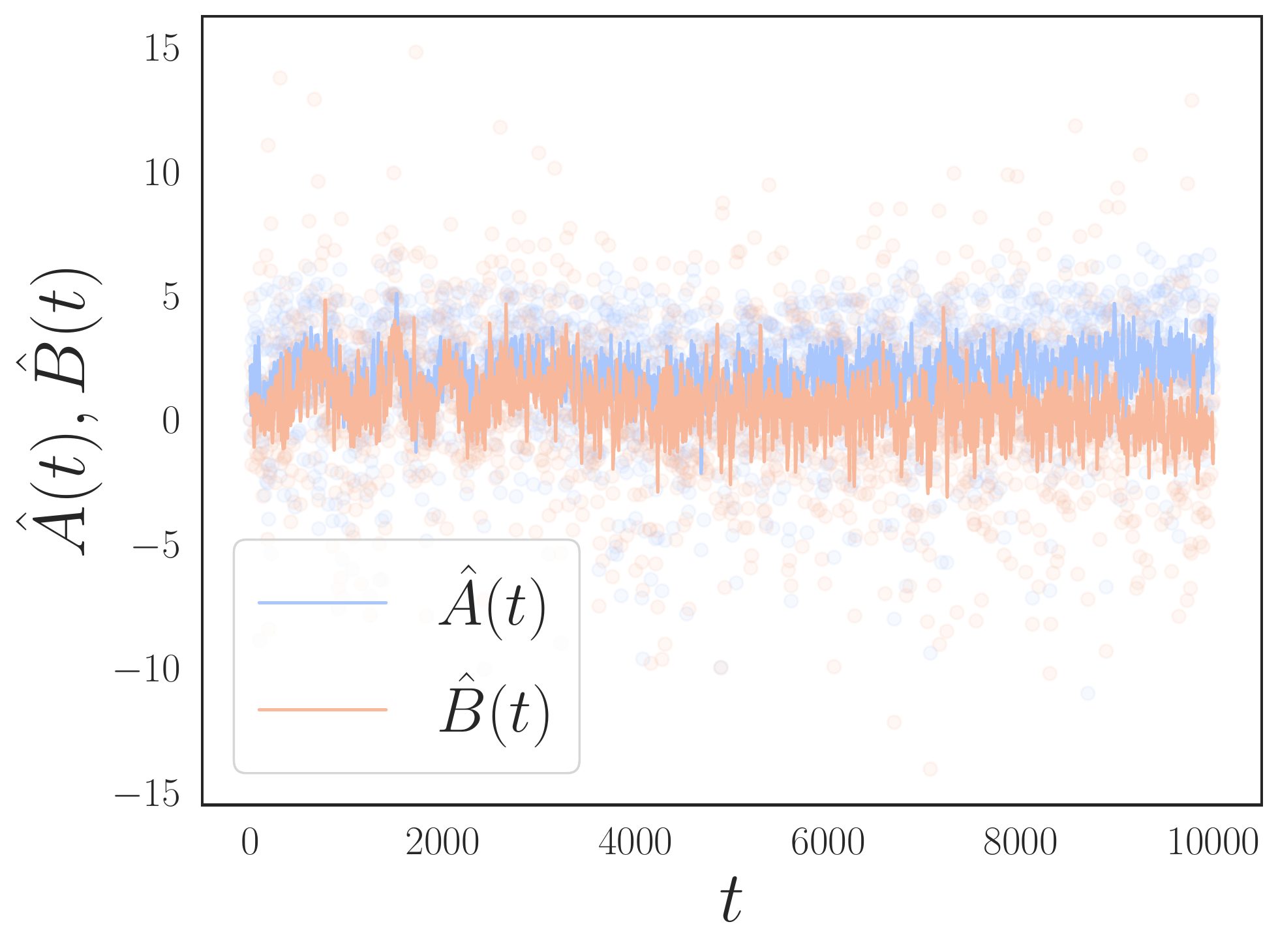}
	    \caption{$\perr=0.8$ (\mL).} \label{fig:supp:geomperr:AB:L:8}
	\end{subfigure}\\
	\begin{subfigure}[t]{0.25\textwidth}
	    \centering
	    \includegraphics[width=\linewidth]{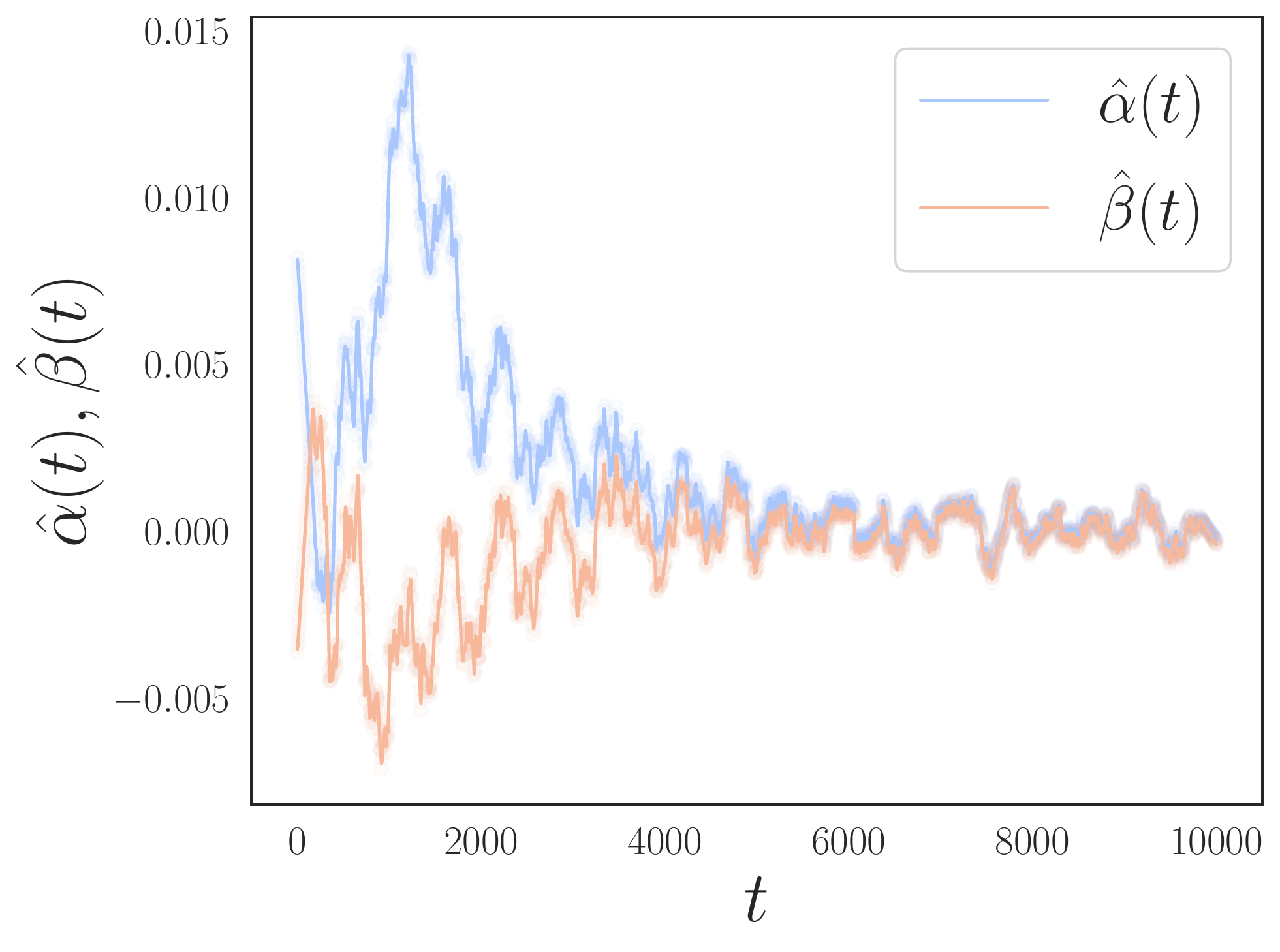}
	    \caption{$\perr=0$ (\mL).} \label{fig:supp:geomperr:dAB:L:0}
	\end{subfigure}%
	\begin{subfigure}[t]{0.25\textwidth}
	    \centering
	    \includegraphics[width=\linewidth]{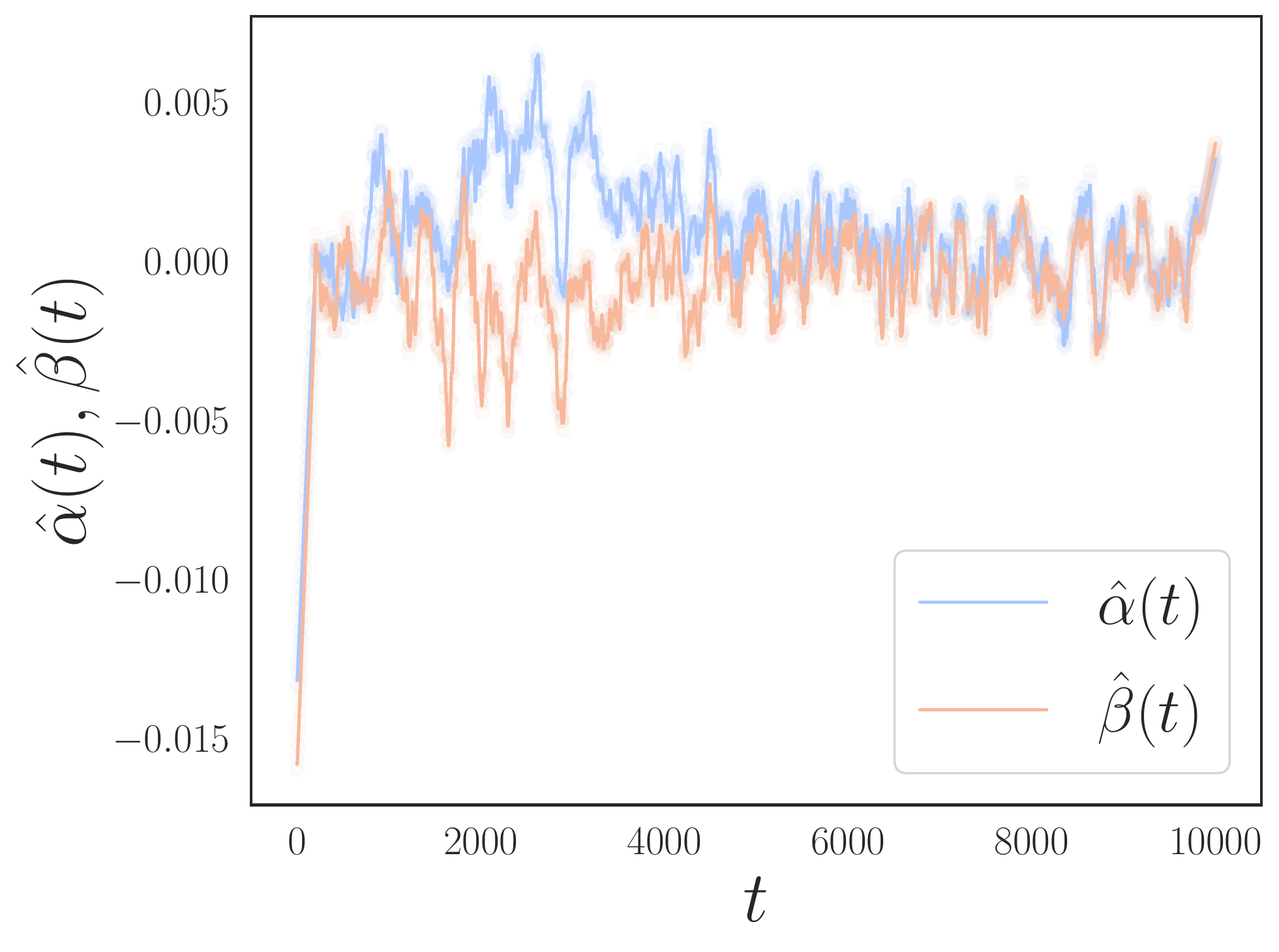}
	    \caption{$\perr=0.3$ (\mL).} \label{fig:supp:geomperr:dAB:L:3}
	\end{subfigure}%
	\begin{subfigure}[t]{0.25\textwidth}
	    \centering
	    \includegraphics[width=\linewidth]{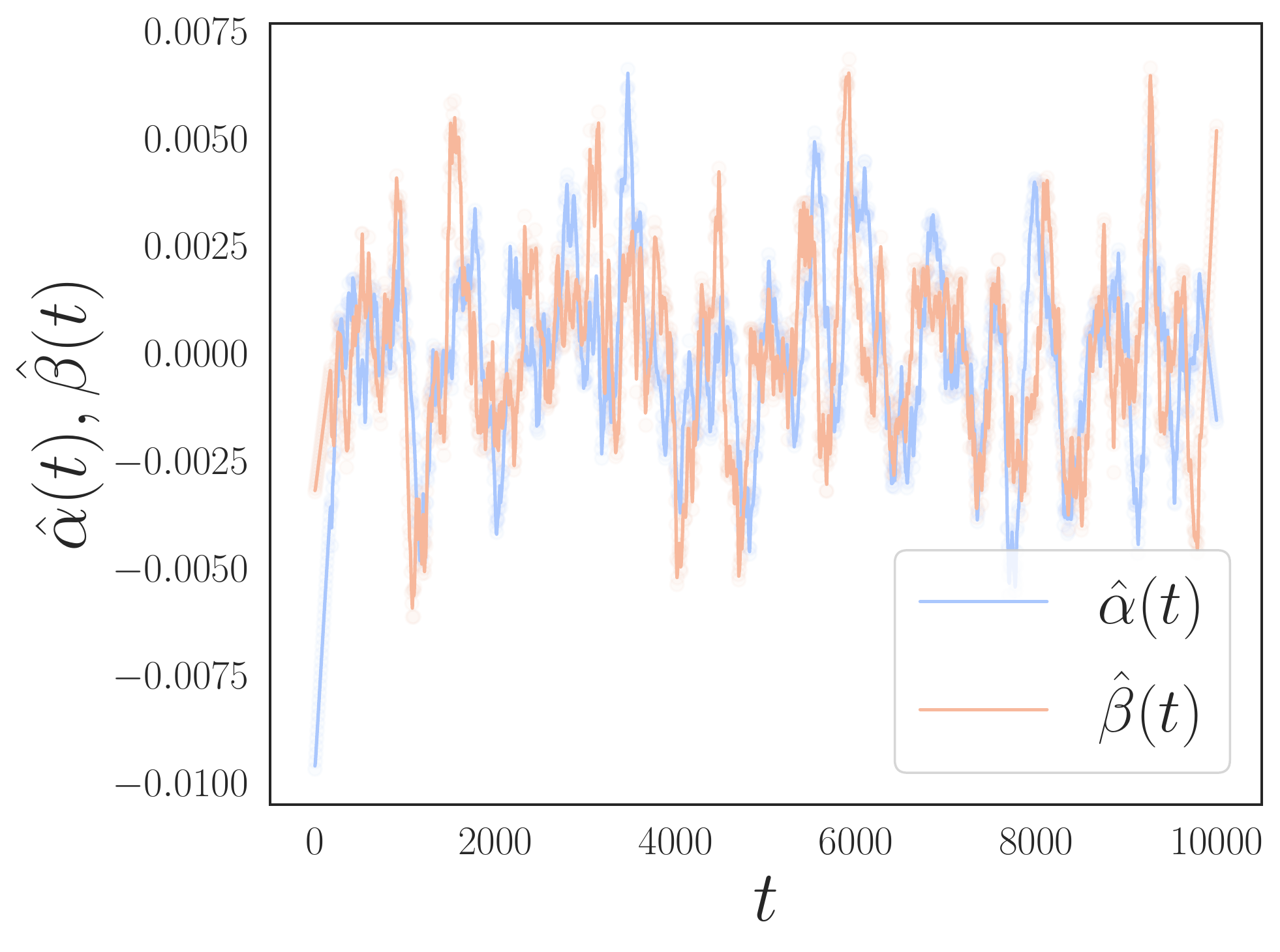}
	    \caption{$\perr=\frac{2}{3}$ (\mL).} \label{fig:supp:geomperr:dAB:L:6}
	\end{subfigure}%
	\begin{subfigure}[t]{0.25\textwidth}
	    \centering
	    \includegraphics[width=\linewidth]{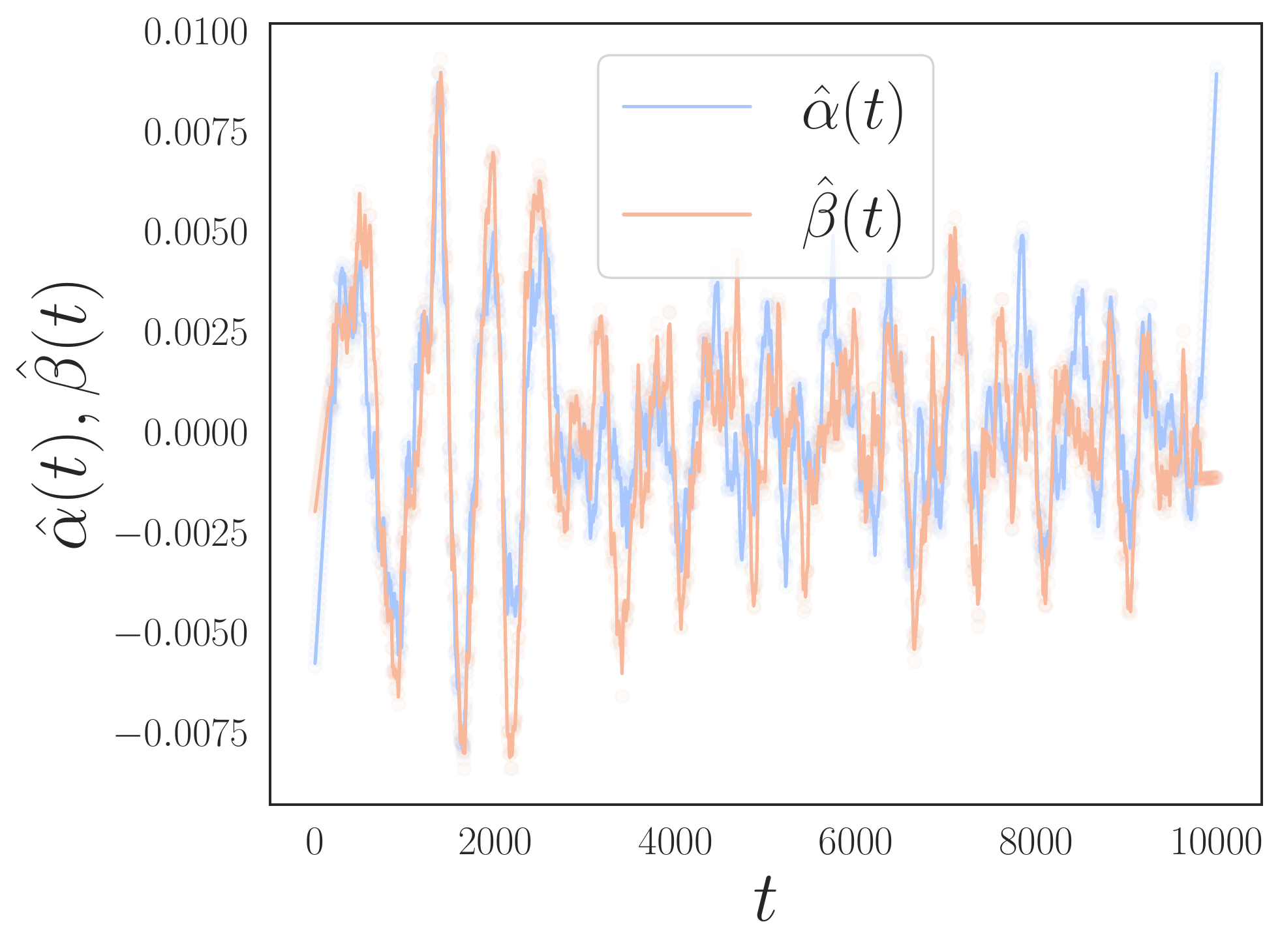}
	    \caption{$\perr=0.8$ (\mL).} \label{fig:supp:geomperr:dAB:L:8}
	\end{subfigure}\\
        \caption{\textbf{Effect of label corruption ratio $\perr$.}
            Estimated $\hat{A}(t)$, $\hat{B}(t)$ (the first and the third rows)
            and $\hat{\alpha}(t)$, $\hat{\beta}(t)$ (the second and the fourth rows), under \mI{} (the first two rows) and \mL{} (the last two rows) on \geomnist{}
            under various label corruptions ratios $\perr \in \{0,0.3, 2/3, 0.8\}$ are shown. Note $\hat{A}(t)$  and $\hat{B}(t)$,
            $\hat{\alpha}(t)$ and $\hat{\beta}(t)$ become indistinguishable
            when $\perr$ is large, supporting \Cref{thm:separation:mean}.
        }
        \label{fig:supp:geomperr}
\end{figure*}

    In \Cref{sec:lesde:general} we have shown that separability depends
    on the tail behavior of $\alpha(t) - \beta(t)$. In \Cref{fig:supp:geomperr}
    we demonstrate this more directly. Here we intentionally
    choose a large window size of $\omega=351$ to highlight the trends
    of $\alpha(t)$ and $\beta(t)$.
    Note that $A(t)$ and $B(t)$
    (and consequently $\alpha(t)$ and $\beta(t)$) become more mixed and
    indistinguishable as we increase the label corruption ratio $\perr$
    from $0$ (no corrupted label) to $0.8$. This also reaffirms
    the important
    rule played by the local elasticity strengths $\alpha(t)$ and $\beta(t)$
    in terms of separability.
    
    We also plot the estimated tail indices versus label corruption ratio
    $\perr$ using $\alpha(t)$ and $\beta(t)$ estimated under both
    \mI{} and \mL{} in \Cref{fig:app:perr}. Although in
    \Cref{fig:app:perr:L} there is no sharp phase transition boundary
    when we increase $\perr$ as \Cref{fig:app:perr:L} does,
    we observe that at around $\perr =2/3$, the estimated tail
    index of $\hat{\beta}(t)$ begins to dominate that of
    $\hat{\alpha}(t)$, which also supports our \Cref{thm:separation:mean}.

\paragraph{Local Elasticity Strengths Adjusted for Logits Norms}
\begin{figure*}
    \centering
	\begin{subfigure}[t]{0.25\textwidth}
	    \centering
	    \includegraphics[width=\linewidth]{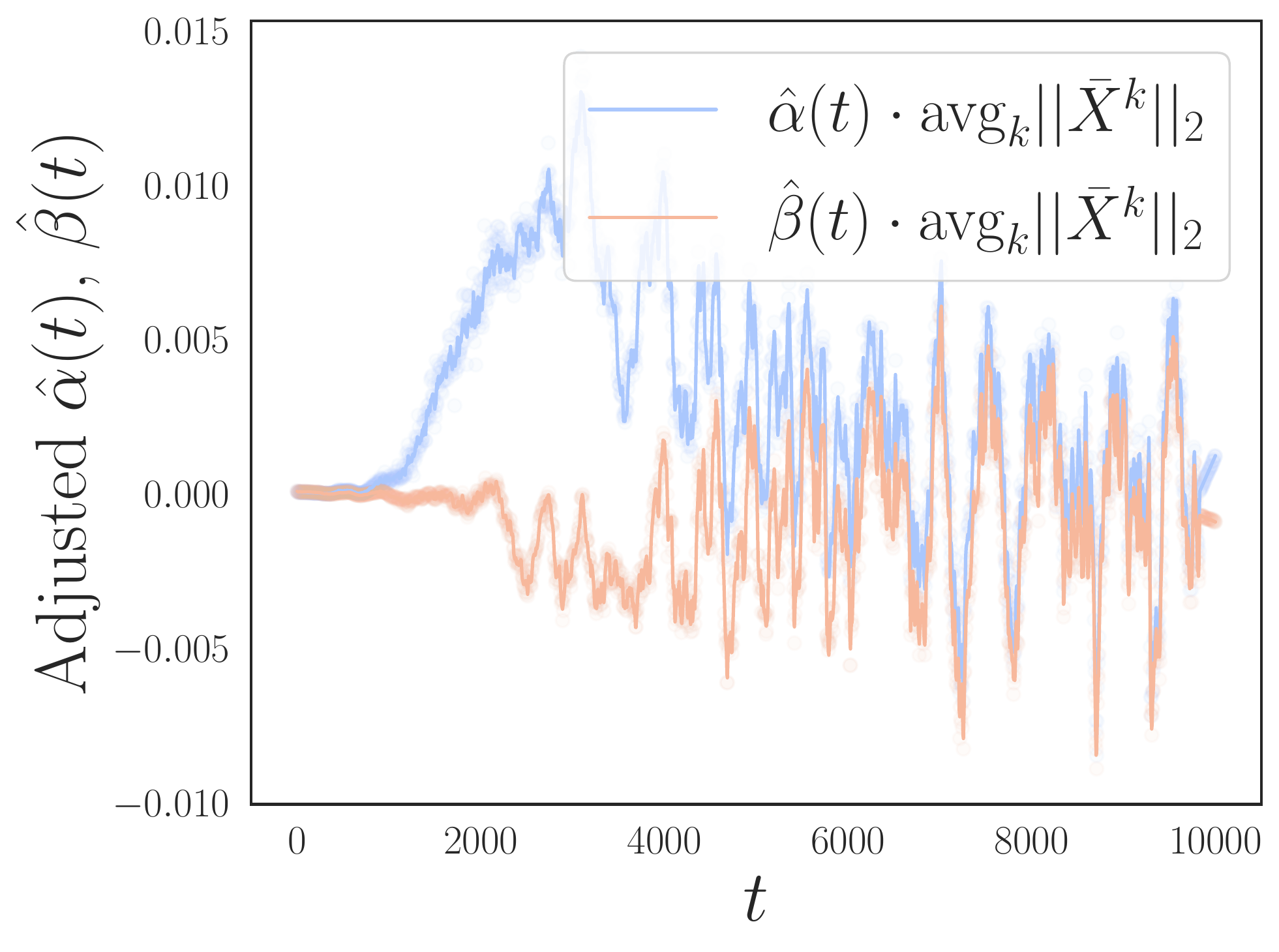}
	    \caption{$\perr=0$ (\mI).} \label{fig:supp:geomxnorm:dAB:I:0}
	\end{subfigure}%
	\begin{subfigure}[t]{0.25\textwidth}
	    \centering
	    \includegraphics[width=\linewidth]{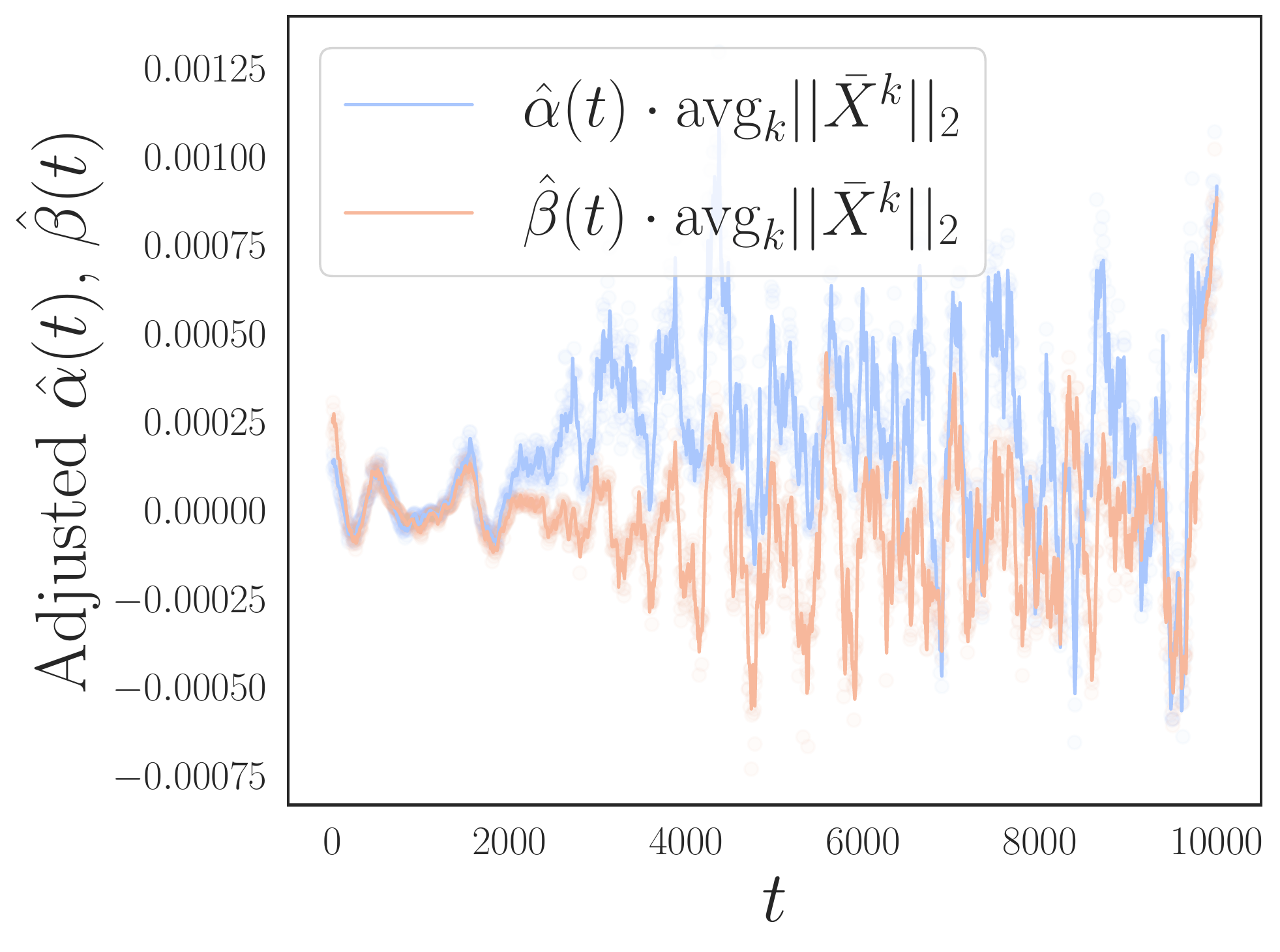}
	    \caption{$\perr=0.3$ (\mI).} \label{fig:supp:geomxnorm:dAB:I:3}
	\end{subfigure}%
	\begin{subfigure}[t]{0.25\textwidth}
	    \centering
	    \includegraphics[width=\linewidth]{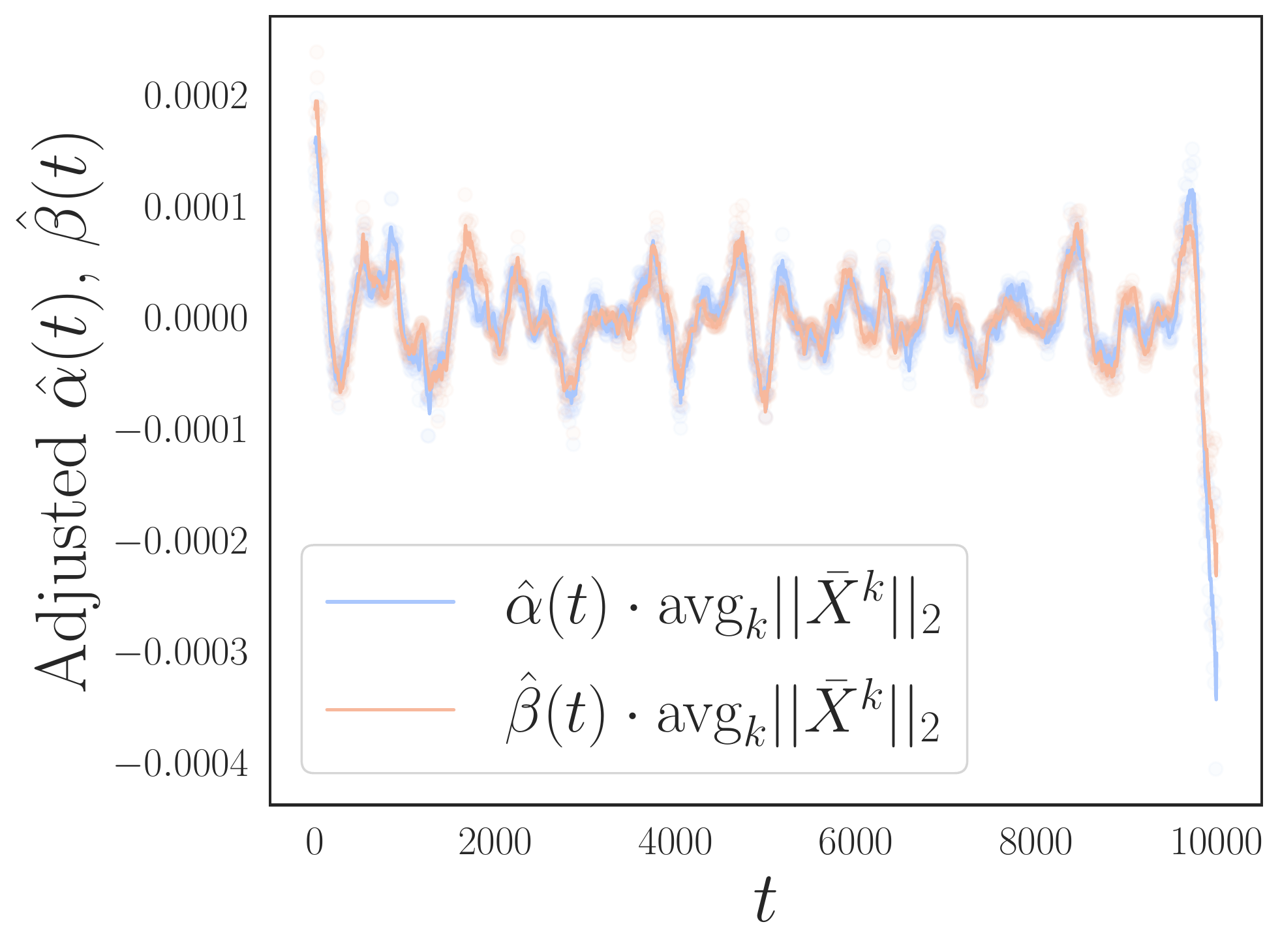}
	    \caption{$\perr=\frac{2}{3}$ (\mI).} \label{fig:supp:geomxnorm:dAB:I:6}
	\end{subfigure}%
	\begin{subfigure}[t]{0.25\textwidth}
	    \centering
	    \includegraphics[width=\linewidth]{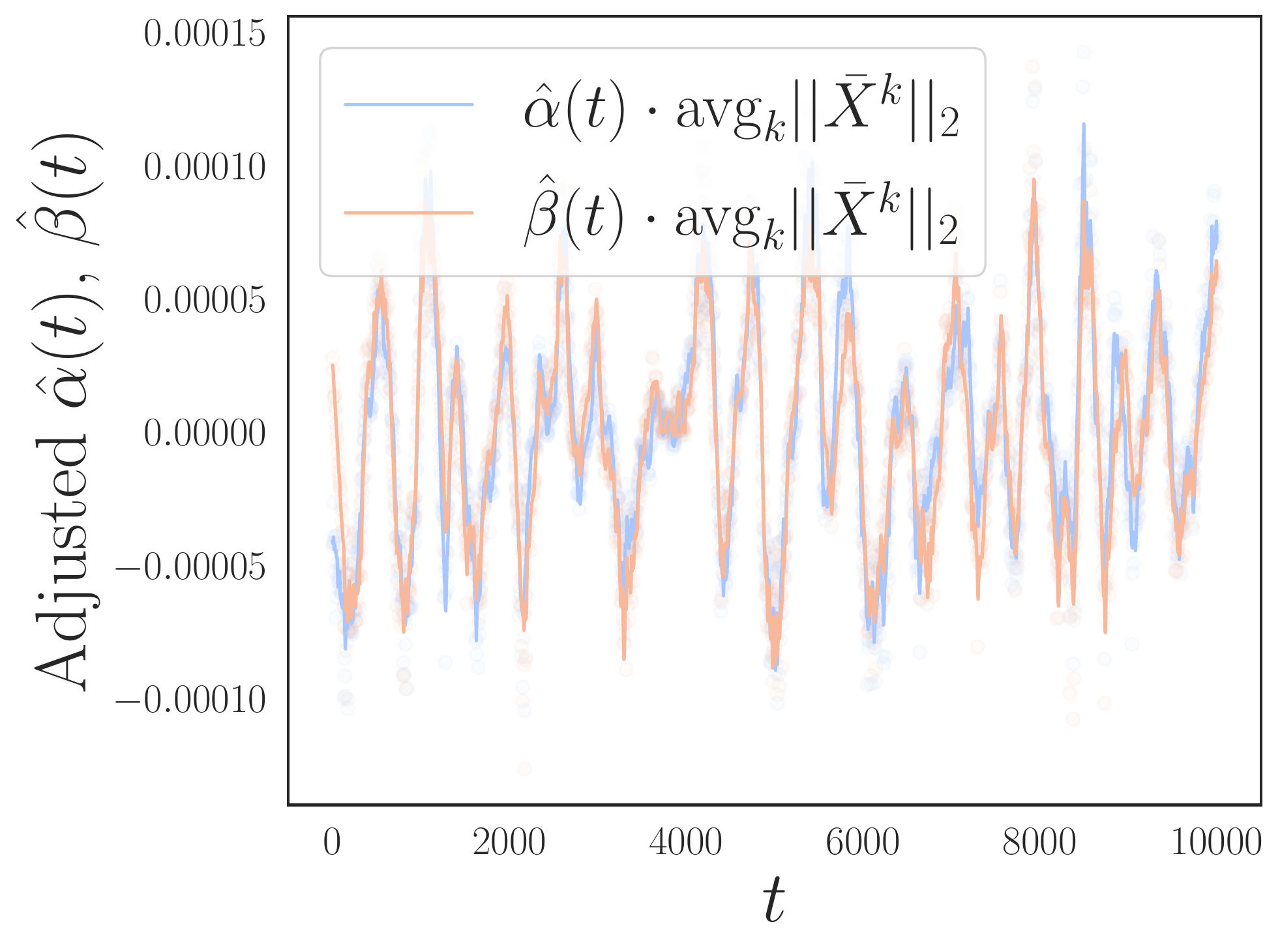}
	    \caption{$\perr=0.8$ (\mI).} \label{fig:supp:geomxnorm:dAB:I:8}
	\end{subfigure}\\
	\begin{subfigure}[t]{0.25\textwidth}
	    \centering
	    \includegraphics[width=\linewidth]{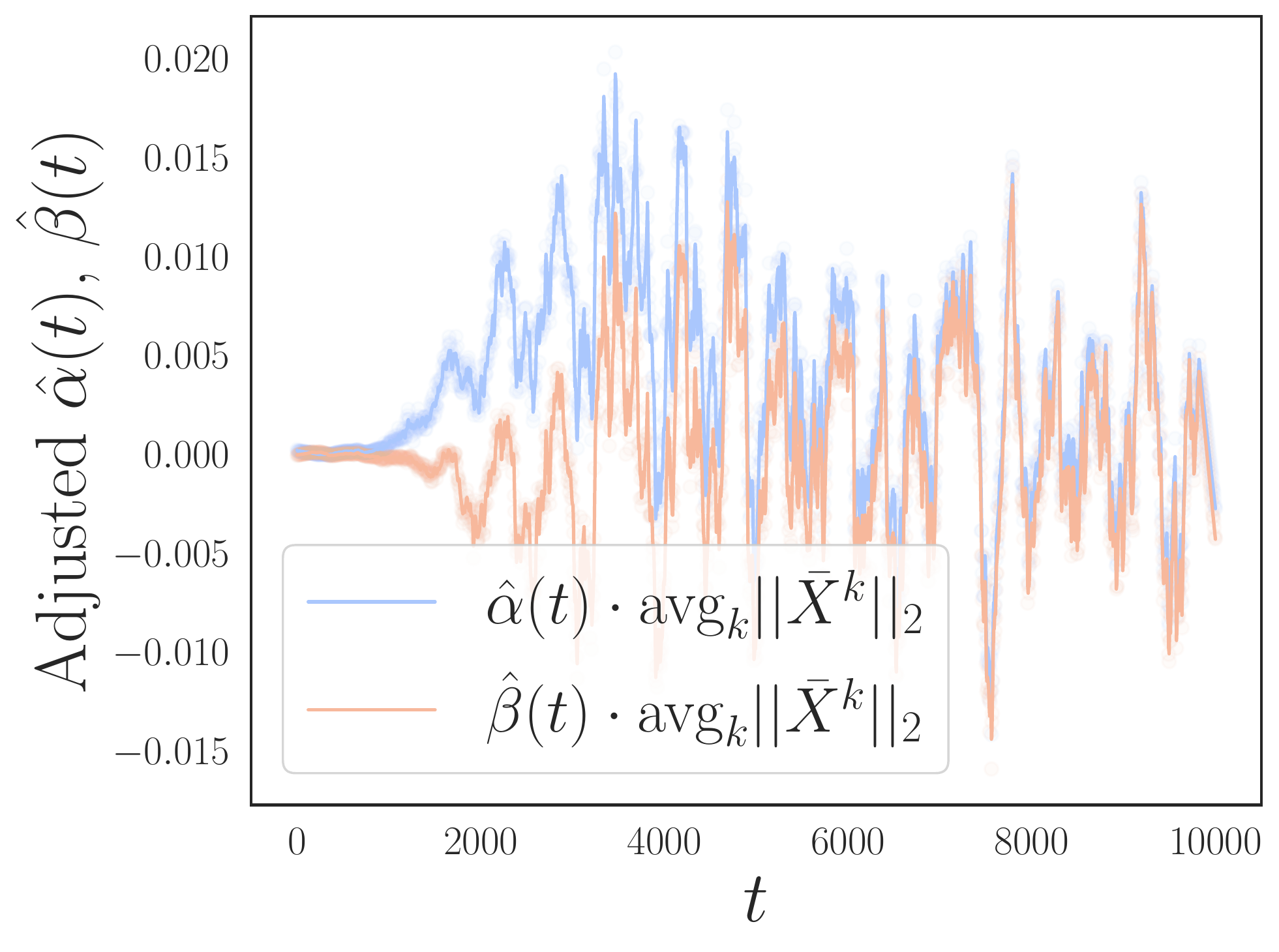}
	    \caption{$\perr=0$ (\mL).} \label{fig:supp:geomxnorm:dAB:L:0}
	\end{subfigure}%
	\begin{subfigure}[t]{0.25\textwidth}
	    \centering
	    \includegraphics[width=\linewidth]{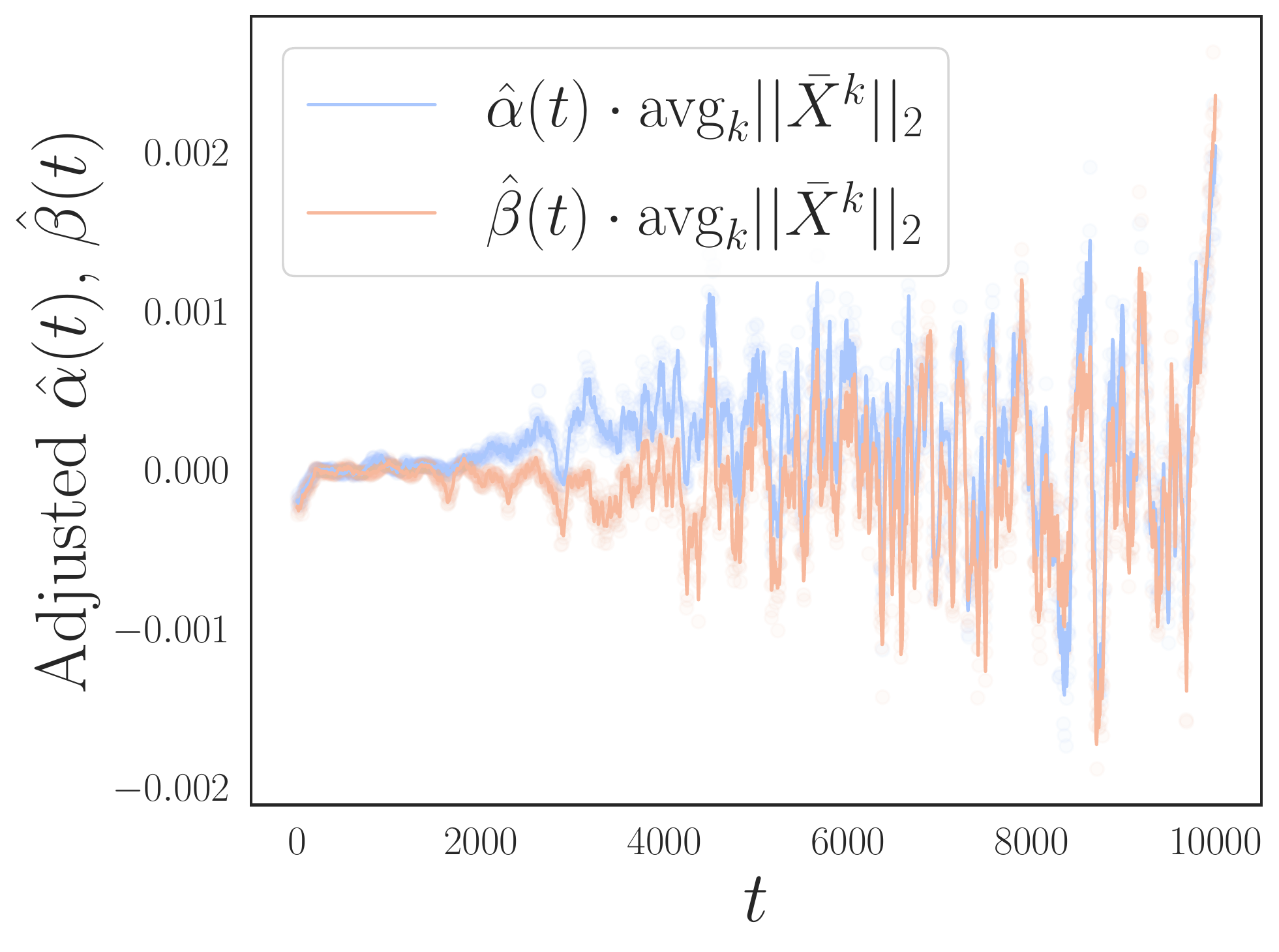}
	    \caption{$\perr=0.3$ (\mL).} \label{fig:supp:geomxnorm:dAB:L:3}
	\end{subfigure}%
	\begin{subfigure}[t]{0.25\textwidth}
	    \centering
	    \includegraphics[width=\linewidth]{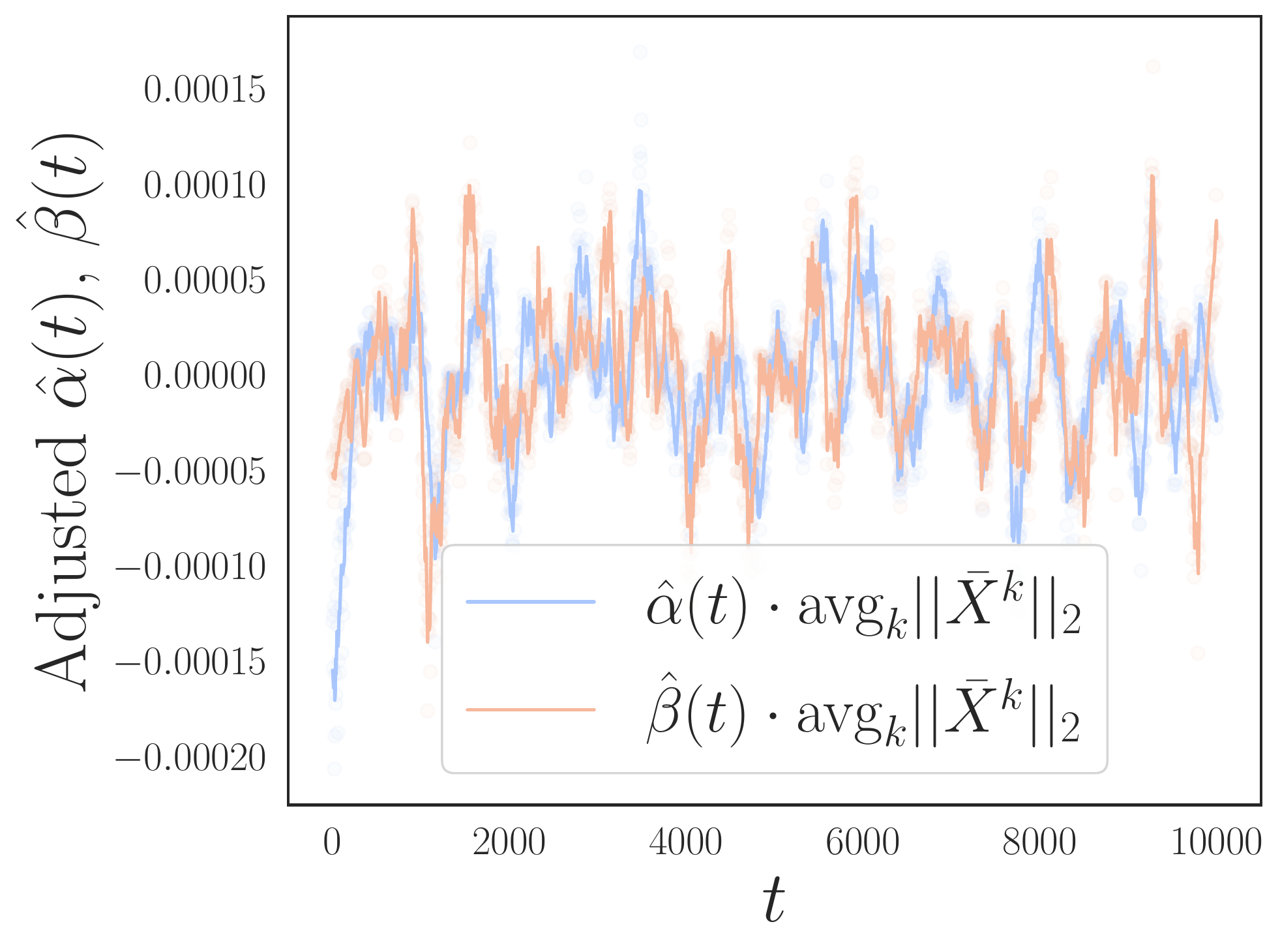}
	    \caption{$\perr=\frac{2}{3}$ (\mL).} \label{fig:supp:geomxnorm:dAB:L:6}
	\end{subfigure}%
	\begin{subfigure}[t]{0.25\textwidth}
	    \centering
	    \includegraphics[width=\linewidth]{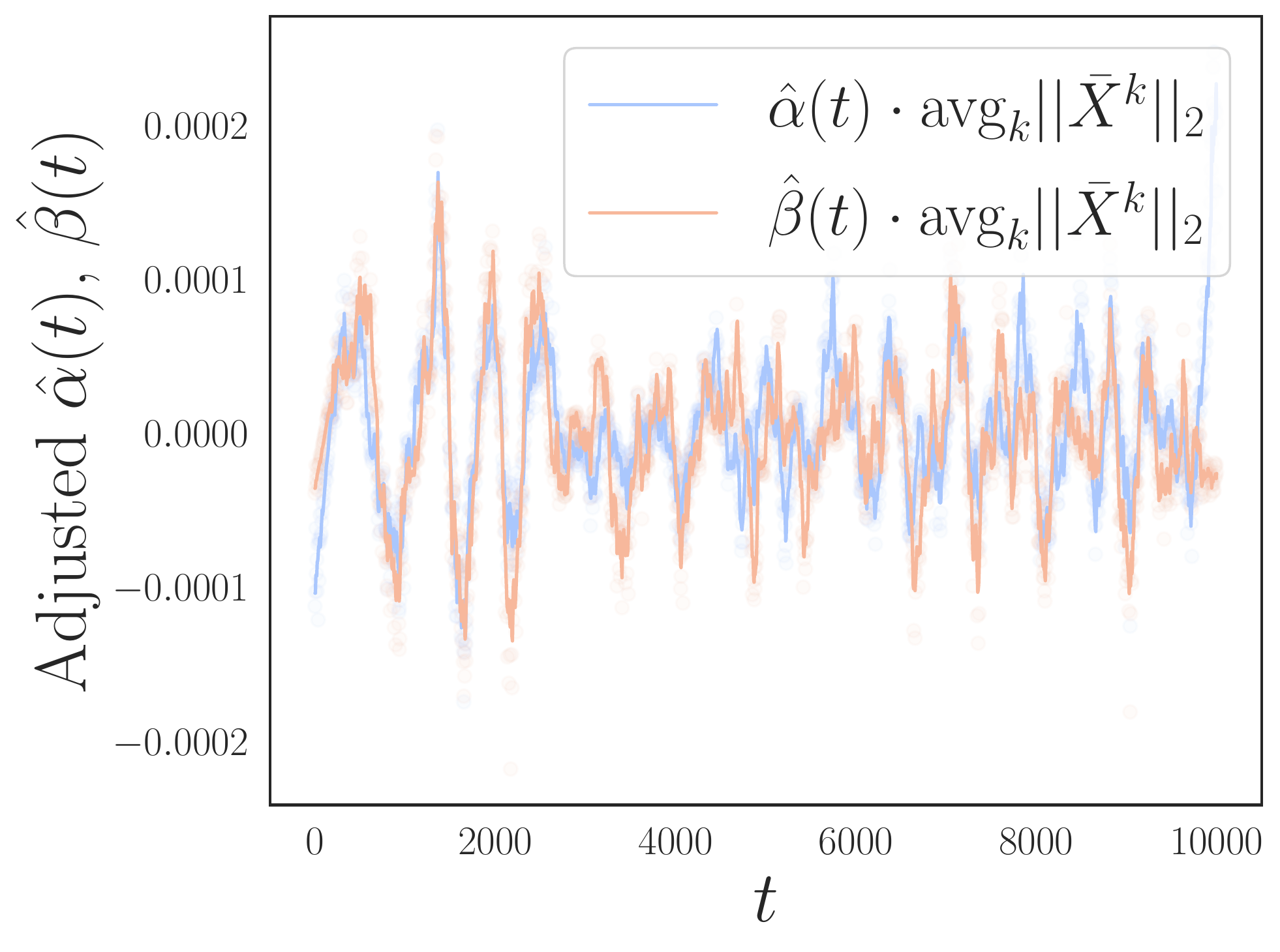}
	    \caption{$\perr=0.8$ (\mL).} \label{fig:supp:geomxnorm:dAB:L:8}
	\end{subfigure}\\
        \caption{\textbf{Effect of label corruption ratio $\perr$ (adjusted for logits norm).}
            Estimated $\hat{\alpha}(t)$, $\hat{\beta}(t)$ under \mI{} (the first two rows) and \mL{} (the last two rows) on \geomnist{}
            under various label corruptions ratios $\perr \in \{0,0.3, 2/3, 0.8\}$ are shown.
            All quantities were multiplied by the average norm of
            per-class means in each iteration, i.e.,
            Note $\hat{A}(t)$  and $\hat{B}(t)$,
            $\hat{\alpha}(t)$ and $\hat{\beta}(t)$ become indistinguishable
            when $\perr$ is large, supporting \Cref{thm:separation:mean}.
        }
        \label{fig:supp:geomxnorm:ABab}
\end{figure*}

    In our LE-SDE/ODE model \mI{} and \mL{}, each block
    of the $\H$ matrix has operator norm $1$. Hence it is of interests
    to inspect the local elasticity strengths
    $\alpha(t)$ and $\beta(t)$, when adjusted for the evolution
    of the norm of $\bar{\X}^k(t)$. As a surrogate, we simply
    multiply the strengths by the average norm, $\avg_k \norm{\bar{\X}^k(t)}_2$ and show the results in
    \Cref{fig:supp:geomxnorm:ABab}. Again we use a large
    window size $\omega=351$ in the Savitzky–Golay filter
    to highlight the general trends.
    We observe that when $\perr=0$, the general trend is similar
    to the unadjusted versions: $\alpha(t)$ has an initial
    increasing stage, and then converges to the vicinity of zero.
    We also note that $\alpha(t)$ and $\beta(t)$ become
    more indistinguishable as we increase $\perr$, similarly
    to the unadjusted cases \Cref{fig:supp:geomperr}.

\paragraph{Simulations of the LE-ODE.}

\begin{figure*}
    \centering
    \begin{subfigure}[t]{0.25\textwidth}
	    \centering
	    \includegraphics[width=\linewidth]{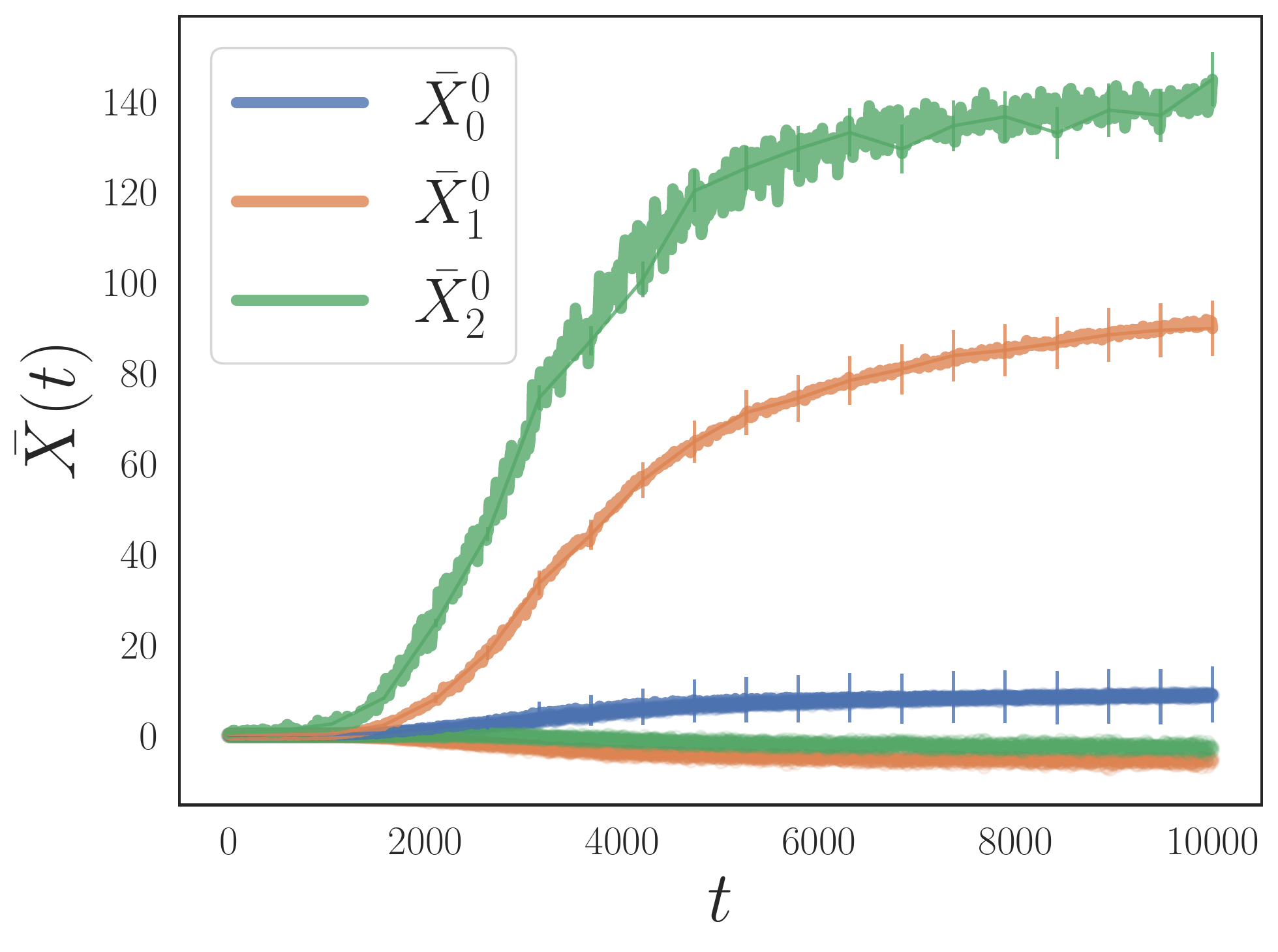}
	    \caption{\geomnist{} (\mI{}). } 
	\end{subfigure}%
	   \begin{subfigure}[t]{0.25\textwidth}
	    \centering
	    \includegraphics[width=\linewidth]{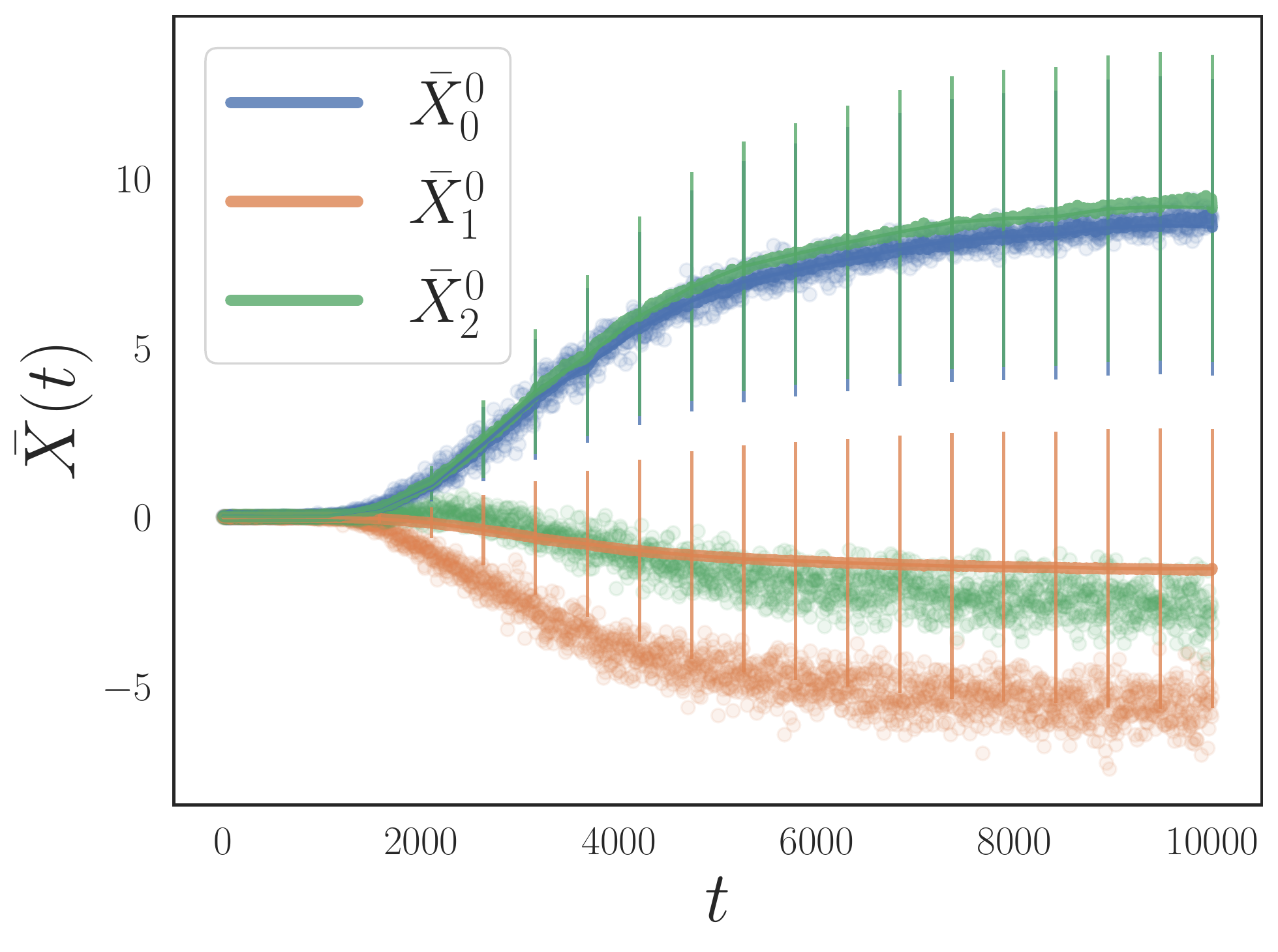}
	    \caption{\geomnist{} (\mH{}).} 
	\end{subfigure}%
    \begin{subfigure}[t]{0.25\textwidth}
	    \centering
	    \includegraphics[width=\linewidth]{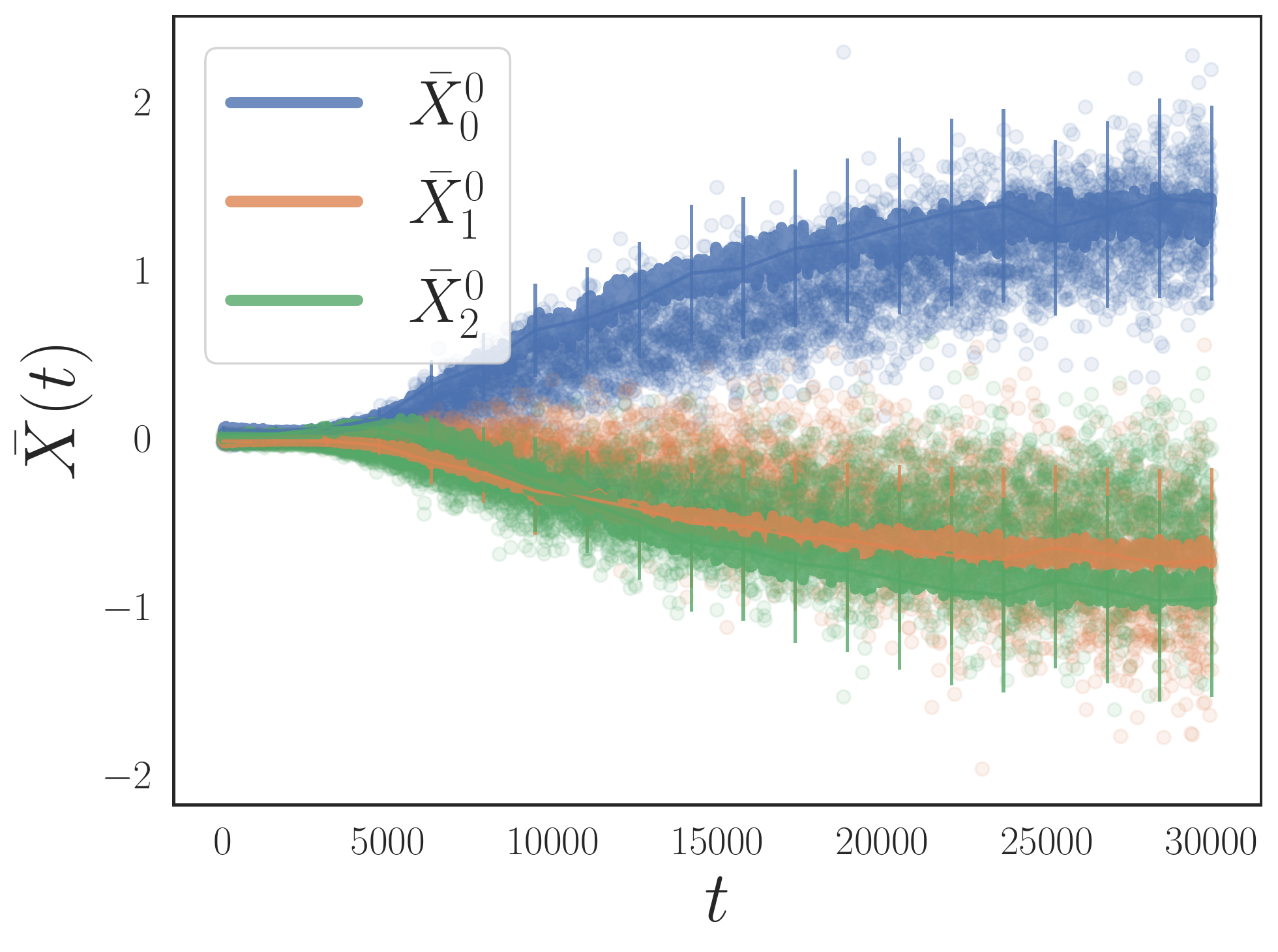}
	    \caption{\cifar{} (\mI).}
	\end{subfigure}%
    \begin{subfigure}[t]{0.25\textwidth}
	    \centering
	    \includegraphics[width=\linewidth]{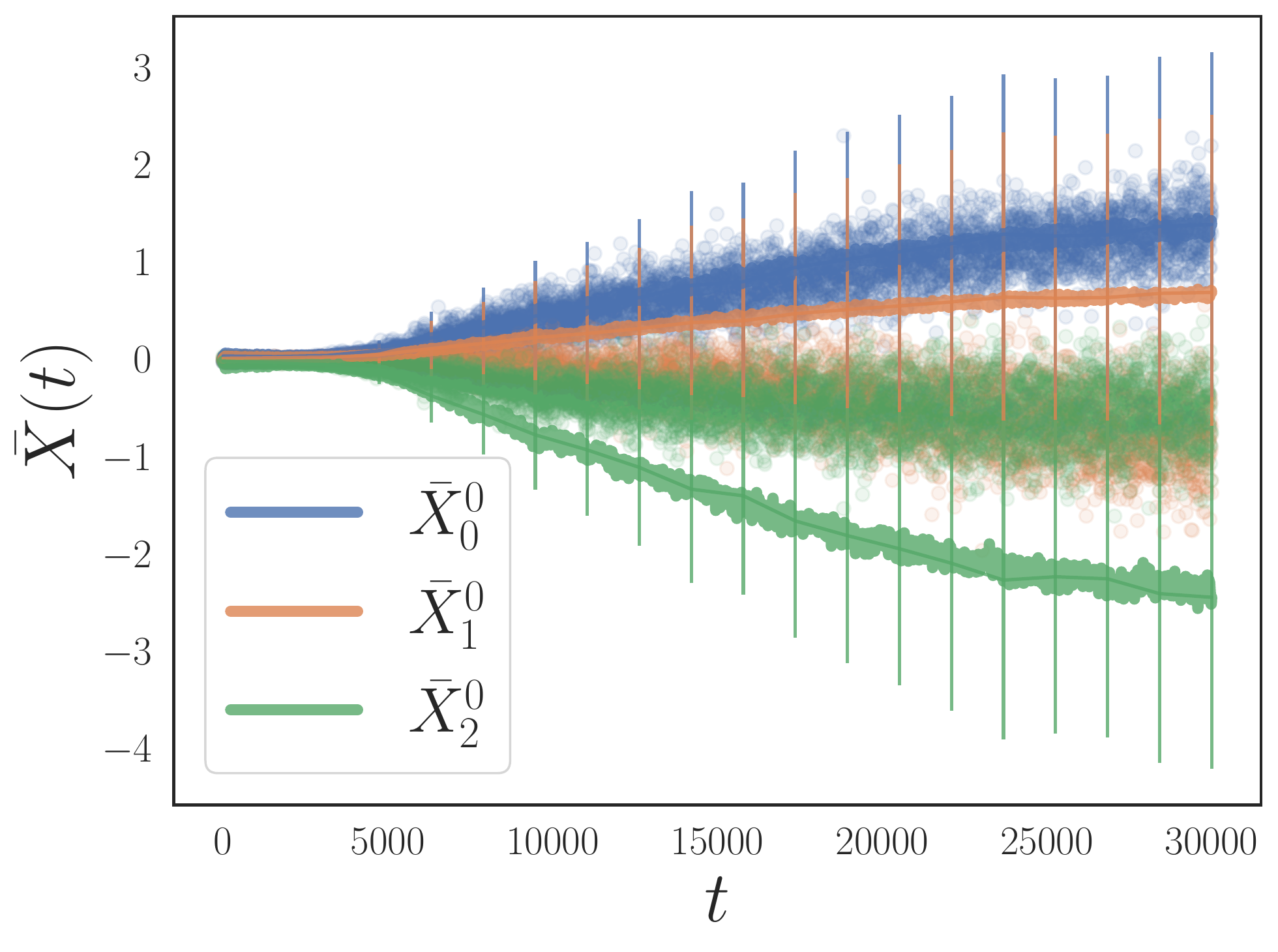}
	    \caption{\cifar{} (\mH).} 
	\end{subfigure}
        \caption{
            \textbf{Simulated LE-ODE solutions under the \mI{}
            versus genuine
            dynamics.} We use $\hat{\alpha}(t)$ and $\hat{\beta}(t)$
            estimated from \mI{} ((a) and (c)) or \mH{} 
            ((b) and (d)) and numerically simulate
            the solution under the \mI{}. The results were
            overlaid with
            true dynamics from neural nets.
            Note that the results are uniformly bad, indicating \mI{}
            is not expressive enough to capture the genuine dynamics
            in deep neural nets.
            }
        \label{fig:supp:leode:I}
        \vspace{-0.6cm}
\end{figure*}

\begin{figure*}
    \centering
    \begin{subfigure}[t]{0.33\textwidth}
	    \centering
	    \includegraphics[width=\linewidth]{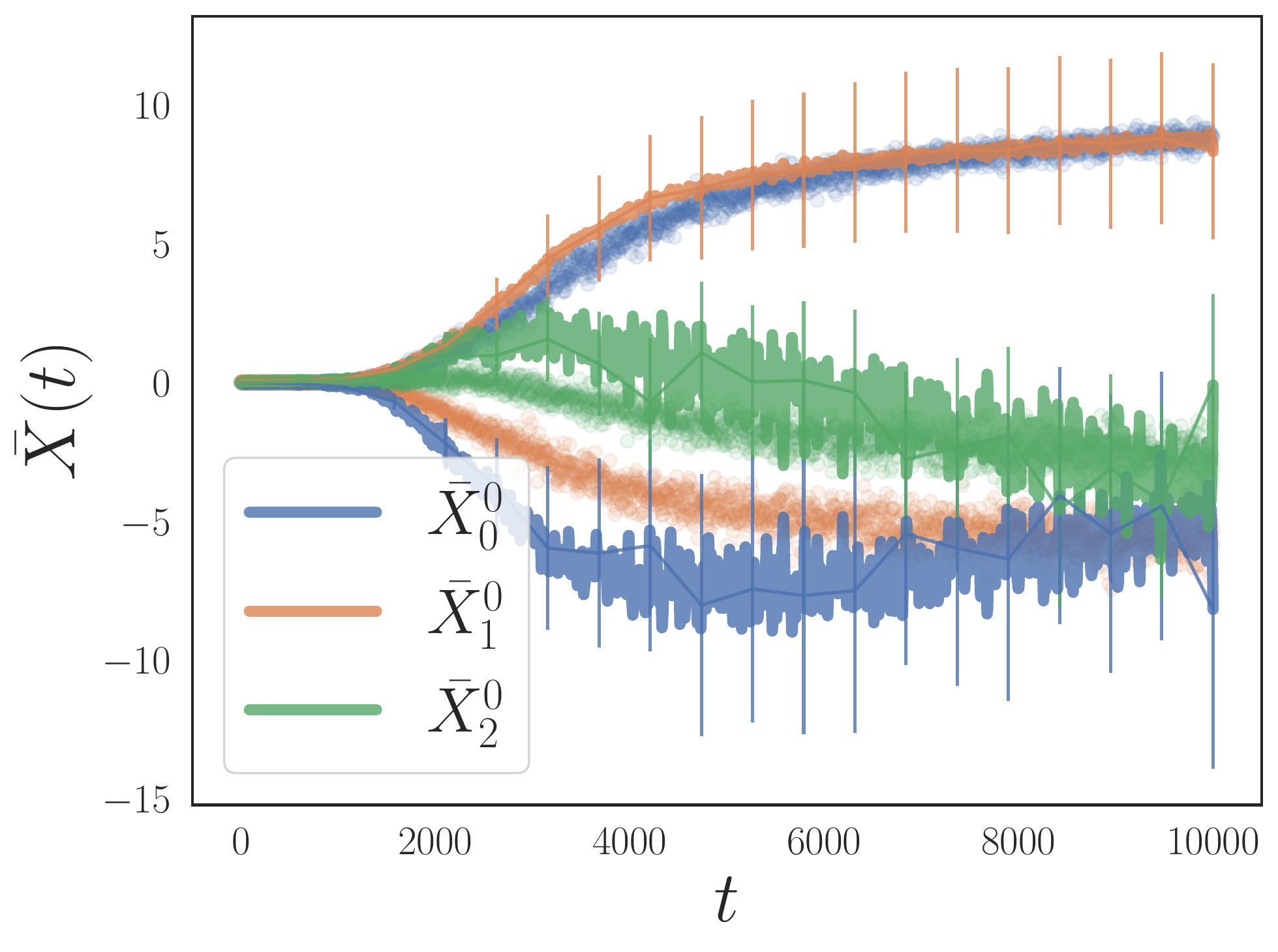}
	    \caption{$k=0$ (\geomnist,\mI).} \label{fig:supp:leodesim:geomnist:I:0}
	\end{subfigure}~
	\begin{subfigure}[t]{0.33\textwidth}
	    \centering
	    \includegraphics[width=\linewidth]{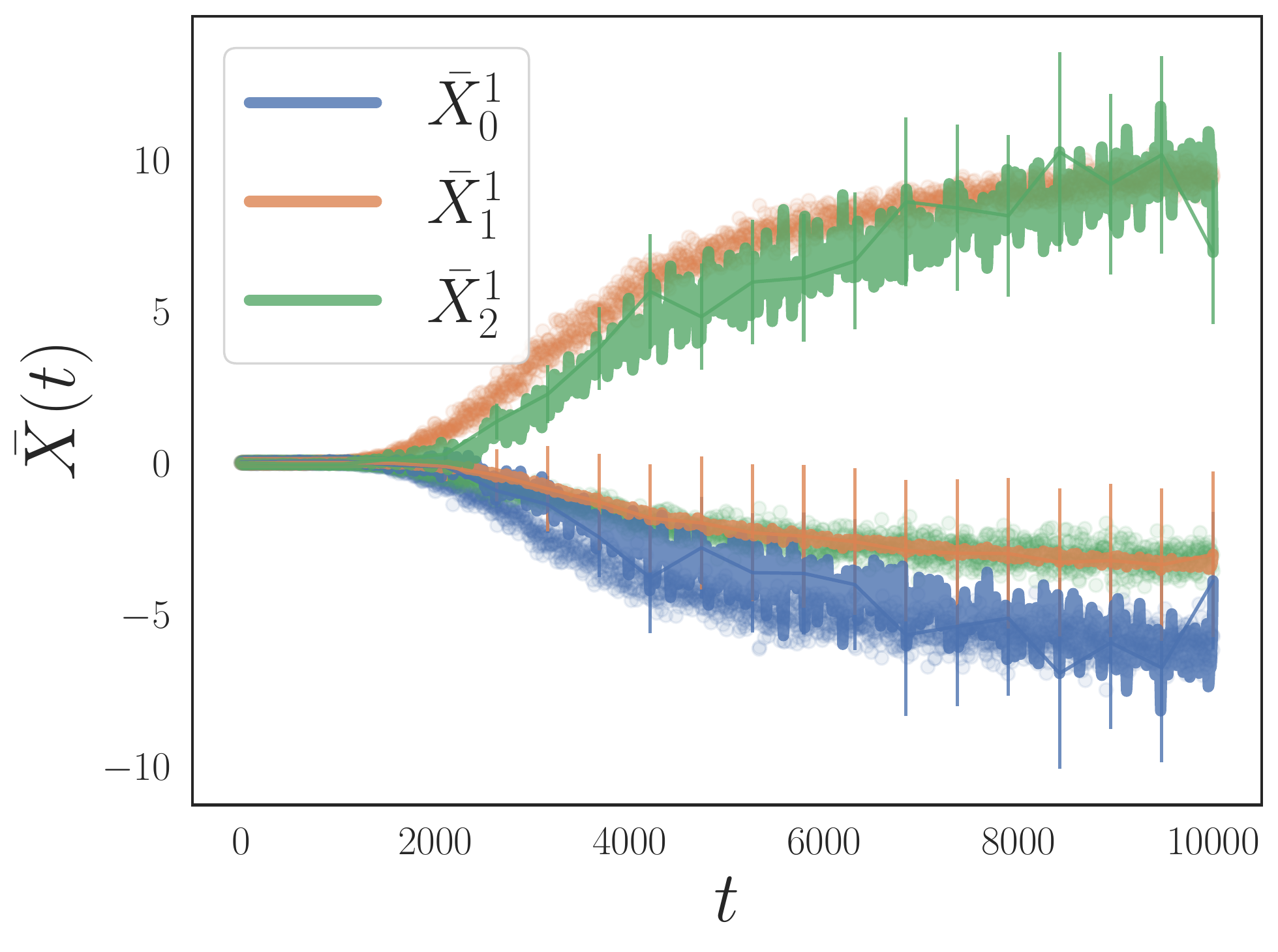}
	    \caption{$k=1$ (\geomnist,\mI).} \label{fig:supp:leodesim:geomnist:I:1}
	\end{subfigure}~
	\begin{subfigure}[t]{0.33\textwidth}
	    \centering
	    \includegraphics[width=\linewidth]{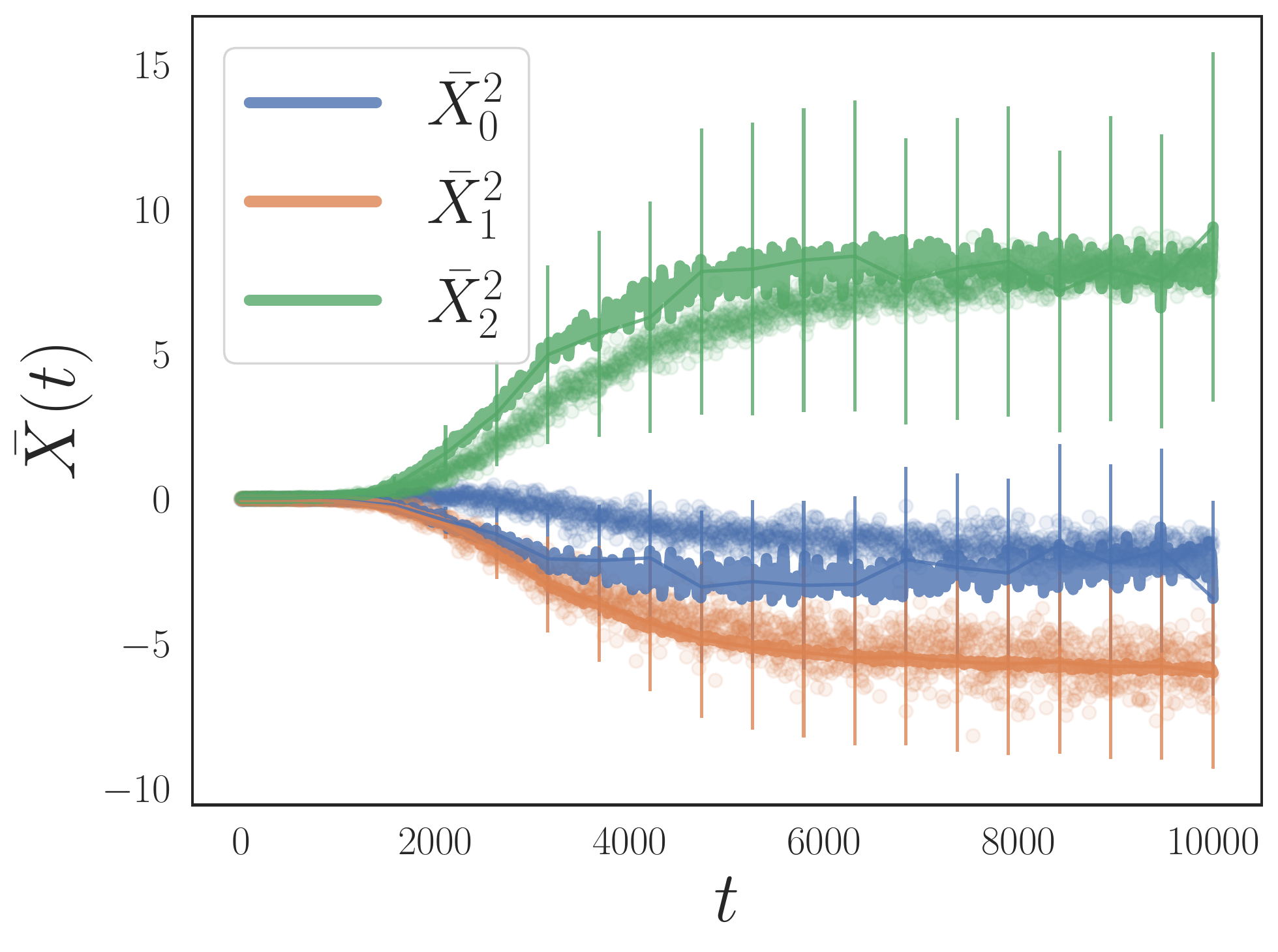}
	    \caption{$k=2$ (\geomnist,\mI).} \label{fig:supp:leodesim:geomnist:I:2}
	\end{subfigure}\\
	\begin{subfigure}[t]{0.33\textwidth}
	    \centering
	    \includegraphics[width=\linewidth]{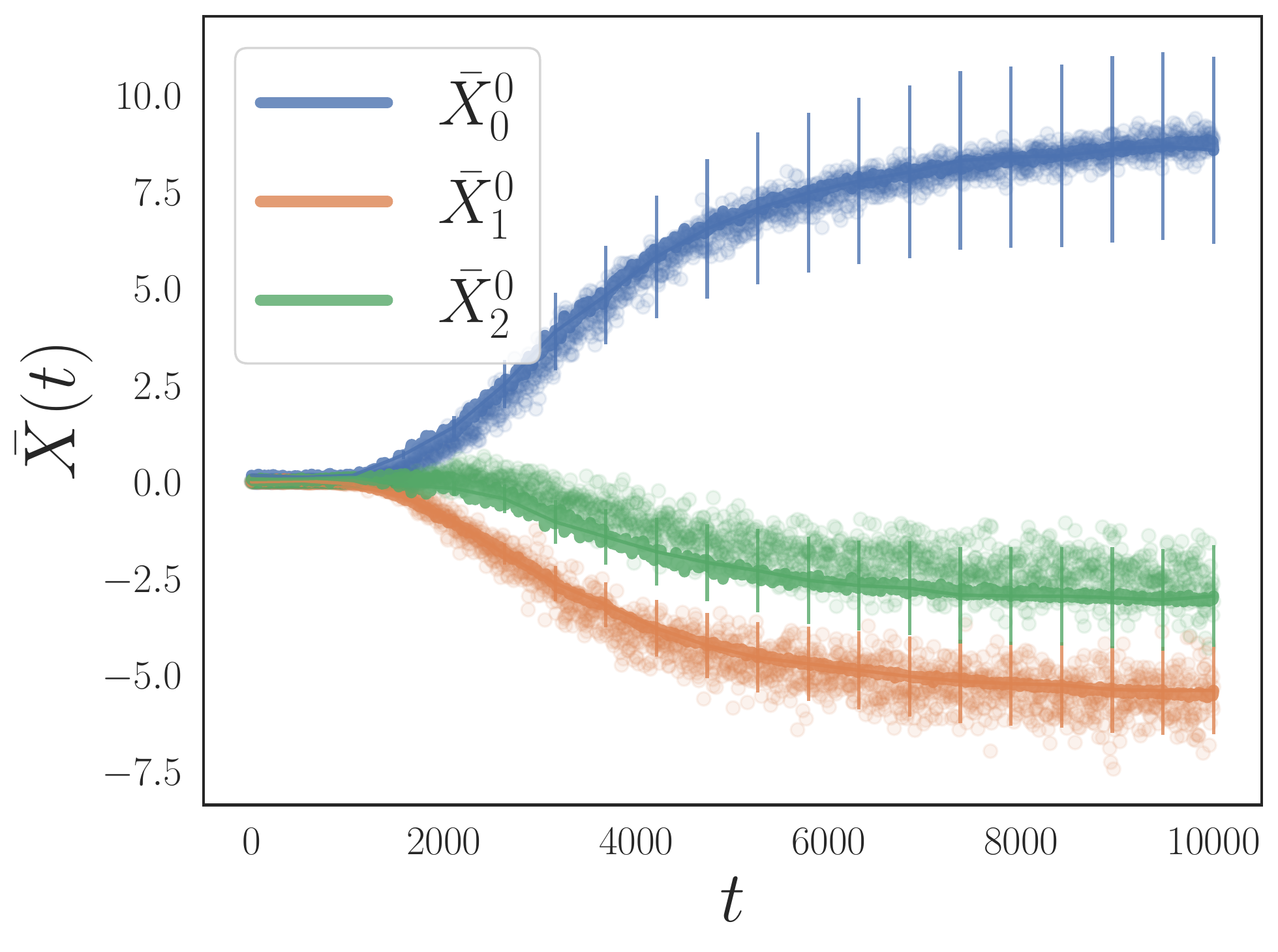}
	    \caption{$k=0$ (\geomnist,\mL).} \label{fig:supp:leodesim:geomnist:L:0}
	\end{subfigure}~
	\begin{subfigure}[t]{0.33\textwidth}
	    \centering
	    \includegraphics[width=\linewidth]{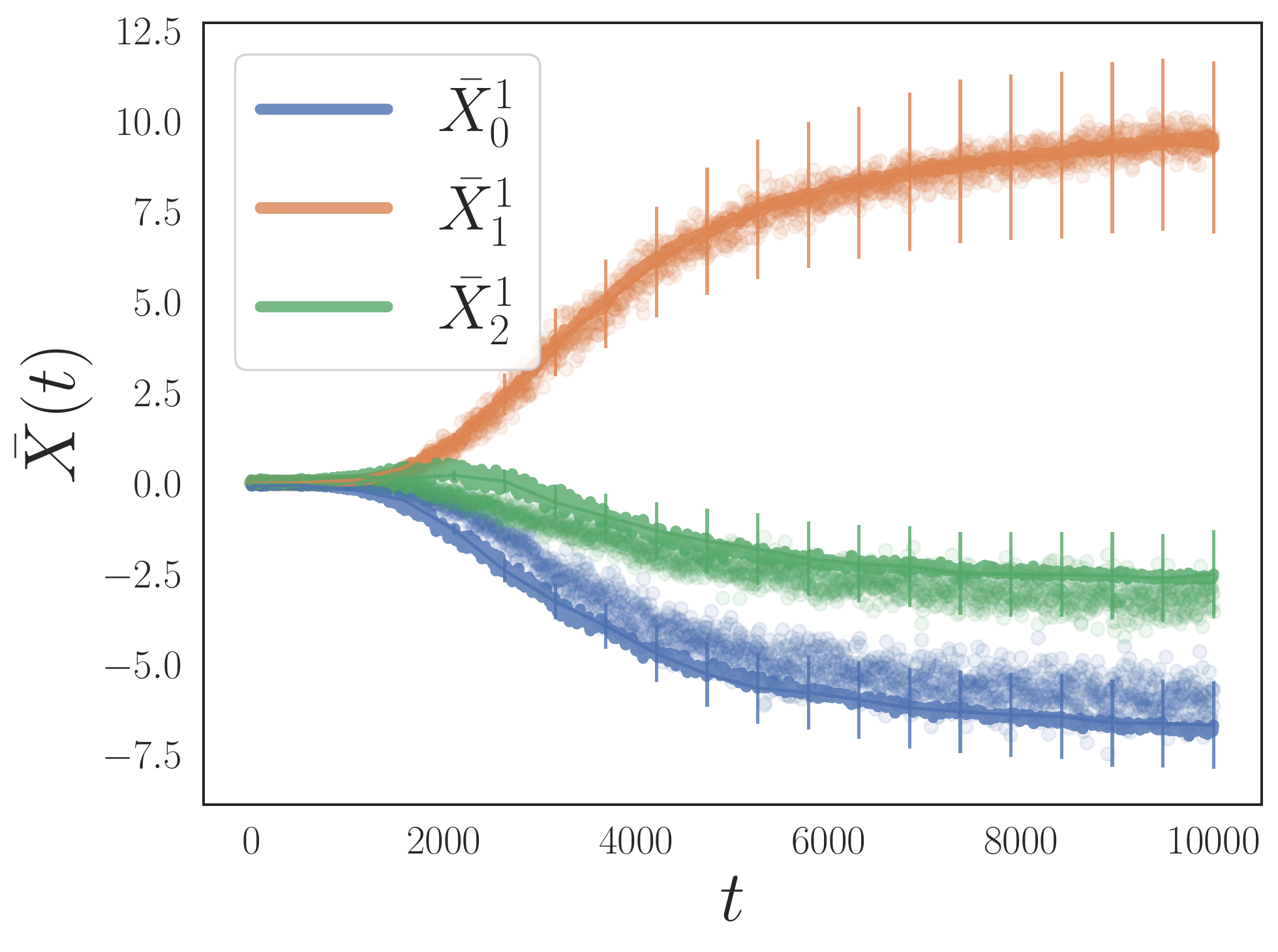}
	    \caption{$k=1$ (\geomnist,\mL).} \label{fig:supp:leodesim:geomnist:L:1}
	\end{subfigure}~
	\begin{subfigure}[t]{0.33\textwidth}
	    \centering
	    \includegraphics[width=\linewidth]{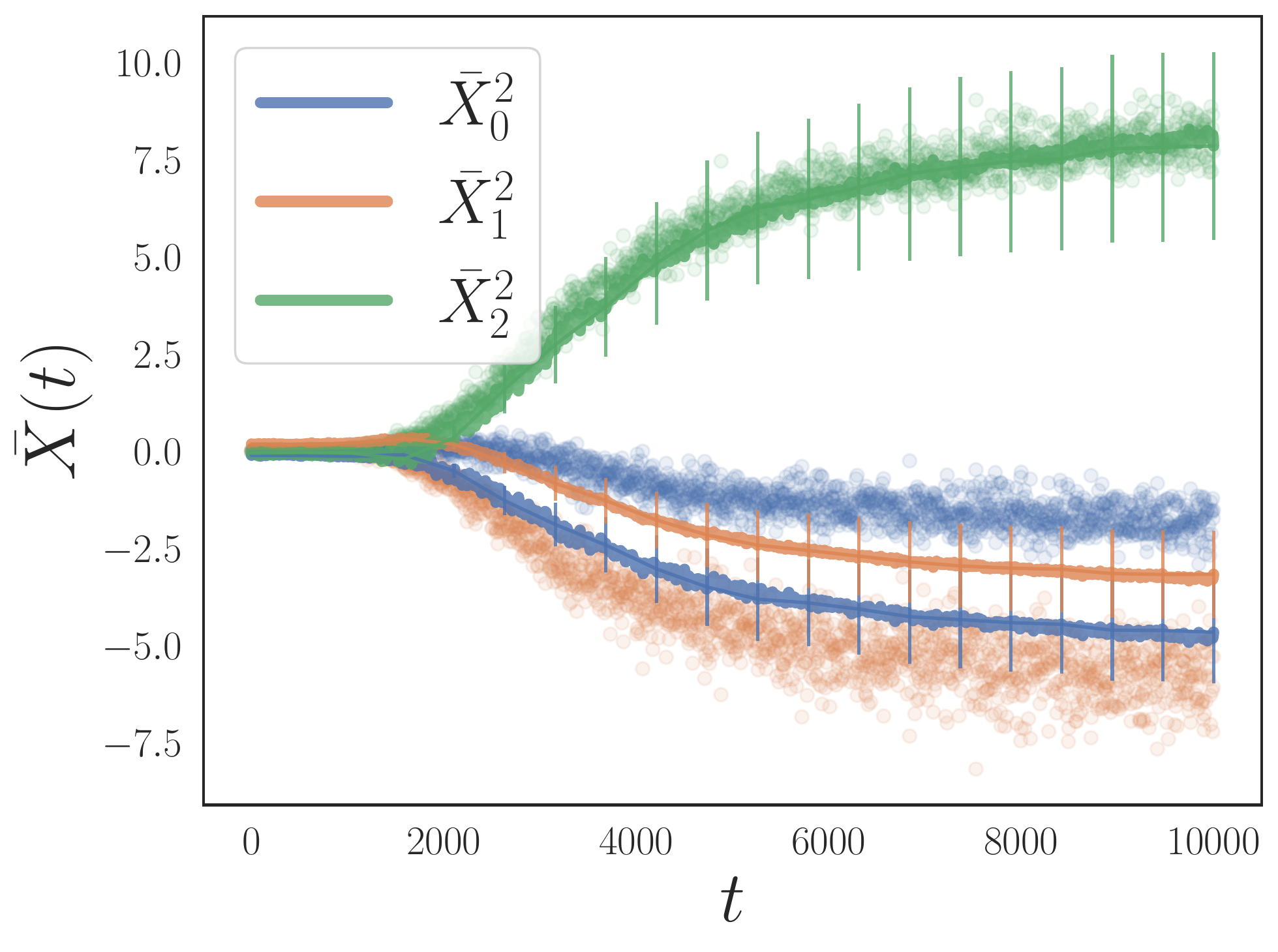}
	    \caption{$k=2$ (\geomnist,\mL).} \label{fig:supp:leodesim:geomnist:L:2}
	\end{subfigure}\\
	\begin{subfigure}[t]{0.33\textwidth}
	    \centering
	    \includegraphics[width=\linewidth]{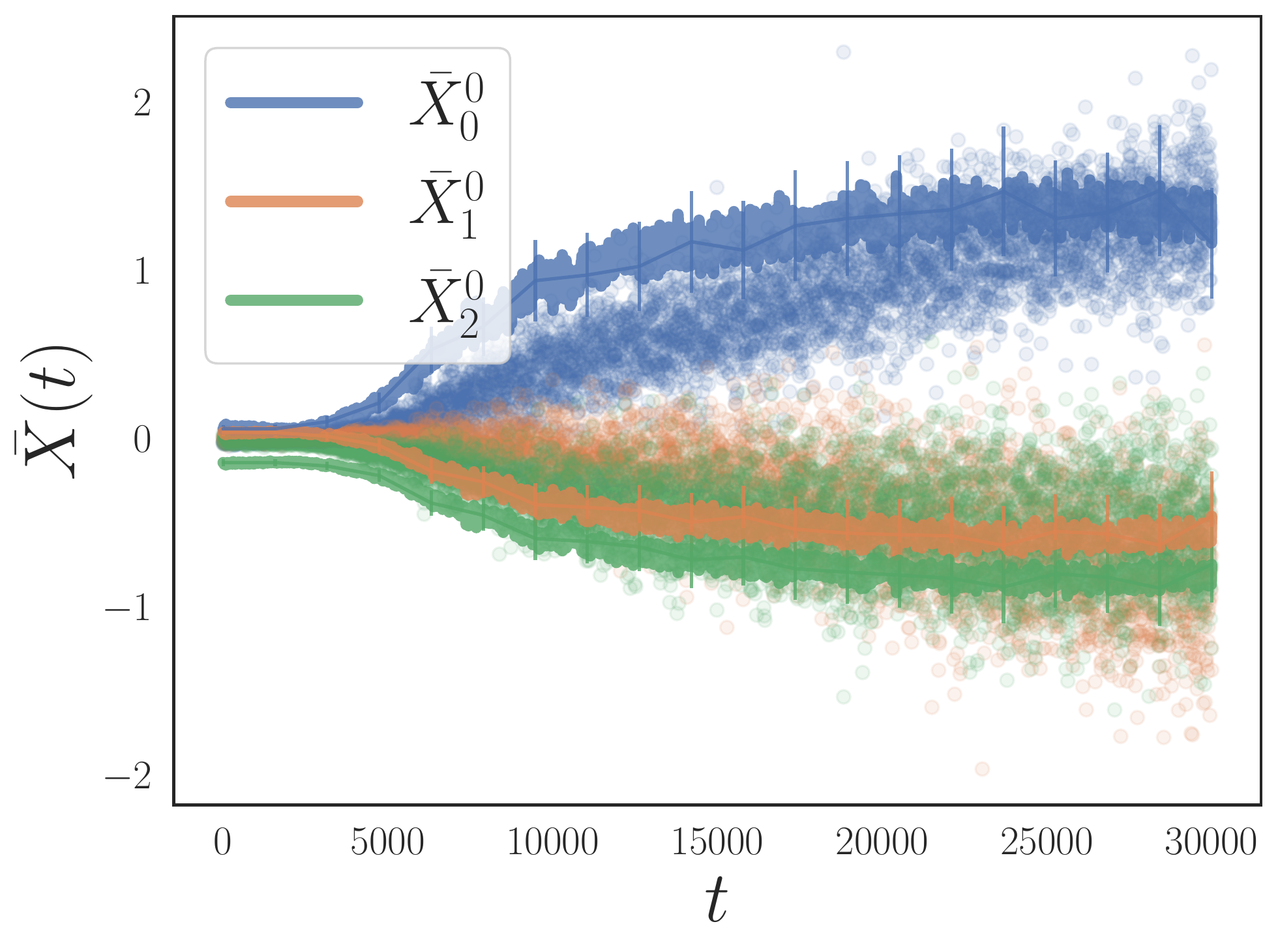}
	    \caption{$k=0$ (\cifar,\mI).} \label{fig:supp:leodesim:cifar10K3:I:0}
	\end{subfigure}~
	\begin{subfigure}[t]{0.33\textwidth}
	    \centering
	    \includegraphics[width=\linewidth]{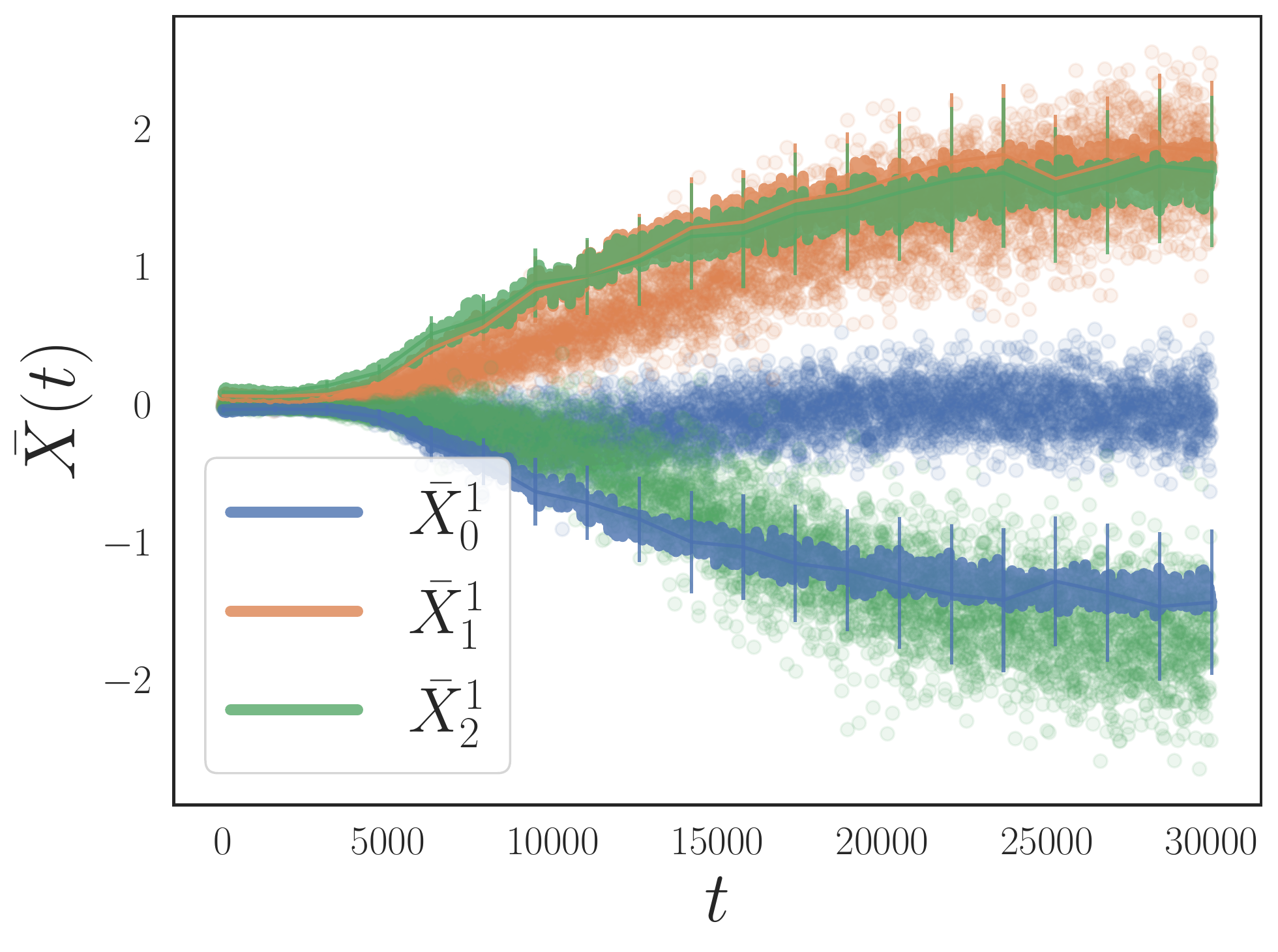}
	    \caption{$k=1$ (\cifar,\mI).} \label{fig:supp:leodesim:cifar10K3:I:1}
	\end{subfigure}~
	\begin{subfigure}[t]{0.33\textwidth}
	    \centering
	    \includegraphics[width=\linewidth]{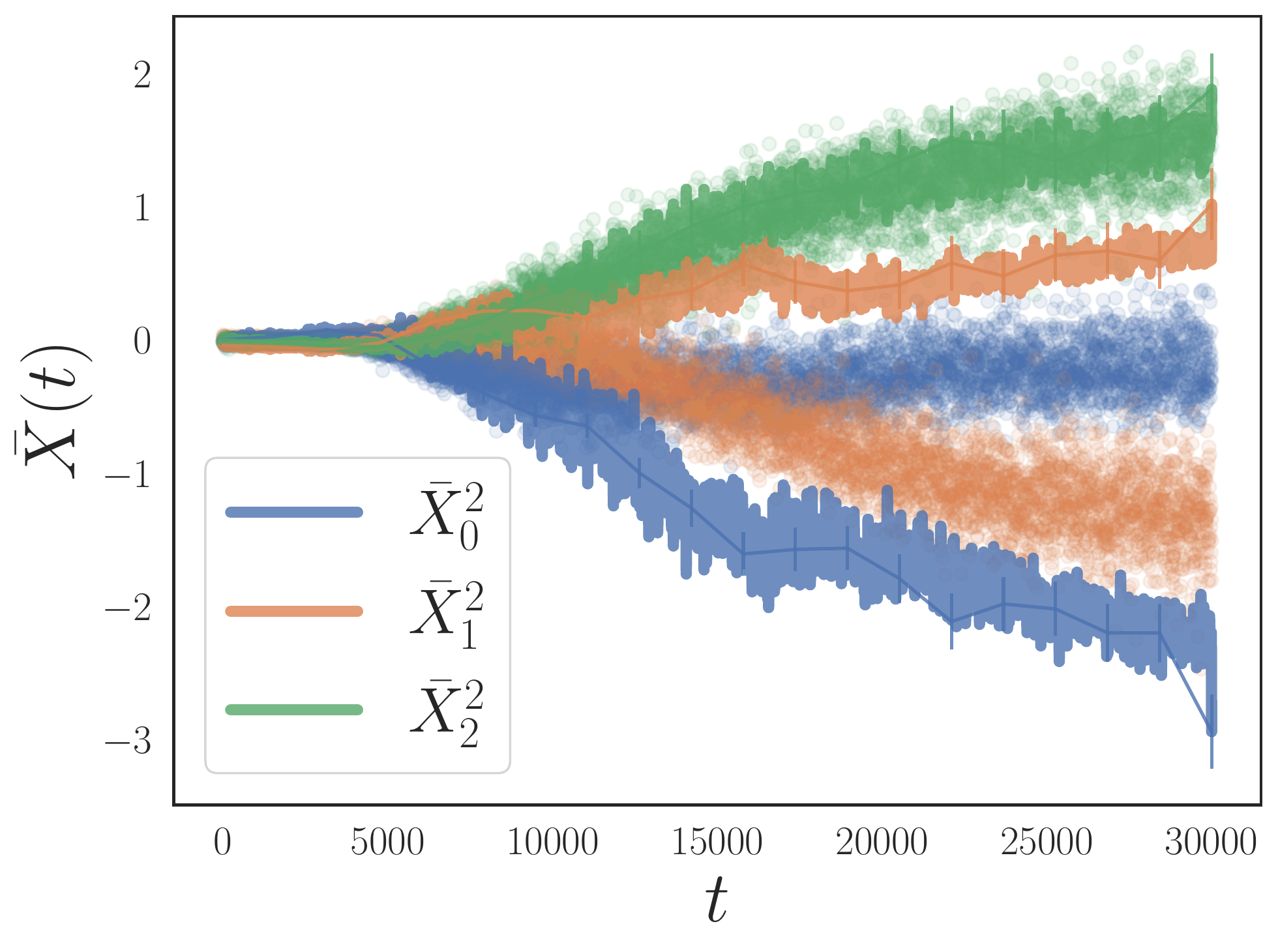}
	    \caption{$k=2$ (\cifar,\mI).} \label{fig:supp:leodesim:cifar10K3:I:2}
	\end{subfigure}\\
	\begin{subfigure}[t]{0.33\textwidth}
	    \centering
	    \includegraphics[width=\linewidth]{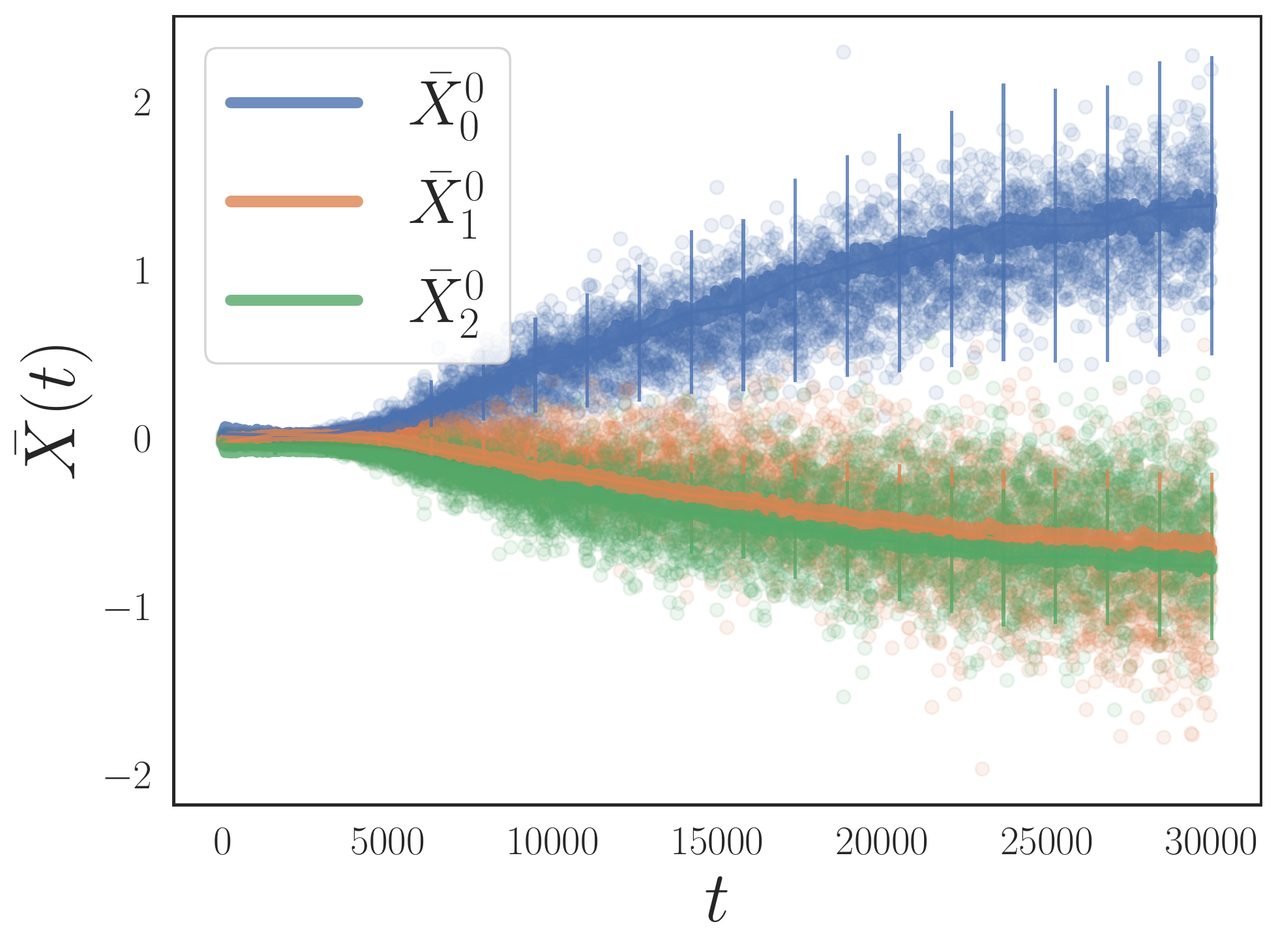}
	    \caption{$k=0$ (\cifar,\mL).} \label{fig:supp:leodesim:cifar10K3:L:0}
	\end{subfigure}~
	\begin{subfigure}[t]{0.33\textwidth}
	    \centering
	    \includegraphics[width=\linewidth]{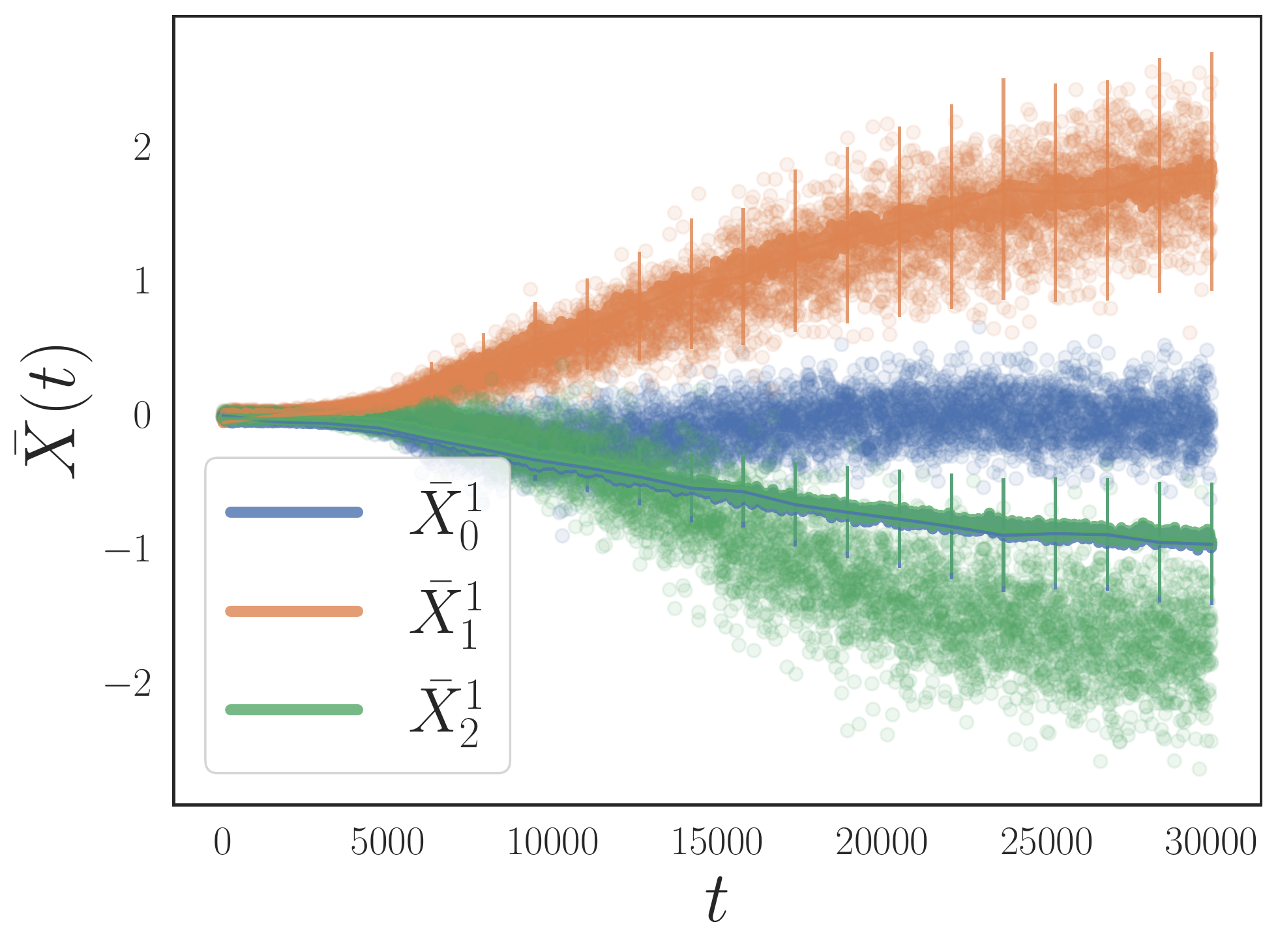}
	    \caption{$k=1$ (\cifar,\mL).} \label{fig:supp:leodesim:cifar10K3:L:1}
	\end{subfigure}~
	\begin{subfigure}[t]{0.33\textwidth}
	    \centering
	    \includegraphics[width=\linewidth]{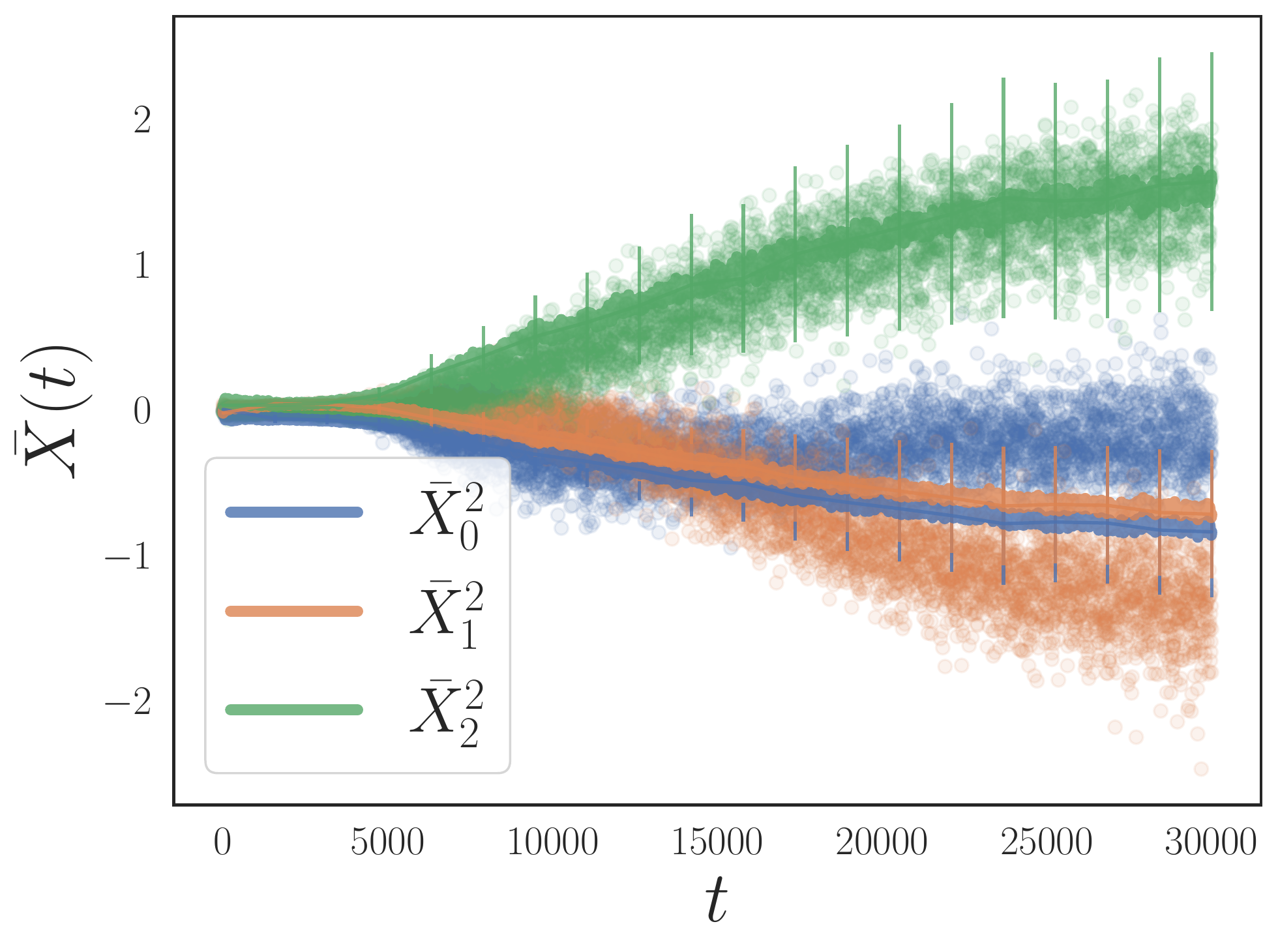}
	    \caption{$k=2$ (\cifar,\mL).} \label{fig:supp:leodesim:cifar10K3:L:2}
	\end{subfigure}\\
        \caption{\textbf{Simulated LE-ODE paths under the \mL{} versus genuine
            dynamics.} The simulation is done under the \mH{} with estimated
        	$\hat{\alpha}(t)$ and $\hat{\beta}(t)$ from the \mI{} (the first and the third rows)
        	and the \mL{} (the second and the fourth rows), on the \geomnist{} (the first two rows)
        	and \cifar{} (the last two rows). We show the trajectories for each class $k\in[3]$,
        	where $\bar{X}^k_l(t)$ denotes the path of the $l$-th logit of the $k$-th class for
        	$k,l\in[3]$.
        }
        \label{fig:supp:leodesim}
\end{figure*}

    With $\alpha(t)$ and $\beta(t)$ estimated (using either the \mI{}
    or the \mL), we can simulate the LE-ODE
    under either \mI{} or \mL{}. 
    In our setup, the initial value for the $k$th class
    is set to be $\bzeta^K\distas \NORMAL_K(\bzero, \sigma_k\I_{K^2})$
    for all $k\in [K]$ with $\sigma_k = \norm{\bar{\X}^k(0)}_2/\sqrt{K}$
    with $\bar{\X}^k(0)$ sampled from simulations in DNNs.
    Empirically, we find that 
    simulations under the \mI{}
    does not generate faithful trajectories compared with the ground truth
    (genuine dynamics from simulations on deep neural nets),
    as shown in \Cref{fig:supp:leode:I}.
    Hence in \Cref{fig:supp:leodesim},
    we only show the case when the simulation is done under the \mL where the
    captions indicate which model (\mI{} or \mL) the estimation
    of $\alpha(t)$ and $\beta(t)$ is performed.
    Here $\bar{X}^k_i(t)\in\sR$ denotes the $i$-th logit from the
    per-class mean logits
    vector of the $k$-th class; 
    as explained in \Cref{sec:app:dynamics},
    a well-trained model should have the $k$-th logit
    being the largest among all $i\in[K]$ in $\bar{X}^k_i$ 
    when $t$ is sufficiently large.
    Since the estimation of $\alpha(t)$ and $\beta(t)$ relies on numerical
    differentiation, which smooths the data and reduces 
    their magnitudes,
    we manually align the 
    $k$-th logit from the $k$-th per-class mean vector,
    $\bar{X}^k_i$, with that from the ground truth.
    All simulations are performed with $K=3$ and
    for $N=500$ trials with the initial data
    being Gaussian random vectors with zero mean and identity covariance such
    that the norm at initialization is approximately equal to that from the ground truth.
    
    We observe the following:
    \textbf{(i) Both \mI{} and \mL{} approximately preserve the relative
    magnitude between different logits.} We find
    that the ratios between converging values of sample paths (dashed lines),
    $\lim_{t\to\infty}\bar{X}^k_i(t)/\bar{X}^k_j(t)$ for $i\ne j$, are roughly equal to the
    ground truth. This indicates that both models can capture the relative
    magnitudes of logits in real dynamics.
    \textbf{(ii) The \mI{} fails to identify the correct class.}
        After manual alignment, we note the \mI{} does not always
        yield faithful results, meaning the largest logit from the $k$-th class,
        $\max_{l\in[K]} \bar{X}^k_l(t)$ for large $t$, is not necessarily $k$.
        This is not surprising though, since the \mI{} itself
        does not differentiate features from different classes,
        and $\alpha(t)$ and $\beta(t)$ thus estimated
        fails to honor the interactions between different classes.
    (iii) \textbf{The \mL{} is able to identify the correct class while oblivious to
    incorrect classes.}
        On the other hand, simulated trajectories from the \mL{} faithfully
        recover the trajectories of correct classes
        $\argmax_{l\in[K]} \bar{X}^k_l$ for each $1\le k \le K = 3$. However, for any $k\in[K]$,
        we note that the trajectories of incorrect classes (i.e., $\{j:j\ne k\}$)
        are sometimes mismatched. This is because that the \mL{}, by
        construction, only uses the information from the correct class, i.e.,
        $\H_{i,j}$ is set to be $\d_j\tp{\d}_j/\tp{\d}_j\d_j$
        while not specifying other directions.
        A model
        that is capable of identifying incorrect classes needs necessarily 
        more information on those classes. We postulate that a better model
        might be
        \ba \label{app:eq:betterH}
            \H_{i,j} = p_j \frac{\d_j\tp{\d}_j}{\tp{\d}_j\d_j}
            +\sum_{l\ne j}^{K} p_l \frac{\d_l\tp{\d}_l}{\tp{\d}_l\d_l},
        \ea
        with $p_j > p_l$ for all $l\ne j$. We leave explorations along
        this direction in future works.
        
\rednote{ 
\paragraph{Residue of LE-ODE Simulations.}

\begin{figure*}
    \centering
    \begin{subfigure}[t]{0.25\textwidth}
	    \centering
	    \includegraphics[width=\linewidth]{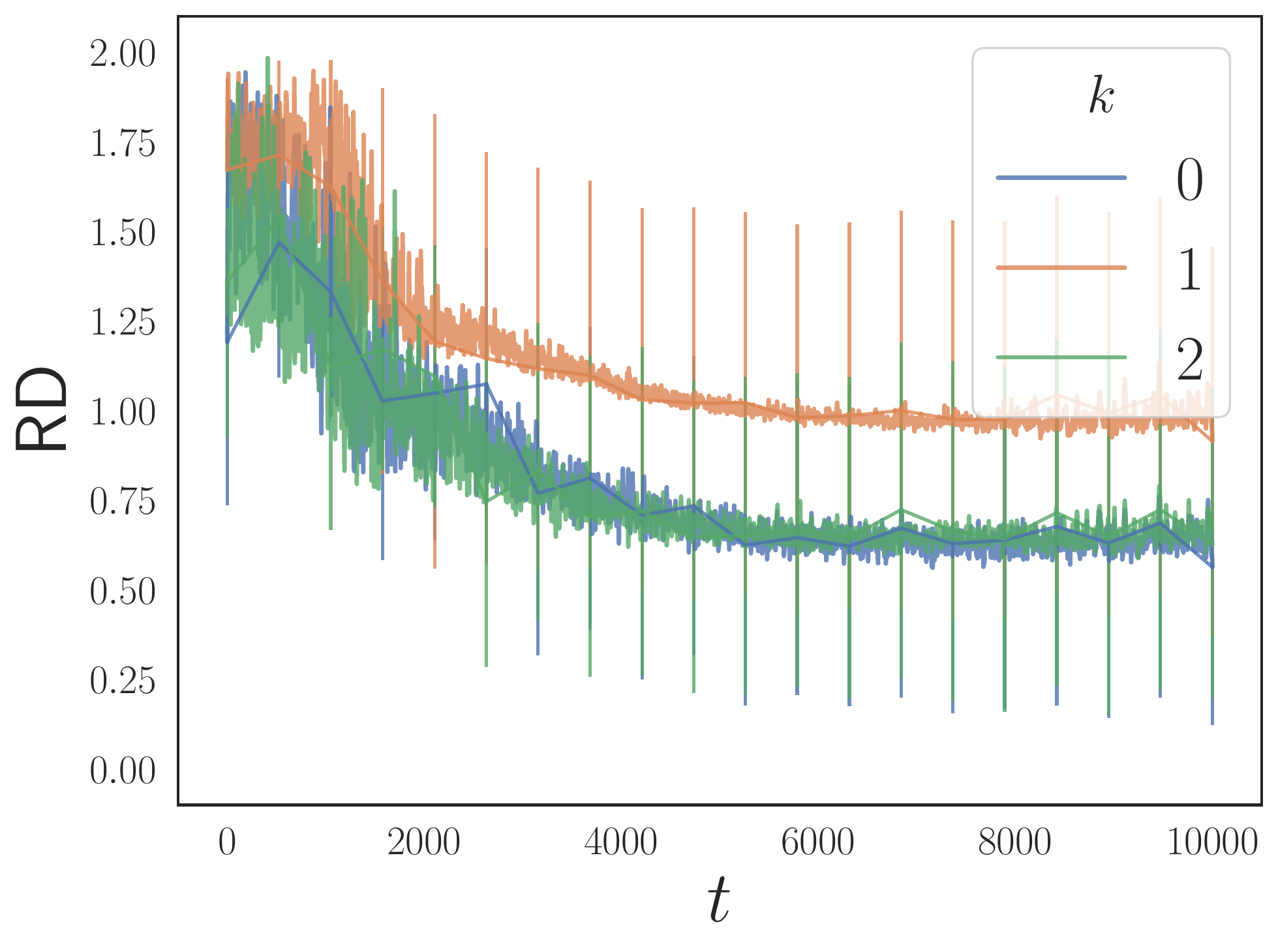}
	    \caption{\geomnist{} (\mI{}). } 
	\end{subfigure}%
	   \begin{subfigure}[t]{0.25\textwidth}
	    \centering
	    \includegraphics[width=\linewidth]{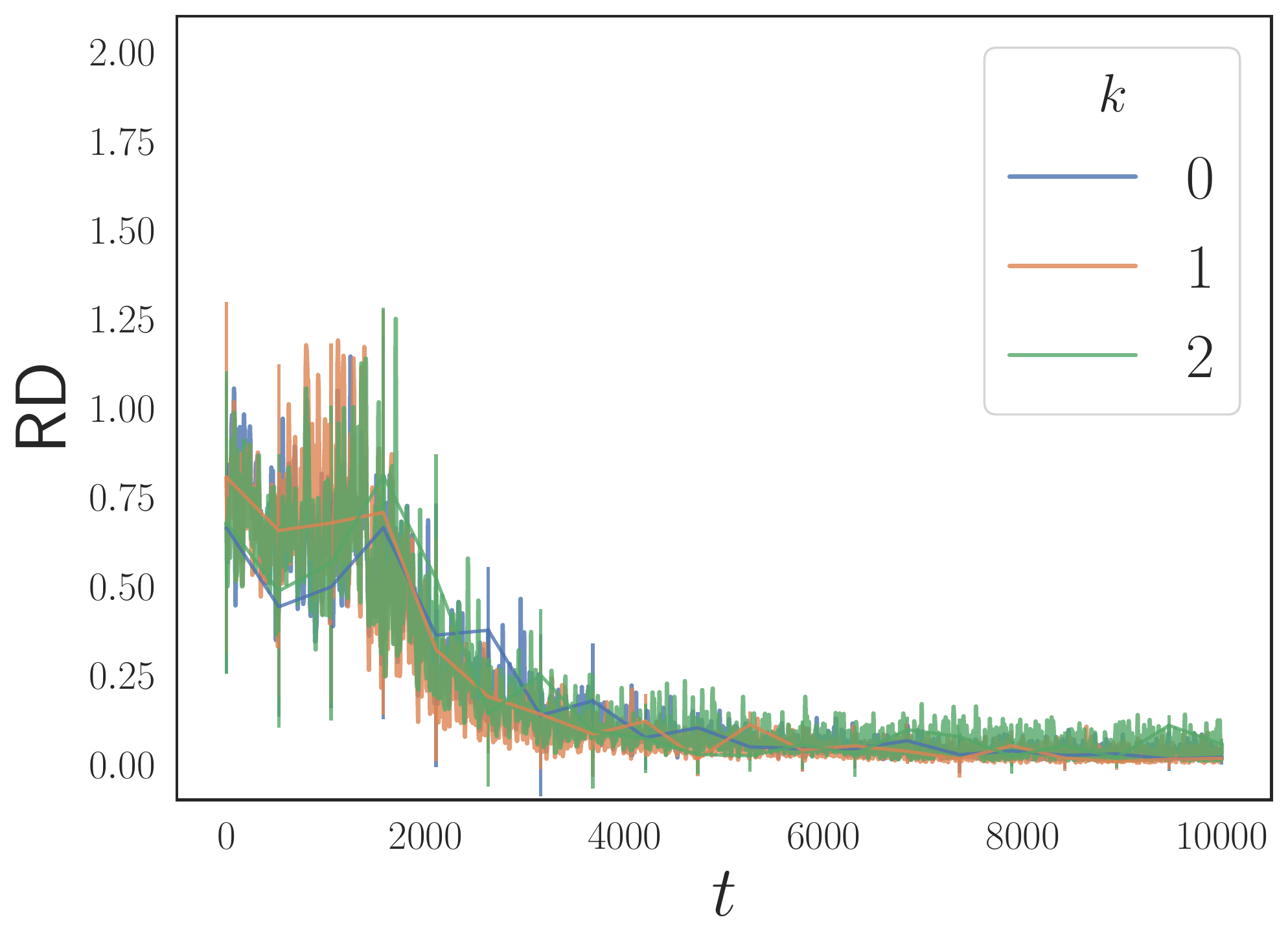}
	    \caption{\geomnist{} (\mH{}).} 
	\end{subfigure}%
    \begin{subfigure}[t]{0.25\textwidth}
	    \centering
	    \includegraphics[width=\linewidth]{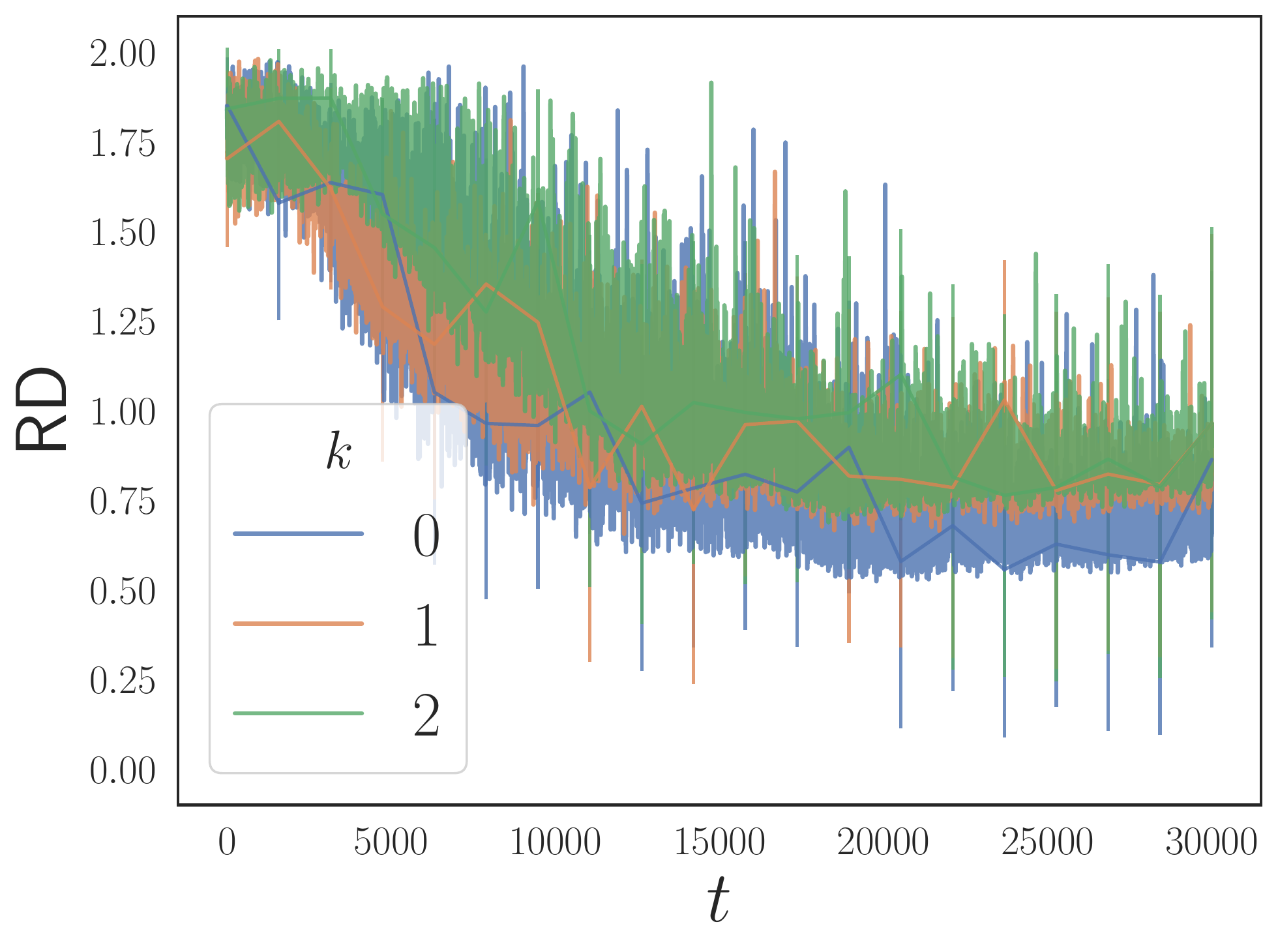}
	    \caption{\cifar{} (\mI).}
	\end{subfigure}%
    \begin{subfigure}[t]{0.25\textwidth}
	    \centering
	    \includegraphics[width=\linewidth]{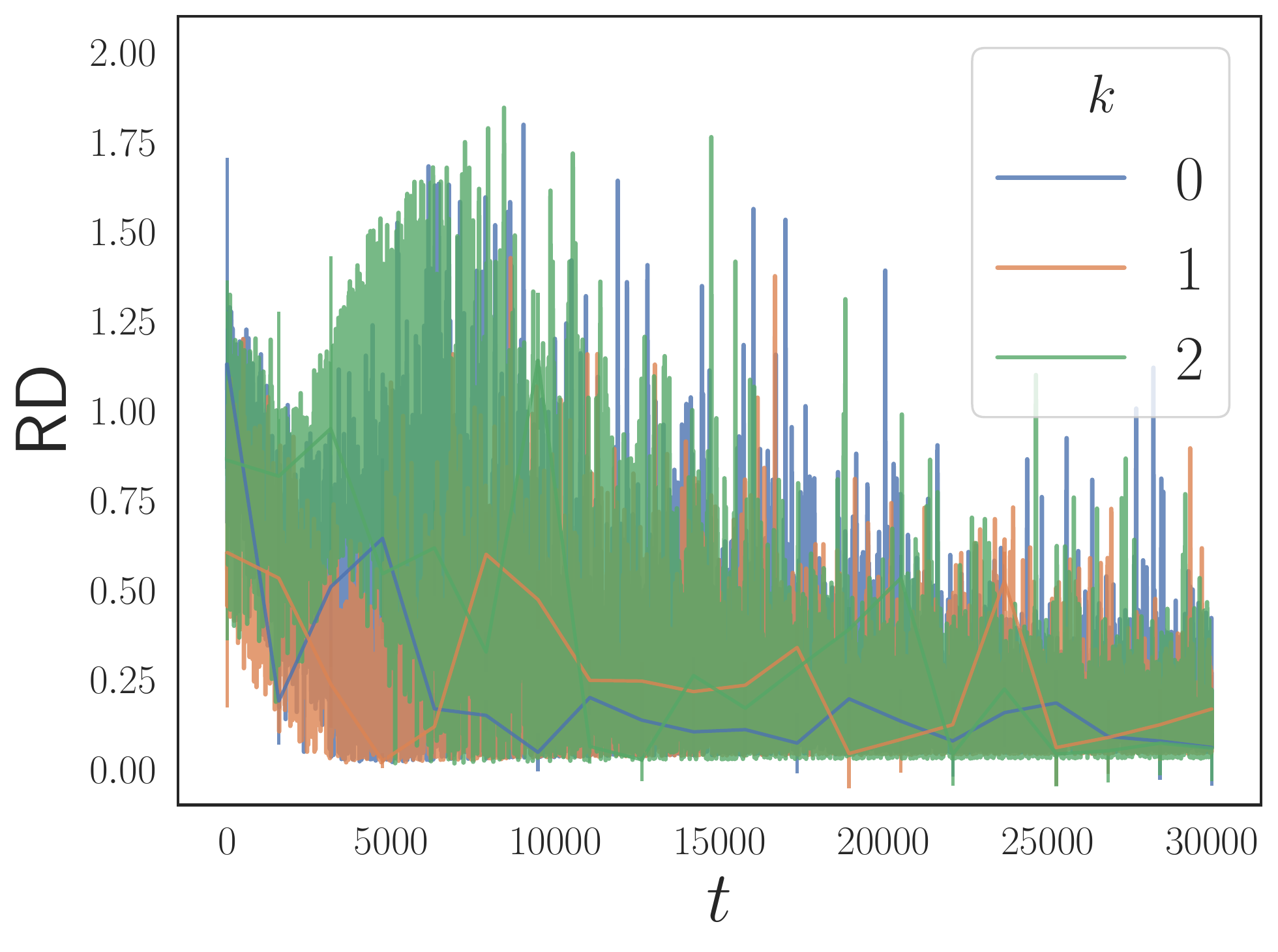}
	    \caption{\cifar{} (\mH).} 
	\end{subfigure}
        \caption{
            \textbf{Relative difference $\textsf{RD}_k$ between
            genuine and simulated
            dynamics.} The RD is computed according to equation~\ref{app:eq:RD} (the lower the better). Note that the \mL{} performs
            better than \mI{} throughout training and better captures the
            later stages of the training (indicated by decreasing RD),
            supporing our discussions in \Cref{sec:app:dynamics:linear}.
            }
        \label{fig:supp:RD}
        \vspace{-0.6cm}
\end{figure*}
    Under the same experiment setup, we also visualize the residue
    of using LE-ODE to imitate the genuine dynamics of neural nets,
    as shown in \Cref{fig:supp:RD}. We measure the goodness-of-fit
    via relative difference (RD) defined for each class $k\in[K]$ 
    as
    \ba \label{app:eq:RD}
        \operatorname{\mathsf{RD} _ k}(t) \coloneqq \frac{\norm{ \bar{\X}^k(t) - \bar{\Y}^k(t) } _ {\H^k} }{\left(\norm{ \bar{\X}^k(t) }_ 2 + \norm{\bar{\Y}^k(t) }_ 2 \right) / 2},
    \ea
    where $\bar{\X}(t)$ and $\bar{\Y}(t)$ are genuine and simulated
    trajectories, respectively, and $\norm{\cdot}_{\H}$ denotes
    the norm induced by the matrix $\H$ that is used to define
    models (i.e., the identity matrix for the \mI{} and $\bar{\H}$ in equation~\ref{eq:H:ce} for the \mH). This choice
    normalizes the difference under the similarity defined by
    $\H$ (which we care the most) and ranges from $0$ to $2$ (the lower
    the better).
    From the results we note that:
    \textbf{(i) The \mL{} performs consistently better than the
    \mI{} throughout training.} We note that the RD under the \mL{}
    are overall smaller and there is no significant differences
    across classes.
    \textbf{(ii) The \mL{} is better suited for capturing later
    stages of training.} This can be seen from a decreasing trend
    of the RD under the \mL{}, which corroborates our discussions
    in \Cref{sec:app:dynamics:linear}. In particular, the approximation 
    becomes better as training progresses (indicated by a decreasing RD). However, the performance of the \mL{} around initialization is still commendable.
    \textbf{(iii) The \mL{} is not perfect.}
    Although RD under \mL{} is small in the terminal stage (of
    the order $10^{-2}$ to $10^{-1}$), it is non-zero, and
    it has a higher RD in the early stage of training.
    This indicates that non-dominant directions are also
    important for the LE-ODE to capture the remainder
    of the feature similarity. A possible avenue for future
    research in this regard is discussed in equation~\ref{app:eq:betterH}.
}




\subsection{The Two-Stage Behavior of Logits Evolution}
An interesting observation that can be made when going through the experiments
is the emergence of a two-stage behavior in many quantities. Specifically:
(i) the training loss and validation loss do not
decrease at a perceivable rate in the first few hundreds (or thousands)
iterations; then they begin to drop at a relatively fast rate until convergence;
(ii) the local elasticity strength $\alpha(t)$ increases at the initial
stage, then drops, which is also manifested by the behavior of $A(t)$,
which resembles roughly a sigmoidal curve;
(iii) the magnitudes of $\avg_{k\in[K]} \norm{\bar{\X}^k}_2$
also resembles that of $A(t)$ (not shown), which has a fast growing stage
and a converging stage.

\begin{figure*}
    \centering
    \begin{subfigure}[t]{0.33\textwidth}
	    \centering
	    \includegraphics[width=\linewidth]{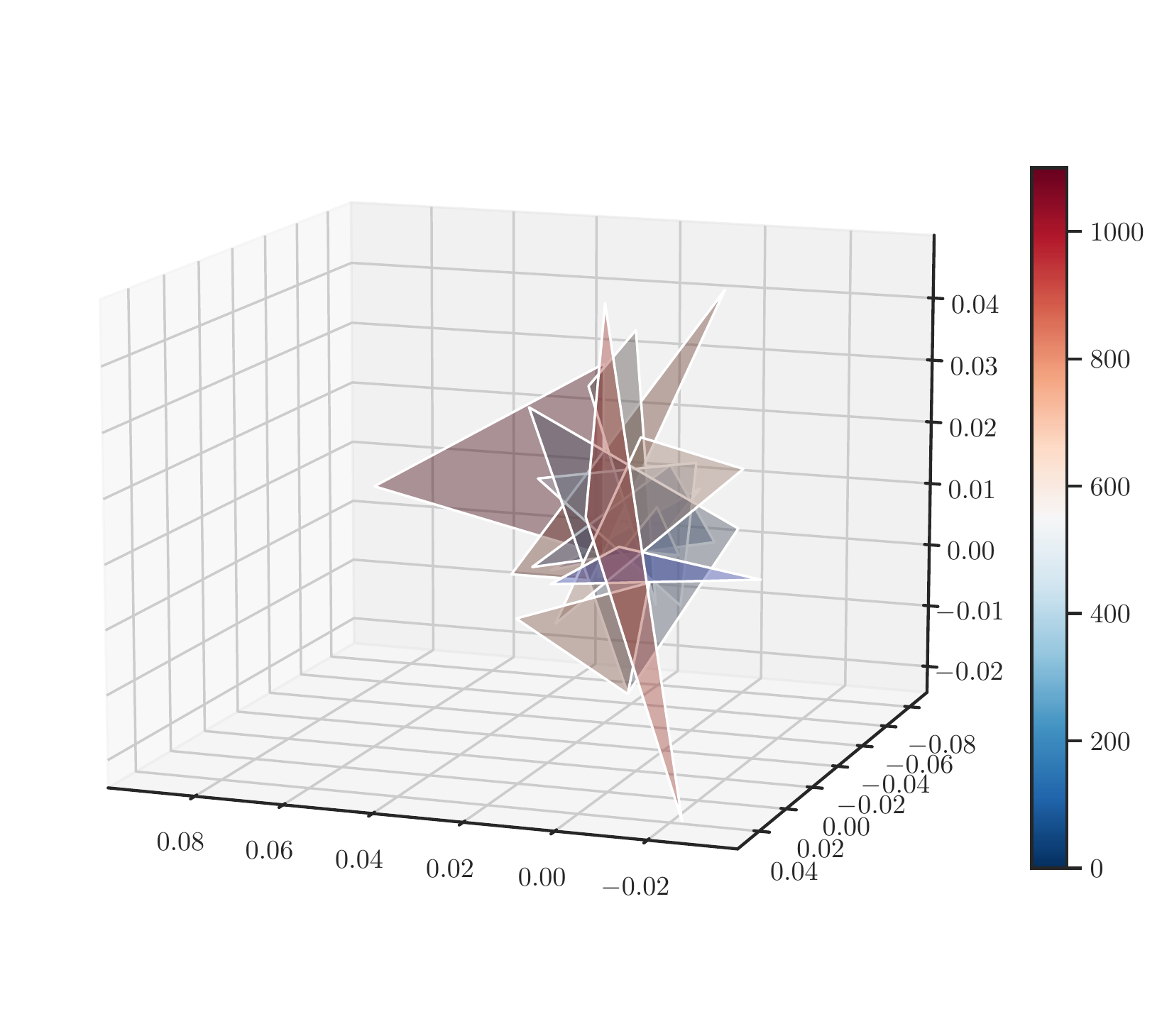}
	    \caption{De-randomization stage.} \label{fig:tri:derand}
	\end{subfigure}~
	\begin{subfigure}[t]{0.33\textwidth}
	    \centering
	    \includegraphics[width=\linewidth]{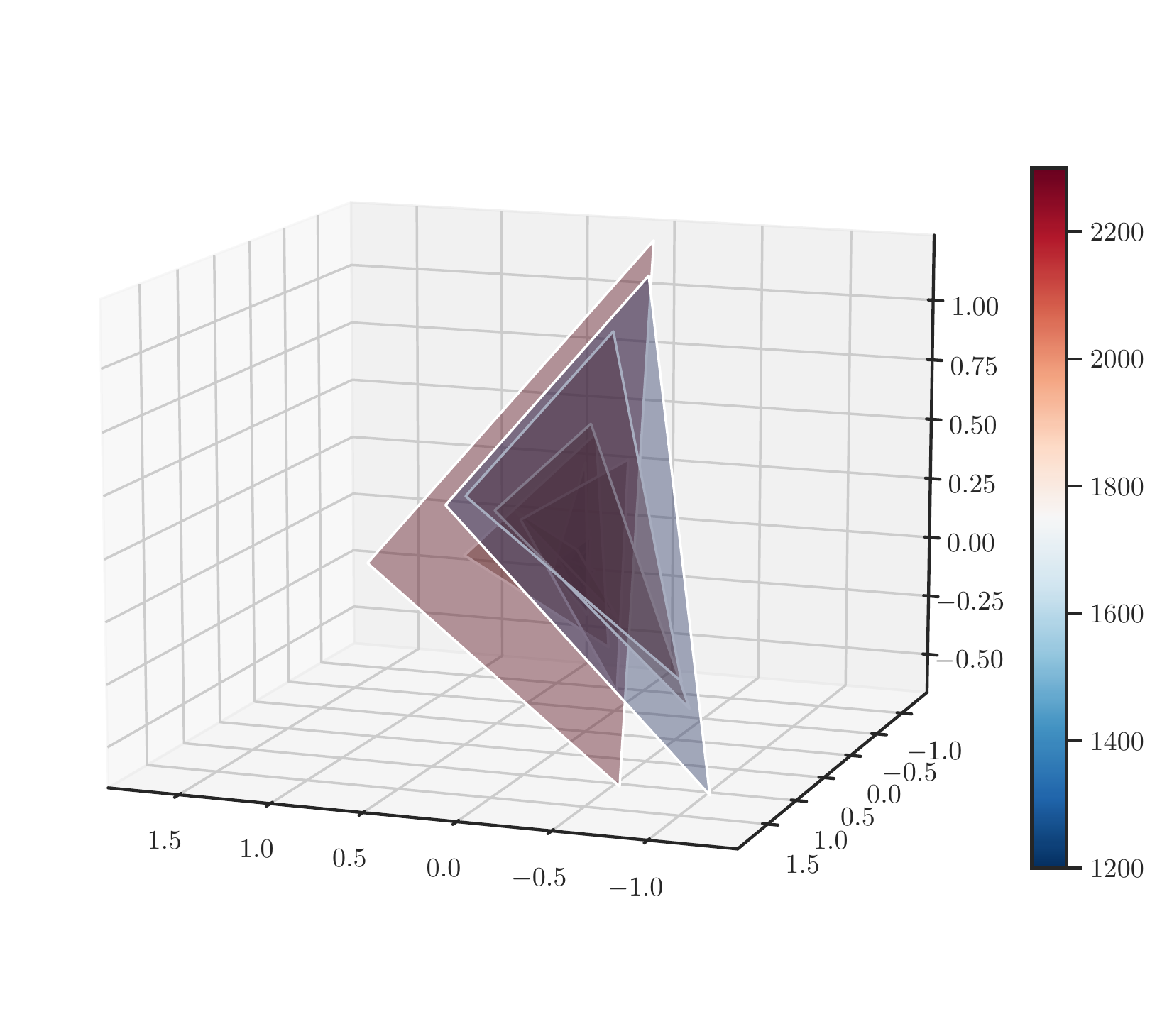}
	    \caption{Amplification stage.} \label{fig:tri:amp}
	\end{subfigure}~
	\begin{subfigure}[t]{0.33\textwidth}
	    \centering
	    \includegraphics[width=\linewidth]{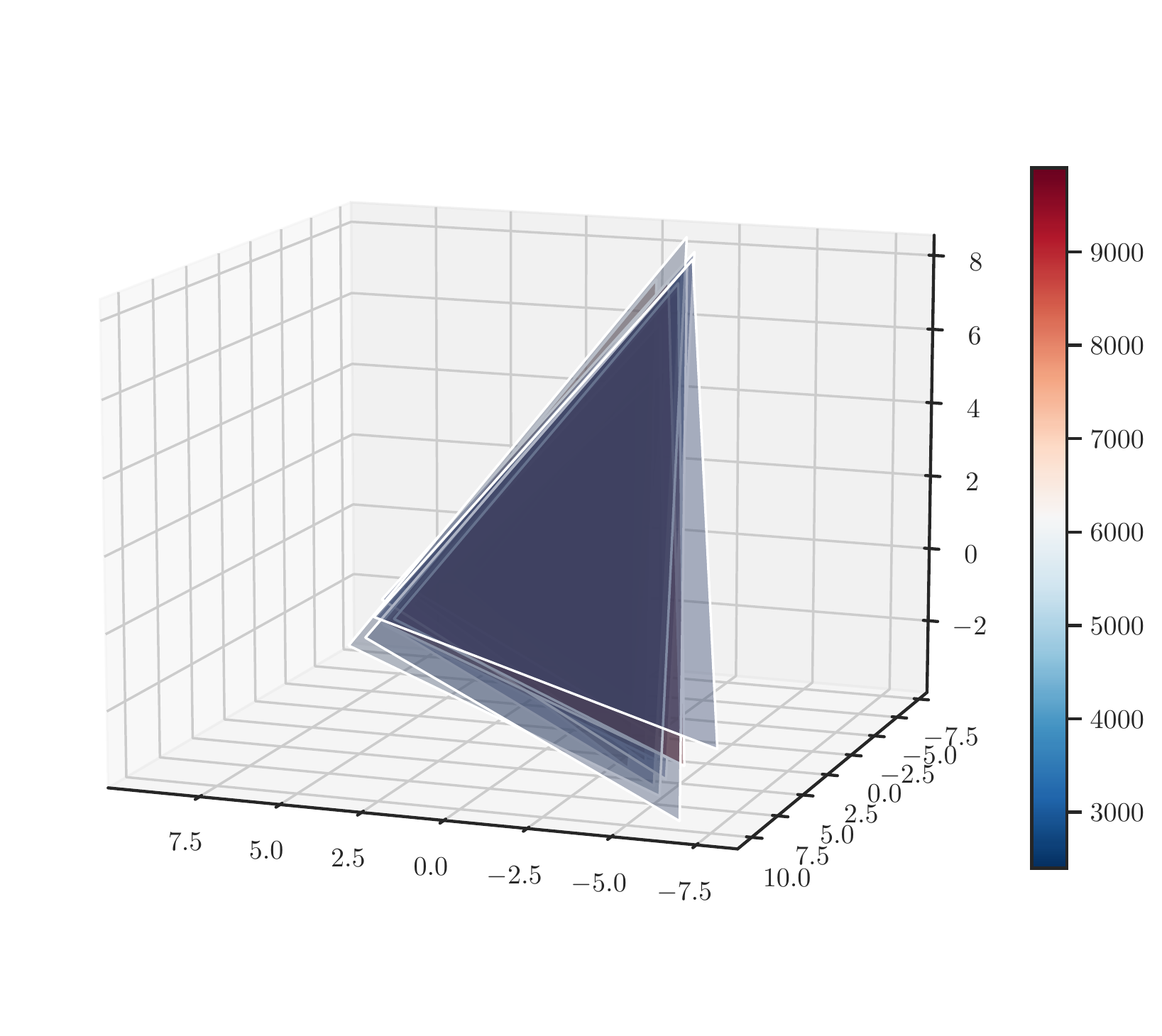}
	    \caption{Amplification stage (cont'd).} \label{fig:tri:amp2}
	\end{subfigure}~
        \caption{
        \textbf{The hyperplanes formed by $\{\bar{X}^k_t\}_{k=1}^3$
        with colors corresponding to iterations.}
        (a) In the first $1200$ iterations, the validation loss
        does not drop perceivably, 
        and the hyperplanes are visibly chaotic.
        (b)-(c) Starting from around the $1200$-th iteration,
        the validation loss drops, and the hyperplanes exhibit
        converging behavior. }
        \label{fig:tri}
\end{figure*}

We coin this seemingly generic phenomenon as the \emph{two-stage behavior}
that consists of a \emph{de-randomization} stage and an \emph{amplification}
stage,
and
demonstrate the supporting experiments
in \Cref{fig:tri}.
Here we trained on \geomnist{} with $K=3$, where
each triangle is a hyperplane spanned by the per-class mean logits vectors, $\bar{\X}^k(t)\in\sR^3$ for
$k\in[3]$.
In \Cref{fig:tri:derand} we plot those hyperplanes for the first $1200$ iterations;
\Cref{fig:tri:amp} iterations from $1200$ to $2400$; and
\Cref{fig:tri:amp2} the remaining iterations.
We observe that in the first $1200$ iterations, the hyperplanes are ``chaotic''
in that their behavior is highly dependent on specific initialization values.
In this de-randomization stage, the supervision guides the dynamics to identify
the correct and \emph{deterministic} (c.f. \Cref{thm:collapse}) direction for the separation of
features from random initialization, thus the name \emph{de-randomization}.
With the correct direction being identified (around iteration $1200$),
the losses begin to drop at a relatively fast speed and the hyperplane remains
approximately the same throughout the training while the magnitude
of logits increases, which pushes the classes to be more discriminative
and further drives down the losses, hence the \emph{amplification} stage.

We believe a more precise characterization of $\alpha(t)$ and $\beta(t)$
and potentially generalization to the $\E$ matrix
would likely
help us to study this two-stage behavior more rigorously
and potentially answer the interesting
questions such as \emph{how long does the first stage take on average?}
\emph{How does local elasticity ($\alpha(t)$ and $\beta(t)$) affect this behavior?} We leave these questions for future works.


\section{Future Work and Extensions}\label{sec:app_ext}

\paragraph{General LE Matrix.}
    Throughout the paper, we have modeled the LE matrix
    $\E$ as $(\alpha(t)-\beta(t))\I_K + \beta(t) \bone_K\bone_K^\top$. Although it depends on $t$,
    this model falls short when we move into the more realistic realm
    where the inter-class and intra-class effects are 
    dependent on the class labels.
    When $\E$ is SPD,
    under the same assumptions as in
    \Cref{thm:separation:mean},
    due to a theorem
    by Schur (Theorem 9.B.1, \cite{marshall11}),
    we know the eigenvalues of $(\E\otimes\I_K)\circ \H$ 
    are bounded within $\ld_{\min}(\E)\min_{i} H_{ii}$
    and $\ld_{\max}(\E) \max_i H_{ii}$, where $\{H_{ii}, i\in [Kp]\}$
    is the diagonal entries of the matrix $\H$, hence we expect
    a very similar result in this case as in \Cref{thm:separation:mean}.
    In the more general case where $\E$ is symmetric but not necessarily
    semi-definite, a more precise analysis on the spectrum
    of $(\E\otimes\I_K)\circ \H$ 
    is needed, though the proof framework would not be too
    different.

\paragraph{Mini-batch Training, Imbalanced Datasets, and Label Corruptions.}
    As discussed in \Cref{sec:lesde:general}, we can incorporate 
    mini-batches and imbalanced datasets in our model easily. 
    Taking imbalanced datasets as an example,
    recall that each block of $\M_t$ in \cref{eq:LE-SDE_general_continuous} takes
    the form of $E_{k, l} \H_{k, l}/K$, 
    where $1/K$ signifies that
    each class has the same $1/K$ probability of 
    being sampled during any iteration in training.
    This can be generalized by changing the $(k,l)$-th block
    to $E_{k, l} \H_{k, l}\cdot p_l$ for $\sum_l p_l = 1$, where $p_l$ is the probability of class $l$ being sampled. More succinctly,
    instead of defining $\M_t = \left(\E_t \otimes \I_K\right)\circ \H/K$, we let
    $\M_t = \left(\E_t \otimes \P\right)\circ \H/K$ for a $K$-by-$K$
    doubly stochastic matrix $\P$ that models this sampling
    effect.
    In the same vein,
    we can also model the case when the data are polluted by
    corrupted labels. 
    Let $\P = (p_{k,l})_{k,l}\in\sR^{K\times K}$
    with $p_{k,l}$ representing
    the probability of a
    sample from class $k$ mis-labelled as class $l$.
    A well-defined model needs further assumption on the structure
    of $\P$ and we leave this theoretical modeling to future work. 

\paragraph{Covariance Structures and Fine-Grained Analyses.} 
    Although our model encompasses a covariance term in the LE-SDE
    model, we do not explicitly use its structure.
    Nonetheless, as indicated from the proof of \Cref{thm:separation:mean}
    (cf.~\Cref{sec:app:pf:separation}), the relative magnitude of the covariance
    to the drift term (i.e., the local elasticity effect) affects the separation
    when $\gamma(t) = \Theta(1/t)$. However, when the order
    of $\gamma(t)$ is guaranteed to be strictly above or below $1/t$ as $t\to\infty$,
    the covariance affects the separation only through the constant factor
    for the separation rate. That said, a more precise analysis of covariance
    would by all means facilitate fine-grained analyses at the edge of separation.

\paragraph{Beyond \mL{} for Imitating Genuine Dynamics of DNNs.}
    We show in \Cref{sec:experiments} that using estimates
    of $\alpha(t)$ and $\beta(t)$, the \mL{} can be used
    to imitate the genuine dynamics of DNNs. As is shown in
    more detail in \Cref{sec:app_sim:results}, we note that
    although simulations under the \mL{} are already
    superior to those under the \mI{} in that the correct
    classes are identified, the \mL{} sometimes still
    fails to identify the correct trajectories for the incorrect
    classes. This is not very surprising though, as the supervision
    from the labels only affects \mL{} though the $\H$ matrix,
    whose $(i,j)$-th block is defined as
    $\bar{\H}^j = \d_j\tp{\d}_j/\tp{\d}_j\d_j$
    where $\d_j = \e_j - \bone_K / K$ --- which
    only encodes information about the correct class. 
    We postulate that a more precise model
    might be to assign the $(i,j)$-th block
    of $\H$ as
    \ba
        \H_{i,j} = p_j^j \bar{\H}^j + \sum_{l\ne j}^K p_l^j \bar{\H}^l, \quad
        p_j^j > p_l^j, \quad
        \sum_{l=1}^K p_l^j = 1,
        \quad \forall l \ne j, \quad
        j \in[K].
    \ea

\paragraph{The Two-Stage Behavior.}
    In \Cref{sec:experiments}, we observed a clear two-stage behavior
    of our numerical simulation of our LE-SDE, as well as in real deep learning dynamics. The first stage is a \emph{de-randomization} stage, which gradually eliminates
    the effect of random initialization and searches for the correct directions
    to be separated. The second stage is an \emph{amplification} stage, where the model amplifies the magnitudes of the features
    in those directions. As can be seen from \Cref{sec:app_sim:results}, these empirical observations naturally lead to many
    interesting questions: 
    Is this a universal phenomenon in deep learning, and what are the conditions to guarantee entering the second stage? Can our LE-SDE predict such a two-stage phenomenon theoretically?
    What is the role of local elasticity in this
    transition? We leave the investigation of these questions to future works.


\end{document}